\documentclass[
	a4paper, %
	fontsize=10pt, %
	twoside=true, %
	numbers=noenddot, %
	pdfa
]{kaobook}

\usepackage{kaobiblio}
\usepackage[framed=true]{kaotheorems}

\usepackage{kaorefs}

\usepackage{soul}
\usepackage{xcolor}
\definecolor{uniknblue}{HTML}{A6E1F4}

\DeclareUnicodeCharacter{2212}{-}

\usepackage{wrapfig}

\newcommand{\tironianEt}{\raisebox{-0.175ex}{\scalebox{0.85}{7}}}

\usepackage{tabto}

\usepackage{csvsimple}
\usepackage{dsfont}

\usepackage[frozencache]{minted}

\newcommand{\mnistimgheight}{.85em}
\newcommand{\mnistimgraise}{-.1em}
\newcommand{\mnistimg}[1]{\raisebox{\mnistimgraise}{\includegraphics[height=\mnistimgheight]{#1}}}

\newcommand{\svhnimg}[1]{\raisebox{-.4em}{\includegraphics[height=1.35em]{#1}}}
\newcommand{\emnistimgraise}{-.1em}
\newcommand{\emnistimg}[1]{\raisebox{\emnistimgraise}{\includegraphics[width=.9\linewidth]{#1}}}

\newcommand{\steepness}{\beta}

\usepackage{pifont}%
\newcommand{\cmark}{\ding{51}}%
\newcommand{\xmark}{\ding{55}}%

\newcommand{\ARTstrength}{\lambda}

\def\IR{\mathbb{R}}
\def\softmin{\mathit{min}}
\def\softmax{\mathit{max}}
\def\softargmin{\mathit{argmin}}
\def\softargmax{\mathit{argmax}}

\newcounter{arrown}

\usepackage{nicefrac}

\newcommand{\subheading}[1]{{\textbf{#1}\hspace{1em}}}

\newcommand{\density}{f}
\newcommand{\cdf}{F}
\newcommand{\cdfsq}{F_\text{sq}}
\newcommand{\cdfrev}{F_\text{Rev.}}

\newcommand{\numrenderers}{1\,242}

\usepackage{justtrees}

\usepackage{amsmath,amsfonts,bm}

\def\1{\bm{1}}

\def\vi{{\bm{i}}}
\def\vj{{\bm{j}}}

\def\vp{{\bm{p}}}

\def\vw{{\bm{w}}}

\def\mP{{\bm{P}}}

\DeclareMathAlphabet{\mathsfit}{\encodingdefault}{\sfdefault}{m}{sl}
\SetMathAlphabet{\mathsfit}{bold}{\encodingdefault}{\sfdefault}{bx}{n}
\newcommand{\tens}[1]{\bm{\mathsfit{#1}}}
\def\tA{{\tens{A}}}

\def\sR{{\mathbb{R}}}

\newcommand{\sigmoid}{\sigma}

\DeclareMathOperator*{\argmax}{arg\,max}

\newcommand{\inlinelogo}[1]{\raisebox{-.3em}{~\includegraphics[height=\baselineskip]{logos/#1_logo.png}~}}

\usepackage{tikzpagenodes}

\newcommand{\pagelogo}[1]{%
\begin{tikzpicture}[remember picture, overlay, shift={(current page.north east)}]
\node[anchor=north east,xshift=-.5cm,yshift=-.5cm]{\includegraphics[width=5.5cm]{logos/#1_logo.png}};
\end{tikzpicture}%
}

\usepackage{adjmulticol}

\begin{document}

\title[Learning with Differentiable Algorithms]{\normalfont\textsf{\setul{0.275ex}{0.15ex}\setulcolor{uniknblue}\ul{Learning with Differentiable Algorithms}}\vspace{2em}}

\subtitle{%
    \normalfont Doctoral Thesis for obtaining the Academic\\%
    Degree Doctor of Natural Sciences (Dr.~rer.~nat.)\\%
    \vspace{3em}%
    submitted by\\%
    \vspace{3em}%
    {\Large Felix Petersen}\\%
    \vspace{3em}%
    at the University of Konstanz\\%
    \vspace{1.125em}%
    \includegraphics[width=.59\linewidth]{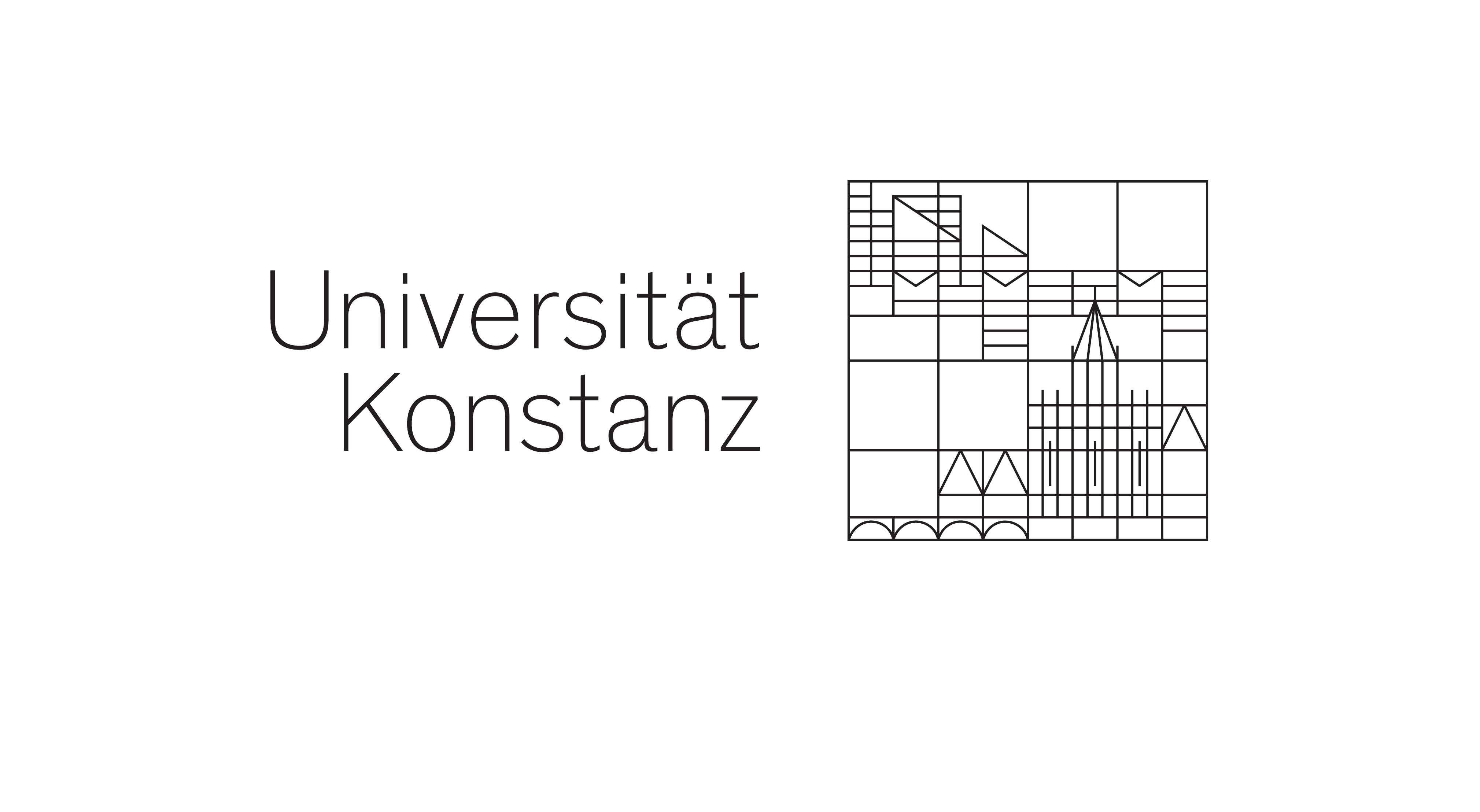}\\%
    \vspace{-.5em}%
    Faculty of Sciences\\%
    Department of Computer and Information Science\\%
    \vspace{3.em}%
    Reviewers\\[.5em]%
    \newdimen\oliverwidth\settowidth\oliverwidth{Oliver Deussen}%
    \newdimen\drwidth\settowidth\drwidth{Prof.~Dr.}%
    \begin{minipage}[t]{\drwidth}
        Prof.~Dr.\linebreak%
        Prof.~Dr.\linebreak%
        Prof.~Dr.\linebreak%
    \end{minipage}~~%
    \begin{minipage}[t]{\oliverwidth}
        Oliver Deussen\linebreak%
        H\kern0.04emi\kern0.04eml\kern0.04emd\kern0.04eme K\kern0.03emu\kern0.03eme\kern0.03emh\kern0.03emn\kern0.03eme\linebreak%
        T\kern0.0emo\kern0.03emb\kern0.04emi\kern0.055ema\kern0.05ems S\kern0.04emu\kern0.055emt\kern0.055emt\kern0.055eme\kern0.055emr\linebreak%
    \end{minipage}%
    \vspace{-1em}
}

\author[Felix Petersen]{}
\date{August 22, 2022}

\frontmatter %

\makeatletter
\uppertitleback{\@titlehead} %

\lowertitleback{
	
	\textbf{Submission \& Defense}\\
	This work was submitted on June 1, 2022, and successfully defended on August 22, 2022.

	\medskip
	
	\textbf{Copyright}\\
	\textcopyright 2022 Felix Petersen. All Rights Reserved.

	\medskip
	
	\textbf{Distribution}\\
	This work will be publicly accessible on arXiv and KOPS (the Institutional Repository of the University of Konstanz).
	
	\medskip
	
	\textbf{Code} \\
	The libraries accompanying this work will be publicly available under:\\
	
	{\Large\inlinelogo{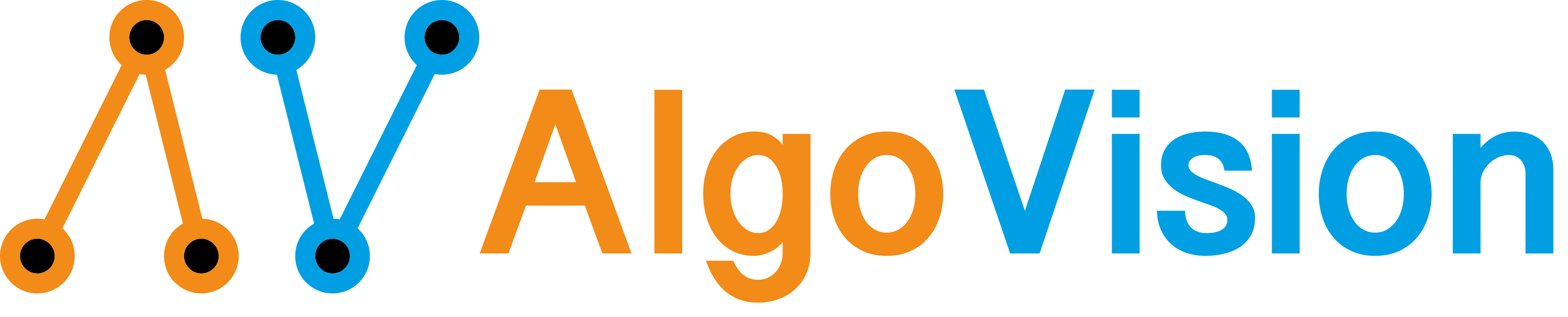}}\tabto{11em}\url{https://github.com/Felix-Petersen/algovision}\\
	
	{\large\inlinelogo{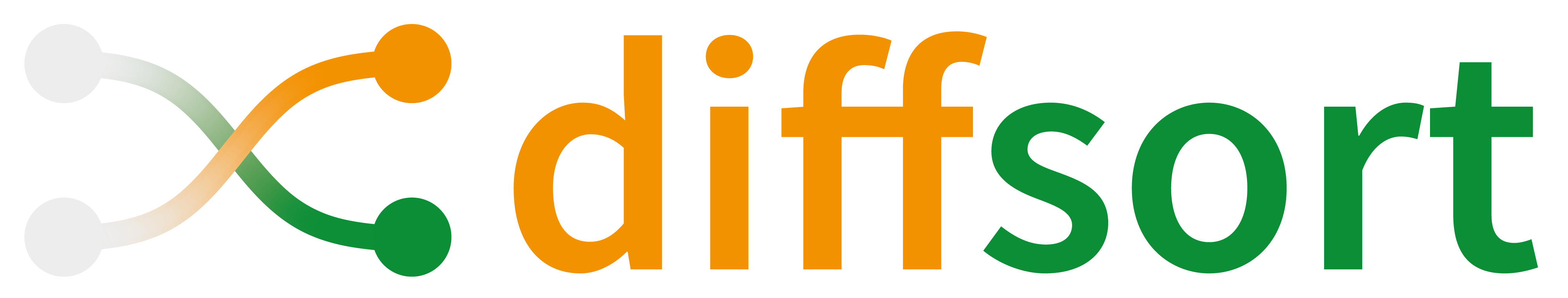}}\tabto{11em}\url{https://github.com/Felix-Petersen/diffsort}\\
	
	{\large\inlinelogo{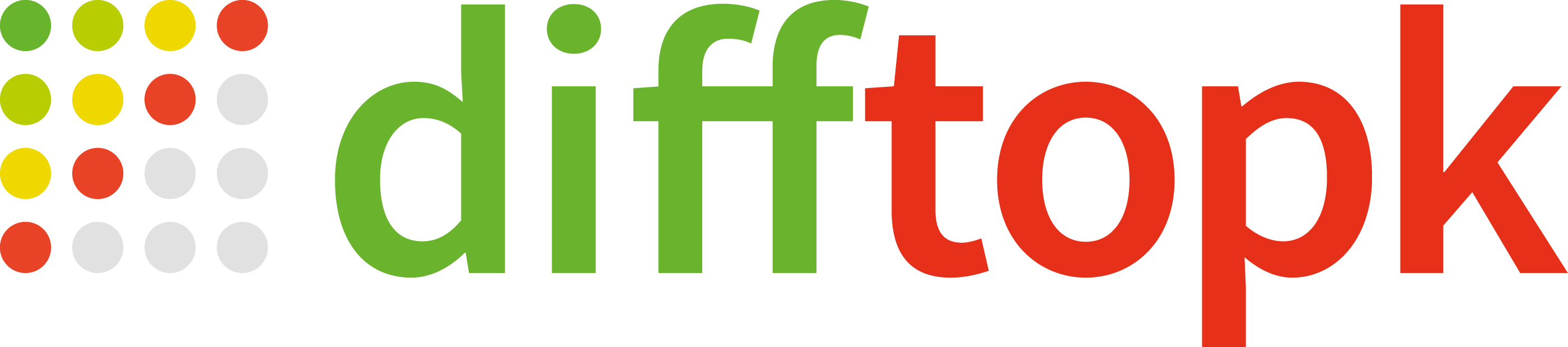}}\tabto{11em}\url{https://github.com/Felix-Petersen/difftopk}\\
	
	\scalebox{1.075}{\large\inlinelogo{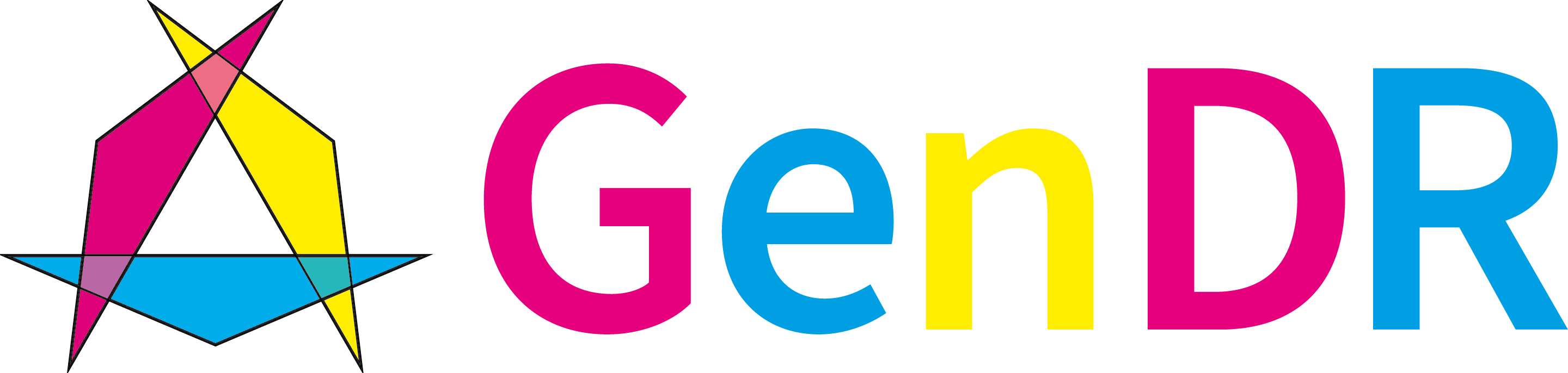}}\tabto{11em}\url{https://github.com/Felix-Petersen/gendr}\\
	
	\scalebox{1.35}{\inlinelogo{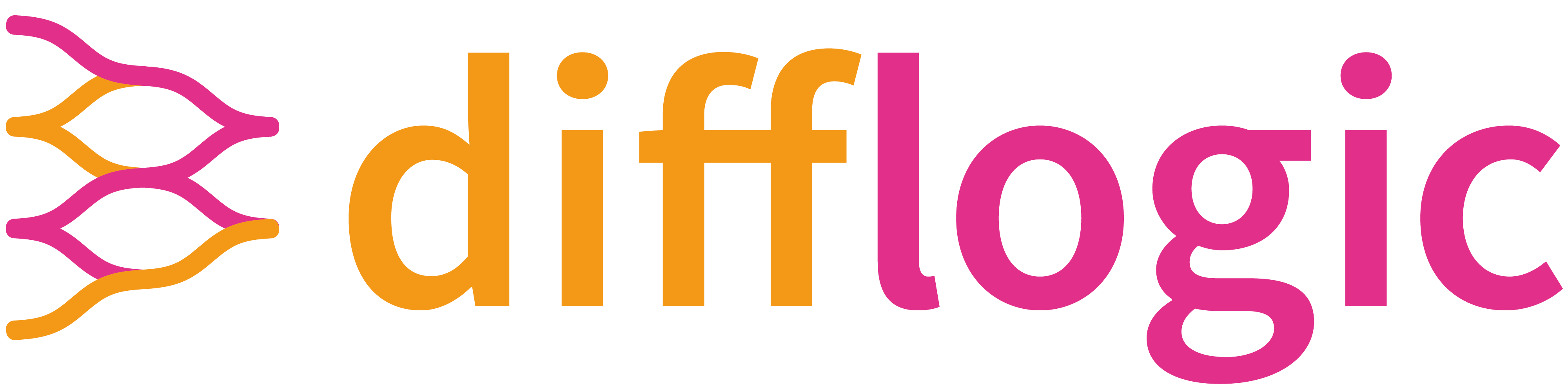}}\tabto{11em}\url{https://github.com/Felix-Petersen/difflogic} \textit{(to be published soon)}\\
	
	{\large\inlinelogo{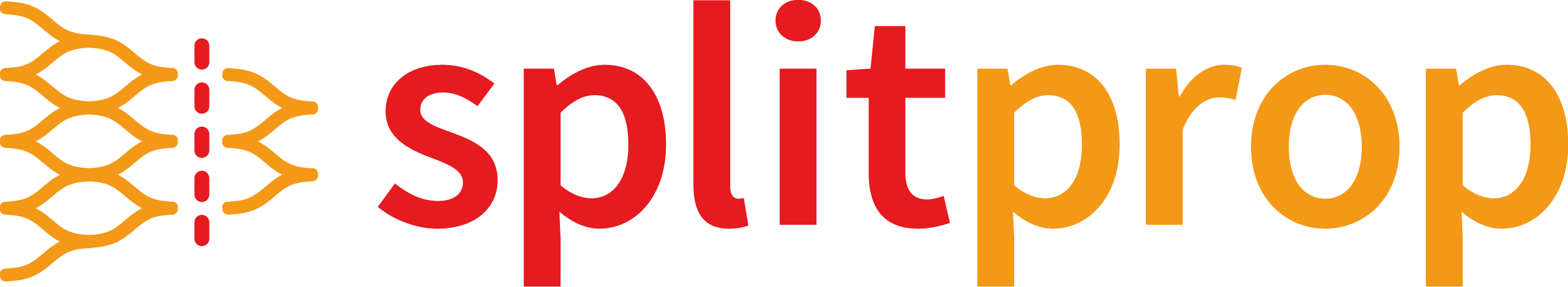}}\tabto{11em}\url{https://github.com/Felix-Petersen/splitprop} \textit{(to be published soon)} \\

	\medskip
	
}
\makeatother

\maketitle

\pagestyle{scrheadings}%
\pagelayout{margin}%
\setchapterstyle{kao} %

\chapter*{Abstract}
\addcontentsline{toc}{chapter}{Abstract}

Classic algorithms and machine learning systems like neural networks are both abundant in everyday life.
While classic computer science algorithms are suitable for precise execution of exactly defined tasks such as finding the shortest path in a large graph, neural networks allow learning from data to predict the most likely answer in more complex tasks such as image classification, which cannot be reduced to an exact algorithm.
To get the best of both worlds, this thesis explores combining both concepts leading to more robust, better performing, more interpretable, more computationally efficient, and more data efficient architectures.
The thesis formalizes the idea of algorithmic supervision, which allows a neural network to learn from or in conjunction with an algorithm.
When integrating an algorithm into a neural architecture, it is important that the algorithm is differentiable such that the architecture can be trained end-to-end and gradients can be propagated back through the algorithm in a meaningful way.
To make algorithms differentiable, this thesis proposes a general method for continuously relaxing algorithms by perturbing variables and approximating the expectation value in closed form, i.e., without sampling.
In addition, this thesis proposes differentiable algorithms, such as differentiable sorting networks, differentiable renderers, and differentiable logic gate networks.
Finally, this thesis presents alternative training strategies for learning with algorithms.

\chapter*{Zusammenfassung}
\addcontentsline{toc}{chapter}{Zusammenfassung}

Klassische Algorithmen und maschinelle Lernsysteme wie neuronale Netze begegnen uns beide häufig im Alltag. %
Während sich klassische Informatik-Algorithmen für die präzise Ausführung genau definierter Aufgaben wie das Finden des kürzesten Weges in einem großen Graphen eignen, ermöglichen neuronale Netze das Lernen aus Daten, um bei komplexeren Aufgaben wie der Bildklassifizierung die wahrscheinlichste Antwort vorherzusagen.
Um das Beste aus beiden Welten herauszuholen, wird die Kombination beider Konzepte untersucht, was zu robusteren, leistungsfähigeren, besser interpretierbaren, recheneffizienteren und dateneffizienteren Architekturen führt.
Diese Dissertation formalisiert die Idee der algorithmischen Überwachung, welche es neuronalen Netzwerken ermöglicht, von einem Algorithmus oder in Verbindung mit einem Algorithmus zu lernen.
Bei der Integration eines Algorithmus in eine neuronale Architektur ist es wichtig, dass der Algorithmus differenzierbar ist, sodass die Architektur Ende-zu-Ende trainiert werden kann und Gradienten auf sinnvolle Weise durch den Algorithmus zurück propagiert werden können.
Um Algorithmen differenzierbar zu machen, präsentiert diese Arbeit ein allgemeines Verfahren zur stetigen Relaxierung von Algorithmen, indem wir Unsicherheit durch eine Verteilung einführen und den Erwartungswert in geschlossener Form approximieren.
Überdies präsentiert diese Arbeit differenzierbare Algorithmen wie differenzierbare Sortier-Netzwerke, differenzierbare Renderer, und differenzierbare Logik-Gatter-Netzwerke.
Abschließend werden alternative Trainingsstrategien für das Lernen mit Algorithmen vorgestellt.

\chapter*{Acknowledgments}
\addcontentsline{toc}{chapter}{Acknowledgments}

First and foremost, I want to thank my advisor, Oliver Deussen, for offering me a position in his research group.
I am most grateful for Oliver's support and encouragement toward pursuing my own research interests, which slowly moved away from computer graphics to computer vision and core machine learning.
While this direction is notably different from the core research direction of his lab, Oliver was a very effective and supportive advisor, providing invaluable guidance not only academically but also personally.

Moreover, I would like to give special thanks to my core collaborators Hilde Kuehne and Christian Borgelt.
It was a great pleasure to collaborate on more than ten publications during the last years.
I especially admire Hilde's compassion for simplifying complex matters and Christian's encyclopedic knowledge of classical machine learning concepts.
I would also like to thank Tobias Sutter for collaborating on four papers, even though we have known each other for only half a year.
I would like to thank my collaborators Mikhail Yurochkin and Yuekai Sun for introducing me to learning theory and reinforcing notation and clarity of theoretical results.

I would like to give special thanks to Bela Gipp for believing in me from the first moment, accepting me as a full member in his group, and offering me a position.
I would like to extend my thanks to all members of Bela's lab.
I would also like to thank Daniel Cohen-Or and his entire group for hosting me as a visiting researcher at the University of Tel Aviv for one semester.

Further, I would like to also give many thanks to all of my other collaborators, colleagues, and students.
Especially, I want to mention Hendrik Strobelt, Bastian Goldluecke, Robert Denk, Debarghya Mukherjee, Nina Shvetsova, and Ruiheng~Wu.

I would like to thank Oliver Seeck, Franz Kaertner, and Ingmar Hartl for hosting me for internships at DESY.
Further, I would like to thank Tilman Irmscher for convincing me to submit my first research paper to Jugend forscht.
I would like to give special thanks to Norina Procopan and Ulrike Leitner for enabling me to pursue my university studies during high school.

I want to thank my father for sparking my critical thinking and my early interest in science, as well as my grandmothers for their generosity and financial support during my early studies.
I want to thank my best friend Nicolas Afritsch for our two decades-long close friendship. 
Further, I want to greatly thank the Welzel family for accepting me as if I were part of their family.

\pagelayout{wide}%
\setchapterstyle{plain} %

\chapter*{Preface}
\addcontentsline{toc}{chapter}{Preface} %

In this thesis, we propose novel approaches for learning with differentiable algorithms and integrating algorithms into neural network architectures.
Specifically, this thesis comprises the following contributions:
\begin{itemize}
    \item We formalize the idea of algorithmic supervision, i.e., settings where an algorithm is applied to the predictions of a model and only the outputs of the algorithm are supervised.
    \item We propose analytical methods (contrasting stochastic methods) for relaxing algorithms and algorithmic concepts.
    \item We propose the idea of variable perturbation (contrasting input perturbations).
    \item We propose a general method for making arbitrary simple algorithms differentiable. 
    \item We propose differentiable sorting networks, a continuous relaxation of sorting networks, and devise a theoretical characterization of monotonic and error-bounded differentiable sorting networks, leading to substantial empirical improvements on ranking supervision tasks.
    \item We propose differentiable top-$k$ classification learning, which relaxes the assumption of optimizing only for the top-1 or top-5 classification accuracy through the integration of differentiable sorting and ranking.
    \item We propose an array of differentiable renderers summarized as GenDR, the generalized differentiable renderer.
    \item We propose differentiable logic gate networks, a relaxation of logic gate networks, which achieve unprecedented inference speeds.
    \item We propose split backpropagation, which allows splitting the training of neural networks into multiple stages while maintaining end-to-end training capabilities.  
    \item We propose Newton losses, an instance of split backpropagation, which incorporate second-order information of algorithmic loss functions into the training while training the actual neural network with fast first-order methods.
    \item We propose the Regularized Sampling Greedy Optimizer (RESGRO), another instance of split backpropagation, which is a very simple yet general method for learning end-to-end through hard-to-optimize functions, including non-differentiable functions.
\end{itemize}

This thesis is based on the following publications, all of which are collaborative works. 
For each publication, the contributions of each author are listed.
In addition, the respective accompanying libraries, including respective implementations, are indicated by their logo.

\newcommand{\defaultauthorcontributions}{
The initial idea was conceived by the author. 
The algorithm, details, and implementation were developed by the author. 
All authors contributed to the text of the final publication.
}

\begin{itemize}
    \item 
    F.~Petersen, C.~Borgelt, H.~Kuehne, and O.~Deussen, 
    \textit{Learning with Algorithmic Supervision via Continuous Relaxations}, 
    in Advances in Neural Information Processing Systems (NeurIPS),
    2021.~\cite{petersen2021learning}~\inlinelogo{algovision}
    \\[.5em]
    This work is covered in Chapter~\ref{ch:algovision}. \defaultauthorcontributions\\
    
    \item 
    F.~Petersen, C.~Borgelt, H.~Kuehne, and O.~Deussen, 
    \textit{Differentiable Sorting Networks for Scalable Sorting and Ranking Supervision}, 
    in Proceedings of the International Conference on Machine Learning (ICML),
    2021.~\cite{petersen2021diffsort}~\inlinelogo{diffsort}
    \\[.5em]
    This work is covered in Chapter~\ref{ch:diffsort}. \defaultauthorcontributions\\
    
    \item 
    F.~Petersen, C.~Borgelt, H.~Kuehne, and O.~Deussen, 
    \textit{Monotonic Differentiable Sorting Networks}, 
    in Proceedings of the International Conference on Learning Representations (ICLR),
    2022.~\cite{petersen2022monotonic}~\inlinelogo{diffsort}
    \\[.5em]
    This work is covered in Chapter~\ref{ch:diffsort}. \defaultauthorcontributions\\
    
    \item 
    F.~Petersen, H.~Kuehne, C.~Borgelt, and O.~Deussen, 
    \textit{Differentiable Top-k Classification Learning}, 
    in Proceedings of the International Conference on Machine Learning (ICML),
    2022.~\cite{petersen2022topk}~\inlinelogo{difftopk}
    \\[.5em]
    This work is covered in Chapter~\ref{ch:difftopk}. \defaultauthorcontributions\\
    
    \item 
    F.~Petersen, B.~Goldluecke, C.~Borgelt, and O.~Deussen, 
    \textit{GenDR: A Generalized Differentiable Renderer}, 
    in Proceedings of the IEEE Conference on Computer Vision and Pattern Recognition (CVPR),
    2022.~\cite{petersen2022gendr}~\inlinelogo{gendr}
    \\[.5em]
    This work is covered in Chapter~\ref{ch:gendr}. \defaultauthorcontributions\\
    
    \item 
    F.~Petersen, C.~Borgelt, H.~Kuehne, and O.~Deussen, 
    \textit{Deep Differentiable Logic Gate Networks}, 
    under review, 2022.~\cite{petersen2022difflogic}~\inlinelogo{difflogic}
    \\[.5em]
    This work is covered in Chapter~\ref{ch:difflogic}. \defaultauthorcontributions\\
    
    \item 
    F.~Petersen, T.~Sutter, C.~Borgelt, H.~Kuehne, and O.~Deussen, 
    \textit{Newton Losses: Efficiently Including Second-Order Information into Gradient Descent}, 
    under review, 2022.~\cite{petersen2022newton}~\inlinelogo{splitprop}
    \\[.5em]
    This work is covered in Chapter~\ref{ch:alternative}. \defaultauthorcontributions\\[.5em]
    \textit{
        We note that we have also prepared a substantially extended article based on this paper for submission to a journal.
        This extension adds RESGRO and a general formulation of splitting backpropagation, which are also included in this thesis.
    }

\end{itemize}

In addition to these publications, the author has also published other closely related works during the author's PhD research, which are not explicitly or only partially covered in this thesis.
For each publication, the contributions of each author are listed. 
Also, the relation of each work to this thesis is discussed.

\begin{itemize}
    \item 
    F.~Petersen, C.~Borgelt, M.~Yurochkin, H.~Kuehne, and O.~Deussen,
    \textit{Propagating Distributions through Neural Networks},
    under review, 2022.~\cite{petersen2022propagating}
    \\[.5em]
    \defaultauthorcontributions
    \\[.5em]
    Making algorithms differentiable requires propagating distributions through them. 
    To demonstrate the utility of propagating distributions, we propose propagating distributions also through neural networks.
    This leads to a variety of benefits, including uncertainty estimation and model robustness.
    The publication strengthens the proposed methods as it uses similar ideas for propagating distributions, and also supported the development of and gave insights about propagating distributions through algorithms.
    \\
    
    \item 
    F.~Petersen, D.~Mukherjee, C.~Borgelt, H.~Kuehne, and O.~Deussen,
    \textit{APE-VAE: Training Variational Auto-Encoders with Approximate Evidence}, 
    under review, 2022.~\cite{petersen2022ape}
    \\[.5em]
    The initial idea was conceived by the author. 
    The algorithm, details, and implementation were developed by the author. 
    The theoretical error-bound was conceived jointly by D.~Mukherjee and the author.
    All authors contributed to the text of the final publication.
    \\[.5em]
    The paper proposes Approximate Evidence (APE), an alternative to the popular VAE training objective Evidence Lower Bound (ELBO).
    The proposed APE objective improves empirically over the ELBO objective and naturally supports sampling-free distribution propagation.
    Thus, the work applies propagating distributions through neural networks, specifically, through the decoder of a VAE, which leads to superior performance compared to sampling methods.
    APE is conceptually different from the ELBO as APE delivers an approximation to the evidence, which has a bounded approximation error, and the ELBO is a lower bound of the evidence, i.e., its optimization only indirectly improves the evidence.
    \\
    
    \item 
    F.~Petersen, B.~Goldluecke, O.~Deussen, and H.~Kuehne,
    \textit{Style Agnostic 3D Reconstruction via Adversarial Style Transfer},
    in Proceedings of the IEEE Winter Conference on Applications of Computer Vision (WACV),
    2022.~\cite{petersen2022style}
    \\[.5em]
    \defaultauthorcontributions
    \\[.5em]
    As the output of differentiable relaxed algorithms usually does not distributionally match the outputs of their ``hard'' counterparts or real data, we propose to bridge this gap using domain adaptation.
    For this, we applied domain adaptation in the form of adversarial style transfer in the task of differentiable renderer supervised 3D geometry reconstruction.
    \\
    
    \item 
    F.~Petersen, A.~H.~Bermano, O.~Deussen, and D.~Cohen-Or,
    \textit{Pix2Vex: Image-to-Geometry Reconstruction using a Smooth Differentiable Renderer},
    in Arxiv,
    2019.~\cite{petersen2019pix2vex}
    \\[.5em]
    \defaultauthorcontributions
    \\[.5em]
    Pix2Vex was an early truly differentiable renderer, and is, to date, the only formally truly differentiable 3D mesh renderer.
    Later, this work inspired ``Style Agnostic 3D Reconstruction via Adversarial Style Transfer''~\cite{petersen2022style} as well as ``GenDR: A Generalized Differentiable Renderer''~\cite{petersen2022gendr}.
    \\
    
    \item 
    F.~Petersen, C.~Borgelt, and O.~Deussen,
    \textit{AlgoNet: $C^\infty$ Smooth Algorithmic Neural Networks},
    in Arxiv,
    2019.~\cite{petersen2019algonet}
    \\[.5em]
    \defaultauthorcontributions
    \\[.5em]
    AlgoNet was a largely theoretical work, describing the original ideas for 
    ``Learning with Algorithmic Supervision via Continuous Relaxations''~\cite{petersen2021learning} and ``Differentiable Sorting Networks for Scalable Sorting and Ranking Supervision''~\cite{petersen2021diffsort}.
    The paper presented smooth WHILE-Programs, a minimalistic precursor of AlgoVision~\cite{petersen2021learning}.
    The work also discussed other concepts such as differentiable iterated function systems, weighted SoftMax, and SoftMedian.
    \\
    
    \item 
    F.~Petersen, C.~Borgelt, and O.~Deussen,
    \textit{$C^\infty$ Smooth Algorithmic Neural Networks for Solving Inverse Problems},
    in the NeurIPS Deep Inverse Workshop,
    2019.~\cite{petersen2019smooth}
    \\[.5em]
    \defaultauthorcontributions
    \\[.5em]
    This workshop paper is largely based on the AlgoNet paper discussed above.
    \\
    
    \item 
    F.~Petersen, T.~Sutter, C.~Borgelt, D.~Huh, H.~Kuehne, Y.~Sun, and O.~Deussen,
    \textit{ISAAC Newton: Input-based Approximate Curvature for Newton's Method},
    under review, 2022.~\cite{petersen2022isaac}
    \\[.5em]
    \defaultauthorcontributions
    \\[.5em]
    ISAAC Newton is a method for more efficient training of neural networks by using selected second-order information.
    ISAAC Newton is a complement to Newton Losses. 
    While Newton Losses incorporates second-order information of the loss into training, ISAAC Newton integrates second-order information derived from the input to each layer into training.
    Core implications of this work, in combination with other works presented in this thesis, are: 
    Integrating second-order information using both ISAAC Newton and Newton Losses at the same time;
    and more efficient training of architectures such as differentiable logic gate networks.
    \\
    
    \item 
    F.~Petersen and T.~Sutter,
    \textit{Distributional Quantization},
    under review, 2022.~\cite{petersen2022distributional}
    \\[.5em]
    \defaultauthorcontributions
    \\[.5em]
    In this work, we present a simple and efficient scalar quantization algorithm based on the distribution of the input signal. 
    We show the quantization algorithm to be error-optimal under some mild assumptions. 
    This work was developed for differentiable logic gate networks and, more generally, for the goal of more efficient neural network training.
    \\
    
    \item 
    F.~Petersen*, D.~Mukherjee*, Y.~Sun, and M.~Yurochkin,
    \textit{Post-processing for Individual Fairness},
    in Advances in Neural Information Processing Systems (NeurIPS),
    2021.~\cite{petersen2021post}
    \\[.5em]
    The initial ideas were conceived by all four authors jointly. 
    D.~Mukherjee and F.~Petersen share joint first-authorship, distributed as follows:
    The theoretical properties were primarily devised and shown by D.~Mukherjee.
    The algorithm, experiments, and implementation were developed by the author. 
    All four authors contributed equally to the text of the final publication.
    \\[.5em]
    In the publication, we consider a setting where the learner only has access to limited information, i.e., the predictions of the original model and a similarity graph between individuals. 
    We supervise the task of individual fairness through algorithmic knowledge about the similarity graph between individuals. 
    For this, we cast the individual fairness problem as a graph smoothing problem corresponding to graph Laplacian regularization that preserves the desired ``treat similar individuals similarly'' interpretation.
    This work is an application of algorithmic supervision to fairness in a post-processing setting.
    The method is especially efficient compared to other individual fairness methods while achieving state-of-the-art on popular individual fairness benchmarks.
    \\
    
    \item 
    D.~Mukherjee*, F.~Petersen*, M.~Yurochkin, and Y.~Sun,
    \textit{Domain Adaptation meets Individual Fairness. And they get along.},
    under review,
    2022.~\cite{mukherjee2022domain}
    \\[.5em]
    The initial ideas were conceived by all four authors jointly. 
    D.~Mukherjee and F.~Petersen share joint first-authorship, distributed as follows:
    The theoretical properties were primarily devised and shown by D.~Mukherjee.
    The algorithm, experiments, and implementation were developed by the author. 
    All four authors contributed equally to the text of the final publication.
    \\[.5em]
    In the publication, we show that domain adaptation methods can enforce individual fairness and vice versa.
    \\
    
    \item 
    F.~Petersen, M.~Schubotz, and B.~Gipp,
    \textit{Towards Formula Translation using Recursive Neural Networks},
    in Proceedings of the 11th Conference on Intelligent Computer Mathematics (CICM), Work-in-Progress Paper Track,
    2018.~\cite{petersen2018towards}
    \\[.5em]
    \defaultauthorcontributions
    \\[.5em]
    In this work, we investigate recursive neural network architectures for machine translation of mathematical formulae from $\LaTeX$ to the semantically-enhanced semantic $\LaTeX$ markup language.
    \\
    
    \item 
    F.~Petersen, M.~Schubotz, A.~Greiner-Petter, and B.~Gipp,
    \textit{Neural Machine Translation for Mathematical Formulae},
    under review,
    2022.~\cite{petersen2022neural}
    \\[.5em]
    \defaultauthorcontributions
    \\[.5em]
    In this work, we apply convolutional sequence-to-sequence models and transformers to machine translation of mathematical formulae between different representations.
    Among others, we consider the translation of formulae from $\LaTeX$ into the format of the computer algebra system Mathematica. %
    \\
    
    \item 
    N.~Shvetsova, F.~Petersen, R.~Feris, and H.~Kuehne,
    \textit{Differentiable K-Nearest Neighbor Sorting for Self-supervised Learning},
    under review,
    2022.~\cite{shvetsova2022differentiable}
    \\[.5em]
    The initial idea was conceived by the authors jointly. 
    All four authors contributed to the text of the final publication.
    \\[.5em]
    In this work, we apply differentiable sorting networks and differentiable top-$k$ to self-supervised learning on images.
    Specifically, we learn by enforcing that the $k$ nearest neighbors of an image in the embedding space are the $k$ image augmentations of the same image and not the negative examples based on other images.
    \\

\end{itemize}

\begingroup %

\setlength{\textheight}{230\hscale} %

\etocstandarddisplaystyle %
\etocstandardlines %

\tableofcontents %

\endgroup

\mainmatter %
\setchapterstyle{kao} %

\setchapterpreamble[u]{\margintoc}
\chapter{Introduction}
\labch{introduction}

Four millennia ago, the Egyptians devised an algorithm for multiplying two numbers, which is the earliest record of any algorithm~\cite{neugebauer1969exact}.
In 1843, Ada Lovelace published the first computer program of an algorithm and envisioned modern applications of computers such as art and music, at a time when such a computer was not even built~\cite{menabrea1843sketch, fuegi2003lovelace}. 
A century later, in 1943, McCulloch and Pitts~\cite{mcculloch1943logical} devised the first mathematical model of neural networks based on observations of the biological processes in the brain.
Within the last decade, artificial neural network-based approaches gained a lot of attention in research. 
This resurgence can be attributed to advances in hardware~\cite{kirk2016programming}, software~\cite{paszke2019pytorch, abadi2016tensorflow, jia2014caffe, bradbury2018jax}, the development of convolutional networks~\cite{fukushima1980self, lecun1999object}, and the supremacy of deep learning on many tasks, such as image classification~\cite{krizhevsky2012imagenet, szegedy2015going}.

Today, classical algorithms and machine learning systems like neural networks are both abundant in everyday life.
While classical computer science algorithms are suitable for precise execution of exactly defined tasks such as finding the shortest path in a large graph, neural networks allow learning from data to predict the most likely answer in more complex tasks such as image classification, which cannot be reduced to an exact algorithm.
To get the best of both worlds, in this thesis, we explore combining \textit{classical computer science algorithms} and \textit{neural networks}, or, more generally, machine learning.
This leads to more robust, better performing, more interpretable, more computationally efficient, and more data efficient architectures.
Herein, model robustness can be achieved through a provably correct algorithm applied to embeddings.
The model performance can be computationally improved when we can reduce the computational complexity of the neural network by replacing part of it by a fast algorithm.
Also, with respect to accuracy, the performance can be improved as there is a smaller potential for errors, and the domain knowledge supports the network.
Correspondingly, these models can also be more interpretable, as the inputs to algorithms are typically (by definition) interpretable.
Finally, as algorithmic supervision is typically a kind of weakly-supervised learning, the level of supervision is reduced and the models are more data / label efficient.

Typically, neural networks are trained with stochastic gradient descent (SGD) or preconditioned SGD methods, such as the Adam optimizer~\cite{kingma2015adam}.
These methods are based on computing the gradient (i.e., derivative) of a loss function with respect to the model's parameters.
This gradient indicates the direction of the steepest ascent of the loss.
As minimizing the loss improves the model, we can optimize the model by going (in the model's parameter space) in the opposite direction of the gradient, i.e., gradient descent.
The derivative of the loss with respect to the model parameters can be efficiently computed using the backpropagation algorithm~\cite{rumelhart1986learning}, which, in today's deep learning frameworks~\cite{paszke2019pytorch, bradbury2018jax}, is implemented as backward-mode automatic differentiation.

Gradient-based learning requires that all involved operations are differentiable; however, many interesting operations like sorting algorithms are non-differentiable.
This is because conditional statements like \texttt{if} are piece-wise constant, i.e., they have a derivative of $0$, with the exception of the transitions (i.e., ``jumps'') between \texttt{true} and \texttt{false}, at which their derivative is undefined.
Accordingly, gradient-based learning with (non-differentiable) algorithms is generally not possible.
Therefore, in this work, we focus on making algorithms differentiable through continuous relaxations.
The primary idea of continuous relaxations is to introduce a level of uncertainty into algorithms, which, e.g., can lead to smooth transitions between \texttt{true} and \texttt{false} in \texttt{if} statements, making the algorithm fully differentiable.
We note that when going beyond backpropagation, e.g., via RESGRO losses, as introduced in Chapter~\ref{ch:alternative}, differentiability and smoothness are not strictly necessary, but still desirable. 
We also note that gradient-free optimization on algorithms in conjunction with gradient-based learning of a neural network is introduced in this work, and that differentiable algorithms typically outperform gradient-free methods.

\begin{figure*}[t]
    \centering
    \includegraphics[width=.499\linewidth, trim={1cm 0cm 1cm 1cm},clip]{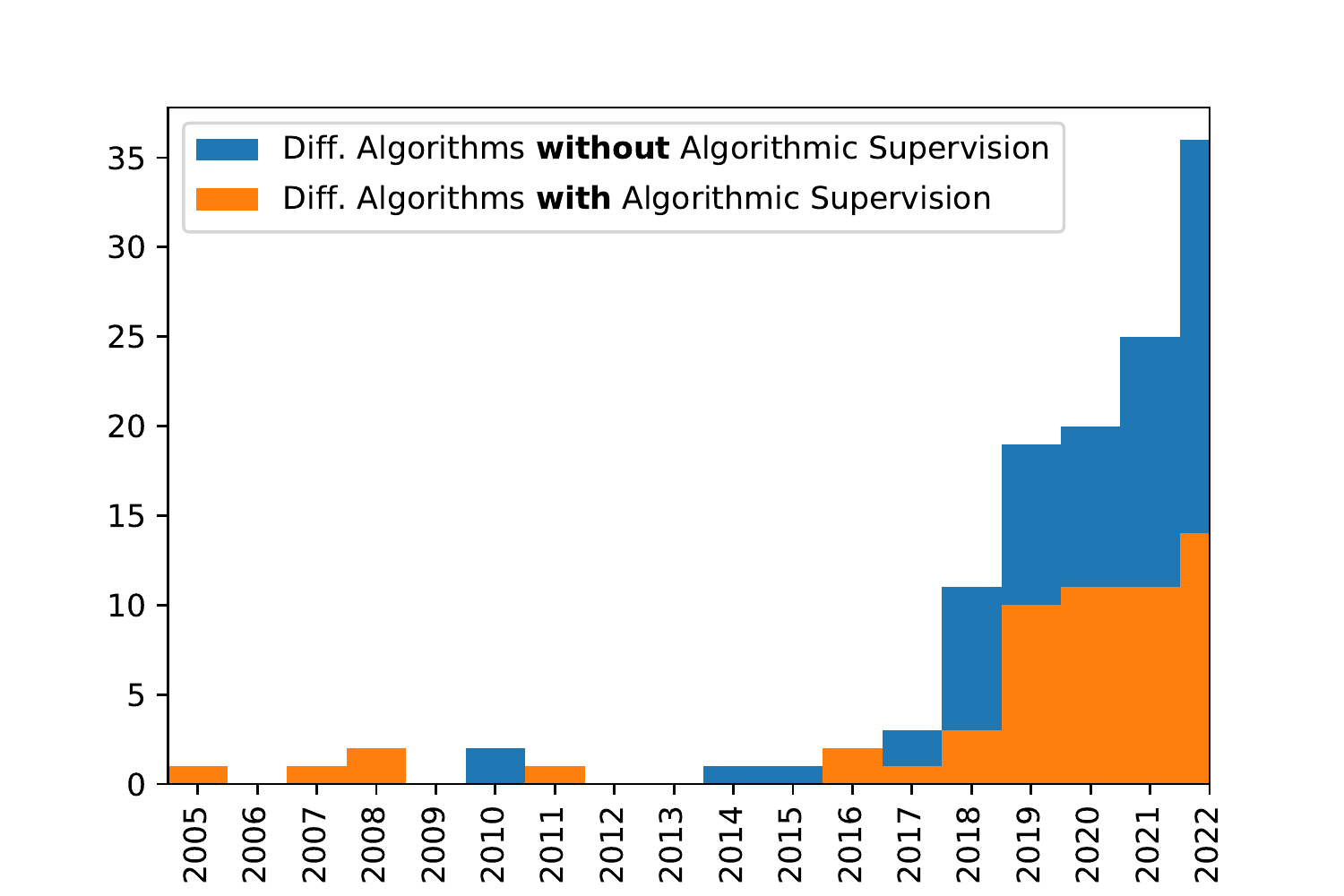}%
    \includegraphics[width=.499\linewidth, trim={1cm 0cm 1cm 1cm},clip]{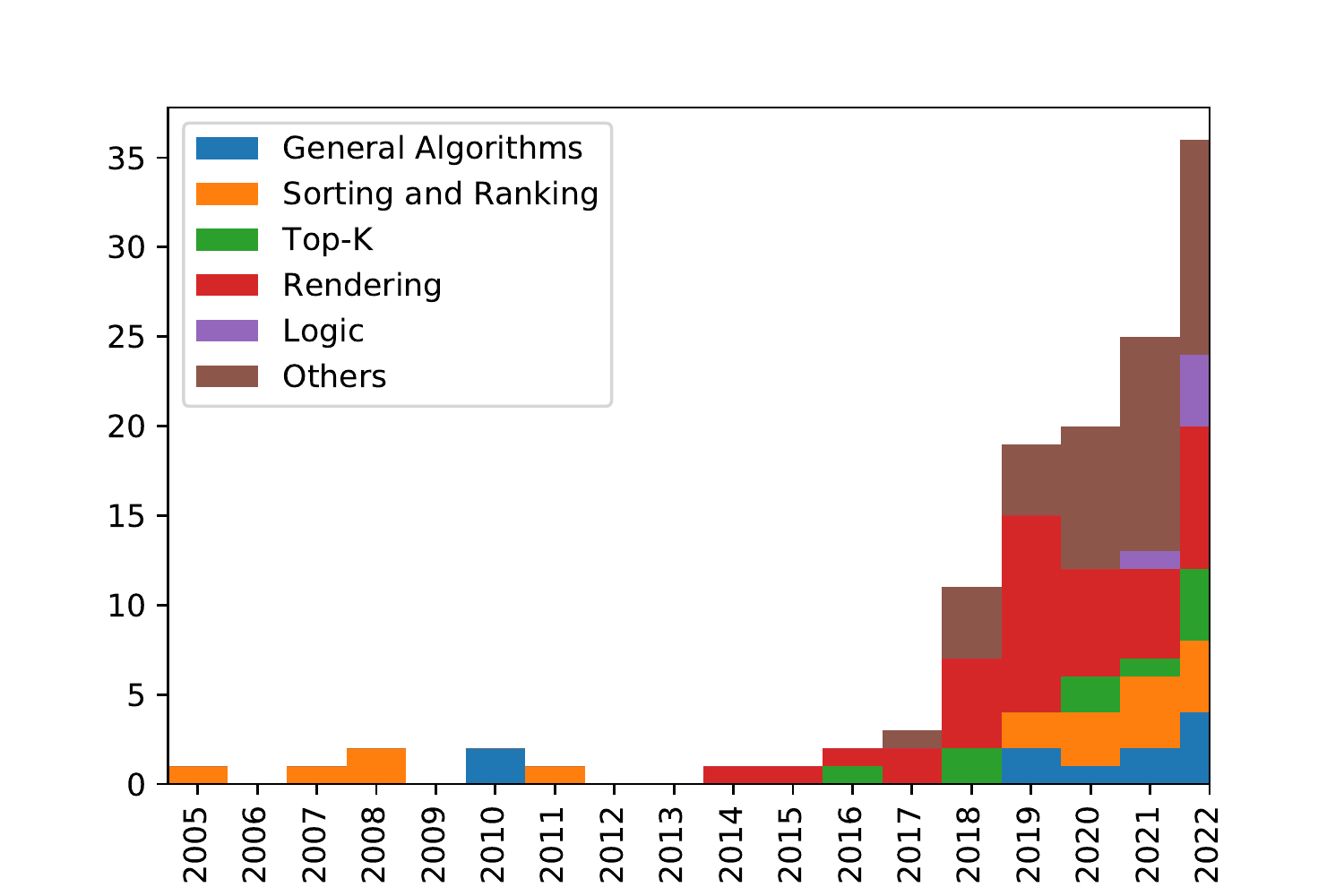}
    \caption{
        Trends in the field of learning with differentiable algorithms based on the references in this work. 
        The histograms show the number of works in each year.
        The field started its formation in 2018 / 2019.
        \textit{Left:} comparing works with and without algorithmic supervision.
        \textit{Right:} categorizing works wrt.~the kind of algorithm that was relaxed. Here, the first 5 categories correspond to Chapters~\ref{ch:algovision}--\ref{ch:difflogic} of this work.
        For 2022, as only 5 months passed at the time of writing, we doubled the height and halved the width of the bars, maintaining the area per paper.
    }
    \label{fig:survey-ref-fig}
\end{figure*}

Learning with differentiable algorithms can be grouped into 2 disciplines:
\begin{itemize}
    \item Differentiable Algorithms, i.e., the study of how to backpropagate through algorithms and obtain meaningful gradients.
    \item Algorithmic Supervision, i.e., incorporating algorithmic knowledge into the training of neural network models.
\end{itemize}

Learning with differentiable algorithms is a relatively new and recent field in machine learning. 
Specifically, apart from a few individual earlier works, the fields of differentiable algorithms and algorithmic supervision gained momentum in 2018.
We have visualized this in the survey histograms in Figure~\ref{fig:survey-ref-fig}. 
Here, we categorized all related works on differentiable algorithms into those which apply algorithmic supervision and those which do not \textit{(left)}.
Further, we categorized them wrt.~the kind of differentiable algorithm they propose or apply \textit{(right)}.
We note that the author proposed and started his work on differentiable algorithms in 2018, i.e., at the beginning of the formation of the field.
We also note that the field is growing and has only recently seen direct application to real-world problems.

The following two sections briefly cover differentiable algorithms and algorithmic supervision.

\section{Differentiable Algorithms}

In general, to allow for end-to-end training of neural architectures with integrated algorithms, the challenge is to estimate the gradients of a respective algorithm, e.g., via a differentiable approximation.

The core idea behind most differentiable algorithms is smoothing, i.e., perturbation by a probability distribution.
Differentiable algorithms can be classified into two broad methodological schools:
\begin{itemize}
    \item smoothing via stochastic perturbation by sampling;
    \item smoothing via analytical perturbation solved in closed form.
\end{itemize}
The primary focus of this work lies in analytical perturbations to be solved in closed form.
However, we also introduce and discuss stochastic sampling methods in detail.

A seminal work in the domain of sampling-based stochastic perturbations is stochastic smoothing by Abernethy~\etal~\cite{abernethy2016perturbation}.
Building on this work, Berthet~\etal~\cite{berthet2020learning} propose extensions of stochastic smoothing for algorithmic supervision.
In the field of reinforcement learning, stochastic smoothing and its derivatives are also known as the score function estimator~\cite{glynn1990likelihood, kleijnen1996optimization} or REINFORCE~\cite{williams1992simple}.

In the domain of analytical perturbations that are solved in closed form, many methods are based on analytical distribution propagation.
A popular method of propagating distributions is moment matching~\cite{frey1999variational, wang2016natural, gast2018lightweight, shekhovtsov2019feedforward}.
However, in the case of correlated random variables, moment matching becomes intractable~\cite{wu2019deterministic} and for distributions like the Cauchy distribution (which has infinite and undefined moments) moment matching is undefined.
But the Cauchy distribution is an especially important distribution, e.g., in the case of differentiable sorting where it leads to monotonic differentiable sorting networks~\cite{petersen2022monotonic}.
Therefore, propagating distributions via local linearization, as discussed by Petersen~\etal~\cite{petersen2022propagating, petersen2022ape} and which also supports distributions like Cauchy, is a suitable alternative to moment matching and has been shown to be optimal in terms of total variation for ReLU neural networks~\cite{petersen2022propagating}.
These methods for propagating distributions are especially important for modeling correlations, which is necessary when modeling input perturbation.

For smoothing algorithms, there are two conceptually different mathematical concepts:
\begin{itemize}
    \item perturbations of inputs to an algorithm;
    \item perturbations of variables / conditions in an algorithm.
\end{itemize}
Smoothing the inputs to an algorithm means relaxing an algorithm $f(x)$ to $\mathbb{E}[f(x+\epsilon)]$ where $\epsilon$ is drawn from some distribution.
Accordingly, smoothing an algorithm's inputs typically does not require assumptions about an algorithm and can be solved with stochastic smoothing: it is independent of the algorithm used, assuming that the algorithm is correct.

On the other hand, smoothing variables and conditions in an algorithm corresponds to relaxing statements like \texttt{if $x$ > 0} to \texttt{if $x+\epsilon$ > 0}.
In this case, the resulting function depends on the choice of algorithm for a given problem. 
This means that for a good choice of algorithm (e.g., Bellman-Ford), smoothing variables and conditions can lead to better relaxations than smoothing of inputs; however, for a poor or inadequate choice of algorithm (e.g., Dijkstra), an uninformed smoothing of inputs would perform better.
Further discussion of this can be found in Chapter~\ref{ch:algovision}.

While smoothing of inputs lends itself to stochastic smoothing and can be intractable for analytical propagation methods, smoothing variables can be difficult for stochastic methods and can be evaluated using analytical propagation methods.
The contributions in this thesis focus on smoothing variables in an algorithm; 
however, we also consider smoothing of inputs.

\begin{marginfigure}
    \centering
    \begin{tabular}{c|m{.3\linewidth}|m{.3\linewidth}|}
        & variables & inputs \\
    \hline
        {\rotatebox[origin=c]{90}{~~~analytical~~~}} 
        & \cite{petersen2021diffsort, petersen2021learning, petersen2022monotonic, petersen2022topk, petersen2022gendr}, \cite{petersen2019pix2vex}, \cite{liu2019soft}, \cite{chen2019learning}
        & (\cite{chaudhuri2010smooth}) 
        \\
    \hline
        {\rotatebox[origin=c]{90}{~~~stochastic~~~}} 
        & \cite{lidec2021differentiable} 
        & \cite{abernethy2016perturbation}, \cite{berthet2020learning}, \cite{nimier2019mitsuba}, \cite{cordonnier2021differentiable}
        \\
    \hline
    \end{tabular}
    \caption{
        Classification of a selection of differentiable algorithms into analytical vs.~stochastic evaluation and modeling of variables vs.~inputs.
        We note that \cite{chaudhuri2010smooth}~does not provide gradients.
    }
    \label{fig:ref-matrix}
\end{marginfigure}

\vspace{.5em}
In Figure~\ref{fig:ref-matrix}, we classify a selection of perturbation-based differentiable algorithms with respect to methodology (analytical vs.~stochastic perturbations) and assumptions (perturbation of inputs vs.~perturbation of variables).
We can observe that the majority of methods are either analytical perturbations of variables or stochastic perturbations of inputs, which is also expected according to the analysis above.

\vspace{.5em}
In the domain of differentiable rendering~\cite{petersen2019pix2vex, kato2020differentiable, zhang2021image}, the method of approximate analytical perturbations is popular~\cite{kato2020differentiable} because methods like stochastic smoothing are typically intractable due to the high dimensionality of the involved spaces (image space and the space of 3D models).
Accordingly, these methods model perturbations of variables or can be considered to model perturbations of variables (as most of them use an ad-hoc relaxation and do not identify their underlying probabilistic model).
Modeling input perturbations using analytical methods is typically intractable due to the complexity of algorithms.
But this thesis provides evidence that variable perturbations might indeed be better for high-quality gradients, given that an adequate algorithm is available for a given problem.
There are also differentiable renderers based on Monte Carlo sampling~\cite{li2018differentiable, nimier2019mitsuba, zhang2021antithetic}. 
Lidec~\etal~\cite{lidec2021differentiable} model variable perturbations using the sampling-based stochastic smoothing technique.

\vspace{.5em}
In the domain of differentiable sorting, there are methods based on entropic regularization rather than modeling perturbations.
Specifically, Cuturi~\etal~\cite{cuturi2019differentiable} reduced sorting to an optimal transport (OT) problem and induce an entropic regularization which relaxes the OT problem and thereby makes sorting smoothly differentiable.
Entropic regularization, in the context of optimal transport, means that we enforce uncertainty in the solution of the OT problem, i.e., we regularize the entropy of the solution to be sufficiently large.
In a seminal work, Cuturi~\cite{cuturi2013sinkhorn} had shown that by introducing this entropic regularization OT can be solved via the iterative Sinkhorn algorithm, which was a milestone in optimal transport as it substantially simplified many OT problems and delivers a fast approximate OT method.
As the entropic regularization enforces a smoothly differentiable distributional solution to an OT problem, it is suitable as a differentiable sorting and ranking algorithm~\cite{cuturi2019differentiable}.

\vspace{.5em}
Apart from smoothing-based differentiable algorithms, there are also heuristic relaxations of algorithms.
These heuristics are usually specific to a certain setting and typically do not come with the benefits of smoothing, such as smooth optimization objectives.
Examples of heuristic-based differentiable sorting algorithms are NeuralSort~\cite{grover2019neuralsort} and SoftSort~\cite{prillo2020softsort}.
In this work, we do not focus on heuristic approaches but include them for completeness.

\section{Algorithmic Supervision}

\begin{figure*}[t]
    \begin{tikzpicture}[scale=1., every node/.style={scale=1.}, gray!50!black, text=gray!25!black]
    \draw[semithick] (0,0) rectangle (1.25,.75) node[pos=.5] {Input};
    \draw[thick] (2.0,0) rectangle (3.25,0.75) node[pos=.5] {Model};
    \draw[thick, dash pattern=on .115625cm off .115625cm, dash phase=.0578125cm] (4.,0) rectangle (5.25,.75) node[pos=.5] {Output};
    \draw[->, thick] (1.35, 0.375) -- (1.9, 0.375);
    \draw[->, thick] (3.35, 0.375) -- (3.9, 0.375);
    \node[gray!50!black] at (4.625, -0.25) {\scriptsize (supervised)};
    \end{tikzpicture}%
    \hfill%
    \begin{tikzpicture}[scale=1., every node/.style={scale=1.}, gray!50!black, text=gray!25!black]
    \draw[semithick] (0,0) rectangle (1.25,.75) node[pos=.5] {Input};
    \draw[thick] (2.0,0) rectangle (3.25,0.75) node[pos=.5] {Model};
    \draw[thick] (4.+1.,0) rectangle (5.75+1.,.75) node[pos=.5] {Algorithm};
    \draw[thick, dashed, dash pattern=on .115625cm off .115625cm, dash phase=.0578125cm] (6.5+1.,0) rectangle (7.75+1.,.75) node[pos=.5] {Output}; %
    \draw[->, thick] (1.35, 0.375) -- (1.9, 0.375);
    \draw[->, thick] (3.35, 0.375) -- node[above] {\scriptsize meaningful} node[below] {\scriptsize embedding} (3.9+1., 0.375);
    \draw[->, thick] (5.85+1., 0.375) -- (6.5+1.-0.1, 0.375);
    \node[gray!50!black] at (7.125+1., -0.25) {\scriptsize (supervised)};
    \end{tikzpicture}%
    \hfill~
    \caption{Direct supervision (on the left) in comparison to algorithmic supervision (on the right).}
    \label{fig:algorithmic-supervision}
\end{figure*}
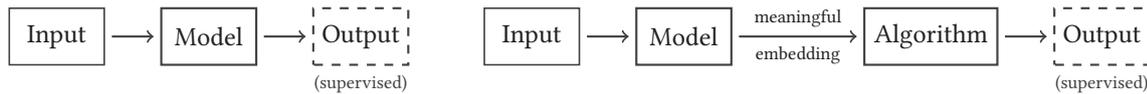

Artificial neural networks have shown their ability to solve various problems, ranging from classical tasks in computer science, such as machine translation \cite{vaswani2017attention} and object detection \cite{redmon2016yolo}, to many other topics in science, such as protein folding \cite{senior2020improved}.
Simultaneously, classical algorithms exist, which typically solve predefined tasks based on a predefined control structure, such as sorting or shortest-path computation, and for which guarantees about their behavior can be deduced.
Recently, research has started to combine both elements by integrating algorithmic concepts into neural network architectures.  
Those approaches allow training neural networks with alternative supervision strategies, such as learning 3D representations via a differentiable renderer \cite{liu2019soft, kato2017neural} or training neural networks with ordering information \cite{grover2019neuralsort, cuturi2019differentiable}. 
We unify these alternative supervision strategies, which integrate algorithms into the training objective, as algorithmic supervision:

Algorithmic supervision is the idea of integrating an algorithm into the training objective of a neural network model.
This lets us directly incorporate algorithmic domain knowledge of the problem, leading, i.a., to weaker supervision requirements and~/~or improved performance.
In this work, we formally define algorithmic supervision as follows:

\begin{definition}[Algorithmic Supervision]
In algorithmic supervision, an algorithm is applied to the predictions of a model and the outputs of the algorithm are supervised. 
In contrast, in direct supervision, the predictions of a model are directly supervised. 
This is illustrated in Figure~\ref{fig:algorithmic-supervision}.
\end{definition}

Examples of algorithmic supervision that we cover in this thesis are:
\begin{itemize}
    \item Sorting and Ranking Supervision
    \item Rendering and Silhouette Supervision
    \item Shortest-Path Supervision
    \item Editing Distance Supervision
    \item Top-$k$ Supervision
\end{itemize}

\textbf{Sorting and Ranking Supervision}\hspace{1em}%
Sorting and Ranking Supervision allow training based on ordering information.
Specifically, it is the setting in which the relative order or ranking of a set of elements is given, while their absolute values remain unsupervised.
This is achieved by using a differentiable ranking algorithm to rank model predictions for a set of elements and training by backpropagating the discrepancy between the resulting ranking and the ground truth ranking back to the neural network.
For differentiable sorting, an array of methods has been proposed in recent years, which includes
NeuralSort~\cite{grover2019neuralsort}, SoftSort~\cite{prillo2020softsort}, Optimal Transport Sort~\cite{cuturi2019differentiable}, and FastSort~\cite{blondel2020fast}, all of which appeared after we started working on differentiable sorting.
This setting is covered in Chapter~\ref{ch:algovision}, Chapter~\ref{ch:diffsort}, and Chapter~\ref{ch:alternative}.

\paragraph{Rendering and Silhouette Supervision}
Rendering and Silhouette Supervision allow learning 3D shapes from 2D information.
The idea is to differentiably render a 3D model and then compare the rendering to a reference image or the rendered silhouette to a reference silhouette.
By reducing the discrepancy between the rendering and a reference image, we can improve the 3D model, or the neural network that predicted the 3D model.
In the domain of differentiable rendering, a body of recent research has considered rendering and silhouette supervision, including but not limited to: \cite{kato2020differentiable, kato2017neural, petersen2019pix2vex, liu2019soft, chen2019learning, petersen2021learning, petersen2022gendr, petersen2022style, nimier2019mitsuba, loper2014opendr}.
This setting is covered in Chapter~\ref{ch:algovision} and Chapter~\ref{ch:gendr}.

\paragraph{Shortest-Path Supervision}
Shortest-Path Supervision allows learning based on shortest-path information.
By applying a differentiable shortest-path algorithm to a cost embedding, we obtain a relaxed shortest path.
Given a reference shortest path, we can compute the discrepancy between the relaxed and the reference shortest paths as our loss.
By minimizing the loss, we improve the cost embedding or the neural network that predicted it.
Recently, a few related works have considered the setting of shortest-path supervision~\cite{vlastelica2019differentiation, berthet2020learning, petersen2021learning}.
This setting is covered in Chapter~\ref{ch:algovision} and Chapter~\ref{ch:alternative}.

\paragraph{Top-$k$ Supervision} 
The idea of top-$k$ supervision is that we want to enforce that a target class is among the top-$k$ classes, e.g., that it is among the top-$5$ classes.
An extension of this core idea is to relax the assumption that $k$ is a constant to $k$ being drawn from a distribution. For example, by setting $k$ with a $50\%$ probability to $1$ and with a $50\%$ probability to $5$, we can optimize both the top-$1$ and top-$5$ objective at the same time and make the model more robust, which even allows improving top-$1$ accuracy.
While the idea of top-$k$ supervision has been established for a long time~\cite{lapin2015topk, goodfellow2016deep, berrada2018smooth}, relaxing $k$ is novel and is covered in Chapter~\ref{ch:difftopk}.

For an illustration of the history of algorithmic supervision as a part of the field of differentiable algorithms, see Figure~\ref{fig:survey-ref-fig}~(left). 

\section{Applications and Additional Related Work}
\label{sec:intro:applications}

There is a large variety of applications and use cases for integrating algorithmic components into machine learning.
In addition to the methods and applications covered above, in the following, we cover additional applications and related works which use differentiable algorithms. 
Some of these build on the work of this thesis, and many others could also be based on the methods proposed in this work (e.g., methods building on top of an ``arbitrary'' differentiable sorting algorithm).
For an illustration of trends in differentiable algorithms, see Figure~\ref{fig:survey-ref-fig}~(right). 

While some of these works cover single algorithmic concepts, e.g., sorting~\cite{grover2019neuralsort, cuturi2019differentiable, blondel2020fast} or rendering~\cite{kato2017neural, liu2019soft, chen2019learning}, others have covered wider areas such as dynamic programming~\cite{mensch2018differentiable, corro2019differentiable} or gradient estimation for general optimizers~\cite{vlastelica2019differentiation, berthet2020learning, rolinek2020optimizing, shirobokov2020black}.
Paulus~\etal~\cite{paulus2020gradient} employ the Gumbel-Softmax~\cite{jang2017categorical} distribution to estimate gradients for concepts such as subset selection.
Cuturi~\etal~\cite{cuturi2017soft} and Blondel~\etal~\cite{blondel2021differentiable} present approaches for differentiable dynamic time warping.
Dasgupta~\etal~\cite{dasgupta2020improving} model box parameters with Gumbel distributions to obtain soft differentiable box embeddings.

In the domain of recommender systems~\cite{chapelle2011yahoo, qin2013introducing}, early learning-to-rank works already appeared in the 2000s~\cite{burges2005ranknet, burges2007lambdarank, taylor2008softrank, hullermeier2008label}, but more recently Lee~\etal~\cite{lee2021differentiable} propose differentiable ranking metrics, Swezey~\etal~\cite{swezey2021pirank} and Pobrotyn~\etal~\cite{pobrotyn2021neuralndcg} propose PiRank and NeuralNDCG, respectively, which both rely on differentiable sorting, and Rolinek~\etal~\cite{rolinek2020optimizing} propose optimizing ranking metrics with their blackbox differentiation method~\cite{vlastelica2019differentiation}.
Adams~\etal~\cite{adams2011ranking} proposed a differentiable ranking method based on the Sinkhorn and Knopp algorithm~\cite{sinkhorn1967concerning, knight2008sinkhorn}.
Kong~\etal~\cite{kong2020rankmax} propose Rankmax, an alternative to Softmax based on an adaptive projection, for learning-to-rank. %

Plötz~\etal~\cite{plotz2018neural} propose a continuous relaxation of the $k$-nearest-neighbor algorithm and propose integrating it into neural architectures.
Grover~\etal~\cite{grover2019neuralsort} apply differentiable sorting to $k$-nearest-neighbor learning in addition to ranking supervision.
Based on ideas from differentiable sorting, Xie~\etal~\cite{xie2020differentiable} propose a differentiable top-$k$ operator based on optimal transport and the Sinkhorn algorithm~\cite{cuturi2013sinkhorn}.
They apply their method to $k$-nearest-neighbor learning ($k$NN), differential beam search with sorted soft top-$k$, and top-$k$ attention for machine translation.
Cordonnier~\etal~\cite{cordonnier2021differentiable} use perturbed optimizers~\cite{berthet2020learning} to derive a differentiable top-$k$ operator, which they use for differentiable image patch selection.
Goyal~\etal~\cite{goyal2018continuous} propose a continuous top-$k$ operator for differentiable beam search.
Pietruszka~\etal~\cite{pietruszka2020successive} propose the differentiable successive halving top-$k$ operator to approximate the normalized Chamfer Cosine Similarity ($nCCS@k$).
Patel~\etal~\cite{patel2022recall} propose a differentiable surrogate for recall@k for image retrieval.

In machine translation, Bahdanau~\etal~\cite{bahdanau2015neural} investigate soft alignments, while Collobert~\etal~\cite{collobert2019fully} present a fully differentiable beam searching algorithm.
In speech recognition, Bahdanau~\etal{}~\cite{bahdanau2016task} propose differentiable surrogate task losses for sequence prediction tasks.
Carr~\etal~\cite{carr2021self} propose self-supervised learning of audio representation via differentiable sorting-based ranking supervision of shuffled patches of spectrograms of the audio. 
Huang~\etal~\cite{huang2022relational} propose learning relational surrogate losses supervised through differentiable sorting networks~\cite{petersen2021diffsort}.

Algorithmic supervision also has similarities to neuro-symbolic systems~\cite{yi2018neural, yi2020clevrer} as they both incorporate algorithmic domain knowledge of the problem.
Yang~\etal~\cite{yang2022safe} combine neuro-symbolic learning with differentiable symbolic execution for safety losses in the context of safe learning.

Over the last years, many differentiable renders for different types of 3D representations have been proposed.
These include renderers for 3D meshes~\cite{loper2014opendr,kato2017neural,liu2019soft,chen2019learning}, voxels~\cite{yan2016perspective}, point clouds~\cite{insafutdinov2018unsupervised}, surfels~\cite{yifan2019differentiable}, signed distance functions~\cite{jiang2020sdfdiff, vicini2022differentiable}, and other implicit representations~\cite{liu2019learning, sitzmann2019scene, niemeyer2020differentiable, yariv2020multiview}.
A notable mention is the work by Sitzmann~\etal~\cite{sitzmann2019scene}, who propose a differentiable ray marching algorithm for learning scene representation networks. 
Deng~\etal~\cite{deng2020cvxnet} propose a method for learning convex decompositions of shapes via a differentiable convex set indicator.
Peng~\etal~\cite{peng2021shape} propose a differentiable Poisson solver, which they apply to various 3D tasks.
Rakotosaona~\etal~\cite{rakotosaona2021differentiable} propose a differentiable surface triangulation algorithm for triangle mesh optimization.

Cho~\etal~\cite{cho2021differentiable} proposed differentiable spline approximations (including differentiable non-uniform rational basis splines (NURBS)) for tasks such as image segmentation, 3D point cloud reconstruction, and finite element analysis.

In another line of work, drawing strokes and sketching have been made differentiable~\cite{mihai2021differentiable} for learning to draw~\cite{mihai2021learning} and, more generally, also rendering of vector graphics has been made differentiable~\cite{li2020differentiable}.
In a different but related line of work, Petersen~\etal~\cite{petersen2019algonet} and Scott~\cite{scott2022differentiable} present differentiable iterated function systems.

There is a line of work on learning with differentiable physics simulations~(\cite{de2018end, hu2019difftaichi, degrave2019differentiable} among many others). 
Holl~\etal~\cite{holl2020phiflow} proposed PhiFlow, a differentiable partial differential equation (PDE) solver for physics simulations like fluid dynamics, which can be used for learning~\cite{holl2020learning}.
Sitzmann~\etal~\cite{sitzmann2018end} propose a differentiable simulation of optics and apply it to achromatic extended depth of field as well as super-resolution.
Ingraham~\etal~\cite{ingraham2018learning} propose learning protein structure with a differentiable simulator.
Fu~\etal~\cite{fu2022differentiable} propose differentiable scaffolding trees for molecular optimization.
Another line of works considered learning with combinatorial solvers for NP-hard problems~\cite{vlastelica2019differentiation, paulus2021comboptnet}.

Bangaru~\etal~\cite{bangaru2021systematically} propose a method for making programs that solve integrals (or may be expressed as solving an integral) differentiable: instead of discretizing the algorithm from an integral to a sum and then differentiating it, they propose to automatically differentiate the symbolic form of the integral and then discretize it.

Domke~\cite{domke2010implicit} proposes a finite difference implicit differentiation method.
Vlastelica~\etal~\cite{vlastelica2019differentiation} build on this one-step linearization method and apply it to a range of algorithmic supervision settings.
Niepert~\etal~\cite{niepert2021implicit} propose a method for backpropagating through discrete structures via sampling from exponential family distributions and implicit differentiation. 
Blondel~\etal~\cite{blondel2021efficient} present an overview of implicit differentiation methods and present an efficient, modular, and general framework for implicit differentiation.

Amos~\etal~\cite{amos2017optnet} and Agrawal~\etal~\cite{agrawal2019differentiable} consider including optimization problems as layers into neural networks; by making the solution of the optimization problem differentiable, these layers can be trained.

Djolonga~\etal~\cite{djolonga2017differentiable} integrate submodular minimization into neural architectures, and specifically, by using a graph cut algorithm, they achieve improved image segmentation performance.
Rolinek~\etal~\cite{rolinek2020deep} apply their blackbox differentiation method~\cite{vlastelica2019differentiation} to graph matching for keypoint correspondence tasks.
Charpentier~\etal~\cite{charpentier2022differentiable} propose a method for differentiable sampling of directed acyclic graphs via differentiable sorting and ranking methods.
Zheng~\etal~\cite{zheng2018dags} and Lorch~\etal~\cite{lorch2021dibs} each propose a relaxation of the constraint of a directed graph being acyclic to a differentiable loss, which they use for structure learning.

Riad~\etal~\cite{riad2022learning} build on spectral pooling~\cite{rippel2015spectral} and make the stride of a convolution differentiable by expressing it as a masking on the Fourier domain, which allows learning the optimal stride during training.
Simultaneously, Romero~\etal~\cite{romero2021flexconv}, propose convolution with differentiable kernel sizes by parameterizing the kernel as an implicit neural representation and masking the kernel using a Gaussian distribution to enforce local support.
Thandiackal~\etal~\cite{thandiackal2022differentiable} propose a differentiable zooming method for multiple instance learning.

Chen~\etal~\cite{chen2022semi} propose a differentiable Voronoi tessellation for semi-discrete normalizing flows.
Corenflos~\etal~\cite{corenflos2021differentiable} propose a differentiable particle filtering algorithm via  optimal transport with entropic regularization.
Engel~\etal~\cite{engel2020ddsp} propose DDSP, a differentiable digital signal processing framework.

\section{Outline}

This thesis is organized into 8 chapters: 
The chapters build chronologically upon each other, specifically, Chapter~\ref{ch:algovision} introduces core ideas and methods on which each of the Chapters~\ref{ch:diffsort}--\ref{ch:difflogic} build. 
Chapter~\ref{ch:alternative} introduces alternative optimization methods and is therefore largely independent of the chronological flow of Chapters~\ref{ch:algovision}--\ref{ch:difflogic}, but knowledge from Chapters~\ref{ch:algovision} and~\ref{ch:diffsort} is recommended for an in-depth understanding of the applications.

\textbf{Chapter~\ref{ch:introduction}} introduces the core ideas of learning with differentiable algorithms and covers important related work and applications.

\textbf{Chapter~\ref{ch:algovision}} details general approaches for differentiable algorithms and algorithmic supervision. %
For that, the chapter gives a general overview of differentiable algorithms and can be seen as an extension to the introduction.
The following chapters build on the ideas presented in this chapter, and each of them considers a specific class of differentiable algorithms in greater depth.
This chapter 
is accompanied by the \inlinelogo{algovision} library.

\textbf{Chapter~\ref{ch:diffsort}} investigates differentiable sorting and ranking approaches, with a focus on differentiable sorting networks.
We start by presenting differentiable sorting approaches and, through careful theoretical analysis, we derive improved differentiable sorting operators.
This chapter 
is accompanied by the \inlinelogo{diffsort} library.

\textbf{Chapter~\ref{ch:difftopk}} presents differentiable top-$k$ approaches, conceptually building on differentiable sorting and ranking methods.
Specifically, we introduce differentiable top-$k$ networks, an improvement over differentiable sorting networks for the top-$k$ operator.
Building on differentiable top-$k$, we present top-$k$ classification learning and achieve competitive performance on the ImageNet classification task.
This chapter 
is accompanied by the \inlinelogo{difftopk} library.

\textbf{Chapter~\ref{ch:gendr}} covers differentiable rendering.
We cover various approaches in differentiable rendering and present GenDR, the generalized differentiable renderer, which (at least approximately) includes most of the existing differentiable renderers and also generalizes to new differentiable renderers.
This chapter 
is accompanied by the \inlinelogo{gendr} library.

\textbf{Chapter~\ref{ch:difflogic}} proposes differentiable logic gate networks, which are a relaxation of logic gate networks and can, therefore, be trained.
This allows extremely fast inference speeds as the resulting logic gate networks can be executed natively on common hardware as such hardware primarily operates on logic gates in the first place.
This is an example of a differentiable algorithm that can be trained and is not necessarily connected to algorithmic supervision.
This chapter 
is accompanied by the \inlinelogo{difflogic} library.

\textbf{Chapter~\ref{ch:alternative}} considers alternative optimization strategies.
Specifically, it discusses splitting backpropagation, a general two-stage optimization algorithm based on regularization, which allows optimizing algorithmic losses with a different optimizer than used for optimizing the neural network.
Split backpropagation also allows splitting a neural network itself into multiple sections and can be extended to multiple splits, which can then all be trained end-to-end, even when subsections are trained with alternative optimizers or are even non-differentiable.
This chapter 
is accompanied by the \inlinelogo{splitprop} library.

We summarize the major contributions of this thesis and discuss directions for future research in \textbf{Chapter~\ref{ch:perspectives}}.

\setchapterpreamble[u]{\pagelogo{algovision}\margintoc}
\chapter{Differentiable Algorithms}
\labch{algovision}

This chapter gives a general overview of differentiable algorithms and is an extension to the introduction by covering general methods for differentiable algorithms.
In the following chapters, we build on these ideas and consider specific differentiable algorithms in greater depth.

In general, to allow for end-to-end training of neural architectures with integrated algorithms, the challenge is to estimate the gradients of a respective algorithm, e.g., by a differentiable approximation.
For this, many existing solutions are tailored to specific problems like, e.g., differentiable sorting or rendering. 
In contrast to these approaches, in this chapter, we cover general methods for differentiable algorithms, which are not limited to specific problems.

In this chapter, we propose a general approach for making algorithms differentiable and estimating their gradients.
Specifically, we propose continuous relaxations of different algorithmic concepts such as comparators, conditional statements, bounded and unbounded loops, and indexing.
For this, we perturb those variables in a discrete algorithm, for which we want to compute a gradient, by logistic distributions. 
This allows estimating the expected value of an algorithm's output sampling-free and in closed form, e.g., compared to methods approximating the distributions via Monte-Carlo sampling (e.g., \cite{berthet2020learning}).
To keep the computation feasible, we approximate the expectation value by merging computation paths after each conditional block in sequences of conditional blocks.
For nested conditional blocks, we compute the exact expectation value without merging conditional cases.
This trade-off allows merging paths in regular intervals and thus alleviates the need to keep track of all possible path combinations.
As we model perturbations of variables when they are accessed, all distributions are independent, which contrasts the case of modeling input perturbations, where all computation paths would have to be handled separately.

To demonstrate the practical aspects, we apply the proposed approach in the context of four tasks that make use of algorithmic supervision to train a neural network, namely sorting supervision \cite{grover2019neuralsort, cuturi2019differentiable, mena2018learning}, shortest-path supervision \cite{vlastelica2019differentiation, berthet2020learning}, silhouette supervision (differentiable rendering) \cite{yan2016perspective, liu2019soft, kato2017neural}, and, finally, Levenshtein distance supervision. 
We show that the proposed method outperforms state-of-the-art methods on sorting supervision and shortest-path supervision, and performs comparably on silhouette supervision.
Later in this work, we will cover sorting supervision and silhouette supervision in greater detail in Chapters~\ref{ch:diffsort} and~\ref{ch:gendr}, respectively.

\section{Related Work} 

\marginnote{
    ``Smooth WHILE-Programs'', which is a minimalistic (yet Turing-complete) precursor of the presented method, was previously proposed by Petersen~\etal~\cite{petersen2019algonet}.
}
In the following, we cover related works in the domain of general differentiable algorithms. 
While Sections~\ref{sec:algovision:related:stoch} and~\ref{sec:algovision:related:blackbox} cover methods that are strongly related and applicable to algorithmic supervision, Sections~\ref{sec:algovision:related:smooth-interpretation} and~\ref{sec:algovision:related:program-synthesis} cover methods that are not directly applicable to algorithmic supervision and which were developed in a different context.
For additional applications of differentiable algorithms, see Section~\ref{sec:intro:applications}, and for specific differentiable algorithms, see the respective subsections in each experimental section of this chapter.

\subsection{Stochastic Gradient Estimation}
\label{sec:algovision:related:stoch}
Stochastic gradient estimation is a popular method for differentiating algorithms. 
These methods focus on modeling perturbations of the inputs to an algorithm via stochastic sampling.
Relevant works in this direction include stochastic smoothing~\cite{abernethy2016perturbation}, perturbed optimizers~\cite{berthet2020learning}, and stochastic softmax tricks~\cite{paulus2020gradient}.

\paragraph{Stochastic Smoothing}
In this paragraph, we cover the method of stochastic smoothing~\cite{abernethy2016perturbation}.
This method regularizes a non-differentiable and discontinuous loss function $\ell(y)$ by randomly perturbing its input with random noise $\epsilon$ (i.e., $\ell(y+\epsilon)$).
The loss function is then approximated as $\ell(y)\approx \ell_\epsilon(y) = \mathbb{E}[\ell(y+\epsilon)]$. 
While $\ell$ is not differentiable, its smoothed stochastic counterpart $\ell_\epsilon$ is differentiable and the corresponding gradients can be estimated via the following result.
\begin{lemma}[{Exponential Family Smoothing, adapted from ~\cite[Lemma 1.5]{abernethy2016perturbation}}]
Given a distribution over $\mathbb{R}^m$ with a probability density function $\mu$ of the form $\mu(\epsilon)=\exp(-\nu(\epsilon))$ for any twice-differentiable~$\nu$, then
\begin{align}
    \nabla_y l_\epsilon(y) = \nabla_y \mathbb{E}\left[\ell(y+\epsilon) \right] 
    &=\, \mathbb{E}\big[\ell(y+\epsilon)\, \nabla_\epsilon\nu(\epsilon) \big] , \label{eq:stoch-smooth-grad} \\
    \nabla_y^2 l_\epsilon(y) = \nabla_y^2 \mathbb{E}\left[\ell(y+\epsilon) \right] 
    &= \mathbb{E}\left[\ell(y+\epsilon)\, \Big(\nabla_\epsilon\nu(\epsilon) \nabla_\epsilon\nu(\epsilon)^\top - \nabla_\epsilon^2\nu(\epsilon)\Big) \right] . \label{eq:stoch-smooth-hess} \notag
\end{align}
\end{lemma}
A \textit{variance-reduced form} of these gradient estimators for symmetric distributions (via the control variates method) is
\begin{align}
    \nabla_y \mathbb{E}\left[\ell(y+\epsilon) \right] 
    &=\, \mathbb{E}\big[(\ell(y+\epsilon) - \ell(y))\, \nabla_\epsilon\nu(\epsilon) \big] , \\
    \nabla_y^2 \mathbb{E}\left[\ell(y+\epsilon) \right] 
    &= \mathbb{E}\left[(\ell(y+\epsilon) - \ell(y))\, \Big(\nabla_\epsilon\nu(\epsilon) \nabla_\epsilon\nu(\epsilon)^\top - \nabla_\epsilon^2\nu(\epsilon)\Big) \right] . \notag
\end{align}
In the field of reinforcement learning, stochastic smoothing and its derivatives are also known as the score function estimator~\cite{glynn1990likelihood, kleijnen1996optimization} or REINFORCE~\cite{williams1992simple}. 
In addition to the above method of control variates, there are also other variance-reduction methods: 
RELAX~\cite{grathwohl2018backpropagation}, which trains a model that predicts the control variate, and REBAR~\cite{tucker2017rebar}, which constructs the control variate based on the difference between the REINFORCE gradient estimator for the relaxed model and the gradient estimator from the reparameterization trick.

\paragraph{Perturbed Optimizers with Fenchel-Young Losses}
Berthet~\textit{et al.}~\cite{berthet2020learning} build on stochastic smoothing and Fenchel-Young losses~\cite{blondel2020learning} to propose perturbed optimizers with Fenchel-Young losses.
For this, they use algorithms, like Dijkstra's shortest-path algorithm, to solve optimization problems of the type $\max_{w\in\mathcal{C}}\langle y, w \rangle$, where $\mathcal{C}$ denotes the feasible set, e.g., the set of valid paths.   
Berthet~\textit{et al.}~\cite{berthet2020learning} identify the argmax to be the differential of max, which allows a simplification of stochastic smoothing.
By identifying similarities to Fenchel-Young losses, they find that the gradient of their loss is 
\begin{equation}
    \nabla_y \ell(y) = \mathbb{E}_\epsilon\left[\arg\max_{w\in\mathcal{C}}\langle y + \epsilon, w \rangle\right] - w^\star
\end{equation}
where $w^\star$ is the ground truth solution of the optimization problem (e.g., the shortest path). 
This formulation allows optimizing the model without the need for computing the actual value of the loss function.

\subsection{One-Step Linearization of Combinatorial Solvers}
\label{sec:algovision:related:blackbox}
The blackbox differentiation method by Vlastelica~\textit{et al.}~\cite{vlastelica2019differentiation} uses a one-step finite differences linearization in the direction of the backpropagated gradient to approximate the first derivative.
The method works for estimating the gradients of combinatorial solvers of linear problems.
If $g: \mathbb{R}^m\to \{0,1\}^m  /\mathbb{R}^m$ is a linear combinatorial solver, $z=g(y)$, and $\nabla_z \ell(y) = \frac{\partial\,\ell(y)}{\partial\,g(y)}$ is given, the derivative wrt.~the input of the solver may be approximated as 
\begin{equation}
    \nabla_y \ell = \frac{\partial\,\ell(y)}{\partial\,y} \approx \frac{1}{\lambda} \left( g\left(y + \lambda \frac{\partial\,\ell(y)}{\partial\,g(y)}\right) - g\big(y\big) \right)
\end{equation}
for some $\lambda>0$.

\subsection{Smooth Interpretation}
\label{sec:algovision:related:smooth-interpretation}
In another line of work, in the field of computer-aided verification,
Chaudhuri~\etal{}~\cite{chaudhuri2010smooth, chaudhuri2011smoothing}
propose a program smoothing method based on randomly perturbing the inputs of a program by a Gaussian distribution. 
Here, an initial Gaussian perturbation is propagated through program transformations, and a final distribution over perturbed outputs is approximated via a mixture of Gaussians.
The smooth function is then optimized using the gradient-free Nelder-Mead optimization method.
The main differences to our method are that we perturb all relevant variables (and not the inputs) with logistic distributions and use this for gradient-based optimization.

\subsection{Neural Programs and Differentiable Program Interpreters}
\label{sec:algovision:related:program-synthesis}
Another line of work deals with the problem of inductive program synthesis by relaxing the discrete space of programs into a continuous one \cite{bosnjak2017programming, gaunt2017differentiable, feser2017neural, shah2020learning}.
Here, neural networks are viewed as continuous relaxations over the space of programs, which can be used to complete partial programs.
Their relaxation of the space of programs allows optimization of the source code or a neural representation of the source code. 
That is, their goal is to make the program interpreter differentiable and not the program itself.
Compared to that, for algorithmic supervision, an algorithm is already given and does not need to be learned. 
Algorithmic supervision supervises an upstream neural network by an algorithm and thus requires the algorithm to be relaxed with respect to its inputs.
In summary, our work focuses on making an algorithm differentiable with respect to its input, while neural programs and differentiable program interpreters make a program differentiable with respect to the formulation of the program itself.

% \section{A General Method for Relaxing Algorithms}
\section{A\texorpdfstring{\kern-.05em~}{ }General\texorpdfstring{\kern-.05em~}{ }Method\texorpdfstring{\kern-.05em~}{ }for\texorpdfstring{\kern-.05em~}{ }Relaxing\texorpdfstring{\kern-.05em~}{ }Algorithms}
\label{sec:algovision:method}

To continuously relax algorithms and, thus, make them differentiable, we relax all values with respect to which we want to differentiate into logistic distributions.
We choose the logistic distribution 
as it provides two distinctive properties that make it especially suited for the task at hand: 
(1) logistic distributions have heavier tails than normal distributions, which allows for larger probability mass and thus larger gradients when two compared values are further away from each other.
(2) the cumulative density function (CDF) of the logistic distribution is the logistic sigmoid function, which can be computed analytically, and its gradient is easily computable. 
This contrasts the CDF of the normal distribution, which has no closed form and is commonly approximated via a polynomial \cite{cody1969rational}.
However, this should not be seen as a restriction of the proposed method as it is also directly applicable to other distributions, as we discuss in Section~\ref{sec:algovision:alt-dist}.

Specifically, we relax any value $x$, for which we want to compute gradients, by perturbing it into a logistic random variable $\tilde x \sim \operatorname{Logistic}(x, 1/\steepness)$, where $\steepness$ is the inverse temperature parameter such that for $\steepness\to\infty : \tilde x = x$. 
Based on this, we can relax a discrete conditional statement, e.g., based on the condition $x<c$ with constant $c\in\sR$, as follows:
\begin{align}
    &\big[ y \texttt{ if } \tilde x < c \texttt{ else } z \big] \\
    &\equiv \int_{-\infty}^{c} f_{\operatorname{log}}(t; x, 1/\steepness) \cdot y\ \mathrm{d}t      +      \int_{c}^{\infty} f_{\operatorname{log}}(t; x, 1/\steepness) \cdot z\ \mathrm{d}t  \\
    & = \qquad \ F_{\operatorname{log}}(c; x, 1/\steepness) \cdot y\quad +\quad (1-F_{\operatorname{log}}(c; x, 1/\steepness))\cdot z \\
    & = \qquad \ \ \, \sigma(c−x) \cdot y \qquad + \quad (1−\sigma(c−x)) \quad\ \cdot z
\end{align}
where $\sigma$ is the logistic (sigmoid) function  $\sigma(x) = 1/(1+e^{-x \steepness})$.
In this example, as $x$ increases, the result smoothly transitions from $y$ to $z$.
Thus, the derivative of the result wrt.~$x$ is defined as
\begin{align}
    \frac{\partial}{\partial x} \big[ y \texttt{ if } \tilde x < c \texttt{ else } z \big]
    &= \frac{\partial}{\partial x} \left( y\cdot\sigma(c−x) + z\cdot(1−\sigma(c−x)) \right) \notag \\ %
    &= (z-y) \cdot \sigma(c-x) \cdot (1-\sigma(c-x))\,.
\end{align}
Hence, the gradient descent method can influence the condition $({\tilde x < c})$ to hold if the \texttt{if} case reduces the loss, or influence the condition to fail if the \texttt{else} case reduces the loss function.

In this example, $y$ and $z$ may not only be scalar values but also results of algorithms or parts of an algorithm themselves.
This introduces a recursive formalism of relaxed program flow, where parts of an algorithm are combined via a convex combination:
\begin{align}
    \big[ f(s) \texttt{ if } a < b \texttt{ else } g(s) \big]\ \equiv\ \sigma(b-a)\cdot f(s) + (1-\sigma(b-a))\cdot g(s)
\end{align}
where $f$ and $g$ denote functions, algorithms, or sequences of statements that operate on the set of all variables $s$ via call-by-value and return the set of all variables $s$. 
The result of this may either overwrite the set of all variables ($s:=[...]$) or be used in a nested conditional statement. %

After introducing \texttt{if-else} statements above, we extend the idea to loops, which extends the formalism of relaxed program flow to Turing-completeness. %
In fixed loops, i.e., loops with a predefined number of iterations, since there is only a single computation path, no relaxation is necessary, and, thus, fixed loops can be handled by unrolling.
 
The more complex case is conditional unbounded loops, i.e., \texttt{While} loops, which are executed as long as a condition holds. 
For that, let $(s_i)_{i\in\mathbb{N}}$ be the sequence of all variables after applying $i$ times the content of a loop.
That is, $s_0 = s$ for an initial set of all variables $s$, and $s_i = f(s_{i-1})$, where $f$ is the content of a loop, i.e., a function, an algorithm, or sequence of statements.
Let $a, b$ denote accessing variables of $s$, i.e., $s[a], s[b]$, respectively.
By recursively applying the rule for \texttt{if-else} statements, we obtain the following rule for unbounded loops:
\begin{align}
    &\big[ \texttt{while } a < b \texttt{ do } s:=f(s) \big]\\
    &\equiv
    \sum_{i=0}^{\infty} \ \ \underbrace{\textstyle\prod_{j=0}^i \left( \sigma(b_j-a_j)\right)}_{\text{(a)}} \cdot \underbrace{(1 - \sigma(b_{i+1}-a_{i+1}))}_{\text{(b)}}\ \cdot\ s_i 
\end{align}
Here, (a) is the probability that the $i$th iteration is reached and (b) is the probability that there are no more than $i$ iterations.
Together, (a) and (b) is the probability that there are exactly $i$ iterations weighing the state of all variables after applying $i$ times $f$, which is $s_i$. 
Computationally, we evaluate the infinite series until the probability of execution (a) becomes numerically negligible or a predefined maximum number of iterations has been reached.
Again, the result may either overwrite the set of all variables ($s:=[...]$) or be used in a nested conditional statement. %

\paragraph{Complexity and Merging of Paths}
To compute the exact expectation value of an algorithm under logistic perturbation of its variables, all computation paths would have to be evaluated separately to account for dependencies.
However, this would result in an exponential complexity.
Therefore, we compute the exact expectation value for nested conditional blocks, but for sequential conditional blocks we merge the computation paths at the end of each block.
This allows for a linear complexity in the number of sequential conditional blocks and an exponential complexity only in the largest depth of nested conditional blocks.
Note that the number of sequential conditional blocks is usually much larger than the depth of conditional blocks, e.g., hundreds or thousands of sequential blocks and a maximum depth of just $2-5$ in our experiments.
An example of a dependency is the expression $\big[ a := (f(x) \texttt{ if } i < 0 \texttt{ else } g(x)) \texttt{; } b := (0 \texttt{ if } a < 0 \texttt{ else } a^2 )\big]$, which contains a dependence between the two sequential conditional blocks, which introduces the error in our approximation.
In general, our formalism also supports modeling dependencies between sequential conditional blocks; however, practically, it might become intractable for entire algorithms.
Also, it is possible to consider relevant dependencies explicitly if an algorithm relies on specific dependencies.

\paragraph{Perturbations of Variables vs.~Perturbations of Inputs}
Note that modeling the perturbation of variables is different from modeling the perturbation of the inputs.
A condition, where the difference becomes clear, is, e.g., $(\tilde x<x)$.
When modeling input perturbations, the condition would have a strong implicit conditional dependency and evaluate to a $0\%$ probability.
However, in this chapter, we do \textit{not} model perturbations of the inputs but instead model perturbations of variables each time they are accessed such that accessing a variable twice is independently identically distributed (iid).
Therefore, $(\tilde x<x)$ evaluates to a $50\%$ probability.
To minimize the approximation error of the relaxation, only those variables should be relaxed for which a gradient is required.

\subsection{Relaxed Comparators}

\begin{marginfigure}
    \centering
    \begin{tikzpicture}[scale=.9]
    \draw[->] (-2.5,0) -- (2.75,0) node[anchor=south, yshift=+.5mm, xshift=-3.mm] {$b-a$};
    \draw	(0,0) node[anchor=north] {0}
    (1,0) node[anchor=north] {2}
    (2,0) node[anchor=north] {4}
    (-1,0) node[anchor=north] {-2}
    (-2,0) node[anchor=north] {-4};
    \draw[->] (0,0) -- (0,1.75) node[anchor=east] {$p$};
    \draw[dotted] (-2.5,1.5) -- (2.5,1.5);

    \draw[thick, cyan] (-2.5,0) -- (0,0);
    \draw[thick, cyan, dashed] (0,0) -- (0,1.5);
    \draw[thick, cyan] (0,1.5) -- (2.5,1.5);
    \node[cyan] at (0,0) {\textbullet};
    \node[cyan] at (0,1.5) {$\circ$};
    
    \draw[thick, magenta]   plot [domain=-2.5:2.5] (\x, {1.5*(0.001 + 1 / (1 + exp(-2*\x)))});
    \end{tikzpicture}\\%
    \begin{tikzpicture}[scale=.9]
    \draw[->] (-2.5,0) -- (2.75,0) node[anchor=south, yshift=+.5mm, xshift=-3.mm] {$b-a$};
    \draw	(0,0) node[anchor=north] {0}
    (1,0) node[anchor=north] {2}
    (2,0) node[anchor=north] {4}
    (-1,0) node[anchor=north] {-2}
    (-2,0) node[anchor=north] {-4};
    
    \draw[->] (0,0) -- (0,1.75) node[anchor=east] {$p$};
    \draw[dotted] (-2.5,1.5) -- (2.5,1.5);

    \csvreader[ head to column names,%
                late after head=\xdef\aold{\a}\xdef\bold{\b},%
                after line=\xdef\aold{\a}\xdef\bold{\b}]%
                {fig_algovision/sechfn2.csv}{}{%
    \draw[thick, magenta] (\aold, {(\bold*\bold)*1.5}) -- (\a, {(\b*\b)*1.5});
    }
    \csvreader[ head to column names,%
                late after head=\xdef\aold{\a}\xdef\bold{\b},%
                after line=\xdef\aold{\a}\xdef\bold{\b}]%
                {fig_algovision/sechfn2.csv}{}{%
    \draw[thick, magenta!40] (\aold, {(1-\bold*\bold)*1.5}) -- (\a, {(1-\b*\b)*1.5});
    }
    
    \draw[thick, cyan] (-2.5,0) -- (0,0);
    \draw[thick, cyan] (0,0) -- (0,1.5);
    \draw[thick, cyan] (0,0) -- (2.5,0);
    \node[cyan] at (0,1.5) {\textbullet};
    \node[cyan] at (0,0) {$\circ$};
    \end{tikzpicture}
    \caption{
        \textit{Top:} hard decision boundary (cyan) and probability under logistic perturbation (magenta).
        \textit{Bottom:} hard equality (cyan), relaxed equality (magenta), and relaxed inequality (light magenta).
    }
    \label{fig:phiplots}
    \label{fig:phiplots2}
    \label{fig:algonet-operators}
    \label{fig:relaxed-comparators}
\end{marginfigure}

So far, we have only considered the comparator $<$.\ \ 
$>$ follows by swapping the arguments:
\begin{align}
    \mathbb{P}\ \big[a < b\big] \ \ &\equiv\ \sigma(b-a)
    &
    \mathbb{P}\ \big[a > b\big] \ \ &\equiv\ \sigma(a-b)
\end{align}

\vspace{-.75em}
\paragraph{Relaxed Equality}
For the equality operator $=$, we consider two distributions $\tilde a \sim \operatorname{Logistic}(a, 1/\steepness)$ and $\tilde b \sim \operatorname{Logistic}(b, 1/\steepness)$, which we want to check for similarity / equality. 
Given value $a$, we compute the likelihood that $a$ is a sample from $\tilde b$ rather than $\tilde a$. 
If $a$ is equally likely to be drawn from $\tilde a$ and $\tilde b$, $\tilde a$ and $\tilde b$ are equal.
If $a$ is unlikely to be drawn from $\tilde b$, $\tilde a$ and $\tilde b$ are unequal.
To compute whether it is equally likely for $a$ to be drawn from $\tilde a$ and $\tilde b$, we take the ratio between the likelihood that $a$ is from $\tilde b$ ($f_{\operatorname{log}}(a; b, 1/\steepness)$) and the likelihood that $a$ is from $\tilde a$ ($f_{\operatorname{log}}(a; a, 1/\steepness)$):
\begin{align}
    \mathbb{P}\ \big[a = b\big] \ \ \equiv\ 
    \frac{ f_{\operatorname{log}}(a; b, 1/\steepness) }{ f_{\operatorname{log}}(a; a, 1/\steepness) }
    \ =\ 
    \frac{ f_{\operatorname{log}}(b; a, 1/\steepness) }{ f_{\operatorname{log}}(b; b, 1/\steepness) }
    \ =\ 
    \operatorname{sech}^2 \left( \frac{b-a}{2/\steepness} \right)
    \label{eq:relaxed-equality}
\end{align}
These relaxed comparators are displayed in Figure~\ref{fig:relaxed-comparators}.
An alternative derivation for \eqref{eq:relaxed-equality} is the normalized conjunction of $\neg (a < b)$ and $\neg (a > b)$.

\vspace{-.5em}
\paragraph{Relaxed Logics}
To compute probabilities of conjunction (i.e., and) or of disjunction (i.e., or), we use the product or probabilistic sum, respectively. 
This corresponds to an intersection / union of independent events.
Alternatives to and extensions of this are discussed throughout this thesis, and a summary of alternatives can be found in Supplementary Material~\ref{sm:tcn}.

\vspace{-.5em}
\paragraph{Relaxed Maximum}
To compare more than two elements and relax the $\arg\max$ function, we use the multinomial logistic distribution, 
which is also known as the softmax distribution.
\begin{align}
    \mathbb{P}(i = \arg\max_j X_j) = \frac{e^{X_i \steepness}}{\sum_j e^{X_j \steepness}}
\end{align}
To relax the $\max$ operation, we use the product of $\arg\max$ / softmax and the respective vector.
The softmax distribution corresponds to an $\arg\max$ under perturbation with a Gumbel distribution.
Alternatives to this are discussed throughout this thesis, and an extensive discussion of $\min$ / $\max$ between two elements can be found in Chapter~\ref{ch:diffsort} and Supplementary Material~\ref{sec:properties-of-min-and-max}.

\vspace{-.5em}
\paragraph{Comparing Categorical Variables}
To compare a categorical probability distribution $X \in [0, 1]^n$ with a categorical probability distribution $Y$, we consider two scenarios:
If $Y\in \{0, 1\}^n$, i.e., $Y$ is one-hot, we can use the inner product of $X$ and $Y$ to obtain their joint probability.
However, if $Y\notin \{0, 1\}^n$, i.e., $Y$ is not one-hot, even if $X=Y$ the inner product can not be $1$, but a probability of $1$ would be desirable if $X=Y$.
Therefore, we ($L_2$) normalize $X$ and $Y$ before taking the inner product, which corresponds to the cosine similarity.
An example of the application of comparing categorical probability distributions is shown in the Levenshtein distance experiments in Section~\ref{sec:exp-ld}.

\subsection{Relaxed Indexing}

As vectors,  arrays,  and tensors are essential for algorithms and machine learning, we also formalize relaxed indexing. 
For this, we introduce real-valued indexing and categorical indexing.

\begin{marginfigure}
    \centering
    \hfill
    \begin{minipage}{\linewidth}
    \centering
        \includegraphics[width=.8\linewidth]{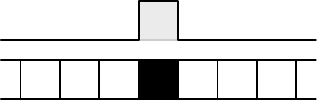}\par
        \vspace{0.0ex}%
        $\downarrow$\par
        \vspace{0.8ex}%
        \includegraphics[width=.8\linewidth]{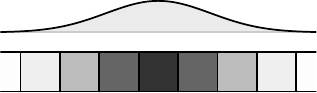}\par
    \end{minipage}\\[1em]
    \caption{
    1D real-valued indexing: the gray-value represents the extent to which each value is used for indexing.
    }
    \label{fig:indexing-1d}
\end{marginfigure}

\begin{marginfigure}
    \centering
    \begin{minipage}[c]{.45\linewidth}
        \includegraphics[height=\linewidth]{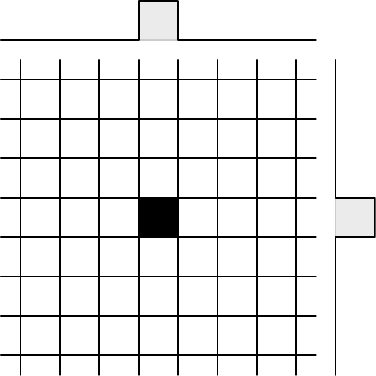}%
    \end{minipage}%
    \begin{minipage}[c]{.1\linewidth}
        ~\\
        $~\rightarrow~$%
    \end{minipage}%
    \begin{minipage}[c]{.45\linewidth}
        \includegraphics[height=\linewidth]{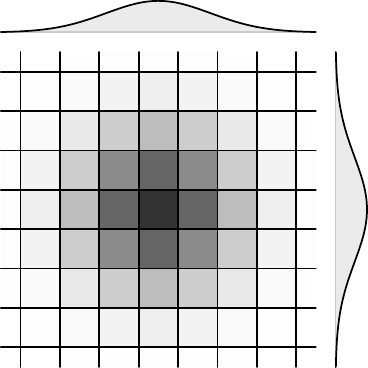}%
    \end{minipage}
    \caption{
    2D real-valued indexing: the gray-value represents the extent to which each value is used for indexing.
    }
    \label{fig:indexing-2d}
\end{marginfigure}

\paragraph{Real-Valued Indexing}
In a relaxed algorithm, indices may be drawn from the set of real numbers as they may be a convex combination of computations or a real-valued input.
This poses a challenge since it requires interpolating between multiple values.
The direct approach would be grid sampling with bilinear or bicubic interpolation to interpolate values.
For example, Neural Turing Machines use linear interpolation for real-valued indexing~\cite{graves2014neural}.
However, in bilinear or bicubic interpolation, relationships exceeding the direct (or next) neighbors in the array are not modeled, and they also do not model logistic perturbations.
Therefore, we index an $n$-dimensional tensor $\tA$ with values $\vi\in\sR^n$ via logistic perturbation by applying a convolution with a logistic filter $g$ and obtain the result as $(g * \tA)(\vi)$.
The convolution of a tensor $\tA$ with a logistic filter $g$ (not to be confused with discrete-discrete convolution in neural networks) yields a function that is evaluated at point $\vi\in\sR^n$. 
We choose the logistic filter over bilinear and bicubic filters because we model logistic perturbation and additionally because bilinear and bicubic filters only have compact support, whereas the logistic filter provides infinite support. 
This allows modeling relationships beyond the next neighbors and is more flexible as the inverse temperature $\steepness$ can be tuned for the respective application.
For stability, we normalize the coefficients used to interpolate the respective indexed values such that they sum up to one: 
instead of computing $(g * \tA)(\vi) = \sum_\vj g(\vj-\vi) \tA_\vj$, we compute $\sum_\vj g(\vj-\vi) \tA_\vj / (\sum_\vj g(\vj-\vi))$ where $\vj$ are all valid indices for the tensor $\tA$.
To prevent optimization algorithms from exploiting aliasing effects, we divide the coefficients by their sum only in the forward pass, but ignore this during the backward pass (computation of the gradient).
Real-valued indexing is displayed in Figures~\ref{fig:indexing-1d} and~\ref{fig:indexing-2d}, comparing it to hard indexing for 1D and 2D arrays.

\paragraph{Relaxed Categorical Indexing}
If a categorical probability distribution over indices is given, e.g., computed by $\operatorname{argmax}$ or its relaxation $\operatorname{softmax}$, categorical indexing can be used.
Here, the marginal categorical distribution is used as weights for indexing a tensor.

Note that real-valued indexing assumes that the indexed array follows a semantic order such as time series, an image, or the position in a grid.
If, in contrast, the array contains categorical information such as the nodes of graphs, values should be indexed with categorical indexing as their neighborhood is arbitrary.

\subsection{Complexity and Runtime Considerations}

In terms of runtime, one has to note that runtime optimized algorithms (e.g., Dijkstra) usually do not improve the runtime for the relaxation because for the proposed continuous relaxations all cases in an algorithm have to be executed. 
Thus, an additional condition to reduce the computational cost does not improve runtime because both (all) cases are executed.
Instead, it becomes beneficial if an algorithm solves a problem in a rather fixed execution order.
On the other hand, optimizing an algorithm with respect to runtime leads to interpolations between the fastest execution paths.
This optimization cannot improve the gradients but rather degrades them as it is an additional approximation and produces {gradients with respect to runtime heuristics}.
For example, when relaxing the Dijkstra shortest-path algorithm, there is an interpolation between different orders of visiting nodes, which is the heuristic that makes Dijkstra fast. 
However, if we have to follow all paths anyway (to compute the relaxation), it can lead to a combinatorial explosion. 
In addition, by interpolating between those alternative orders, a large amount of uncertainty is introduced, and the gradients will also depend on the orders of visiting nodes, both of which are undesirable.
Further, algorithms with rather strict execution can be executed in parallel on GPUs such that they can be faster than runtime-optimized sequential algorithms on CPUs.
Therefore, we prefer simple algorithms with a largely fixed execution structure and without runtime optimizations from both a runtime and gradient quality perspective.

\subsection{Alternative Distributions}
\label{sec:algovision:alt-dist}
Above, we focused on perturbing by logistic distributions; however, the formulation is not limited to this setting.
Specifically, in Chapters~\ref{ch:diffsort} and~\ref{ch:gendr}, we cover a variety of alternative distributions.
For a given problem setting, the optimal choice of distribution depends on the specific used algorithm as well as the data, which we specifically discuss for the setting of differentiable renderers in Chapter~\ref{ch:gendr}.
A collection of possible distributions is discussed in Supplementary Material~\ref{sm:dist}.

\subsection{The \texttt{AlgoVision} Library}
We published the proposed method in the form of the \inlinelogo{algovision} library to facilitate easy application and further research in this direction.
The library is based on Python and PyTorch~\cite{paszke2019pytorch}. 
It is publicly available under \url{https://github.com/Felix-Petersen/algovision}. 
The accompanying documentation is published under \url{https://felix-petersen.github.io/algovision-docs/} and also archived in the Internet Archive.
Examples for algorithms can be found in Sections~\ref{sec:bubble-sort} and~\ref{sec:exp-ld}.

\section{Sorting Supervision}\label{sec:bubble-sort}

\begin{marginfigure}
    \centering
    \includegraphics[]{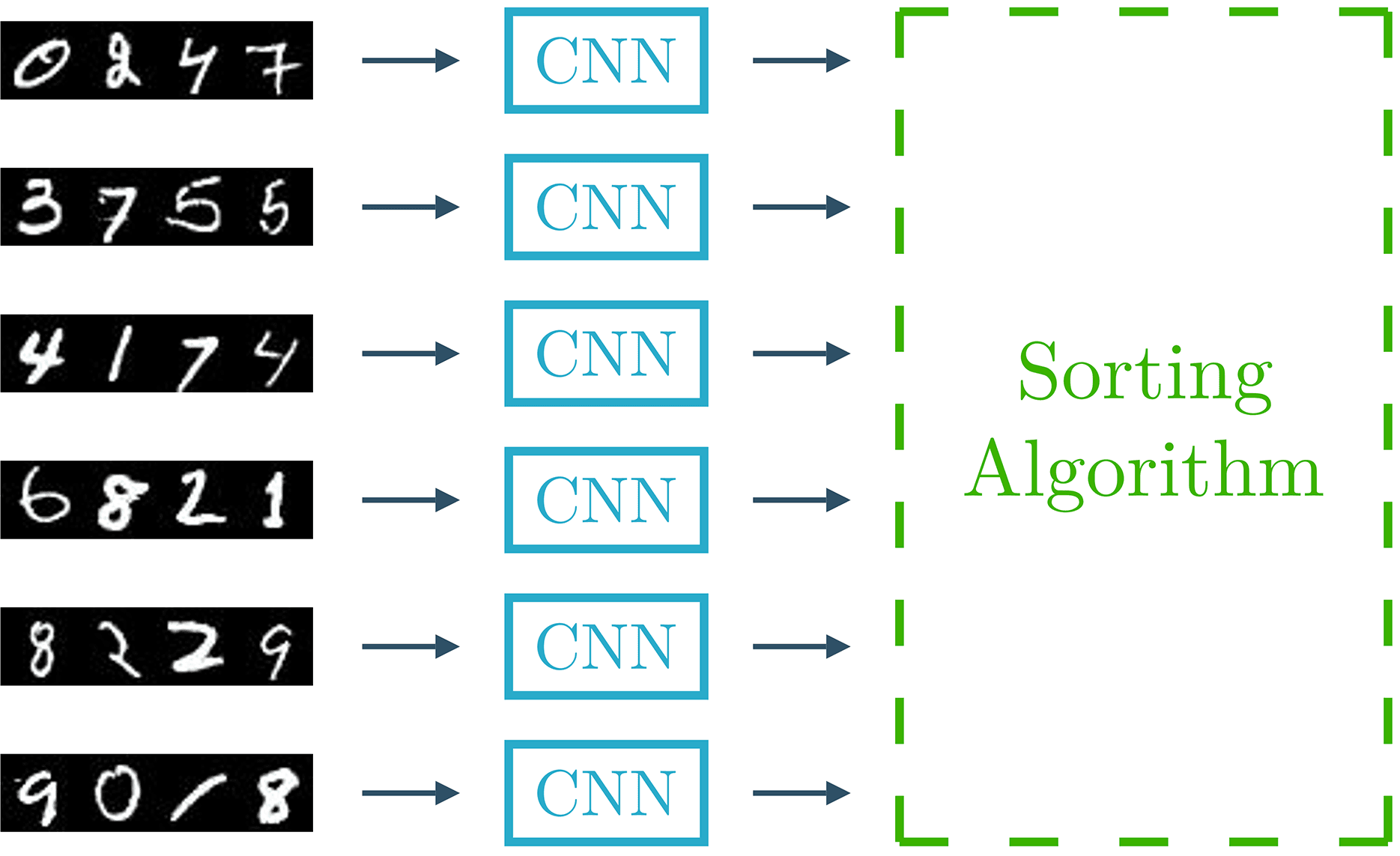}
    \caption{Illustration of Sorting Supervision}
    \label{fig:illustration-sorting}
\end{marginfigure}
In this section, we cover the first application of algorithmic supervision, namely sorting and ranking supervision.
Here, the idea is to train a model based on relative ordering information instead of absolute values or labels.
This can be done using a differentiable sorting algorithm that enforces the model to return values corresponding to the ground truth ordering, as illustrated in Figure~\ref{fig:illustration-sorting}.

\paragraph{Related Work}
The four-digit MNIST sorting benchmark task for sorting supervision has been first proposed by Grover \etal~\cite{grover2019neuralsort}. 
Grover~\etal~\cite{grover2019neuralsort} address the task by relaxing the permutation matrices to double stochastic matrices.
Cuturi~\etal~\cite{cuturi2019differentiable} pick up on this benchmark and propose a differentiable proxy by approximating the sorting problem with a regularizing optimal transport algorithm.
These methods are discussed in greater detail in the differentiable sorting-focused Chapter~\ref{ch:diffsort}.\\

In the sorting supervision four-digit MNIST experiment, a set of four-digit numbers based on concatenated MNIST digits~\cite{lecun2010mnist} is given, and the task is to find an order-preserving mapping from the images to scalars.
A CNN learns to predict a scalar for each of $n$ four-digit numbers such that their order is preserved among the predicted scalars.
For this, only sorting supervision in the form of the ground truth order of input images is given, while their absolute values remain unsupervised. 
This follows the protocol of Grover~\etal{}~\cite{grover2019neuralsort} and Cuturi~\etal{}~\cite{cuturi2019differentiable}. 
An example of a concatenated MNIST image is~~\mnistimg{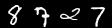}~.

Using our method, we relax the well established stable sorting algorithm Bubble sort~\cite{astrachan2003bubble}, which works by iteratively going through a list and swapping any two adjacent elements if they are not in the correct order until there are no more swap operations in one iteration. 
In the following, we display the annotated AlgoVision code of the relaxed algorithm.
The library provides a set of modules, a subset of which is imported at the beginning.
These modules can then be used to define the algorithm, which is then automatically relaxed and supports backpropagation, parallelization, and CPU as well as GPU execution.

\definecolor{modulecolor}{HTML}{F27100}  %
\begin{widepar}
\begin{minted}[escapeinside=||]{python}
from algovision import (
    |\textcolor{modulecolor}{Algorithm}|, |\textcolor{modulecolor}{Input}|, |\textcolor{modulecolor}{Output}|, |\textcolor{modulecolor}{Var}|, |\textcolor{modulecolor}{VarInt}|,                                          # core
    |\textcolor{modulecolor}{Let}|, |\textcolor{modulecolor}{LetInt}|, |\textcolor{modulecolor}{Print}|,                                                     # instructions
    |\textcolor{modulecolor}{Eq}|, |\textcolor{modulecolor}{NEq}|, |\textcolor{modulecolor}{LT}|, |\textcolor{modulecolor}{LEq}|, |\textcolor{modulecolor}{GT}|, |\textcolor{modulecolor}{GEq}|, |\textcolor{modulecolor}{CatProbEq}|, |\textcolor{modulecolor}{CosineSimilarity}|, |\textcolor{modulecolor}{IsTrue}|, |\textcolor{modulecolor}{IsFalse}|,  # conditions
    |\textcolor{modulecolor}{If}|, |\textcolor{modulecolor}{While}|, |\textcolor{modulecolor}{For}|,                                                   # control_structures
    |\textcolor{modulecolor}{Min}|, |\textcolor{modulecolor}{ArgMin}|, |\textcolor{modulecolor}{Max}|, |\textcolor{modulecolor}{ArgMax}|                                                   # functions
)
import torch

bubble_sort = |\textcolor{modulecolor}{Algorithm}|(
    # Define the variables the input corresponds to
    |\textcolor{modulecolor}{Input}|('array'),
    # Declare and initialize all differentiable variables 
    |\textcolor{modulecolor}{Var}|('a',        torch.tensor(0.)),
    |\textcolor{modulecolor}{Var}|('b',        torch.tensor(0.)),
    |\textcolor{modulecolor}{Var}|('swapped',  torch.tensor(1.)),
    |\textcolor{modulecolor}{Var}|('loss',     torch.tensor(0.)),
    # Declare and initialize a hard integer variable (VarInt) for the control flow.
    # It can be defined in terms of a lambda expression. The required variables
    # are automatically inferred from the signature of the lambda expression.
    |\textcolor{modulecolor}{VarInt}|('n', lambda array: array.shape[1] - 1),
\end{minted}
\end{widepar}
\newpage
\begin{widepar}
\begin{minted}[escapeinside=||]{python}
    # Start a relaxed While loop:
    |\textcolor{modulecolor}{While}|(|\textcolor{modulecolor}{IsTrue}|('swapped'),
        # Set `swapped` to 0 / False
        |\textcolor{modulecolor}{Let}|('swapped', 0),
        # Start an unrolled For loop. Corresponds to `for i in range(n):`
        |\textcolor{modulecolor}{For}|('i', 'n',
            # Set `a` to the `i`th element of `array`
            |\textcolor{modulecolor}{Let}|('a', 'array', ['i']),
            # Using an inplace lambda expression, we can include computations 
            # based on variables to obtain the element at position i+1. 
            |\textcolor{modulecolor}{Let}|('b', 'array', [lambda i: i+1]),
            # An If-Else statement with the condition a > b
            |\textcolor{modulecolor}{If}|(|\textcolor{modulecolor}{GT}|('a', 'b'),
               if_true=[
                   # Set the i+1 th element of array to a
                   |\textcolor{modulecolor}{Let}|('array', [lambda i: i + 1], 'a'),
                   # Set the i th element of array to b
                   |\textcolor{modulecolor}{Let}|('array', ['i'], 'b'),
                   # Set swapped to 1 / True
                   |\textcolor{modulecolor}{Let}|('swapped', 1.),
                   # Increment the loss by 1 using a lambda expression
                   |\textcolor{modulecolor}{Let}|('loss', lambda loss: loss + 1.),
               ]
           ),
        ),
        # Decrement the hard integer variable n by 1
        |\textcolor{modulecolor}{LetInt}|('n', lambda n: n-1),
    ),
    # Define what the algorithm should return
    |\textcolor{modulecolor}{Output}|('array'),
    |\textcolor{modulecolor}{Output}|('loss'),
    # Set the inverse temperature beta
    beta=5,
)
\end{minted}
\end{widepar}

We include a variable (\texttt{swapped}) that keeps track of whether the input sequence is in the correct order by setting it to true if a swap operation occurs.
Due to the relaxation, this variable is a floating-point number between $0$~and~$1$ corresponding to the probability that the predictions are sorted correctly (under perturbation of variables).
We use this probability as the loss function. 
This variable equals the probability that no swap operation was necessary, and thus $\mathcal{L} = 1 - \prod_{p\in P} (1-p)$ for probabilities $p$ of each potential swap $p\in P$.
By that, the training objective enforces the input sequence to be sorted and, thus, enforces that the scores predicted by the neural network correspond to the supervised partial order.
In fact, the number of swaps in bubble sort corresponds to the Kendall's $\tau$ coefficient, which indicates to which degree a sequence is sorted.
Note that, e.g., QuickSort does not have this property.

We emphasize that the task of the trained neural network is \textit{not} to sort a sequence but instead to predict a score for each element such that the ordering / ranking corresponds to the supervised ordering / ranking.
While the relaxed algorithm can sort the inputs correctly, at evaluation time, following the setup of \cite{grover2019neuralsort} and \cite{cuturi2019differentiable} we use an $\operatorname{argsort}$ method to test whether the outputs produced by the neural network are in accordance with the ground truth partial order.
We use the same network architecture as \cite{grover2019neuralsort} and \cite{cuturi2019differentiable}.
Here, we only optimize the inverse temperature for $n=5$, resulting in $\steepness=8$, and fix this for all other $n$ . For training, we use the Adam optimizer \cite{kingma2015adam} with a learning rate of $10^{-4}$ for a number of iterations between $1.5\cdot10^5$ and $1.5\cdot10^6$.
For comparability to Grover~\etal~\cite{grover2019neuralsort} and Cuturi~\etal~\cite{cuturi2019differentiable}, we use the same network architecture.
That is, two convolutional layers with a kernel size of $5\times5$, $32$ and $64$ channels respectively, each followed by a ReLU and MaxPool layer; 
after flattening, it is followed by a linear layer with a size of $64$, a ReLU layer, and a linear output layer mapping to a scalar.

We evaluate our method against state-of-the-art hand-crafted relaxations of the sorting operation using the same network architecture and evaluation metrics as Grover~\etal{}~\cite{grover2019neuralsort} and Cuturi~\etal{}~\cite{cuturi2019differentiable}.
As displayed in Table~\ref{table:results-bubble}, our general formulation outperforms state-of-the-art hand-crafted relaxations of the sorting operation for sorting supervision.
In Chapter~\ref{ch:diffsort}, we show that differentiable sorting networks improve upon this result.

\begin{table}[t]
    \centering
    \caption{
        Results for the 4-digit MNIST sorting task, averaged over 10 runs.
        Baselines as reported by Cuturi~\etal{}~\cite{cuturi2019differentiable}.
        Trained and evaluated on sets of $n$ elements, the \hbox{displayed metrics} are exact matches (and element-wise correct ranks). 
        \label{table:results-bubble}
    }
    \begingroup
    \setlength{\tabcolsep}{3pt}
    \footnotesize
    \begin{tabular}{lccc}
        \toprule
        Method & $n=3$ & $n=5$ & $n=7$ \\
        \midrule
        Stoch. NeuralSort~\cite{grover2019neuralsort}          & $0.920$ ($0.946$) & $0.790$ ($0.907$) & $0.636$ ($0.873$)  \\
        Det. NeuralSort~\cite{grover2019neuralsort}       & $0.919$ ($0.945$) & $0.777$ ($0.901$) & $0.610$ ($0.862$)  \\
        Optimal Transport~\cite{cuturi2019differentiable}               & $0.928$ ($0.950$) & $0.811$ ($0.917$) & $0.656$ ($0.882$)  \\
        \midrule
        Relaxed Bubble Sort                 & $\pmb{0.944}$ ($\pmb{0.961}$) & $\pmb{0.842}$ ($\pmb{0.930}$) & $\pmb{0.707}$ ($\pmb{0.898}$)  \\
        \bottomrule
    \end{tabular}
    \endgroup
\end{table}

\section{Shortest-Path Supervision}\label{sec:exp-shortest-path}

\begin{figure}[h]
    \centering
    \includegraphics[]{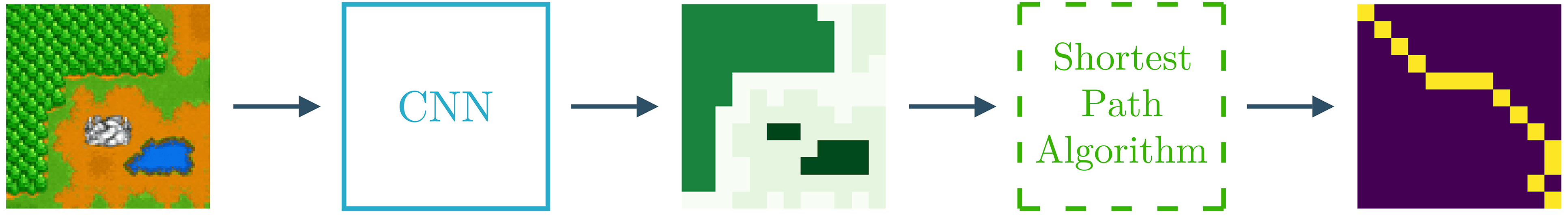}
    \caption{Illustration of Shortest-Path Supervision}
    \label{fig:illustration-shortest-path}
\end{figure}

The setting of shortest-path supervision is illustrated in Figure~\ref{fig:illustration-shortest-path}.
Here, an image is given, a cost embedding is predicted by a neural network, and the algorithm yields a differentiable shortest path.
After comparing the predicted shortest path to a ground truth shortest path and backpropagating the error through the algorithm, we can train the neural network.

\begin{figure*}[t]
    \centering
    \includegraphics[width=\linewidth]{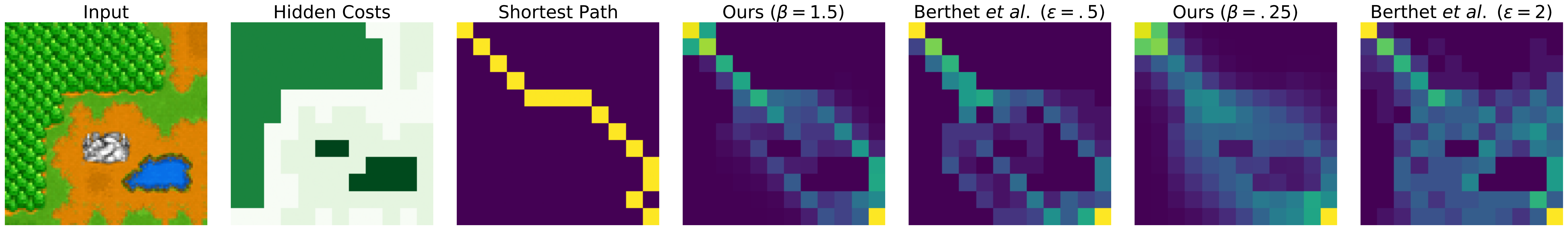}
    \caption{
    \textit{(From left to right.)} Example input image of the Warcraft terrain data set with the hidden cost matrix and the resulting shortest path.
    Shortest paths relaxed with the proposed method for $\steepness\in\{1.5, 0.25\}$, which correspond to the perturbed paths by Berthet \etal~\cite{berthet2020learning} with $\epsilon\in\{0.5,2.0\}$.
    }
    \label{fig:shortest-path-supervision}
\end{figure*}

For the 2D Warcraft terrains shortest-path supervision benchmark, we follow the setup by Vlastelica~\etal~\cite{vlastelica2019differentiation} and Berthet~\etal~\cite{berthet2020learning} and use the data set of $10\,000$ patches of Warcraft terrains of size $96\times96$ representing terrain grids of size $12\times12$.
Given an image of a terrain (e.g., Figure~\ref{fig:shortest-path-supervision} first), the goal is to predict the shortest path from the top-left to the bottom-right (e.g., Figure~\ref{fig:shortest-path-supervision} third) according to a hidden cost matrix (e.g., Figure~\ref{fig:shortest-path-supervision} second).
For this, $12\times12$ binary matrices of the shortest path are supervised, while the hidden cost matrix is used only to determine the shortest path.
Integrating a shortest-path algorithm into a neural architecture lets the neural network produce a cost embedding of the terrain, which the shortest-path algorithm uses for predicting the shortest path.

\paragraph{Related Work}
Vlastelica~\etal{}~\cite{vlastelica2019differentiation} tackle this task by finding a linearization of the Dijkstra algorithm \cite{dijkstra1959note}, which they can differentiate.
Berthet \etal{}~\cite{berthet2020learning} take up this problem and produce gradient estimates for the Dijkstra algorithm by stochastically perturbing the inputs to the shortest-path optimization problem. 
In their works, Vlastelica~\etal~\cite{vlastelica2019differentiation} and Berthet~\etal~\cite{berthet2020learning} show that integrating and differentiating a shortest-path algorithm can improve the results by allowing the neural network to predict a cost matrix from which the shortest path can be computed via the algorithm.
This performs significantly better than a ResNet baseline, where the shortest paths have to be predicted by the neural network alone (see Table~\ref{tab:shortest-path-supervision}).\\

\begin{margintable}
    \caption{
    Results for the Warcraft shortest-path task using shortest-path supervision, averaged over 10 runs.
    Reported is the exact match accuracy (EM). %
    For additional and more extensive results, see Chapter~\ref{ch:alternative}.
    }
    \label{tab:shortest-path-supervision}
    \centering
    \begingroup
    \footnotesize
    \begin{tabular}{lc}
    \toprule
        Method & EM \\
        \midrule
        ResNet Baseline  & $40.2\%$  \\
        Black-Box Loss   & $86.6\%$  \\
        Perturbed Opt.   & $80.6\%$  \\ %
        \midrule
        AlgoVision       & $\pmb{95.8\%}$  \\
    \bottomrule
    \end{tabular}
    \endgroup
\end{margintable}
For this task, we relax the Bellman-Ford algorithm \cite{bellman1958routing} with 8-neighborhood, node weights, and path reconstruction.
For the loss between the supervised shortest paths and the paths produced by the relaxed Bellman-Ford algorithm, we use the $\ell^2$ loss.
To illustrate the shortest paths created by our method and to compare them to those created through the Perturbed Optimizers by Berthet~\etal~\cite{berthet2020learning}, we display examples of back-traced shortest paths for two inverse temperatures in Figure~\ref{fig:shortest-path-supervision} (center to right).

We use the same ResNet network architecture as Vlastelica~\etal~\cite{vlastelica2019differentiation} and Berthet~\etal~\cite{berthet2020learning}.
That is, the first five layers of ResNet18 followed by an adaptive max pooling to the size of $12\times12$ and an averaging over all features.
As in previous works, we train for 50 epochs with batch size 70 and decay the learning rate by a factor of $0.1$ after $60\%$ as well as after $80\%$ of training.

As shown in Table~\ref{tab:shortest-path-supervision}, our relaxation outperforms all baselines. %

\section{Silhouette Supervision}\label{sec:exp-3d}
\marginnote{
The silhouette supervision experiment in this chapter is to demonstrate the flexibility and applicability of the proposed approach even to complex tasks like silhouette supervision.
In Chapter~\ref{ch:gendr}, we discuss a natively implemented differentiable renderer, a special case of which is equivalent to the differentiable renderer proposed here. 
}

\begin{figure}[h]
    \centering
    \includegraphics[]{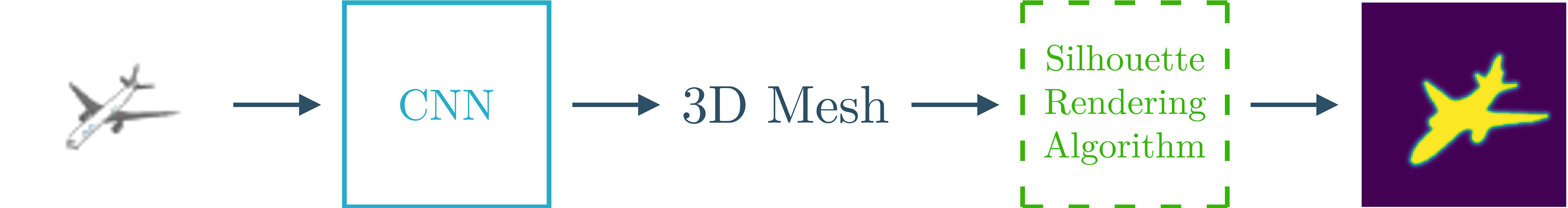}
    \caption{Illustration of Silhouette Supervision}
    \label{fig:illustration-silhouette-supervision}
\end{figure}

Reconstructing a 3D model from a single 2D image is an important task in computer vision.
This task is frequently solved using silhouette supervision as illustrated in Figure~\ref{fig:illustration-silhouette-supervision}.
Here, a single image is processed by a neural network, which returns a 3D mesh; this mesh is then rendered back into the image space and the predicted image can be compared to the silhouette of the input, facilitating training. 

\paragraph{Related Work}
Recent works~\cite{yan2016perspective, kato2017neural, liu2019soft} have benchmarked their differentiable renderers on a data set of 13 object classes from ShapeNet~\cite{chang2015shapenet} that have been rendered from $24$ azimuths at a resolution of $64\times64$~\cite{kato2017neural}.
Kato~\etal~\cite{kato2017neural} propose a renderer where surrogate gradients of rasterization are approximated to perform 3D mesh reconstruction via silhouette supervision as well as 3D style transfer.
Liu~\etal~\cite{liu2019soft} propose a differentiable renderer without surrogate gradients by using a differentiable aggregating process and apply it to 3D mesh reconstruction as well as pose / shape optimization.
More details can be found in the differentiable rendering-focused Chapter~\ref{ch:gendr}.\\

For training, the silhouette of the predicted mesh is rendered from two view-points by a differentiable renderer and the intersection-over-union between the rendered and predicted meshes is used as a training objective to update the neural network \cite{kato2017neural, liu2019soft}.
For training, we also use the same neural network architecture as Kato~\etal~\cite{kato2017neural} as well as Liu~\etal~\cite{liu2019soft}.
While some differentiable renderers also render RGB images, in these experiments, only the silhouette is used for supervision.
Specifically, public implementations of \cite{kato2017neural, liu2019soft} only use silhouette supervision.

\begin{marginfigure}
    \centering
    \includegraphics[width=\linewidth]{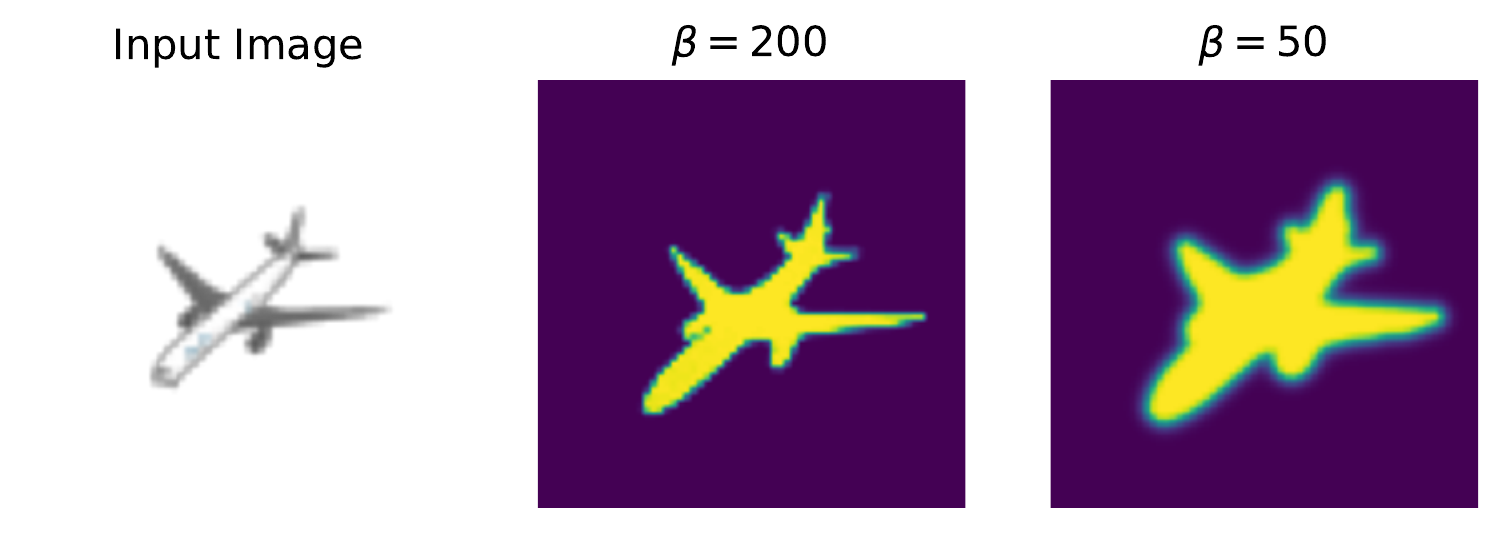}
    \caption{
    An input image from the data set \textit{(left)}.
    Silhouette of a prediction rendered with directed Euclidean distance approach for two different inverse temperatures $\beta = 200$ \textit{(center)} and $\beta=50$ \textit{(right)}.
    }
    \label{fig:algovision:silhouette}
\end{marginfigure}

For this task, we relax two silhouette rendering algorithms.
The algorithms rasterize a 3D mesh as follows:
For each pixel and for each triangle of the mesh, if a pixel lies inside a triangle, the value of the pixel in the output image is set to $1$.
The condition of whether a pixel lies inside a triangle is checked in two alternative fashions:
(1) by three nested \texttt{if} conditions that check on which side of each edge the pixel lies.
(2) by checking whether the directed euclidean distance between a pixel and a triangle is positive.
Note that, by relaxing these algorithms using our framework, we obtain differentiable renderers equivalent to Pix2Vex~\cite{petersen2019pix2vex} for (1) and a specific instance of GenDR for (2).
A greater discussion of relations between differentiable renderers can be found in Chapter~\ref{ch:gendr}.
Examples of relaxed silhouette renderings and an example image from the data set are displayed in Figure~\ref{fig:algovision:silhouette}.

\begin{table*}[]
    \centering
    \caption{
    Single-view 3D reconstruction results using silhouette supervision. Reported is the 3D IoU.
    }
    \label{tab:silhouette}
    \resizebox{\linewidth}{!}{
\centering
\setlength{\tabcolsep}{2pt}
\begin{tabular}{lcccccccccccccc} 
\toprule
Method    & Airplane & Bench  & Dresser & Car    & Chair  & Display & Lamp & Speaker & Rifle & Sofa & Table & Phone & Vessel & \textit{Mean}       \\ 
\midrule
\textbf{With a batch size of 64} \\
Yan \textit{et al.} 
\cite{yan2016perspective} (retrieval)                      & 0.5564   & 0.4875 & 0.5713  & 0.6519 & 0.3512 & 0.3958  & 0.2905 & 0.4600   & 0.5133 & 0.5314  & 0.3097 & 0.6696 & 0.4078  & 0.4766       \\
Yan \textit{et al.} 
\cite{yan2016perspective} (voxel)                          & 0.5556   & 0.4924 & 0.6823  & 0.7123 & 0.4494 & 0.5395  & 0.4223 & 0.5868   & 0.5987 & 0.6221  & 0.4938 & 0.7504 & 0.5507  & 0.5736       \\
Kato \textit{et al.} 
\cite{kato2017neural} (NMR)                          & 0.6172   & 0.4998 & 0.7143  & 0.7095 & 0.4990 & 0.5831  & 0.4126 & 0.6536   & 0.6322 & 0.6735  & 0.4829 & 0.7777 & 0.5645  & 0.6015       \\
Liu \textit{et al.} 
\cite{liu2019soft} (SoftRas)          & 0.6419   & 0.5080 & 0.7116  & 0.7697 & 0.5270 & 0.6156  & {0.4628} & 0.6654   & {0.6811} & 0.6878  & 0.4487 & 0.7895 & 0.5953  &  0.6234       \\ 
\midrule
\textbf{With a batch size of 2} \\
Liu \textit{et al.} 
\cite{liu2019soft} (SoftRas)   & $\pmb{0.5741}$ & $\pmb{0.3746}$ & ${0.6373}$ & $\pmb{0.6939}$ & $\pmb{0.4220}$ & ${0.5168}$ & ${0.4001}$ & ${0.6068}$ & $\pmb{0.6026}$ & $\pmb{0.5922}$ & ${0.3712}$ & $\pmb{0.7464}$ & $\pmb{0.5534}$ & $\pmb{0.5455}$ \\
Relaxed Sil. via Three Edges  & ${0.5418}$ & ${0.3667}$ & $\pmb{0.6626}$ & ${0.6546}$ & ${0.3899}$ & ${0.5229}$ & ${0.4105}$ & $\pmb{0.6232}$ & ${0.5497}$ & ${0.5639}$ & ${0.3580}$ & ${0.6609}$ & ${0.5279}$ & ${0.5256}$  \\  %
Relaxed Sil. via Euclid. Dist & ${0.5399}$ & ${0.3698}$ & ${0.6503}$ & ${0.6524}$ & ${0.4044}$ & $\pmb{0.5261}$ & $\pmb{0.4247}$ & ${0.6225}$ & ${0.5723}$ & ${0.5643}$ & $\pmb{0.3829}$ & ${0.7265}$ & ${0.5180}$ & ${0.5349}$  \\  %
\bottomrule
\end{tabular}}
\end{table*}

As the simple silhouette renderer does not have any optimizations, such as discarding pixels that are far away from a triangle or triangles that are occluded by others, it is not very efficient.
Thus, due to limited resources, we are only able to train with a maximum batch size of $2$ while previous works used a batch size of $64$.
Therefore, we reproduce the recent best performing work by Liu~\etal~\cite{liu2019soft} on a batch size of only~$2$ to allow for a fair comparison. 
For directed Euclidean distance, we use an inverse temperature of $\steepness=2\,000$; for three edges, $\steepness=10\,000$.
For comparability to Liu~\etal~\cite{liu2019soft}, we use the same network architecture.
That is, three convolutional layers with a kernel size of $5\times5$, $64$, $128$, and $256$ channels respectively, each followed by a ReLU;
after flattening, this is followed by 6 ReLU-activated fully connected layers with the following output dimensions:
$1024, 1024, 512, 1024, 1024, 642\times3$.
The $642\times3$ elements are interpreted as three-dimensional vectors that displace the vertices of a sphere with $642$ vertices.
We train the Three Edges approach with Adam ($\eta=5\cdot10^{-5}$) for $2.5\cdot10^6$ iterations, and train the directed Euclidean distance approach with Adam ($\eta=5\cdot10^{-5}$) for $10^6$ iterations, as each of them took around 6 days of training.
We decay the learning rate by a factor of $0.3$ after $60\%$ as well as after $80\%$ of training.

We report the average as well as class-wise 3D IoU results in Table~\ref{tab:silhouette}. 
Our relaxations outperform the retrieval baseline by Yan~\etal~\cite{yan2016perspective} even though we use a batch size of only~$2$.
In direct comparison to the SoftRas renderer at a batch size of $2$, our relaxations achieve the best accuracy for 5 out of 13 object classes.
However, on average, our methods, although they do not outperform SoftRas, show comparable performance with a drop of only $1\%$.
It is notable that the directed Euclidean distance approach performs better than the three edges approach.
The three edges approach is faster by a factor of $3$ because computing the directed Euclidean distance is more expensive than three nested \texttt{if} clauses (even in the relaxed case.)

\section{Levenshtein Distance Supervision}\label{sec:exp-ld}

\begin{marginfigure}
    \centering
    \includegraphics[]{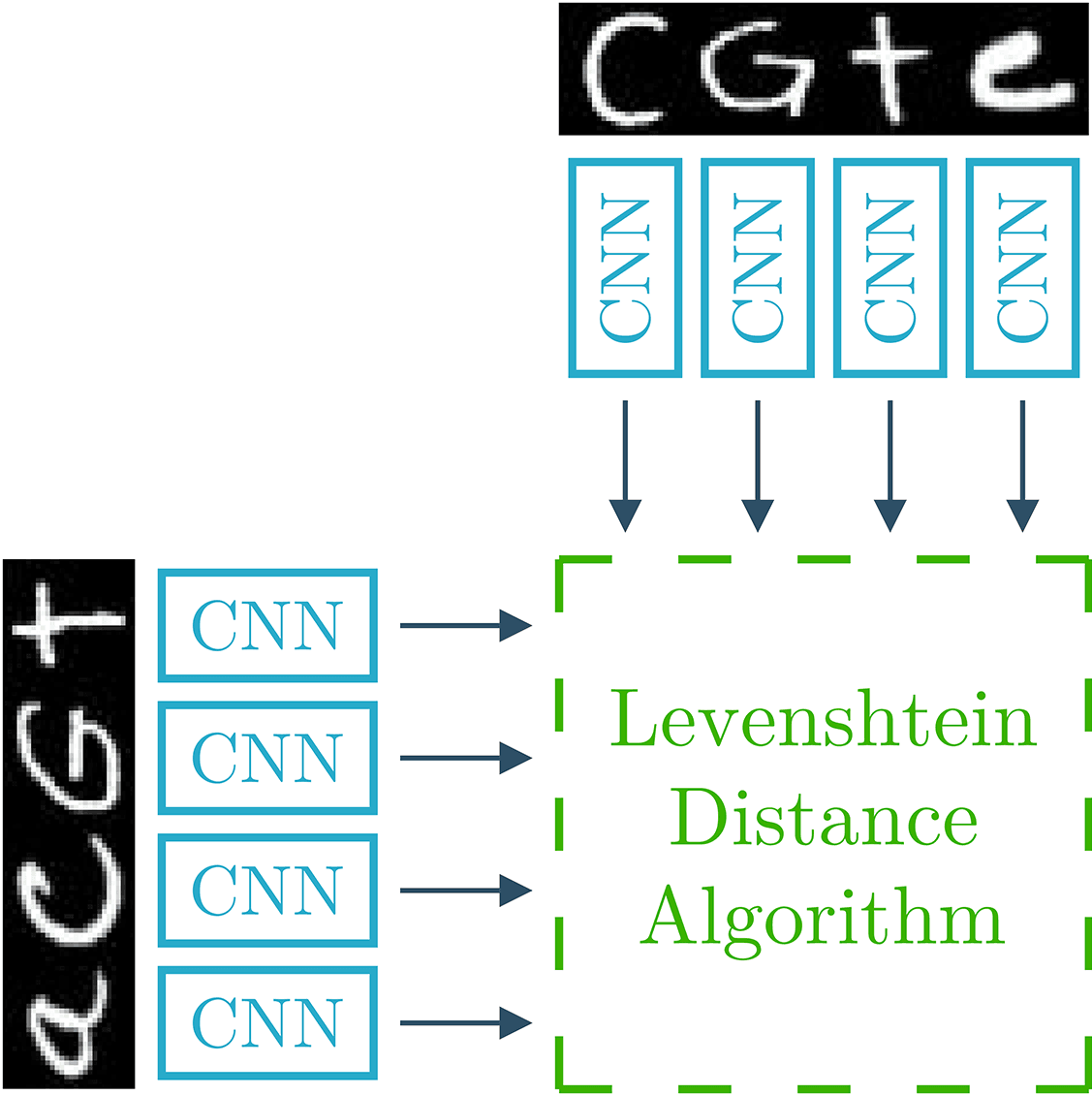}
    \caption{Illustration of Levenshtein Supervision}
    \label{fig:illustration-levenshtein}
\end{marginfigure}

The idea behind the Levenshtein distance supervision is that we want to learn an image classifier, but only the Levenshtein distance between pairs of strings of images is given, as illustrated in Figure~\ref{fig:illustration-levenshtein}.
Here, we predict a probability distribution over classes for each image using a CNN, feed them into the differentiable Levenshtein distance algorithm, and compare the predicted editing distance to the ground truth editing distance.
By propagating the error back through the algorithm, we can train the CNN.

In the experiment, we learn a classifier for handwritten EMNIST letters \cite{cohen2017emnist} and use pairs of handwritten character strings of length $32$.
We use the Levenshtein distance $\operatorname{LD}$ \cite{levenshtein1966binary} which is defined as $\operatorname{LD}(a,b) = \operatorname{LD}_{|a|,|b|}(a,b)$:
\begin{equation} \textstyle
\kern-1em\operatorname{LD}_{i,j}(a,b) =
\begin{cases}
  \quad j &   \kern-1emi=0\rule{2em}{0pt} \\
  \quad i &   \kern-1emj=0 \\
  \min \left\{\begin{array}{l@{}}
          \operatorname{LD}_{i-1,j}(a,b) + 1 \\
          \operatorname{LD}_{i,j-1}(a,b) + 1 \\
          \operatorname{LD}_{i-1,j-1}(a,b)+ \mathds{1}_{(a_{i-1} \neq b_{j-1})}
      \end{array}\right.
        & \kern-1em\hbox to0pt{\vbox to0pt{\vss\hbox{\raise 1.2ex\hbox{\textrm{else.}}}}}
\end{cases}
\end{equation}%

We relax the Levenshtein distance using its classic dynamic programming algorithm.
Here, distance matrix \texttt{d} of size $(|a|+1)\times (|b|+1)$ keeps track of the current Levenshtein distance when $i$ elements of $a$ and $j$ elements of $b$ have been processed.
That is, we iteratively fill matrix \texttt{d} with $\texttt{d}_{i,j}=\operatorname{LD}_{i,j}(a,b)$ such that we do not need to recompute intermediate values.
Since $\operatorname{LD}(a,b) = \texttt{d}_{|a|,|b|}$, we can read out the result from the matrix.
Using the \inlinelogo{algovision} library, we can implement it as follows:

\begin{widepar}
\begin{minted}[escapeinside=||]{python}
from algovision import (
    |\textcolor{modulecolor}{Algorithm}|, |\textcolor{modulecolor}{Input}|, |\textcolor{modulecolor}{Output}|, |\textcolor{modulecolor}{Var}|,                                        # core
    |\textcolor{modulecolor}{CosineSimilarity}|,                                               # conditions
    |\textcolor{modulecolor}{If}|, |\textcolor{modulecolor}{For}|,                                                # control_structures
    |\textcolor{modulecolor}{Let}|, |\textcolor{modulecolor}{Min}|,                                                        # functions
)
import torch
\end{minted}
\end{widepar}
\newpage
\begin{widepar}
\begin{minted}[escapeinside=||]{python}
levenshtein = |\textcolor{modulecolor}{Algorithm}|(
    # Define the variables the input corresponds to.
    |\textcolor{modulecolor}{Input}|('array_s'),
    |\textcolor{modulecolor}{Input}|('array_t'),
    # Declare and initialize all differentiable variables.
    |\textcolor{modulecolor}{Var}|('d', lambda array_s, array_t:
             torch.zeros(array_s.shape[1] + 1, array_t.shape[1] + 1)
    ),
    |\textcolor{modulecolor}{Var}|('subs_cost', torch.tensor(0.)),
    |\textcolor{modulecolor}{Var}|('return_d', torch.tensor(0.)),
    # Initialize the borders of the cost matrix.
    |\textcolor{modulecolor}{For}|('i', lambda d: d.shape[1]-1,
        |\textcolor{modulecolor}{Let}|('d', [lambda i: i+1, 0], lambda i: i+1)
    ),
    |\textcolor{modulecolor}{For}|('j', lambda d: d.shape[2]-1,
        |\textcolor{modulecolor}{Let}|('d', [0, lambda j: j+1], lambda j: j+1)
    ),
    # Run the dynamic programming algorithm.
    |\textcolor{modulecolor}{For}|('i', lambda d: d.shape[1]-1,
        |\textcolor{modulecolor}{For}|('j', lambda d: d.shape[2]-1,
            # Differentiable check for equality of two categorical embeddings.
            |\textcolor{modulecolor}{If}|(|\textcolor{modulecolor}{CosineSimilarity}|(lambda array_s, i: array_s[:, i],
                                lambda array_t, j: array_t[:, j]),
                if_true=Let('subs_cost', 0),
                if_false=Let('subs_cost', 1),
            ),
            # Correspond to 
            # `d[i+1, j+1] = min(d[i,j+1]+1, d[i+1,j]+1, d[i+1,j+1]+subs_cost`.
            |\textcolor{modulecolor}{Let}|('d', [lambda i: i+1, lambda j: j+1],
                lambda d, i, j, subs_cost: |\textcolor{modulecolor}{Min}|(beta=5)(d[:,i,j+1]+1,
                                                       d[:,i+1,j]+1,
                                                       d[:,i,j]+subs_cost)
            )
        ),
    ),
    # Return the distance and the cost matrix.
    |\textcolor{modulecolor}{Let}|('return_d', 'd', [lambda d: d.shape[1]-1, lambda d: d.shape[2]-1]),
    |\textcolor{modulecolor}{Output}|('return_d'),
    |\textcolor{modulecolor}{Output}|('d'),
    beta=5,
)
\end{minted}
\end{widepar}

An example Levenshtein distance matrix and its relaxation are displayed in Figure~\ref{fig:ld-plots}.

\begin{marginfigure}
    \centering
    \let\pgfimageWithoutPath\pgfimage 
    \renewcommand{\pgfimage}[2][]{\pgfimageWithoutPath[#1]{fig_algovision/#2}}
    \resizebox{1.\linewidth}{!}{
        \input{fig_algovision/levenshtein_dist_only_2.pgf}
    }
    \caption{
    Levenshtein distance matrix.
    \textit{Left:} hard matrix and alignment path.
    \textit{Right:} relaxed matrix with inverse temperature $\steepness=1.5$.
    }
    \label{fig:ld-plots}
\end{marginfigure}%

For learning, pairs of strings of images of $32$ handwritten characters $a,b$ as well as the ground truth Levenshtein distance $\operatorname{LD}(\mathbf{y}_a,\mathbf{y}_b)$ are given. %
We sample pairs of strings $a,b$ from an alphabet of $2$ or $4$ characters.
For sampling the second string given the first one, we uniformly choose between two and four insertion and deletion operations.
Thus, the average editing distance for strings, which use two different characters, is $4.25$ and for four characters is $5$.
We process each letter, using a CNN, which returns a categorical distribution over letters, which is then fed to the algorithm.
An example of a pair of strings based on $\{\text{A, C, G, T}\}$ is \\
\emnistimg{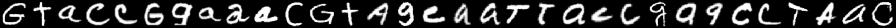} and\\
\emnistimg{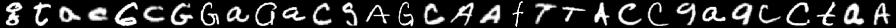}~.\\
Our training objective is minimizing the $\ell^2$ loss between the predicted distance and the ground truth distance:
\begin{equation}
    \mathcal{L} = \left\| 
    \operatorname{LD}\left(\left(\operatorname{CNN}(a_i)\right)_{i\in\{1..32\}},\left(\operatorname{CNN}(b_i)\right)_{i\in\{1..32\}}\right) \,-\, \operatorname{LD}(\mathbf{y}_a,\mathbf{y}_b) 
    \right\|_2.
\end{equation}
For training, we use an inverse temperature of $\steepness=9$ and Adam ($\eta =10^{-4}$) for $128-512$ iterations.
The CNN consists of two convolutional layers with a kernel size of $5$ and hidden sizes of $32$ and $64$, each followed by a ReLU, and a max-pooling layer. 
The convolutional layers are followed by two fully connected layers with a hidden size of $64$ and a ReLU activation.

\begin{table}[]
    \caption{
    EMNIST classification results with Levenshtein distance supervision averaged over 10 runs. %
    }
    \label{table:results-align}
    \begingroup
    \setlength{\tabcolsep}{3pt}
    \newcommand{\emnistLDFormatBestValue}[1]{\pmb{#1}}
    \centering
    \footnotesize
    \begin{tabular}{lc!{\color{black!40}\vrule}ccccc!{\color{black!40}\vrule}cc!{\color{black!40}\vrule}cc}
        \toprule
        Method & & \texttt{AB} & \texttt{BC} & \texttt{CD} & \texttt{DE} & \texttt{EF} & \texttt{IL} & \texttt{OX} & \texttt{ACGT} & \texttt{OSXL} \\
        \midrule
        \multirow{ 2}{*}{Baseline}  & Top-1 acc.      & $.616$ & $.651$ & $.768$ & $.739$ & $.701$ & $.550$ & $.893$ & $.403$ & $.448$  \\
                                    & F1-score       & $.581$ & $.629$ & $.759$ & $.711$ & $.674$ & $.490$ & $.890$ & $.336$ & $.384$  \\
        \midrule
        \multirow{ 2}{*}{Relaxed LD}   & Top-1 acc.   & $\emnistLDFormatBestValue{.671}$ & $\emnistLDFormatBestValue{.807}$ & $\emnistLDFormatBestValue{.816}$ & $\emnistLDFormatBestValue{.833}$ & $\emnistLDFormatBestValue{.847}$ & $\emnistLDFormatBestValue{.570}$ & $\emnistLDFormatBestValue{.960}$ & $\emnistLDFormatBestValue{.437}$ & $\emnistLDFormatBestValue{.487}$  \\
                                       & F1-score    & $\emnistLDFormatBestValue{.666}$ & $\emnistLDFormatBestValue{.805}$ & $\emnistLDFormatBestValue{.815}$ & $\emnistLDFormatBestValue{.831}$ & $\emnistLDFormatBestValue{.845}$ & $\emnistLDFormatBestValue{.539}$ & $\emnistLDFormatBestValue{.960}$ & $\emnistLDFormatBestValue{.367}$ & $\emnistLDFormatBestValue{.404}$  \\
        \bottomrule
    \end{tabular}
    \endgroup
\end{table}

For evaluation, we use Hungarian matching with Top-1 accuracy as well as the F1-score.
We compare it against a baseline, which uses the $\ell_1$ distance between encodings instead of the editing distance:
\begin{equation}
    \mathcal{L} = \left\| 
    \left\| \left(\operatorname{CNN}(a_i)\right)_{i\in\{1..32\}} - \left(\operatorname{CNN}(b_i)\right)_{i\in\{1..32\}} \right\|_1 \,-\, \operatorname{LD}(\mathbf{y}_a,\mathbf{y}_b) 
    \right\|_2.
\end{equation}
Table~\ref{table:results-align} shows that our method consistently outperforms the baseline on both metrics in all cases.
The character combinations \texttt{AB}, \texttt{BC}, \texttt{CD}, \texttt{DE}, and \texttt{EF} are a canonical choice for random combinations.
The characters \texttt{IL} are the hardest combination of two letters as they even get frequently confused by supervised neural networks \cite{cohen2017emnist} and can also be indistinguishable for humans.
The characters \texttt{OX} represent the easiest case as supervised classifiers can perfectly distinguish them on the test dataset \cite{cohen2017emnist}.
For two letter combinations, we achieve Top-1 accuracies between $57\%$ (\texttt{IL}) and $96\%$ (\texttt{OX}).
Even for four letter combinations (\texttt{ACGT} and \texttt{OSXL}), we achieve Top-1 accuracies of up to $48.7\%$.
Note that, as we use strings of length $32$, in the Levenshtein algorithm, more than $1\,000$ statements are relaxed. %

\subsection*{Conclusion}

In this chapter, we covered and proposed general approaches for continuous relaxations of algorithms that allow their integration into end-to-end trainable neural network architectures.
For that, we use convex combinations of execution paths of algorithms that are parameterized by smooth functions.
We found that the proposed general framework can compete with continuous relaxations of specific algorithms as well as gradient estimation methods on a variety of algorithmic supervision tasks.
While this chapter covered a general approach, in the following chapters, we focus on relaxing specific algorithms.

\setchapterpreamble[u]{\pagelogo{diffsort}\margintoc}
\chapter[Differentiable Sorting and Ranking]{Differentiable\\ Sorting and Ranking}
\labch{diffsort}

After presenting general methods for relaxing algorithms, in this chapter, we focus on differentiable sorting and ranking, which allows us to explore it in much greater detail.   

Sorting and ranking, i.e., the ability to score elements by their relevance, is essential in various applications.
It can be used for choosing the best results to display by a search engine or organizing data in memory, among many others.
As sorting a sequence of values requires finding the respective ranking order, we use the terms ``sorting'' and ``ranking'' interchangeably.

Recently, the idea of end-to-end training of neural networks with sorting and ranking supervision by a continuous relaxation of the sorting and ranking functions has been presented by Grover~\etal~\cite{grover2019neuralsort}.
The idea of ordering supervision is that the ground truth order of some samples is known while their absolute values remain unsupervised. 
This is achieved by integrating a sorting algorithm in the neural architecture.
As for training with a sorting algorithm in the architecture, the error needs to be propagated in a meaningful way back to the neural network, it is necessary to use a differentiable sorting function.
Several such differentiable relaxations of the sorting and ranking functions have been introduced, e.g., by Adams~\etal~\cite{adams2011ranking}, Grover~\etal{}~\cite{grover2019neuralsort}, Cuturi~\etal{}~\cite{cuturi2019differentiable}, and Blondel~\etal{}~\cite{blondel2020fast}.
These methods enable training a neural network based on ordering and ranking information instead of absolute ground truth values.

Starting in the 1950s, sorting networks have been presented to address the sorting task~\cite{knuth1998sorting}.
Sorting networks are sorting algorithms with a fixed execution structure, which makes them suitable for hardware implementations, e.g., as part of circuit designs.
They are oblivious to the input, i.e., their execution structure is independent of the data to be sorted.
They allow for fast hardware-implementation, e.g., in application-specific integrated circuits (ASICs), as well as on highly parallelized general-purpose hardware like GPUs.
As such hardware implementations are significantly faster than conventional multipurpose hardware, they are of interest for sorting in high-performance computing applications~\cite{govindaraju2006gputerasort}.
This motivated the optimization of sorting networks toward faster networks with fewer layers, which is a still-standing problem \cite{bidlo2019evolutionary}.
Note that, although the name is similar, sorting networks are \emph{not} neural networks that sort.

In this chapter, we propose to combine traditional sorting networks and differentiable sorting functions by presenting smooth differentiable sorting networks.
Later in this chapter, we focus on analyzing differentiable sorting functions \cite{grover2019neuralsort,cuturi2019differentiable,blondel2020fast} and demonstrate how monotonicity improves differentiable sorting networks.

\begin{figure*}[t]
    \centering
    \includegraphics[width=\linewidth]{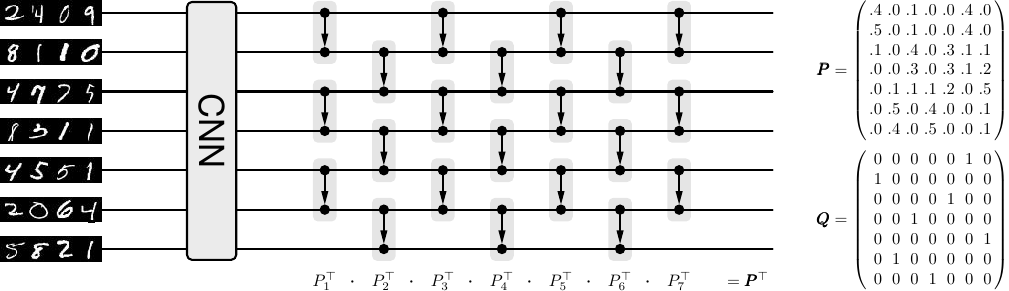}
    \caption{
    Overview of the system for training with sorting supervision.
    \textit{Left:} input images are fed separately / independently into a Convolutional Neural Network (CNN) that maps them to scalar values.
    \textit{Center:} the odd-even sorting network sorts the scalars by parallel conditional swap operations.
    \textit{Right:} the sorting network produces a differentiable permutation matrix $\pmb{P}$ which can then be compared to the ground truth permutation matrix $\pmb{Q}$ using binary cross-entropy to produce the training loss.
    By propagating this error backward through the sorting network, we can train the CNN.
    }
    \label{fig:overall-architecture-odd-even}
\end{figure*}

Sorting networks are a family of sorting algorithms that consist of two basic components: so-called ``wires'' (or ``lanes'') carrying values and conditional swap operations that connect pairs of wires~\cite{knuth1998sorting}.
An example of such a sorting network is shown in the center of Figure~\ref{fig:overall-architecture-odd-even}.
The conditional swap operations swap the values carried by these wires if they are not in the desired order.
As sorting networks are data-oblivious, i.e., the program flow of the algorithm is independent of the input data, they are especially suitable for continuous relaxation. 
Sorting networks are conventionally non-differentiable as they use $\min$ and $\max$ operators for conditionally swapping elements.
Thus, we relax these operators by perturbation with a probability distribution, e.g., with the logistic distribution.

One problem that arises in this context is that using a logistic sigmoid function does not preserve the monotonicity of the relaxed sorting operation, which can cause gradients with the wrong sign.
In Section~\ref{sec:monotonic-diffsort}, we present a family of sigmoid functions that preserve the monotonicity of differentiable sorting networks.
These include the CDF of the Cauchy distribution, as well as a function that minimizes the error-bound and thus induces the smallest possible approximation error.
For all sigmoid functions, we prove and visualize the respective properties and validate their advantages empirically.
In fact, by making the sorting function monotonic, it also becomes quasiconvex, which has been shown to produce favorable convergence rates \cite{kiwiel2001convergence}.
In Figure~\ref{fig:softminx0}, we demonstrate monotonicity for different choices of sigmoid functions.
As can be seen in Figure~\ref{fig:softminx0-others}, existing differentiable sorting operators are either non-monotonic or have an unbounded error.
We show that sigmoid functions with specific characteristics produce monotonic and error-bounded differentiable sorting networks.
We provide theoretical guarantees for these functions and also give the monotonic function that minimizes the approximation error, and demonstrate that the proposed functions improve empirical performance.

To validate the proposed idea and to show its generalization, we evaluate it for two sorting network architectures, the odd-even as well as the bitonic sorting network.
The idea of odd-even sort is to iteratively compare adjacent elements and swap pairs that are in the wrong order.
The method alternately compares all elements at odd and even indices with their successors.
To make sure that the smallest (or greatest) element will be propagated to its final position for any possible input of length $n$, we need $n$ exchange layers.
An odd-even network is displayed in Figure~\ref{fig:overall-architecture-odd-even} (center).
Odd-even networks can be seen as the most generic architectures, and are mainly suitable for small input sets as their number of layers directly depends on the number of elements to be sorted.

Bitonic sorting networks \cite{batcher1968sorting} use bitonic sequences to sort based on the Divide-and-Conquer principle and allow sorting in only $\mathcal{O}({\log^2n})$ parallel time.
Bitonic sequences are twice monotonic sequences, i.e., they consist of a monotonically increasing and a monotonically decreasing sequence.
Bitonic sorting networks recursively combine pairs of monotonic sequences into bitonic sequences and then merge them into single monotonic sequences.
Starting at single elements, they eventually end up with one sorted monotonic sequence.
With the bitonic architecture, we can sort large numbers of input values as we only need ${\log_2n \cdot ((\log_2n)+1)}/{2}$ layers to sort $n$ inputs.
As a consequence, the proposed architecture provides good accuracy even for large input sets and allows scaling up sorting and ranking supervision to large input sets of up to $1024$ elements.

Following the experiments in the previous chapter as well as Grover~\etal~\cite{grover2019neuralsort} and Cuturi~\etal~\cite{cuturi2019differentiable}, we benchmark our continuous relaxation of the sorting function on the four-digit MNIST \cite{lecun2010mnist} sorting supervision benchmark.
To evaluate the performance in the context of a real-world application, we apply our continuous relaxation to the multi-digit images of the Street View House Number (SVHN) data set.
We compare the performance of both sorting network architectures and evaluate their characteristics under different conditions.
We show that both differentiable sorting network architectures outperform existing continuous relaxations of the sorting function on the four-digit MNIST sorting benchmark and also perform well on the more realistic SVHN benchmark.
Further, we show that our model scales and achieves performance gains on larger sets of ordered elements and confirm this up to $n=1024$ elements.
An overview of the overall architecture is shown in Figure~\ref{fig:overall-architecture-odd-even}.

In the next chapter, we apply differentiable sorting and ranking to top-$k$ classification learning.

\section{Related Work}

In this section, we discuss related work wrt.~sorting and ranking.
We begin by discussing differentiable sorting and ranking approaches.
We continue by discussing learning-to-rank, an application in the domain of recommender systems.
After that, we cover neural networks that sort.
Finally, we discuss sorting networks, which are the basis for the differentiable sorting networks that we propose in this chapter.

\subsection{SoftSort and NeuralSort}

A (hard) permutation matrix is a square matrix with entries $0$ and $1$ such that every row and every column sums up to $1$, which defines the permutation necessary to sort a sequence.
To make the sorting operation differentiable, Grover~\textit{et al.}~\cite{grover2019neuralsort} proposed NeuralSort, which relaxes permutation matrices to unimodal row-stochastic matrices.
For this, they use the softmax of pairwise differences of (cumulative) sums of the top elements.
This relaxation allows for gradient-based stochastic optimization.
They prove that this, for the temperature parameter approaching $0$, is the correct permutation matrix, and propose multiple deep learning differentiable sorting benchmark tasks.
Note that NeuralSort is not based on sorting networks.
On various tasks, including the four-digit MNIST sorting benchmark, they evaluate their relaxation against the Sinkhorn and Gumbel-Sinkhorn approaches proposed by Mena~\textit{et al.}~\cite{mena2018learning}.

Prillo~\textit{et al.}~\cite{prillo2020softsort} build on this idea but simplify the formulation and provide SoftSort, a faster alternative to NeuralSort.
They show that it is sufficient to build on pairwise differences of elements of the vectors to be sorted instead of the cumulative sums.
Specifically, given a set of $n$ inputs, i.e., $y\in\mathbb{R}^n$, SoftSort is defined as
\begin{align}
    P(y) = \operatorname{softmax} \left(- \left|y^\top \ominus \operatorname{sort} (y)\right| / \tau\right) = \operatorname{softmax} \left(- \left|y^\top \ominus S y\right| / \tau\right)
\end{align} 
where $\tau$ is a temperature parameter, ``$\operatorname{sort}$'' sorts the entries of a vector in non-ascending order, $\ominus$ is the element-wise broadcasting subtraction, $|\cdot|$ is the element-wise absolute value, and ``$\operatorname{softmax}$'' is the row-wise softmax operator.
NeuralSort is defined similarly and omitted for the sake of brevity.
In the limit of $\tau\to0$, SoftSort and NeuralSort converge to the exact ranking permutation matrix~\cite{grover2019neuralsort, prillo2020softsort}. 
Prillo~\textit{et al.}~\cite{prillo2020softsort} find that SoftSort achieves approximately the same accuracy as NeuralSort, while being faster to compute.

\subsection{Optimal Transport / Sinkhorn Sort}
Cuturi~\textit{et al.}~\cite{cuturi2019differentiable} propose an entropy regularized optimal transport (OT) formulation of the sorting operation.
They use the idea that sorting can be achieved by minimizing the matching cost between elements and an auxiliary target of increasing values.
That is, the smallest element is matched to the first value, the second smallest to the second value, etc.
They make this differentiable by regularizing the OT problem with an entropic penalty, solve it by applying Sinkhorn iterations~\cite{cuturi2013sinkhorn} %
and compute the gradients via automatic differentiation rather than the implicit function theorem, which resolves the need for solving a linear equation system.
The Sinkhorn algorithm produces a transport matrix, which can be used as a relaxed permutation matrix.

\subsection{FastSort}
Blondel~\textit{et al.}~\cite{blondel2020fast} propose a differentiable sorting and ranking operator by casting sorting and ranking as linear programs over the permutahedron and regularizing them, which turns them into projections onto the permutahedron.
They solve this by reducing it to isotonic optimization and make it differentiable by considering the Jacobians of the isotonic optimization and the projection.
While FastSort is asymptotically (to date) the fastest differentiable sorting operator with a complexity of $\mathcal{O}(n \log n)$, it does not produce a differentiable permutation matrix.
However, in sorting and ranking supervision, we are typically primarily interested in the differentiable permutation matrix, which makes FastSort rather unsuitable for the applications explored in this chapter.
Accordingly, they evaluate the proposed approach in the context of top-$k$ classification, and label ranking accuracy via a soft Spearman's rank correlation coefficient.

\subsection{Learning-to-Rank}
RankNet was one of the first approaches to use a neural network for learning to rank \cite{burges2005ranknet}.
RankNet uses ordered pairs of elements and optimizes a neural network to predict their relevance scores.
This was followed shortly after by LambdaRank \cite{burges2007lambdarank} which incorporates the influence of swapping the considered elements on an information retrieval metric.
In 2008, SoftRank \cite{taylor2008softrank} was introduced, which uses pairwise distances between the scores of elements to elevate this method to supervising the order of arbitrary many elements.
Adams~\textit{et al.}~\cite{adams2011ranking} propose relaxing permutation matrices to doubly-stochastic matrices based on marginals of distributions over permutation matrices and apply their method to the LETOR learning-to-rank benchmark~\cite{liu2011learning}.
We note that these works on neural-network based information retrieval typically used two-layer or similar fully connected networks, as such shallow networks were common at the time of publishing.
Recently, Lee~\etal~\cite{lee2021differentiable} propose differentiable ranking metrics, and Swezey~\etal~\cite{swezey2021pirank} propose PiRank, a learning-to-rank method which is based on the differentiable sorting method NeuralSort.

\subsection{Neural Networks that Sort}
In the past, neural networks that sort have been proposed, e.g., by Ceterchi~\etal~\cite{ceterchi2008spiking}, who proposed simulating sorting networks with spiking neural P systems.
Spiking neural P systems are predecessors of current spiking networks, a form of computational models inspired by biological neurons.
This was later adapted by Metta~\etal~\cite{metta2015sorting} for a spiking neural P system with anti-spikes and rules on synapses.

Graves~\etal~\cite{graves2014neural} raised the idea of integrating sorting capabilities into neural networks in the context of Neural Turing Machines (NTM).
The NTM architecture contains two basic components: a neural network controller based on a long short-term memory network (LSTM) and a memory bank with an attention mechanism, both of which are differentiable.
The authors use this architecture to sort sequences of binary vectors according to given priorities.
Vinyals~\etal~\cite{vinyals2016order} address the problem of the order of input and output elements in LSTM sequence-to-sequence models by content-based attention.
To show the effect of the proposed model, they apply it to the task of sorting numbers and formulate the task of sorting as an instance of the set2seq problem.
Mena~\etal~\cite{mena2018learning} introduce the Gumbel-Sinkhorn, a Sinkhorn-operator--based analog of the Gumbel-Softmax distribution for permutations.
They evaluate the proposed approach, i.a., on the task of sorting up to $120$ numbers.
Note that these architectures learn to sort, while sorting networks and differentiable sorting functions sort provably correct.
These methods allow sorting input values, as an alternative to classical sorting algorithms, but not training with sorting supervision because they are not meaningfully differentiable.

\subsection{Sorting Networks} 
The goal of research on sorting networks is to find optimal sorting networks, i.e., networks that can sort an input of $n$ elements in as few layers of parallel swap operations as possible.
Initial attempts to sorting networks required $\mathcal{O}(n)$ layers, each of which requires $\mathcal{O}(n)$ operations (examples are bubble and insertion sort \cite{knuth1998sorting}).
With parallel hardware, these sorting algorithms can be executed in $\mathcal{O}(n)$ time.
Further research led to the discovery of the bitonic sorting network (aka.~bitonic sorter), which requires only $\mathcal{O}(\log^2n)$ layers~\cite{knuth1998sorting, batcher1968sorting}.
Using genetic and evolutionary algorithms, slightly better optimal sorting networks were found for specific $n$ \cite{bidlo2019evolutionary, baddar2012designing}.
However, these networks do not exhibit a simple, regular structure.
Ajtai, Koml\'os, and Szemer\'edi \cite{ajtai1983sorting} presented the AKS sorting network which can sort in $\mathcal{O}(\log n)$ parallel time, i.e., using only $\mathcal{O}(n\log n)$ operations.
However, the complexity constants for the AKS algorithm are to date unknown and optimistic approximations assume that it is faster than bitonic sort if and only if $n \gg 10^{80}$.
Today, sorting networks are still in use, e.g., for fast sorting implementations on GPU accelerated hardware as described by Govindaraju~\etal~\cite{govindaraju2006gputerasort} and in hybrid systems as described by Gowanlock~\etal~\cite{gowanlock2019hybrid}.

\section{Sorting Networks}

In this section, we introduce two common sorting networks: the simple odd-even sorting network as well as the more complex but also more efficient bitonic sorting network.

\subsection{Odd-Even Sorting Network}

One of the simplest sorting networks is the odd-even sorting network~\cite{habermann1972parallel}.
Here, neighboring elements are swapped if they are in the wrong order.
As the name implies, this is done in a fashion alternating between comparing odd and even indexed elements with their successors.
In detail, for sorting an input sequence $a_1a_2...a_n$, each layer updates the elements such that $a'_i = \min(a_i, a_{i+1})$ and $a'_{i+1} = \max(a_i, a_{i+1})$ for all odd or even indices $i$, respectively.
Using $n$ of such layers, a sequence of $n$ elements is sorted as displayed in Figure~\ref{fig:overall-architecture-odd-even}~(center).

\begin{figure*}[t]
\centering
\includegraphics[height=41mm]{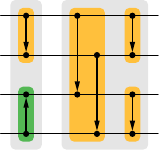}\hfill
\includegraphics[height=41mm]{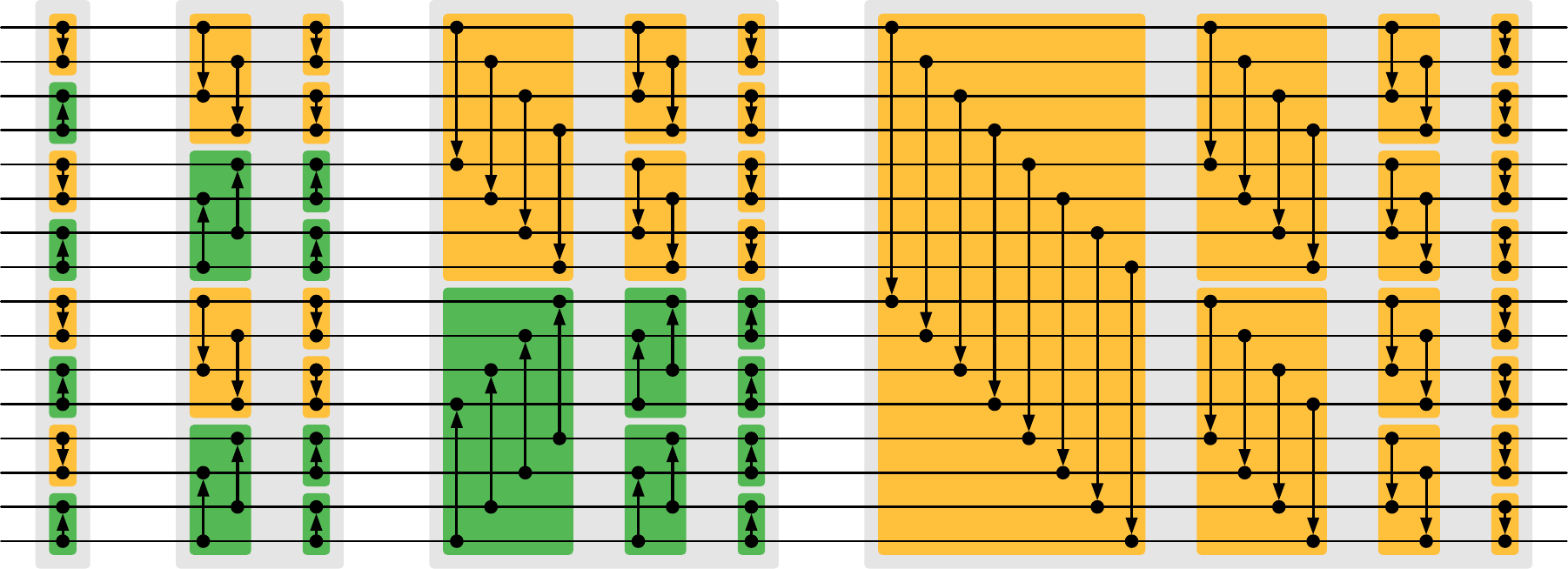}
\caption{
    Bitonic sorting networks for 4 and 16 lanes, consisting of bitonic merge blocks (colored). Arrows pointing toward the maximum.
    \label{fig:bitonic_sort} \label{fig:bitonic-schema}
    }
\end{figure*}

\subsection{Bitonic Sorting Network}

Second, we review the bitonic sorting network for sorting $n=2^k$ elements where $k\in\mathbb{N}_+$.
The sorting network can be extended to $n\in \mathbb{N}_+$~\cite{knuth1998sorting}, which is also supported in the \inlinelogo{diffsort} library. 
The bitonic sorting network builds on bitonic sequences: a sequence $(a_i)_{1\leq i < n}$ is called bitonic if (after an appropriate circular shift) $a_1\leq...\leq a_j \geq...\geq a_{n}$ for some $j$. 

Following the Divide-and-Conquer principle, in analogy to merge sort, bitonic sort recursively splits the task of sorting a sequence into the tasks of sorting two subsequences of equal length, which are then combined into a bitonic sequence.
Like merge sort, bitonic sort starts by merging individual elements, to obtain sorted lists of length~$2$ (first gray block in Figure~\ref{fig:bitonic_sort}).
Pairs of these are then combined into bitonic sequences and then merged into monotonic sequences (second gray block in Figure~\ref{fig:bitonic_sort}). %
This proceeds, doubling the length of the sorted sequences with each (gray) block, until the entire sequence is sorted.
The difference to merge sort lies in the bitonic merge operation, which merges two sequences sorted in opposite order (i.e., a single bitonic sequence) into a single sorted (monotonic) sequence.

\paragraph{Proof Sketch} In the following, we detail the bitonic sorting network and sketch a proof of why the bitonic sorting networks sorts:

A bitonic sequence is sorted by several bitonic merge blocks, shown in orange and green in Figure~\ref{fig:bitonic_sort}.
Each block takes a bitonic input sequence $a_1a_2\ldots a_{2m}$ of length~$2m$ and
turns it into two bitonic output sequences $\ell_1\ell_2\ldots\ell_m$ and $u_1u_2\ldots u_m$ of length~$m$ that satisfy
$\max_{i=1}^m l_i \le \min_{i=1}^m u_i$. 
These subsequences are recursively processed by bitonic merge blocks, until the output sequences are of length~$1$.
At this point, the initial bitonic sequence has been turned into a monotonic sequence due to the minimum/maximum conditions that hold between the output sequences (and thus elements).

A bitonic merge block computes its output as
$\ell_i = \min(a_i,a_{m+i})$ and $u_i = \max(a_i,a_{m+i})$.
This is depicted in Figure~\ref{fig:bitonic-schema} by the arrows pointing from the minimum to the maximum.
To demonstrate that bitonic merge works, we show that this
operation indeed produces two bitonic output sequences for which the relationship
$\max_{i=1}^m l_i \le \min_{i=1}^m u_i$ holds. 

\begin{figure*}[t]
\newdimen\myunit\myunit1.15mm
\centering
\includegraphics[height=3.5cm]{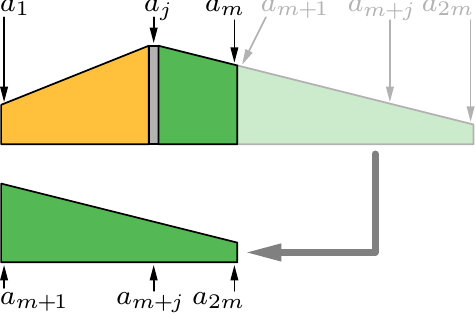}\hfill
\includegraphics[height=3.5cm]{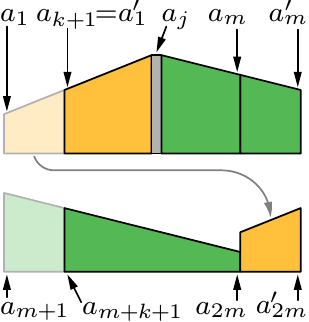}\hfill
\includegraphics[height=3.5cm]{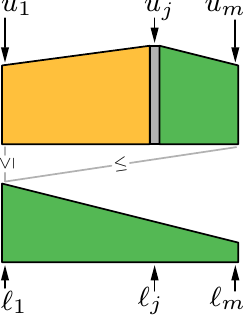}\hskip4mm
\includegraphics[height=3.5cm]{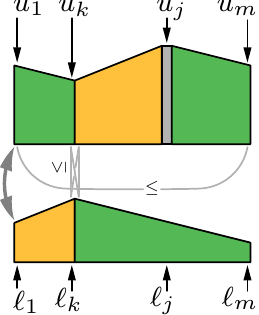} \\
\hbox to\textwidth{\footnotesize
\hbox to48.23\myunit{\hss (a)\hss}\hfill
\hbox to38.00\myunit{\hss (b)\hss}\hfill
\hbox to27.00\myunit{\hss (c)\hss}\hskip4mm
\hbox to24.00\myunit{\hss (d)\hss}}
\caption{Bitonic merge turns a bitonic input sequence into two
         bitonic output sequences, with all elements in the
         one (upper, $u_i$) sequence larger than all elements
         in the other (lower, $\ell_i$) sequence.
         The diagrams show the vertical alignment of elements to compare (a) and the 
         invariance to cyclic permutations (b).
         Depending on the values in (a), no exchanges (c) or exchanges (d) are executed.%
         \label{fig:bitonic_merge}}
\end{figure*}

Note that neither a cyclic permutation of the sequence (Figure~\ref{fig:bitonic_merge}~(b), $a_i' = a_{(i+k-1 \bmod 2m)+1}$ for some $k$), nor
a reversal, change the bitonic character of the sequence.
As can be seen in Figure~\ref{fig:bitonic_merge}~(b), even under cyclic permutation, still the same pairs of elements are considered for a potential swap. 
Thus, as a cyclic permutation or a reversal only causes the output sequences to be analogously cyclically permuted or reversed, this changes neither the bitonic character of these sequences nor the relationship between them. 
Therefore, it suffices to consider the special case shown
in Figure~\ref{fig:bitonic_merge}a, with a monotonically increasing sequence (orange)
followed by a monotonically decreasing sequence (green) and the maximum
element~$a_j$ (gray) in the first half. Note that in this case
$\forall i; j \le i \le m: a_i \ge a_{m+i}
\wedge u_i = a_i \wedge \ell_i = a_{m+i}$.

For this case, we distinguish two subcases: 
$a_1 \ge a_{m+1}$ and $a_1 < a_{m+1}$.

If, on one hand, $a_1 \ge a_{m+1}$, we have the situation shown in Figure \ref{fig:bitonic_merge}~(c): the output sequence $u_1u_2\ldots u_m$ is simply the first
half of the sequence, the output sequence $\ell_1\ell_2\ldots\ell_m$
is the second half. 
Thus, both output sequences are bitonic (since
they are subsequences of a bitonic input sequence) and
$\min_{i=1}^m u_i = \min(u_1,u_m) \ge \ell_1 = \max_{i=1}^m \ell_i$.

If, on the other hand, $a_1 < a_{m+1}$, we can infer
$\exists k; 1 \le k < j: a_k > a_{m+k} \wedge a_{k+1} \le a_{m+k+1}$.
This situation is depicted in Figure~\ref{fig:bitonic_merge}~(d). 
Thus,
$\forall i; 1 \le i \le k: u_i = a_{m+i} \wedge \ell_i = a_i$ and
$\forall i; k {\,<\,} i {\,\le\,} m: u_i {\,=\,} a_i \wedge \ell_i {\,=\,} a_{m+i}$. Since
$u_k = a_{m+k} > a_k = \ell_k$,
$u_k = a_{m+k} \ge a_{m+k+1} = \ell_{k+1}$,
$u_{k+1} \kern-0.8pt=\kern-0.8pt a_{k+1} \ge a_{m+k+1} \kern-0.8pt=\kern-0.8pt \ell_{k+1}$,
$u_{k+1} \kern-0.8pt=\kern-0.8pt a_{k+1} \ge a_k \kern-0.8pt=\kern-0.8pt \ell_k$, we obtain
$\max_{i=1}^m l_i \le \min_{i=1}^m u_i$.
Figure~\ref{fig:bitonic_merge}~(d) shows that the two output sequences are bitonic and that all elements of the upper output sequence are greater than or equal to all elements of the lower output sequence.

\section{Differentiable Sorting Networks}

To relax sorting networks, we need to relax the $\min$ and $\max$ operators, which are used as a basis for the swap operations in sorting networks.
By perturbing the difference between the two inputs with a distribution, $\min$ and $\max$ can be relaxed to differentiable $\softmin$ and $\softmax$.
Note that we denote the differentiable relaxations in $\mathit{italic}$ font and their hard counterparts in $\mathrm{roman}$ font.
The differentiable relaxations $\softmin$ and $\softmax$ differ from the commonly used $\mathrm{softmin}$ and $\mathrm{softmax}$, which are relaxations of $\mathrm{argmin}$ and $\mathrm{argmax}$~\cite{goodfellow2016deep}.

One example of such a relaxation of $\min$ and $\max$ is the logistic relaxation
\begin{align}
    \softmin_\sigma(a, b) &= a \cdot \sigmoid(b-a) + b \cdot \sigmoid(a-b),\\
    \softmax_\sigma(a, b) &= a \cdot \sigmoid(a-b) + b \cdot \sigmoid(b-a)
\end{align}
where $\sigma$ is the logistic sigmoid function with inverse temperature $\beta>0$:
\begin{equation}
    \sigma : x\mapsto \frac{1}{1+e^{-\beta x}}
    \mbox{\hbox to0pt{$\displaystyle
    \phantom{\frac{1}{{|x|}^\ARTstrength}}$\hss}}.
    \label{eq:logistic-background}
\end{equation}
In the limit, for $\steepness\to\infty$, $\sigma$ converges to the Heaviside function and the smooth operators converge to the discrete operators.
We formalize the notion of a sigmoid function and generalize it beyond the example of the logistic distribution.\\

\begin{definition}[Sigmoid Function]
We define a (unipolar) sigmoid (i.e., s-shaped) function as a function~$f$ that is continuous, monotonically non-decreasing, and odd-symmetric (around $\frac{1}{2}$) with
\[ f: \IR \to [0,1] \quad \mbox{with} \quad
   \lim_{x \to -\infty} f(x) = 0  \quad \mbox{and} \quad
   \lim_{x \to  \infty} f(x) = 1. \]
\end{definition}

Based on a sigmoid function $f$, we can define a continuous conditional swap as follows.\\

\begin{definition}[Continuous Conditional Swaps]
    Following~\cite{petersen2021diffsort}, we define a continuous conditional swap in terms of a sigmoid function $f:\mathbb{R}\to [0,1]$ as
    \vspace{-1em}
    \begin{align}
        \softmin_f(a, b) &= a \cdot f(b-a) + b \cdot f(a-b)\ \\
        \softmax_f(a, b) &= a \cdot f(a-b) + b \cdot f(b-a)\ \\
        \softargmin_f(a, b) &= \left(\ f(b-a),\quad f(a-b)\ \right) \\
        \softargmax_f(a, b) &= \left(\ f(a-b),\quad f(b-a)\ \right) .
    \end{align}
\end{definition}
We require a continuous odd-symmetric sigmoid function to preserve most of the properties of $\min$ and $\max$, while also making $\softargmin$ and $\softargmax$ continuous as discussed in Supplementary Material~\ref{sec:properties-of-min-and-max}.

\subsection{Differentiable Permutation Matrices}

For sorting and ranking supervision, i.e., training a neural network to predict scalars, where only the order of these scalars is known, we use the ground truth permutation matrix as supervision.
Thus, to train an underlying neural network end-to-end through the differentiable sorting network, we need to return the underlying permutation matrix rather than the actual sorted scalar values.
For that, we compute the permutation matrices for the swap operations for each layer as shown in Figure~\ref{fig:overall-architecture-odd-even}.
Here, for all swap operations between any elements $a_i$ and $a_j$ that are to be ordered in non-descending order, the layer-wise permutation matrix is
\begin{align}
    P_{l,ii} = P_{l,jj} &= \phantom{1-{}}\alpha_{ij} = \phantom{1-{}} f(a_j - a_i), \\
    P_{l, ij}= P_{l,ji} &= 1-\alpha_{ij} = 1 - f(a_j - a_i)\,
\end{align}
where all other entries of $P_l$ are set to $0$.
By multiplication, we compute the complete relaxed permutation matrix $\mathbf{P}$ as
\begin{equation}
    \mathbf{P} = P_{n} \cdot ...\cdot P_2\cdot P_1 = \bigg(\prod_{l=1}^{n} P_l^\top\bigg)^{\!\top}\,.
    \label{eq:p-mult}
\end{equation}
A column in the relaxed permutation matrix can be seen as a distribution over possible ranks for the corresponding input value.
Multiplying $\mathbf{P}$ with an input $x$ yields the differentiably sorted vector $\hat x = \mathbf{P}x$, which is also the output of the differentiable sorting network.
Note that computing $\mathbf{P}$ is optional, as we can compute $\hat x$ faster without computing it via $\mathbf{P}$ because $\hat x$ is just the output of the differentiable sorting network.
Whether it is necessary to compute $\mathbf{P}$, or whether $\hat x$ suffices, depends on the specific application.
For example, for a cross-entropy ranking / sorting loss as used in the experiments in Section~\ref{sec:ranking-supervision-experiments}, $\mathbf{P}$ can be used to compute the cross-entropy to a ground truth permutation matrix $\mathbf{Q}$ as
\begin{equation}
    \mathcal{L} := \sum_{c=1}^n \left( \frac{1}{n} \operatorname{CE}\left(\mathbf{P}_c, \pmb{Q}_c\right) \right)
    \label{eq:loss}
\end{equation}
where $\mathbf{P}_c$ and $\mathbf{Q}_c$ denote the $c$th columns of $\mathbf{P}$ and $\mathbf{Q}$, respectively.
Note that, as the cross-entropy loss is, by definition, computed element-wise, the column-wise cross-entropy is equivalent to the row-wise cross-entropy.

In the following, we establish doubly-stochasticity and differentiability of $\mathbf{P}$, which are important properties for differentiable sorting and ranking operators.

\begin{lemma}[Doubly-Stochasticity and Differentiability of $\mathbf{P}$]
    (i) The relaxed permutation matrix $\mathbf{P}$, produced by a differentiable sorting network, is doubly-stochastic.
    (ii) $\mathbf{P}$ has the same differentiability as $f$, e.g., if $f$ is continuously differentiable in the input, $\mathbf{P}$ will be continuously differentiable in the input to the sorting network. If $f$ is differentiable almost everywhere (a.e.), $\mathbf{P}$ will be differentiable a.e.
    \begin{proof}
    (i) For each conditional swap between two elements $i,j$, the relaxed permutation matrix is $1$ at the diagonal except for rows $i$ and $j$: at points $i,i$ and $j,j$ the value is~$v\in [0, 1]$, at points $i,j$ and $j,i$ the value is $1-v$ and all other entries are $0$. This is doubly-stochastic as all rows and columns add up to $1$ by construction.
    As the product of doubly-stochastic matrices is doubly-stochastic, the relaxed permutation matrix $\mathbf{P}$, produced by a differentiable sorting network, is doubly-stochastic.

    (ii) The composition of differentiable functions is differentiable and the addition and multiplication of differentiable functions is also differentiable. Thus, a sorting network is differentiable if the employed sigmoid function is differentiable. ``Differentiable'' may be replaced with any other form of differentiability, such as ``differentiable a.e.''
    \end{proof}
\end{lemma}

\subsection{The Activation Replacement Trick \texorpdfstring{$\protect\varphi$}{Phi}}

In this subsection, we present the activation replacement trick (ART), which can greatly enhance the performance of differentiable sorting networks in the case of using the logistic sigmoid function / distribution.
In the next subsection, we present monotonic differentiable sorting networks, an alternative to the ART and derived from a different viewpoint. 
As we will see in the experimental evaluation, monotonic differentiable sorting networks typically outperform differentiable sorting networks with the ART; however, there are also cases in which the ART performs better, specifically for large~$n$.

Due to the nature of the sorting network, values with large as well as very small differences are compared in each layer.
Comparing values with large differences causes vanishing gradients, while comparing values with very small differences can modify, i.e., blur, values as they are only partially swapped.
Specifically, in the case of the logistic function, it is saturated for large inputs but also returns a value close to the mean for inputs that are close to each other.
Based on these observations, we propose an activation replacement trick, which avoids vanishing gradients as well as blurring.
That is, we modify the distribution of the differences between compared values to avoid small differences close to $0$ as well as large differences.

Assuming that the inputs to a sorting network are normally distributed, there are many cases in which the differences of two values $|a_j -a_i|$ are very small as well as many cases in which the differences are very large.
For the relaxation of sorting networks, this poses two problems:

If $|a_j -a_i|$ is close to $0$, while we obtain large gradients, this also blurs the two values to a great extent, modifying them considerably.
Thus, it is desirable to avoid $|a_{j} - a_i| \approx 0$.\\
On the other hand, if $|a_{j} - a_i|$ is large, vanishing gradients occur, which hinders training.

\begin{marginfigure}
    \centering
    \setcounter{arrown}{0}
    \footnotesize
    \begin{tikzpicture}[node distance=50pt, p/.style={circle, minimum size=3pt, inner sep=0pt, outer sep=0pt, fill, anchor=center}]
        \colorlet{arrowColor}{black!65}
        \definecolor{originalDist}{RGB}{243, 155, 35}
        \definecolor{newDist}{RGB}{2, 115, 85}
        \definecolor{newDist2}{RGB}{115, 198, 54}
        \colorlet{newDist}{newDist!80!newDist2}
        \node[anchor=south] (sigmoid) at (0,0) {
            \begin{tikzpicture}[scale=.7, every node/.style={scale=1.}]
                \draw[->] (-3,0) -- (3.2,0) node[right] {$x$};
                \draw[->] (0,0) -- (0,1.2) node[above] {$\sigma(x)$};
                \draw[scale=1.,domain=-3:3,smooth,variable=\x] plot ({\x},{1/(1+exp(-\x*1.25))});

                \foreach \x in {-2,...,2} {
                    \draw (\x,0) -- (\x,-.07) node[anchor=north] {$\x$};
                };

                \setcounter{arrown}{0}
                \foreach \x in {-2.5, -.5,.6,1.4} {
                    \node[p, originalDist] (begin\thearrown) at (\x, {1/(1+exp(-\x*1.25))}) {};
                    \node[p, newDist] (end\thearrown) at ({\x/abs(\x)^.6}, {1/(1+exp(-(\x/abs(\x)^.6)*1.25))}) {};
                    \stepcounter{arrown}
                };
                \draw[->,>=latex, arrowColor] (begin0) edge[bend left] (end0) ;
                \draw[->,>=latex, arrowColor] (begin1) edge[in=100, out=80, looseness=3.5] (end1) ;
                \draw[->,>=latex, arrowColor] (begin2) edge[in=100, out=135, looseness=4.5] (end2) ;
                \draw[->,>=latex, arrowColor] (begin3) edge[in=100, out=70, looseness=4] (end3) ;
            \end{tikzpicture}
        };
        \node[anchor=south] (distributions) at (0,-1.8) {
            \begin{tikzpicture}[scale=.7, every node/.style={scale=1.}]
                \draw[->] (-3,0) -- (3.2,0) node[right] {$x$};
                \draw[->] (0,0) -- (0,1.2);
                \draw[scale=1.,domain=-3:3,smooth,variable=\x, originalDist] plot ({\x},{exp(-1/2*(\x/1.)^2)});
                \draw[scale=1.,domain=-3:3,smooth,variable=\x, newDist] plot ({\x},{exp(-1/2*(\x/1.)^2)*(2*abs(\x)^.5)*0.60814});

                \foreach \x in {-2,...,2} {
                    \draw (\x,0) -- (\x,-.07) node[anchor=north] {$\x$};
                };

                \setcounter{arrown}{0}
                \foreach \x in {-2.5, -.5,.6,1.4} {
                    \node[p, originalDist] (begin\thearrown) at ({\x},{exp(-1/2*(\x/1.)^2)}) {};
                    \node[p, newDist] (end\thearrown) at ({(\x/abs(\x)^.6)},{exp(-1/2*((\x/abs(\x)^.6)/1.)^2)*(2*abs((\x/abs(\x)^.6))^.5)*0.60814}) {};
                    \stepcounter{arrown}
                };
                \draw[->,>=latex, arrowColor] (begin0) edge[bend left] (end0) ;
                \draw[->,>=latex, arrowColor] (begin1) edge[in=100, out=80, looseness=3.5] (end1) ;
                \draw[->,>=latex, arrowColor] (begin2) edge[in=80, out=110, looseness=4.5] (end2) ;
                \draw[->,>=latex, arrowColor] (begin3) edge[in=65, out=55, looseness=3.5] (end3) ;

                \node[originalDist, anchor=east] at (3.5, .7) {$p(x)$};
                \node[newDist, anchor=east]      at (3.5, 1.25) {$p(\varphi (x))$};
            \end{tikzpicture}
        };
    \end{tikzpicture}
    
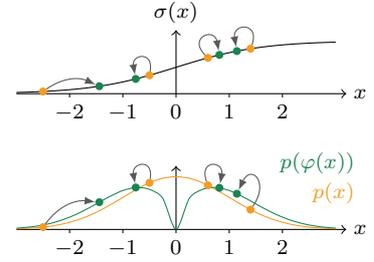
\captionof{figure}{
    The Activation Replacement Trick.
    \textit{Top:} on the logistic sigmoid function, the input values $x$ (orange) are mapped to $\varphi(x)$ (green) and are thus closer to $-1$ and $+1$.
    \textit{Bottom:} probability density functions of Gaussian distributed input values $x$ (orange) and the distribution of replaced input values $\varphi(x)$ (green).
    }
    \label{fig:act-replacement-trick}
\end{marginfigure}

To counter these two problems at the same time, we propose the activation replacement trick.
We transform the differences between two values to be potentially swapped (e.g., $x=(a_j - a_i)$) from a unimodal Gaussian distribution into a bimodal distribution, which has a low probability density around $0$.
To this end, we apply the transformation\\[-.4em]
\begin{equation}
    \varphi : x\mapsto \frac{x}{{|x|}^\ARTstrength + \epsilon}
    \label{eq:varphi-def}
\end{equation}\\[-.4em]
to the differences $x$, where $\ARTstrength \in [0, 1]$ and $\epsilon\approx 10^{-10}$.
$\varphi$ pushes all input values (depending on the sign) toward $-1$ and $+1$, respectively.
Thus, by applying $\varphi$ before $\sigma$, we move the input values outside $[-1,+1]$ to positions at which they have a larger gradient, thus mitigating the problem of vanishing gradients.
Simultaneously, we achieve a probability density of $0$ at $|a_j - a_i| = 0$ (i.e., here $p(\varphi(0)) = 0$) as all values close to zero are mapped toward $-1$ and $+1$, respectively.
This is displayed in Figure~\ref{fig:act-replacement-trick}.

As we multiply by the inverse temperature parameter~$\steepness$, we map the input to the sigmoid function toward $-\steepness$ and $+\steepness$, respectively.
Thus, when replacing $\sigma(x)$ by $\sigma(\varphi(x))$, we push the output values toward $\frac{1}{1+e^{-1 \cdot \beta}}$ or $\frac{1}{1+e^{1 \cdot \beta}}$.
This increases the gradient $\frac{\partial \sigmoid(\varphi(x))}{\partial x}$ for large $\operatorname{abs}(x)$ which are those values causing the vanishing gradients, addressing the problem of vanishing gradients.
Further, for all $x\in (-1, +1)$ this pushes the output values away from $\nicefrac{1}{2}$, addressing the problem of blurring of values.

This leads to a sigmoid function of $\sigma\circ\varphi$.
Empirically, the activation replacement trick accelerates the training through our sorting network.
We observe that, while sorting networks up to $21$~layers (i.e., bitonic networks with $n\leq 64$) can operate with moderate inverse temperature (i.e., $\steepness \leq 15$) and without the activation replacement trick (i.e., $\ARTstrength=0$), for more layers, the activation replacement trick becomes necessary for good performance.
Notably, the activation replacement trick also improves the performance for sorting networks with fewer layers.
Further, the activation replacement trick allows training with smaller inverse temperature $\steepness$, which makes training more stable specifically for long sequences as it avoids exploding gradients.

Note that, in the case of bitonic sorting networks, in the first layer of the last merge block, $\nicefrac{n}{2}$ elements in non-descending order are element-wise compared to $\nicefrac{n}{2}$ elements in non-ascending order.
Thus, in this layer, we compare the minimum of the first sequence to the maximum of the second sequence and vice versa.
At the same time, we also compare the median of both sequences as well as values close to the median to each other.
While we consider very large differences as well as very small differences in the same layer, the activation replacement trick achieves an equalization of the mixing behavior, reducing blurring and vanishing gradients.

\marginnote{
We note that the activation replacement trick and monotonicity are mutually exclusive.
They both have independent theoretical motivations, justifications, and advantages.
While the activation replacement trick has its merits in sorting large numbers of elements $n$, in general, monotonic differentiable sorting networks perform better and convey more stable behavior.
}

\subsection{Monotonic Differentiable Sorting Networks}
\label{sec:monotonic-diffsort}

In this subsection, we formalize the notion of monotonic and error-bounded differentiable sorting networks and characterize families of monotonic and error-bounded differentiable sorting networks.
For this, we start by defining monotonicity in continuous conditional swap operations.

\begin{definition}[Monotonic Continuous Conditional Swaps]
    \label{def:monotonic}
    We say $f$ produces monotonic conditional swaps if $\softmin_f(x, 0)$ is non-decreasingly monotonic in $x$, i.e., $\softmin'_f(x, 0) \geq 0$ for all $x$.
\end{definition}
It is sufficient to define it w.l.o.g.~in terms of $\softmin_f(x, 0)$ due to its commutativity, stability, and odd-symmetry of the operators (cf.~Supplementary Material~\ref{sec:properties-of-min-and-max}).

\begin{theorem}[Monotonicity of Continuous Conditional Swaps]
    \label{theorem:monotonic-swap}
    A continuous conditional swap (in terms of a differentiable sigmoid function $f$) being non-decreasingly monotonic in all arguments and outputs requires that the derivative of $f$ decays no faster than $1/x^2$, i.e.,
    \begin{equation}
        f^\prime(x) \in \Omega\left(\frac{1}{x^2}\right)\,.
        \label{eq:omega-req}
    \end{equation}
    \begin{proof}

        We show that Equation~\eqref{eq:omega-req} is a necessary criterion for monotonicity of the conditional swap.
        Because $f$ is a continuous sigmoid function with $f:\mathbb{R}\to [0,1]$, $\softmin_f(x, 0) = f(-x)\cdot x > 0$ for some $x>0$.
        Thus, monotonicity of $\softmin_f(x, 0)$ implies $\limsup_{x\to\infty} \softmin_f(x, 0) > 0$ (otherwise the value would decrease again from a value $>0$.)
        Thus,
        \begin{equation}
            \lim_{x\to\infty} \softmin_f(x, 0) = \lim_{x\to\infty} f(-x)\cdot x = \lim_{x\to\infty} \frac{f(-x)}{1/x}
        \end{equation}
        \begin{equation}
            \overset{\text{(L'Hôpital's rule)}}{=} \lim_{x\to\infty} \frac{-f'(-x)}{-1/x^2} = \lim_{x\to\infty} \frac{f'(-x)}{1/x^2} = \lim_{x\to\infty} \frac{f'(x)}{1/x^2} 
        \end{equation}
        \begin{equation}
            = \limsup_{x\to\infty} \frac{f'(x)}{1/x^2} > 0
            \Longleftrightarrow f'(x) \in \Omega\left(\frac1{x^2}\right). 
        \end{equation}
        assuming $\lim_{x\to\infty} \frac{f'(x)}{1/x^2}$ exists. Otherwise, it can be proven analogously via a proof by contradiction.
    \end{proof}
\end{theorem}

After defining the monotonicity of conditional swap operators, we can extend this to differentiable sorting networks.

\begin{corollary}[Monotonic Sorting Networks]
    If the individual conditional swaps of a sorting network are monotonic, the sorting network is also monotonic.
    \begin{proof}
        If single layers $g, h$ are non-decreasingly monotonic in all arguments and outputs, their composition $h\circ g$ is also non-decreasingly monotonic in all arguments and outputs.
        Thus, a network of arbitrarily many layers is non-decreasingly monotonic.
    \end{proof}
\end{corollary}

Above, we formalized the property of monotonicity.
Another important aspect is whether the error of the differentiable sorting network is bounded. %
It is very desirable to have a bounded error because without bounded errors, the error, i.e., the difference between the result of the differentiable sorting network and the result of the hard sorting function, diverges to infinity.
Minimizing this error is desirable.

\begin{definition}[Error-Bounded Continuous Conditional Swaps]
    A continuous conditional swap has a bounded error if and only if $\sup_{x} \softmin_f(x, 0)\allowbreak = c$ is finite.
    The continuous conditional swap is therefore said to have an error bounded by $c$.
\end{definition}
It is sufficient to define it w.l.o.g.~in terms of $\softmin_f(x, 0)$ due to its commutativity, stability, and odd-symmetry of the operators (cf.~Supplementary Material~\ref{sec:properties-of-min-and-max}).
In general, for better comparability between functions, we assume a Lipschitz continuous function $f$ with Lipschitz constant $1$.

\begin{theorem}[Error-Bounds of Continuous Conditional Swaps]~\\%
    \label{theorem:error-bounds-swap}
    (i) A differentiable continuous conditional swap has a bounded error if
    \begin{equation}
        f^\prime(x) \in \mathcal{O}\left(\frac{1}{x^2}\right)\,.
        \label{eq:error-bounds-o-notation}
    \end{equation}
    (ii) If it is additionally monotonic, the error-bound can be found as $\lim_{x\to\infty}\allowbreak \softmin_f(x, 0)$ and additionally the error is bound only if Equation \eqref{eq:error-bounds-o-notation} holds.
    \begin{proof}
        (i)
        W.l.o.g.~we consider $x>0$. Let $g(z) := f(-1/z), g(0)=0$. Thus, $g'(z) = 1/z^2 \cdot f'(-1/z) \leq c$ according to Equation~\eqref{eq:error-bounds-o-notation}.
        Thus, $g(z) = g(0) + \int_0^z g'(t) dt \leq c\cdot z$.
        Therefore, $f(-1/z) \leq c\cdot z \implies 1/z \cdot f(-1/z) \leq c$ and with $x=1/z$ $\implies x\cdot f(-x) = \softmin_f(x, 0) \leq c$.

        (ii)
        Let $\softmin_f(x, 0)$ be monotonic and bound by $\softmin_f(x, 0)\leq c$.
        For $x>0$ and $h(x):= \softmin_f(x, 0)$, \\[-1.em]
        \begin{align}
            h'(x) = -x \cdot f'(-x) + f(-x) \implies & x^2 f'(-x) = \underbrace{- x h'(x)}_{\leq 0} + x\cdot f(-x) \notag \\
            & \leq x\cdot f(-x) \leq c\,.
        \end{align}\\[-1.5em]
        Thus, $f^\prime(x) \in \mathcal{O}\left(\frac{1}{x^2}\right)$.
    \end{proof}
\end{theorem}

After characterizing a family of distributions leading to error-bounded conditional swap operators, we can also extend this property to differentiable sorting networks.

\begin{theorem}[Error-Bounds of Diff.~Sorting Networks]
    If the error of individual conditional swaps of a sorting network is bounded by $\epsilon$ and the network has $\ell$ layers, the total error is bounded by $\epsilon\cdot\ell$.
    \begin{proof}
        Induction over number $k$ of executed layers.
        Let $x^{(k)}$ be input $x$ differentially sorted for $k$ layers and $\mathbf{x}^{(k)}$ be input $x$ hard sorted for $k$ layers as an anchor.
        We require this anchor, as it is possible that $\mathbf{x}_i^{(k)} < \mathbf{x}_j^{(k)}$ but $x_i^{(k)} > x_j^{(k)}$ for some $i, j, k$.

        Begin of induction: $k=0$.
        Input vector $x$ equals the vector $x^{(0)}$ after 0 layers. Thus, the error is equal to $0 \cdot \epsilon$.

        Step of induction:
        Given that after $k-1$ layers the error is smaller than or equal to $(k-1)\epsilon$, we need to show that the error after $k$ layers is smaller than or equal to $k\epsilon$.

        The layer consists of comparator pairs $i, j$.
        W.l.o.g.~we assume $\mathbf{x}_i^{(k-1)} \leq \mathbf{x}_j^{(k-1)}$.
        W.l.o.g.~we assume that wire $i$ will be the $\min$ and that wire $j$ will be the $\max$, therefore $\mathbf{x}_i^{(k)} \leq \mathbf{x}_j^{(k)}$.
        This implies $\mathbf{x}_i^{(k-1)} = \mathbf{x}_i^{(k)}$ and $\mathbf{x}_j^{(k-1)} = \mathbf{x}_j^{(k)}$.
        We distinguish two cases:

        $\bullet~~\Big(\mathbf{x}_i^{(k-1)} \leq \mathbf{x}_j^{(k-1)}$ and $x_i^{(k-1)} \leq x_j^{(k-1)}\Big)~\quad$
        According to the assumption, $\big|x_i^{(k-1)} - x_i^{(k)}\big| \leq \epsilon$ and $\big|x_j^{(k-1)} - x_j^{(k)}\big| \leq \epsilon$.
        Thus, $
        \big|x_i^{(k)} - \mathbf{x}_i^{(k)}\big|
        \leq \big|x_i^{(k-1)} - \mathbf{x}_i^{(k-1)}\big| + \big|x_i^{(k-1)} - x_i^{(k)}\big|
        \leq (k-1)\epsilon + \epsilon = k\epsilon
        $.\\

        $\bullet~~\Big(\mathbf{x}_i^{(k-1)} \leq \mathbf{x}_j^{(k-1)}$ but $x_i^{(k-1)} > x_j^{(k-1)}\Big)~\quad$
        This case can only occur if $\big|x_j^{(k-1)} - \mathbf{x}_i^{(k-1)}\big| \leq (k-1)\epsilon$ and $\big|x_i^{(k-1)} - \mathbf{x}_j^{(k-1)}\big| \leq (k-1)\epsilon$ because $\mathbf{x}_i^{(k-1)}$ and $\mathbf{x}_j^{(k-1)}$ have to be so close that within margin of error such a reversed order is possible.
        According to the assumption, $\big|x_j^{(k-1)} - x_i^{(k)}\big| \leq \epsilon$ and $\big|x_i^{(k-1)} - x_j^{(k)}\big| \leq \epsilon$.
        Thus, $
        \big|x_i^{(k)} - \mathbf{x}_i^{(k)}\big|
        \leq \big|x_j^{(k-1)} - \mathbf{x}_i^{(k-1)}\big| + \big|x_j^{(k-1)} - x_i^{(k)}\big|
        \leq (k-1)\epsilon + \epsilon = k\epsilon
        $.
    \end{proof}
\end{theorem}

\paragraph{Discussion}
Monotonicity is highly desirable because, otherwise, adverse effects can occur, e.g., an input requiring to be decreased in order to increase the output.
In gradient-based training, non-monotonicity is problematic as it produces gradients with the opposite sign.
In addition, as monotonicity is also given in hard sorting networks, it is desirable to preserve this property in the relaxation.
Further, monotonic differentiable sorting networks are quasiconvex and quasiconcave as any monotonic function is both quasiconvex and quasiconcave, which leads to favorable convergence rates \cite{kiwiel2001convergence}.
Bounding and reducing the deviation from its hard counterpart reduces the relaxation error, and is therefore desirable.
\vspace{2em}

\begin{corollary}[Monotonic and Error-Bounded Diff.~Sorting Networks]
    We can approximately characterize the sigmoid function that we are interested in, i.e., those which lead to monotonic and error-bounded differentiable sorting networks as
    \begin{equation}
        f^\prime(x) \in \Theta\left(\frac{1}{x^2}\right)\,.
        \label{eq:theta-notation}
    \end{equation}
    This characterization is approximate as for Theorem~\ref{theorem:monotonic-swap} an asymptotic characterization can only be a requirement for monotonicity but cannot be sufficient. 
    This is because a local non-monotonic behavior cannot be prevented through an asymptotic requirement.
    \begin{proof}
        Equation \eqref{eq:theta-notation} ($\Theta$) is the intersection of \eqref{eq:omega-req} ($\Omega$) and \eqref{eq:error-bounds-o-notation} ($\mathcal{O}$).
    \end{proof}
\end{corollary}

\newcommand{\sigmoidReciprocal}[0]{f_\mathcal{R}}
\newcommand{\sigmoidCauchy}[0]{f_\mathcal{C}}
\newcommand{\sigmoidOptimal}[0]{f_\mathcal{O}}

\subsubsection{Examples}

\label{sec:sigmoid-functions}

\begin{marginfigure}
    \centering
    \includegraphics[width=\linewidth]{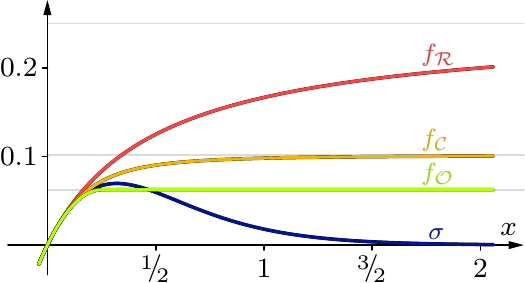}
    \caption{\label{fig:softminx0}$\softmin_f(x,0)$ for different sigmoid functions~$f$; color coding as in Table~\ref{tab:sigmoid-functions-overview}.}
\end{marginfigure}

\begin{marginfigure}
    \centering
    \includegraphics[width=\linewidth]{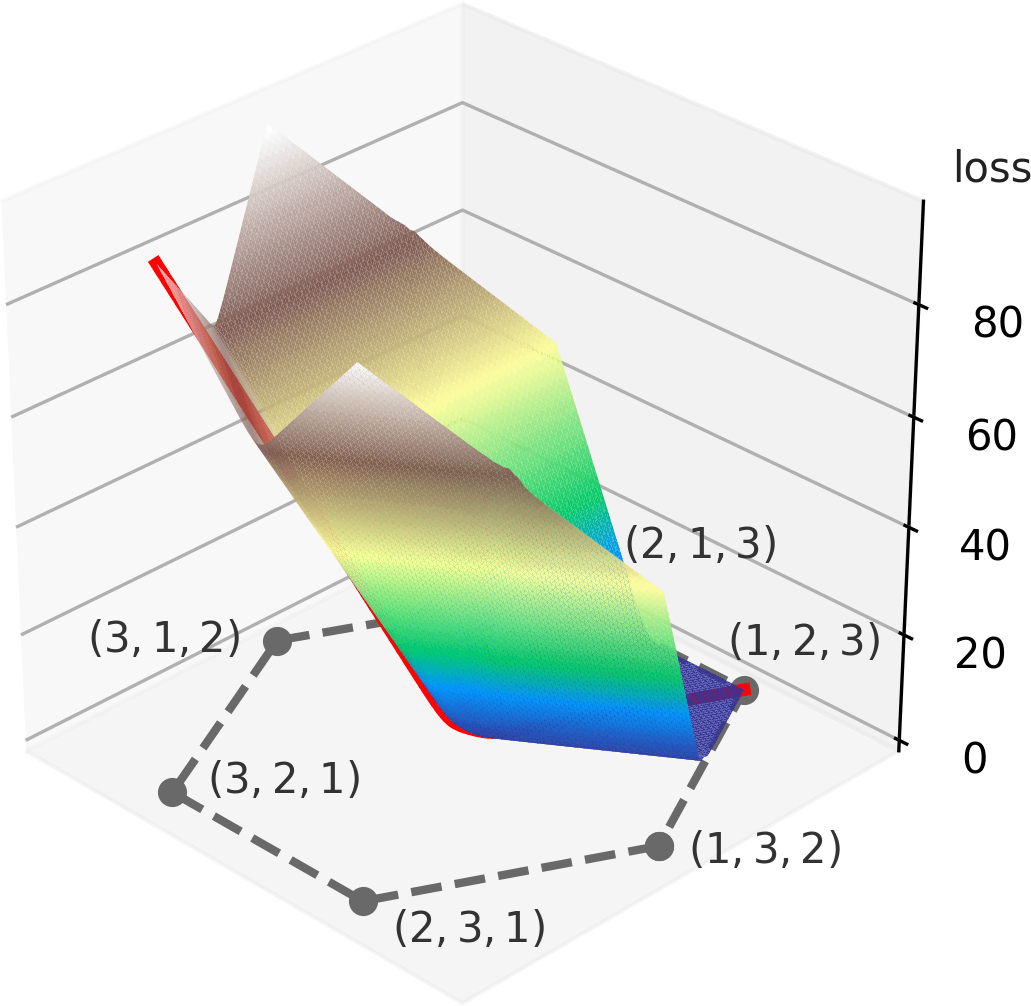}\\[1em]
    \includegraphics[width=\linewidth]{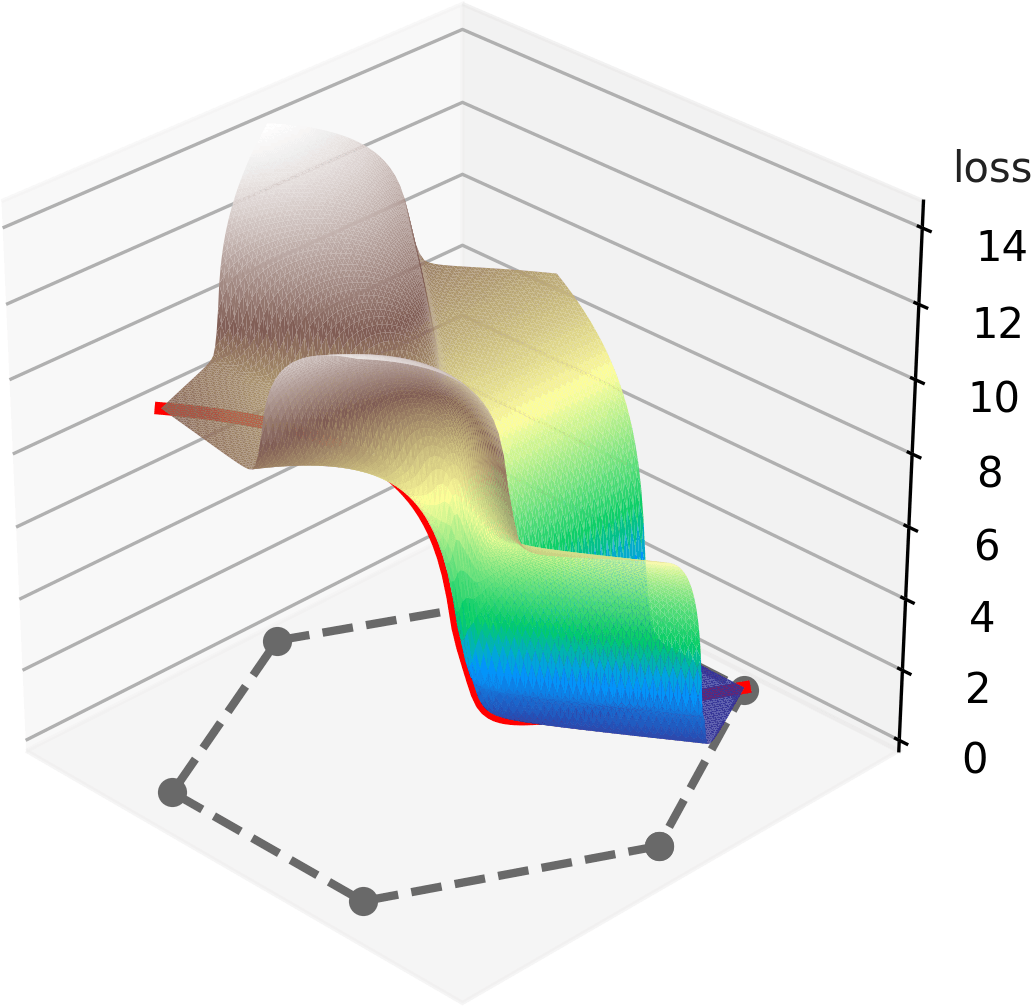}%
    \caption{Loss for a 3-wire odd-even sorting network, drawn over a permutahedron projected onto the $x$-$y$-plane. For logistic sigmoid \textit{(top)} and optimal sigmoid \textit{(bottom)}.
    }
    \label{fig:loss-3d-permutahedron}
\end{marginfigure}

Above, we have specified the space of functions for the differentiable swap operation, as well as their desirable properties.
In the following, we discuss four notable candidates as well as their properties.
The properties of these functions are visualized in Figures~\ref{fig:softminx0}~and~\ref{fig:loss-3d-permutahedron} and an overview of their properties is given in Table~\ref{tab:sigmoid-functions-overview}.

\paragraph{Logistic Distributions} The first candidate is the logistic sigmoid function (the CDF of a logistic distribution) as proposed earlier:
\begin{equation}
    \label{eq:logistic}
    \sigma (x) = \operatorname{CDF}_\mathcal{L} \big({\beta x}\big) = \frac{1}{1+e^{-\beta x}}
\end{equation}
This function is the de-facto default sigmoid function in machine learning.
It provides a continuous, error-bounded, and Lipschitz continuous conditional swap.
However, for the logistic function, monotonicity is not given, as displayed in Figure~\ref{fig:softminx0}.

\paragraph{Reciprocal Sigmoid Function}
To obtain a function that yields a monotonic as well as error-bound differentiable sorting network, a necessary criterion is $f'(x)\in \Theta(1/x^2)$ (the intersection of Equations~\eqref{eq:omega-req}~and~\eqref{eq:error-bounds-o-notation}.)
A natural choice is, therefore, $\sigmoidReciprocal'(x) = \frac{1}{(2|x|+1)^2}$, which produces
\begin{equation}
    \sigmoidReciprocal(x) = \int_{-\infty}^x \frac{1}{(2\beta|t|+1)^2} dt = \frac{1}{2}\frac{2\beta x}{1+2\beta|x|}+\frac{1}{2}.
    \label{eq:reciprocal}
\end{equation}
$\sigmoidReciprocal$ fulfills all criteria, i.e., it is an adequate sigmoid function and produces monotonic and error-bound conditional swaps.
It has an $\epsilon$-bounded-error of $\epsilon=0.25$.
It is also an affine transformation of the elementary bipolar sigmoid function $x\mapsto \frac{x}{|x| + 1}$.
Properties of this function are visualized in Table~\ref{tab:sigmoid-functions-overview} and Figure~\ref{fig:softminx0}.
Proofs for monotonicity can be found in Supplementary Material~\ref{apx:proofs:sec:monotonic}.

\begin{table}[t]
    \centering
    \caption{
        For each function, we display the \textit{function}, its \textit{derivative}, and indicate whether the respective relaxed sorting network is \textit{monotonic} and has a \textit{bounded error}.
    }
    \label{tab:sigmoid-functions-overview}
    \begingroup
    \newcommand{\raiseamount}{-0.25em}
    \newcommand{\figureheight}{2.1em}
    \addtolength{\tabcolsep}{-3pt}
    \footnotesize
    \begin{tabular}{lcclllll}
        \toprule
        Function$\quad$                 & $f$ (CDF) & $f'$ (PDF) &  Eq.
                                        & Mono. & Bounded Error \\
        \midrule
        $\displaystyle \sigma$               & \raisebox{\raiseamount}{\includegraphics[height=\figureheight]{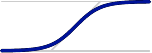}} & \raisebox{\raiseamount}{\includegraphics[height=\figureheight]{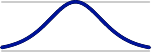}}
        & (\ref{eq:logistic})
        & $\quad$\xmark & \cmark\ ($\approx .0696 / \alpha$)  \\
        $\displaystyle \sigma \circ \varphi$ & \raisebox{\raiseamount}{\includegraphics[height=\figureheight]{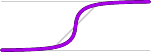}} & \raisebox{\raiseamount}{\includegraphics[height=\figureheight]{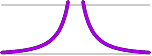}}
        & (\ref{eq:logistic}, \ref{eq:varphi-def})
        & $\quad$\xmark & \cmark\ ($\approx .0302^\ast$)  \\
        \midrule
        $\displaystyle \sigmoidReciprocal$   & \raisebox{\raiseamount}{\includegraphics[height=\figureheight]{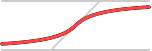}} &  \raisebox{\raiseamount}{\includegraphics[height=\figureheight]{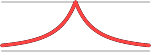}}
        & (\ref{eq:reciprocal})
        & $\quad$\cmark & \cmark\ ($1/4/\alpha$) \\
        $\displaystyle \sigmoidCauchy$       & \raisebox{\raiseamount}{\includegraphics[height=\figureheight]{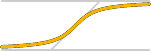}} &  \raisebox{\raiseamount}{\includegraphics[height=\figureheight]{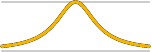}}
        & (\ref{eq:cauchy})
        & $\quad$\cmark & \cmark\ ($1/\pi^2/\alpha$)  \\
        $\displaystyle \sigmoidOptimal$      & \raisebox{\raiseamount}{\includegraphics[height=\figureheight]{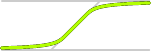}} &  \raisebox{\raiseamount}{\includegraphics[height=\figureheight]{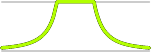}}
        & (\ref{eq:f-opt})
        & $\quad$\cmark & \cmark\ ($1/16/\alpha$)  \\
        \bottomrule
    \end{tabular}
    \endgroup
\end{table}

\paragraph{Cauchy Distributions}
By using the CDF of the Cauchy distribution, we maintain monotonicity while reducing the error-bound to $\epsilon=1/\pi^2\approx0.101$.
It is defined as
\begin{equation}
    \sigmoidCauchy(x)= \operatorname{CDF}_\mathcal{C} \big({\beta x}\big) = \frac{1}{\pi} \int_{-\infty}^x \frac{\beta}{1 + (\beta t)^2} dt = \frac{1}{\pi} \arctan \big( {\beta x} \big) + \frac{1}{2}
    \label{eq:cauchy}
\end{equation}
In the experimental evaluation, we find that tightening the error improves the performance.

\paragraph{Optimal Monotonic Sigmoid Function}
At this point, we are interested in the monotonic swap operation that minimizes the error-bound.
Here, we set \textit{$1$-Lipschitz continuity} again as a requirement to make different relaxations of conditional swaps comparable.
We show that $\sigmoidOptimal$ is the best possible sigmoid function achieving an error-bound of only $\epsilon=1/16$
\begin{theorem}[Optimal Sigmoid Function]
    The optimal sigmoid function minimizing the error-bound, while producing a monotonic and $1$-Lipschitz continuous (with $\beta =1$) conditional swap operation, is
    \begin{equation}
        \sigmoidOptimal(x) =
        \left\{ \begin{array}{ll}
           \phantom{1}-\frac{1}{16\beta x}
             & \mbox{\text{if} $\beta x < -\frac{1}{4}$,} \\[1ex]
           1-\frac{1}{16\beta x}
             & \mbox{\text{if} $\beta x > +\frac{1}{4}$,} \\[1ex]
           \beta x +\frac{1}{2}
             & \mbox{\text{otherwise.}}
        \end{array}\right.
        \label{eq:f-opt}
    \end{equation}
    \vspace*{-1em}
    \begin{proof}
        Given the above conditions, the optimal sigmoid function is uniquely
        determined and can easily be derived as follows:
        Due to stability,
        it suffices to consider $\softmin_f(x,0) = x \cdot f(-x)$
        or $\softmax_f(0,x) = -x \cdot f(x)$. Due to symmetry and inversion,
        it suffices to consider $\softmin_f(x,0) = x \cdot f(-x)$ for $x > 0$.

        Since $\min(x,0) = 0$ for $x > 0$, we have to choose $f$ in such a
        way as to make $\softmin_f(x,0) = x \cdot f(-x)$ as small as possible,
        but not negative. For this, $f(-x)$ must be made as small as possible.
        Since we know that $f(0) = \frac{1}{2}$ and we are limited to
        functions~$f$ that are Lipschitz continuous with $\alpha = 1$,
        $f(-x)$ cannot be made smaller than $\frac{1}{2}-x$, and hence
        $\softmin_f(x,0)$ cannot be made smaller than
        $x\cdot\left(\frac{1}{2} -x\right)$. To make $\softmin_f(x,0)$ as small as possible, we have to follow $x\cdot\left(\frac{1}{2}-x\right)$ as far as possible (i.e., to values $x$ as large as possible).
        Monotonicity requires that this function can be followed only
        up to $x = \frac{1}{4}$, at which point we have
        $\softmin_f(\frac{1}{4},0) =
        \frac{1}{4}\left(\frac{1}{2}-\frac{1}{4}\right) = \frac{1}{16}$.
        For larger~$x$, that is, for $x > \frac{1}{4}$, the value of
        $x\cdot\left(\frac{1}{2} -x\right)$ decreases again and hence the functional
        form of the sigmoid function~$f$ has to change at $x = \frac{1}{4}$
        to remain monotonic.

        The best that can be achieved for $x > \frac{1}{4}$ is to make it
        constant, as it must not decrease (due to monotonicity) and should
        not increase (to minimize the deviation from the crisp / hard version).
        That is, $\softmin_f(x,0) = \frac{1}{16}$ for $x > \frac{1}{4}$.
        It follows $x \cdot f(-x) = \frac{1}{16}$ and hence
        $f(-x) = \frac{1}{16x}$ for $x > \frac{1}{4}$.
        Note that, if the transition from the linear part to the hyperbolic part were at $|x| < \frac{1}{4}$, the function would not be Lipschitz continuous with $\alpha=1$.
    \end{proof}
\end{theorem}

An overview of the selection of sigmoid functions we consider is shown in Table~\ref{tab:sigmoid-functions-overview}.
Note how $f_{\cal R}$, $f_{\cal C}$ and $f_{\cal O}$ in this order get closer to $x +\frac{1}{2}$ (the light gray diagonal line) and hence steeper in their middle part.
This is reflected by a widening region of values of the derivatives that are close to or even equal to~1.

Table~\ref{tab:sigmoid-functions-overview} also indicates whether a sigmoid function yields a \textit{monotonic} swap operation or not, which is visualized in Figure~\ref{fig:softminx0}: clearly $\sigma$-based sorting networks are not monotonic, while all others are.
It also states whether the \textit{error is bounded}, which for a monotonic swap operation means $\lim_{x\to\infty}\softmin_f(x,0) < \infty$, and gives their bound relative to the Lipschitz constant $\alpha$.

Figure~\ref{fig:loss-3d-permutahedron} displays the loss for a sorting network with $n=3$ inputs.
We project the hexagon-shaped
3-value permutahedron onto the $x$-$y$-plane,
while the $z$-axis indicates the loss. Note that, at the rightmost point $(1,2,3)$, the loss is $0$ because all elements are in the correct order, while at the left front $(2,3,1)$ and rear $(3,1,2)$ the loss is at its maximum because all elements are at the wrong positions.
Along the red center line, the loss rises logarithmic for the optimal sigmoid function on the right.
Note that the monotonic sigmoid functions produce a loss that is larger when more elements are in the wrong order.
For the logistic function, $(3, 2, 1)$ has the same loss as $(2, 3, 1)$ even though one of the ranks is correct at $(3, 2, 1)$, while for $(2, 3, 1)$ all three ranks are incorrect.

\section{Monotonicity and Error-Boundedness of Differentiable Sorting Operators}

\begin{marginfigure}
    \centering
    \includegraphics[width=\linewidth]{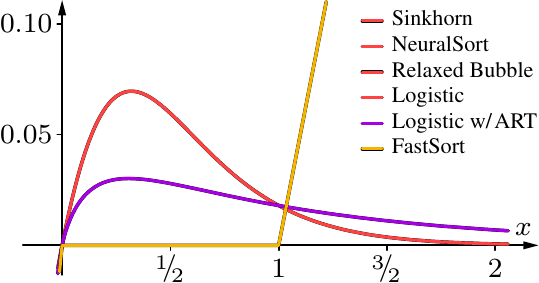}
    \caption{\label{fig:softminx0-others}
    $\softmin(x,0)$ for SinkhornSort (red), 
    NeuralSort (red), 
    Relaxed Bubble sort (red), 
    diffsort with logistic sigmoid (red), 
    diffsort with activation replacement trick (purple), 
    and FastSort (orange).
    }
\end{marginfigure}

The \textit{Relaxed Bubble sort}, as proposed in the previous chapter, uses logistic distributions, which indicates a lack of monotonicity. 
In fact, for $n=2$, Relaxed Bubble sort is equivalent to differentiable sorting networks with the logistic distribution, i.e., it is non-monotonic.

For \textit{differentiable sorting networks with the activation replacement trick}, the asymptotic character of $\sigma \circ \varphi$ does not fulfill the requirement set by Theorem~\ref{theorem:monotonic-swap}, and is, therefore, non-monotonic as also displayed in Figure~\ref{fig:softminx0-others} (purple).

For the special case of $n=2$, i.e., for sorting two elements, \textit{NeuralSort}~\cite{grover2019neuralsort} is equivalent to differentiable sorting networks with the logistic sigmoid function.
Thus, it is non-monotonic as displayed in Figure~\ref{fig:softminx0-others}.
The same also applies to SoftSort~\cite{prillo2020softsort}.

For \textit{SinkhornSort}, we can simply construct an example of non-monotonicity by keeping one value fixed, e.g., at zero, and varying the second value ($x$) as in Figure~\ref{fig:softminx0-others} and displaying the minimum.
Interestingly, for the case of $n=2$, this function is numerically equal to NeuralSort and differentiable sorting networks with the logistic function.

\begin{margintable}
    \centering
    \captionof{table}{
        For each differentiable sorting operator, whether it is monotonic (M), and whether it has a bounded error (BE).
        \label{tab:mono-bound-all}
    }
    \begingroup
    \footnotesize
    \addtolength{\tabcolsep}{-4pt}
    \begin{tabular}{llcc}
        \toprule
        Method & & M & BE \\
        \midrule
        NeuralSort & & \xmark & -- \\
        SoftSort & & \xmark & -- \\
        SinkhornSort & & \xmark & -- \\
        FastSort & & \cmark & \xmark \\
        Relaxed Bubble Sort & & \xmark & -- \\
        Diff. Sorting Networks & $\sigma$ & \xmark & \cmark \\
        Diff. Sorting Networks & $\sigma\circ \varphi$ & \xmark & \cmark \\
                         \midrule
        Diff. Sorting Networks & $\sigmoidReciprocal$     & \cmark & \cmark \\
        Diff. Sorting Networks & $\sigmoidCauchy$         & \cmark & \cmark \\
        Diff. Sorting Networks & $\sigmoidOptimal$        & \cmark & \cmark \\
        \bottomrule
    \end{tabular}
    \endgroup
\end{margintable}

For \textit{FastSort}, we follow the same principle and find that it is indeed monotonic (in this example); however, the error is unbounded, which is undesirable.

We summarize monotonicity and error-boundedness for all differentiable sorting functions in Table~\ref{tab:mono-bound-all}.

\section{Experiments}
\label{sec:ranking-supervision-experiments}

The experimental section is structured as follows: 
first, we give all details on the overall training setting;
second, we evaluate differentiable sorting networks with the ART, which allows us to do large-scale experiments;
third, we benchmark all methods and focus on the benefits of monotonic differentiable sorting networks;
finally, we perform a runtime and memory analysis.
We evaluate the proposed differentiable sorting networks on the four-digit MNIST sorting benchmark \cite{grover2019neuralsort,cuturi2019differentiable} as well as on the real-world SVHN data set. %

\paragraph{MNIST}
For the four-digit MNIST sorting benchmark, MNIST digits are concatenated to four-digit numbers, e.g., \mnistimg{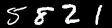}.
A CNN then predicts a scalar value corresponding to the value displayed in the four-digit image.
For training, $n$ of those four-digit images are separately processed by the CNN and then sorted by the relaxed sorting network as shown in Figure~\ref{fig:overall-architecture-odd-even}.
Based on the permutation matrix produced by the sorting network and the ground truth ranking, the training objective is computed (Equation~\eqref{eq:loss}) and the CNN is updated.
At test time, we forward single images of four-digit numbers from the test data set.
For evaluation, the discrete rankings of the predicted values are compared to the rankings of their ground truth.
Note that the $n$ used for testing and evaluation can be independent of the $n$ used for training because the $n$ images are processed independently.

\paragraph{SVHN}
Since the multi-digit MNIST data set is an artificial data set, we also evaluate our technique on the SVHN data set \cite{netzer2011svhn}.
This data set comprises house numbers collected from Google Street View and provides a larger variety wrt.~different fonts and formats than the MNIST data set.
We use the published ``Format~$1$'' and preprocess it as described by Goodfellow~\etal~\cite{goodfellow2013svhn}, cropping the centered multi-digit numbers with a boundary of $30\%$, resizing it to a resolution of $64\times64$, and then selecting $54\times54$ pixels at a random location.
As SVHN contains $1-5$ digit numbers, we can avoid the concatenation and use the original images directly.
Example images are
~\svhnimg{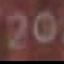}
~\svhnimg{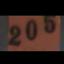}
~\svhnimg{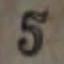}
~\svhnimg{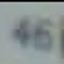}
~\svhnimg{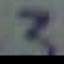}
~\svhnimg{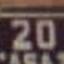}
~\svhnimg{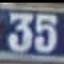}
~\svhnimg{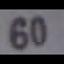}
~\svhnimg{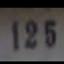}
~\svhnimg{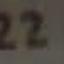}%
\;.
Otherwise, the experimental setup is as for the four-digit MNIST data set.

\paragraph{Network Architecture} For the MNIST sorting task,  we use the same convolutional neural network (CNN) architecture as Grover~\etal~\cite{grover2019neuralsort} and Cuturi~\etal~\cite{cuturi2019differentiable} to allow for comparability.
This architecture consists of two convolutional layers with a kernel size of $5\times5$, $32$ and $64$ channels respectively, each followed by a ReLU and MaxPool layer;
this is (after flattening) followed by a fully connected layer with a size of $64$, a ReLU layer, and a fully connected output layer mapping to a scalar.~~~~
For the SVHN task, we use a network with four convolutional layers with a kernel size of $5\times5$ and ($32, 64, 128, 256$) filters, each followed by a ReLU and a max-pooling layer with stride $2\times2$; followed by a fully connected layer with size $64$, a ReLU, and a layer with output size~$1$.

\paragraph{Evaluation Metrics}
For evaluation, discrete rankings based on the scalar predictions are computed and compared to the discrete ground truth rankings.
As in previous works, we use the evaluation metrics of exact match (EM) of the predicted ranking, and fraction of element-wise correct ranks (EW) in the predicted ranking.
For EM and EW, we follow Grover~\etal~\cite{grover2019neuralsort} and Cuturi~\etal~\cite{cuturi2019differentiable}, and use the same $n$ for training and evaluation.
However, this can be a problem in the context of large input sets as these evaluation metrics become unreliable as $n$ increases.
For example, the difficulty of exact matches rises with the factorial of $n$, which is why they become too sparse to allow for valid conclusions for large $n$.
To allow for a comparison of the performance independent of the number of elements~$n$ used for training, for some experiments, we also evaluate the models based on the EM accuracy for $n=5$ (EM5).
That is, the network can be trained with an arbitrary $n$, but the evaluation is done for $n=5$.

\paragraph{Training Settings}
We use the Adam optimizer~\cite{kingma2015adam} with a learning rate of $10^{-3.5}$.
We use a constant batch size of $100$ as in previous works unless denoted otherwise.
Note that, although $\ARTstrength$ is chosen as a constant value for all $n$, a higher accuracy is possible when optimizing $\ARTstrength$ for each $n$ separately.
The fast sorting and ranking method by Blondel~\etal~\cite{blondel2020fast} does not produce permutation matrices; thus, we cannot use the cross-entropy loss, and therefore, we use the mean-squared-error loss between predicted and ground truth ranks.

\subsection{The Activation Replacement Trick}

We start by evaluating the performance of differentiable sorting networks with the ART. 
For this, we train each model for up to $10^6$ steps.
Furthermore, we set $\ARTstrength=0.25$ and use an inverse temperature of two times the number of layers ($\steepness=2n$ for odd-even and $\steepness=(\log_2n)(1+\log_2n)$ for bitonic.)

\paragraph{Comparison to State-of-the-Art (MNIST)}
We first compare our approach to the methods proposed by Grover~\etal~\cite{grover2019neuralsort} and Cuturi~\etal~\cite{cuturi2019differentiable}.
Here, we follow the setting that the $n$ used for evaluation is the same as the $n$ used for training.
The evaluation is shown in Table~\ref{table:results-mnist}.
We report results for exact match, correct ranks, and EM5, respectively.
For the odd-even architecture, we compare results for the original $n\in\{3,5,7,9,15\}$.
Our approach outperforms current methods on all metrics and input set sizes.
In addition, we extend the original benchmark set sizes by $n\in\{2,4,8,16,32\}$, allowing for the canonical version of the bitonic sorting network which requires input size of powers of $2$.
We apply $n\in\{2,4,8,16,32\}$ to the odd-even as well as the bitonic sorting network.
In this direct comparison, we can see that the bitonic and the odd-even architectures perform similarly.
Notably, the EM and EW accuracies do not always correlate as can be seen for $n=32$.
Here, the EM accuracy is greater for the bitonic network and the EW accuracy is greater for the odd-even network.
We attribute this to the odd-even network's gradients causing swaps of neighbors while the bitonic network's gradients provide a holistic approach favoring exact matches.

\label{sec:evaluation_svhn}
\paragraph{SVHN}
The results in Table~\ref{tab:svhn} show that the real-world SVHN task is significantly harder than the MNIST task.
On this data set, differentiable sorting networks are also better than current methods on all metrics and input set sizes.
Here, the performance of odd-even and bitonic are similar.
Notably, the EM5 accuracy is largest for the bitonic sorting network at $n=32$, which demonstrates that the method benefits from longer input sets.
Further, for $n\in\{8, 16, 32\}$, the bitonic sorting network marginally outperforms the odd-even sorting network on all metrics.

\begin{table*}[t]
    \caption{
    Results for comparing the ART to state-of-the-art~\cite{grover2019neuralsort, cuturi2019differentiable} using the same network architectures averaged over 5 runs.
    The first three rows are duplicated from Cuturi~\etal~\cite{cuturi2019differentiable}.
    The metrics are \hbox{(EM | EW | EM5)}. 
    The models are trained for up to $10^6$ steps.
    }
    \label{table:results-mnist}
    \centering
    \newcommand{\emfivespacer}[0]{\phantom{\ 00.0}}
    \footnotesize
    \begin{tabular}{lccccc}
        \toprule
        \textbf{MNIST} & $n=\pmb{3}$ & $n=\pmb{5}$ & $n=\pmb{7}$ & $n=\pmb{9}$ & $n=\pmb{15}$ \\
        \midrule
        Stoch.~NeuralSort       & $92.0\ |\ 94.6\ |\emfivespacer{}$ & ${79.0}\ |\ 90.7\ |\ {79.0}$ & $63.6\ |\ 87.3\ |\emfivespacer{}$ & $45.2\ |\ 82.9\ |\emfivespacer{}$ & $12.2\ |\ 73.4\ |\emfivespacer{}$  \\
        Det.~NeuralSort         & $91.9\ |\ 94.5\ |\emfivespacer{}$ & ${77.7}\ |\ 90.1\ |\ {77.7}$ & $61.0\ |\ 86.2\ |\emfivespacer{}$ & $43.4\ |\ 82.4\ |\emfivespacer{}$ & $\phantom{0}9.7\ |\ 71.6\ |\emfivespacer{}$  \\
        Optimal Transport       & $92.8\ |\ 95.0\ |\emfivespacer{}$ & ${81.1}\ |\ 91.7\ |\ {81.1}$ & $65.6\ |\ 88.2\ |\emfivespacer{}$ & $49.7\ |\ 84.7\ |\emfivespacer{}$ & $12.6\ |\ 74.2\ |\emfivespacer{}$  \\
        \midrule
        Fast Sort \& Rank       & $90.6\ |\ 93.5\ |\ 73.5$  &  $71.5\ |\ 87.2\ |\ 71.5$  &  $49.7\ |\ 81.3\ |\ 70.5$  &  $29.0\ |\ 75.2\ |\ 69.2$  &  $\phantom{0}2.8\ |\ 60.9\ |\ 67.4$ \\
        \midrule
            $\sigma\circ\varphi$: ART (Odd-Even)      & $\pmb{95.2}\ |\ \pmb{96.7}\ |\ \pmb{86.1}$ & $\pmb{86.3}\ |\ \pmb{93.8}\ |\ \pmb{86.3}$ & $\pmb{75.4}\ |\ \pmb{91.2}\ |\ \pmb{86.4}$ & $\pmb{64.3}\ |\ \pmb{89.0}\ |\ \pmb{86.7}$ & $\pmb{35.4}\ |\ \pmb{83.7}\ |\ \pmb{87.6}$  \\
        \bottomrule
        \toprule
        \textbf{MNIST} & $n=\pmb{2}$ & $n=\pmb{4}$ & $n=\pmb{8}$ & $n=\pmb{16}$ & $n=\pmb{32}$  \\
        \midrule
            $\sigma\circ\varphi$: ART (Odd-Even)        & $98.1\ |\ 98.1\ |\ 84.3$ & $90.5\ |\ 94.9\ |\ 85.5$ & $63.6\ |\ 87.9\ |\ 83.6$ & $31.7\ |\ 82.8\ |\ 87.3$ & $\phantom{0}1.7\ |\ 69.1\ |\ 86.7$  \\
            $\sigma\circ\varphi$: ART (Bitonic)      & $98.1\ |\ 98.1\ |\ {84.0}$ & $91.4\ |\ 95.3\ |\ 86.7$ & $70.6\ |\ 90.3\ |\ {86.9}$ & $30.5\ |\ 81.7\ |\ 86.6$ & $\phantom{0}2.7\ |\ 67.3\ |\ 85.4$  \\
        \bottomrule
    \end{tabular}%
\end{table*}

\begin{table*}[t]
    \caption{
    Results for training on the SVHN data set averaged over 5 runs. 
    The metrics are \hbox{(EM | EW | EM5)}.
     The models are trained for up to $10^6$ steps.
    }
    \label{tab:svhn}
    \centering
    \footnotesize
    \newcommand{\emfivespacer}[0]{\phantom{\ 00.0}}
    \newcommand{\minispace}[0]{}
    \begin{tabular}{lccccc}
        \toprule
        \textbf{SVHN} & $n=\pmb{2}$ & $n=\pmb{4}$ & $n=\pmb{8}$ & $n=\pmb{16}$ & $n=\pmb{32}$  \\
        \midrule
        Det.~NeuralSort     & $90.1\ |\ 90.1\ |\ 39.9$  &  $61.4\ |\ 78.1\ |\ 45.4$  &  $15.7\ |\ 62.3\ |\ 48.5$  &  $\phantom{0}0.1\ |\ 45.7\ |\ 51.0$  &  $\phantom{0}0.0\ |\ 29.9\ |\ 52.7$  \\
        Optimal Transport   & $85.5\ |\ 85.5\ |\ 25.9$  &  $57.6\ |\ 75.6\ |\ 41.6$  &  $19.9\ |\ 64.5\ |\ 51.7$  &  $\phantom{0}0.3\ |\ 47.7\ |\ 53.8$  &  $\phantom{0}0.0\ |\ 29.4\ |\ 53.3$  \\
        Fast Sort \& Rank   & $93.4\ |\ 93.4\ |\ 57.6$  &  $58.0\ |\ 75.8\ |\ 41.5$  &  $\phantom{0}8.6\ |\ 52.7\ |\ 34.4$  &  $\phantom{0}0.3\ |\ 36.5\ |\ 41.6$  &  $\phantom{0}0.0\ |\ 14.0\ |\ 27.5$  \\
        \midrule
            $\sigma\circ\varphi$: ART (Odd-Even)      & \minispace{}$93.4\ |\ 93.4\ |\ 58.0$\minispace{} & \minispace{}$\pmb{74.8}\ |\ \pmb{85.5}\ |\ \pmb{62.6}$\minispace{} & \minispace{}$35.2\ |\ 73.5\ |\ 63.9$\minispace{} & \minispace{}$\phantom{0}1.8\ |\ 54.4\ |\ 62.3$\minispace{} & \minispace{}$\phantom{0}0.0\ |\ 36.6\ |\ 62.6$\minispace{}  \\
            $\sigma\circ\varphi$: ART (Bitonic)       & $\pmb{93.8}\ |\ \pmb{93.8}\ |\ \pmb{58.6}$ & $74.4\ |\ 85.3\ |\ 62.1$ & $\pmb{38.3}\ |\ \pmb{75.1}\ |\ \pmb{66.8}$ & $\pmb{\phantom{0}3.9}\ |\ \pmb{59.6}\ |\ \pmb{66.8}$ & $\phantom{0}0.0\ |\ \pmb{42.4}\ |\ \pmb{67.7}$  \\
        \bottomrule
    \end{tabular}%
\end{table*}

\subsubsection{Large-Scale Sorting and Ranking Supervision}

We are interested in the effect of training with larger input set sizes $n$.
As the bitonic sorting network requires significantly fewer layers than odd-even and is (thus) faster, we use the bitonic sorting network for the scalability experiments.
Here, we evaluate for $n=2^k, k\in\{5,6,7,8,9,10\}$ on the MNIST sorting benchmark, comparing the EM5 accuracy as shown in Table~\ref{table:results-bitonic-large}.

For this experiment, we consider inverse temperature values of $\steepness\in\{30,\allowbreak 32.5,\allowbreak 35,\allowbreak 37.5,\allowbreak 40\}$ and report the mean, best, and worst over all inverse temperature values for each $n$.
We set $\ARTstrength$ to $0.4$ as this allows for stable training with $n>128$.
To keep the evaluation feasible, we reduce the number of steps during training to $10^4$, compared to the $10^6$ iteration in Table~\ref{table:results-mnist}.
Again, we use the Adam optimizer with a learning rate of $10^{-3.5}$.

In the first two columns of Table~\ref{table:results-bitonic-large}, we show a head-to-head comparison with the setting in Table~\ref{table:results-mnist} with $\ARTstrength=0.25$ and $\ARTstrength=0.4$ for $n=32$.
Trained for $10^6$ steps, the EM5 accuracy is $85.4\%$, while it is $78.2\%$ after $10^4$ steps.
Increasing $\ARTstrength$ from $0.25$ to $0.4$ improves the EM5 accuracy from $78.2\%$ to $80.97\%$.

This also demonstrates that already at this scale, a larger $\ARTstrength$, i.e., a stronger activation replacement trick, can improve the overall accuracy of a bitonic sorting network compared to training with $\ARTstrength=0.25$.

As the size $n$ of training tuples increases, this also increases the overall number of observed images during training.
Therefore, in the left half of Table~\ref{table:results-bitonic-large}, we consider the accuracy for a constant total of observed images per iteration, i.e., for $n\times\mathrm{batch~size} = 4096$ (e.g., for $n=32$ this results in a batch size of $128$, while for $n=1024$, the batch size is only $4$).
In the right half of Table~\ref{table:results-bitonic-large}, we consider a constant batch size of $4$.

With increasing $n$, the accuracy of our model increases for a constant number of observed images, even though it has to operate on very small batch sizes.
This shows that training with larger ordered sets results in better accuracy.
This suggests that, if possible, larger $n$ should be prioritized over larger batch sizes and that good results can be achieved by using the largest possible $n$ for the available data to learn from all available information.

\begin{table*}[t]
    \centering
    \caption{
    Results the ART for large $n$ measured using the EM5 metric with fixed number of samples as well as a fixed batch size.
    Independent of the batch size, the model always performs better for larger $n$.
    Trained for $10^4$ steps \& averaged over $10$ runs.
    }
    \label{table:results-bitonic-large}
    \footnotesize
    \begin{tabular}{l|c|cccccc|cccccc}
        \toprule
        $\qquad\ARTstrength$      & $0.25$    & $0.4$     & $0.4$     & $0.4$     & $0.4$     & $0.4$     & $0.4$     & $0.4$     & $0.4$     & $0.4$     & $0.4$     & $0.4$     & $0.4$  \\
        \midrule
        $\qquad n$          & $32$      & $32$      & $64$      & $128$     & $256$     & $512$     & $1024$    & $32$      & $64$      & $128$     & $256$     & $512$     & $1024$  \\
        \midrule
        batch size          & $128$     & $128$     & $64$      & $32$      & $16$      & $8$       & $4$       & $4$       & $4$       & $4$       & $4$       & $4$       & $4$  \\
        \midrule
            $\steepness=30$   & $\pmb{78.20}$ & $79.89$ & $81.25$ & $\pmb{82.50}$ & $\pmb{82.05}$ & $82.50$ & $\pmb{82.80}$ & $71.08$ & $\pmb{75.88}$ & $79.43$ & $\pmb{81.46}$ & $82.98$ & $\pmb{82.80}$  \\
            $\steepness=32.5$ & $76.98$ & $79.62$ & $\pmb{81.66}$ & $80.15$ & $81.87$ & $82.64$ & $81.63$ & $\pmb{72.31}$ & $75.59$ & $\pmb{79.71}$ & $81.36$ & $\pmb{82.99}$ & $81.63$  \\
            $\steepness=35$   & $77.45$ & $80.93$ & $81.26$ & $80.72$ & $81.42$ & $81.51$ & $81.15$ & $71.15$ & $75.73$ & $78.81$ & $79.32$ & $82.30$ & $81.15$  \\
            $\steepness=37.5$ & $76.40$ & $80.02$ & $80.05$ & $81.50$ & $80.05$ & $\pmb{82.67}$ & $80.07$ & $70.69$ & $75.80$ & $79.11$ & $80.64$ & $82.70$ & $80.07$  \\
            $\steepness=40$   & $77.69$ & $\pmb{80.97}$ & $80.23$ & $81.55$ & $79.75$ & $81.89$ & $81.15$ & $70.20$ & $74.67$ & $78.14$ & $80.06$ & $81.39$ & $81.15$  \\
            \midrule
            mean              & $77.35$ & $80.29$ & $80.89$ & $81.28$ & $81.03$ & $82.24$ & $81.36$ & $71.09$ & $75.53$ & $79.04$ & $80.57$ & $82.47$ & $81.36$  \\
            best $\steepness$          & $\pmb{78.20}$ & $\pmb{80.97}$ & $\pmb{81.66}$ & $\pmb{82.50}$ & $\pmb{82.05}$ & $\pmb{82.67}$ & $\pmb{82.80}$ & $\pmb{72.31}$ & $\pmb{75.88}$ & $\pmb{79.71}$ & $\pmb{81.46}$ & $\pmb{82.99}$ & $\pmb{82.80}$  \\
            worst $\steepness$         & $76.40$ & $79.62$ & $80.05$ & $80.15$ & $79.75$ & $81.51$ & $80.07$ & $70.20$ & $74.67$ & $78.14$ & $79.32$ & $81.39$ & $80.07$  \\
        \bottomrule
    \end{tabular}%
\end{table*}

\begin{margintable}
    \centering
    \caption{
    Ablation Study: Evaluation of the ART ($\ARTstrength=0$ vs.~$\ARTstrength=0.25$) for $n=4$ and $n=32$ on the MNIST and the SVHN data set using odd-even and bitonic sorting networks.
    The displayed metric is EW.
    \label{tab:ablation-study}
    }
    {\footnotesize
    \addtolength{\tabcolsep}{-4pt}
    \begin{tabular}{lcccc}
        \toprule
        & \multicolumn{2}{c}{$n = 4$} & \multicolumn{2}{c}{$n = 32$} \\
        \cmidrule(l){2-3}\cmidrule(l){4-5}
        Setting~~/~~$\ARTstrength$ & $0$ & $0.25$ & $0$ & $0.25$ \\
        \midrule
        Odd-E (MNIST)    & $94.5$ & $\pmb{94.9}$ & $61.5$ & $\pmb{69.1}$  \\
        Bitonic (MNIST)     & $93.6$ & $\pmb{95.3}$ & $62.8$ & $\pmb{67.3}$  \\
        \midrule
        Odd-E (SVHN)     & $77.3$ & $\pmb{85.5}$ & $28.5$ & $\pmb{36.6}$  \\
        Bitonic (SVHN)      & $78.1$ & $\pmb{85.3}$ & $35.0$ & $\pmb{42.4}$  \\
        \bottomrule
    \end{tabular}}
\end{margintable}
\begin{marginfigure}
    \caption{
    Comparing different ART strengths $\ARTstrength$ for $n=8$ (top) and $n=16$ (bottom). Training with $\lambda \leq 0.5$ is stable.%
    \label{fig:different-art-strengths}%
    }
    \hspace{-.1\linewidth}%
    \resizebox{1.2\linewidth}{!}{
    \input{fig_diffsort/plot_rebuttal_2.pgf}
    }
\end{marginfigure}

\subsubsection{Ablation Study and Hyperparameter Sensitivity}
To assess the impact of the proposed activation replacement trick (ART), we evaluate both architectures with and without ART at $\ARTstrength=0.25$ in Table~\ref{tab:ablation-study}.
The accuracy improves by using the ART for small as well as for large $n$.
For large $n$, the activation replacement trick has a greater impact on the performance of both architectures.
For the bitonic sorting network at $n=32$, even a stronger ART with $\ARTstrength=0.4$ is beneficial as demonstrated in Table~\ref{table:results-bitonic-large}.
In Figure~\ref{fig:different-art-strengths}, we evaluate both differentiable sorting networks for varying ART intensities~$\ARTstrength$.
Here, performance increases with larger $\ARTstrength$s (i.e., with a stronger ART). For $\ARTstrength > 0.5$, the performance drops as $\varphi$ converges to a discrete step function for $\ARTstrength \to 1$.

\subsection{Monotonic Differentiable Sorting Networks}

In this section, we evaluate the performance of monotonic differentiable sorting networks in comparison to existing non-monotonic differentiable sorting methods as well as non-monotonic differentiable sorting networks like the ART variant.

\begin{figure*}[p]
    \centering
    \includegraphics[width=.49\linewidth]{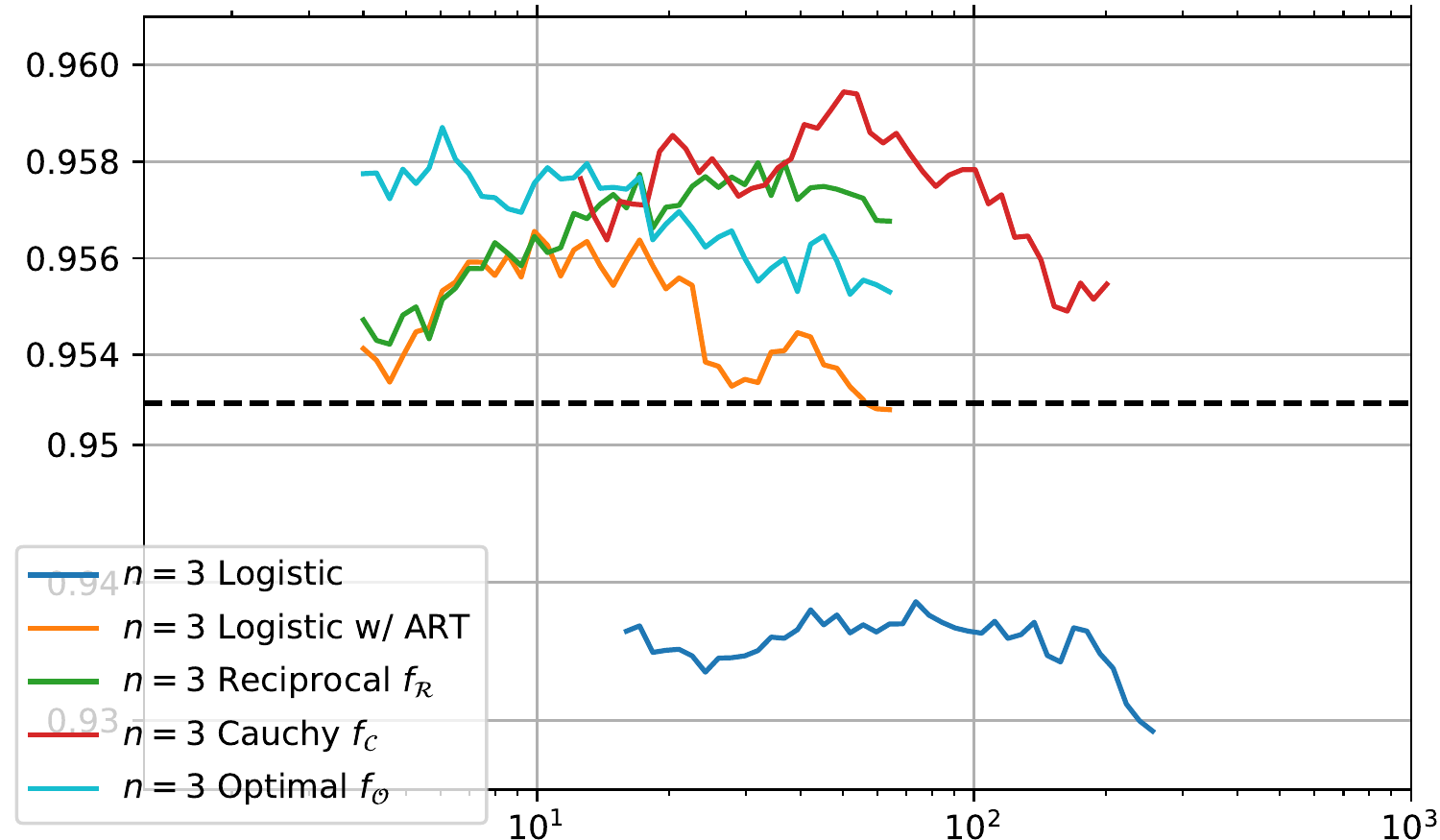}\hfill
    \includegraphics[width=.49\linewidth]{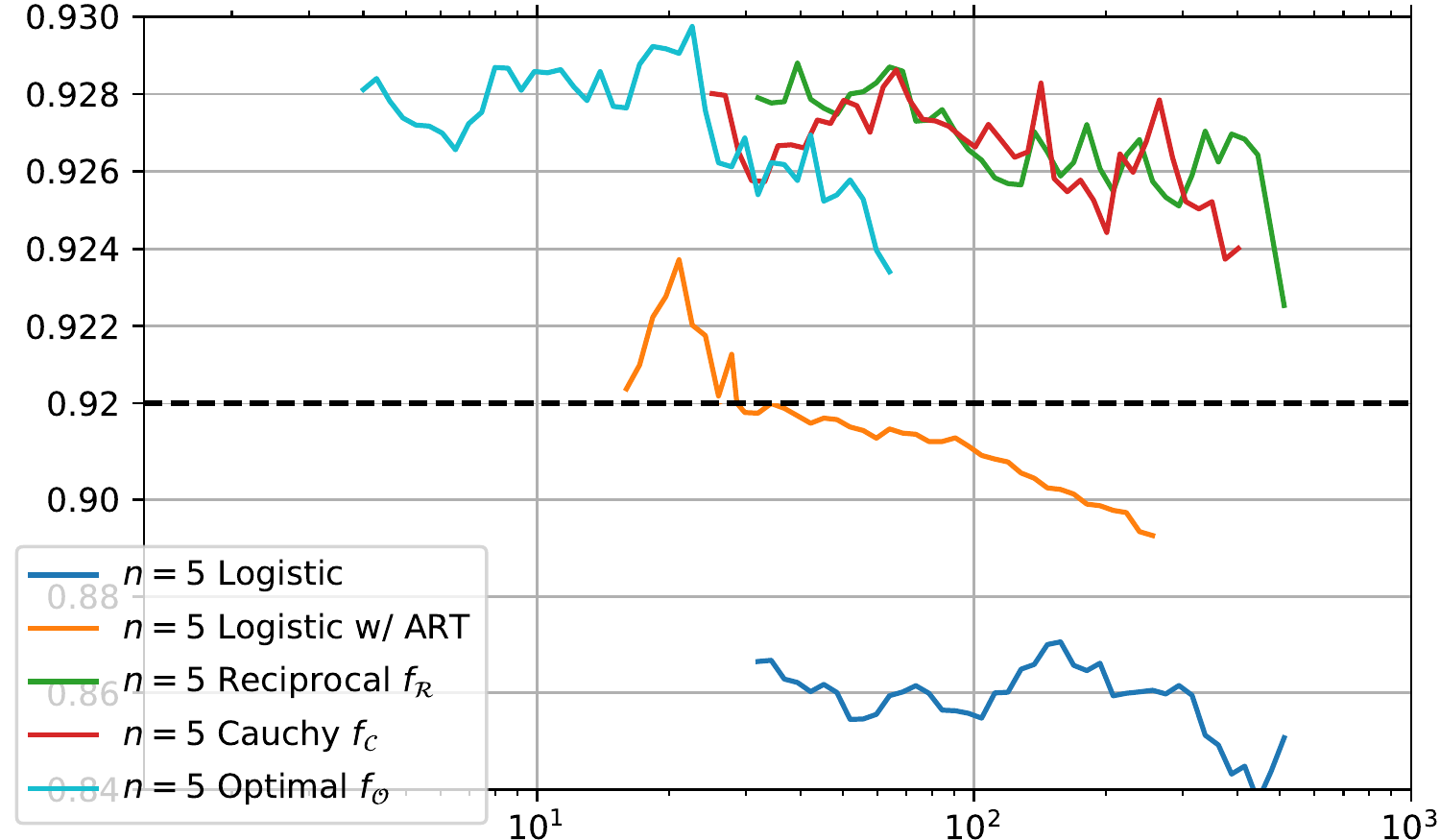}\\
    ~\hfill\resizebox{0.4em}{!}{$\beta$}\hfill\hfill\resizebox{0.4em}{!}{$\beta$}\hfill~\\
    \includegraphics[width=.49\linewidth]{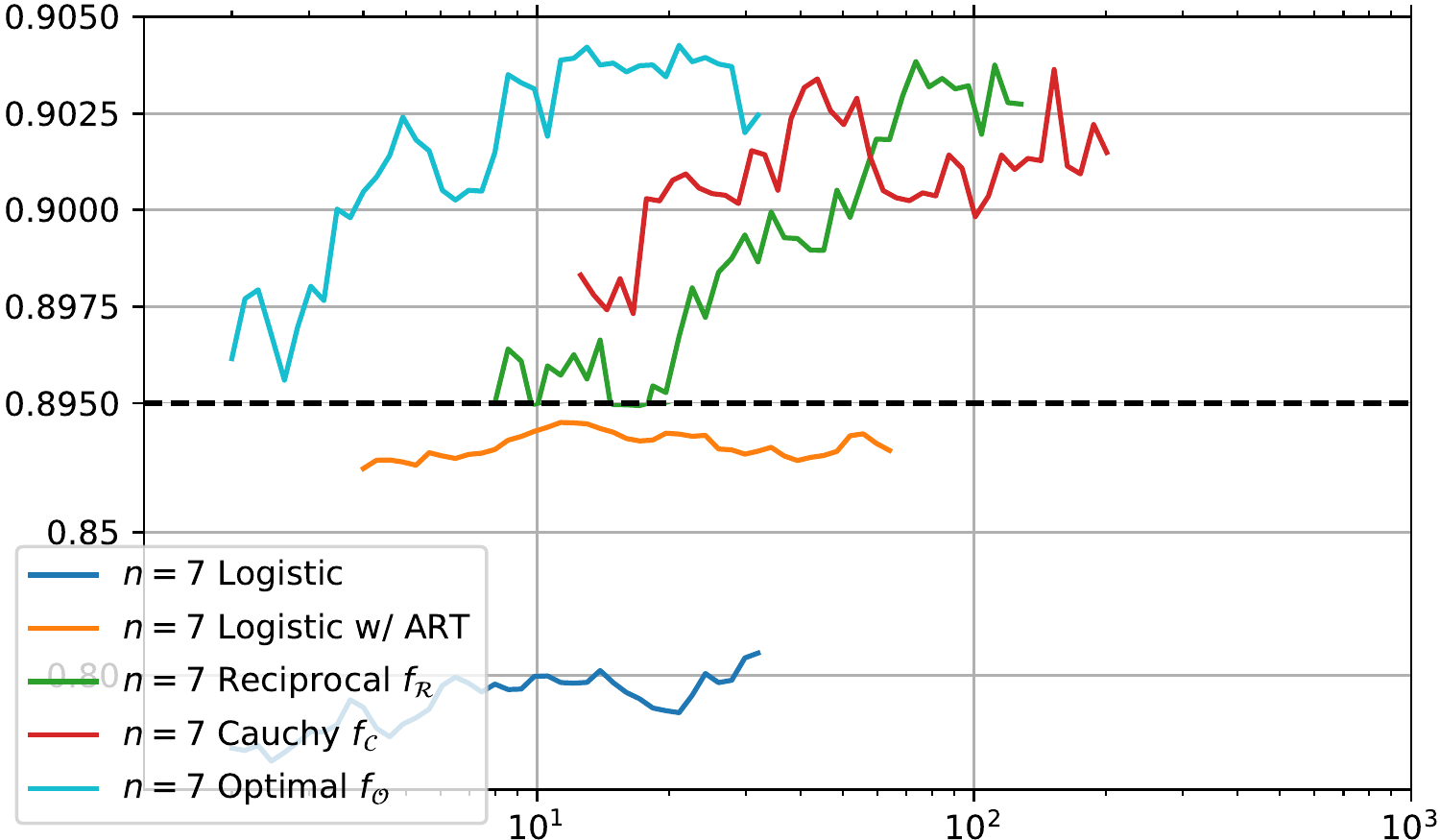}\hfill
    \includegraphics[width=.49\linewidth]{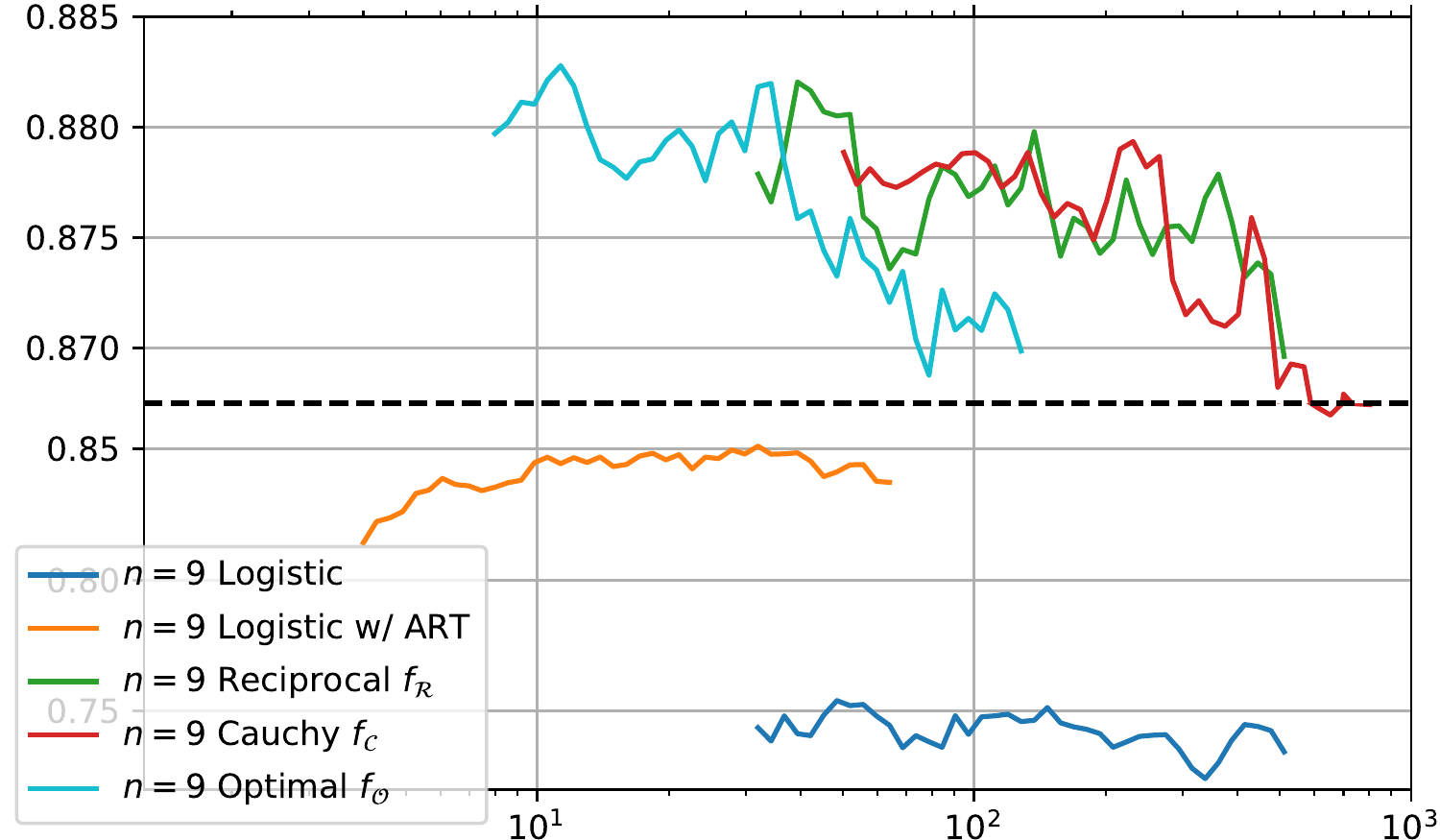}\\
    ~\hfill\resizebox{0.4em}{!}{$\beta$}\hfill\hfill\resizebox{0.4em}{!}{$\beta$}\hfill~\\
    \includegraphics[width=.49\linewidth]{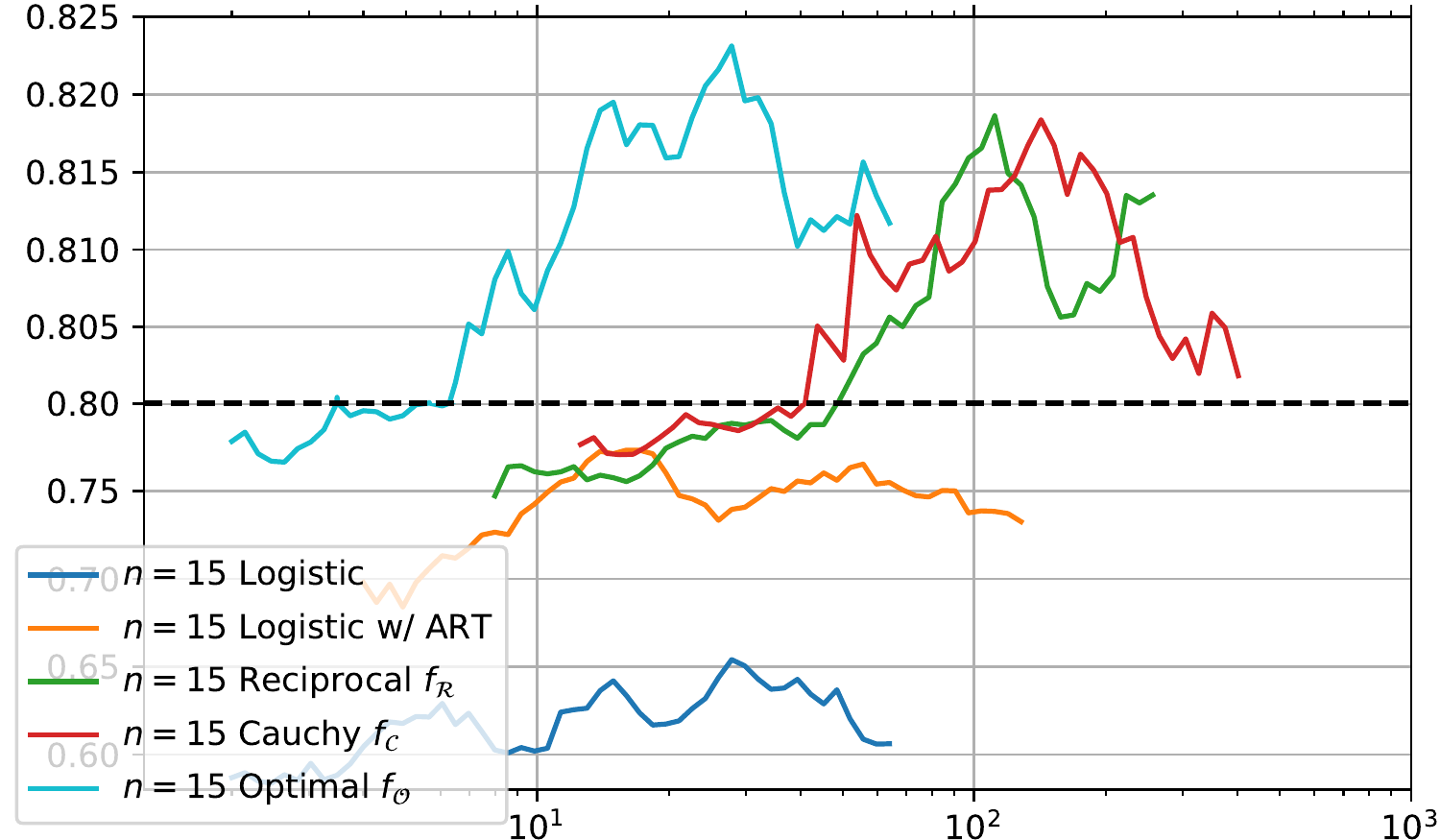}\hfill
    \includegraphics[width=.49\linewidth]{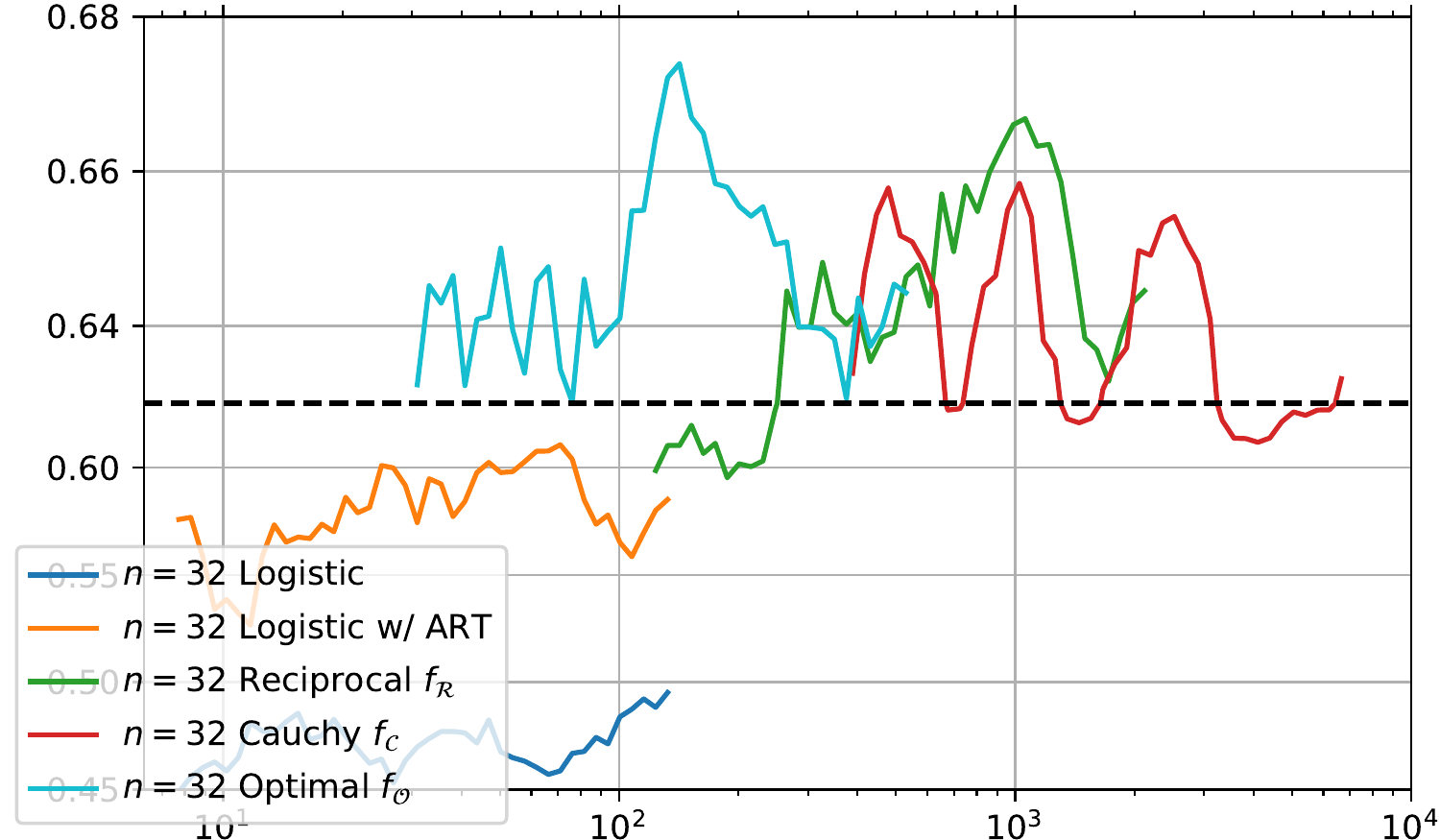}\\
    ~\hfill\resizebox{0.4em}{!}{$\beta$}\hfill\hfill\resizebox{0.4em}{!}{$\beta$}\hfill~\\
    \includegraphics[width=.49\linewidth]{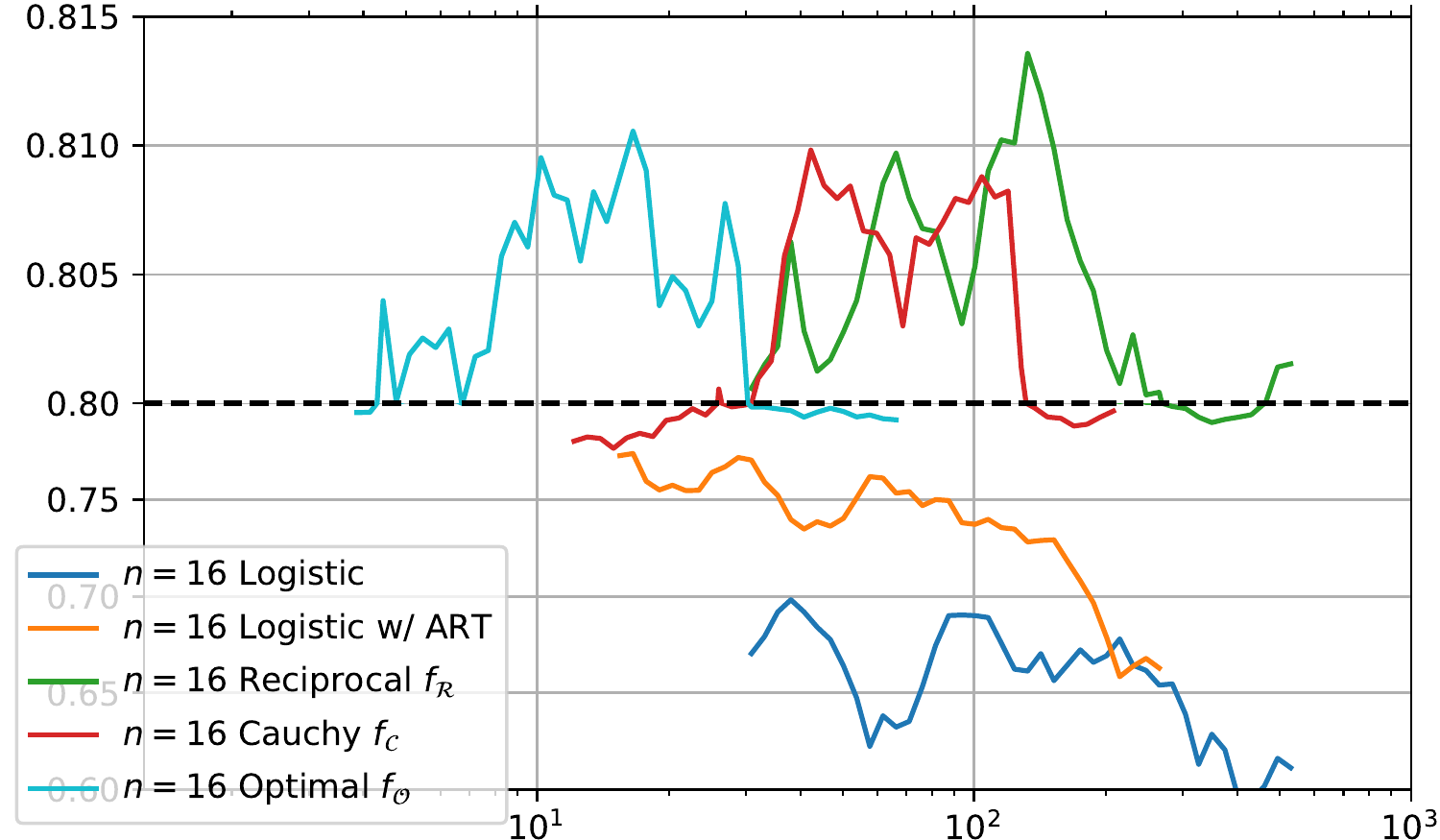}\hfill
    \includegraphics[width=.49\linewidth]{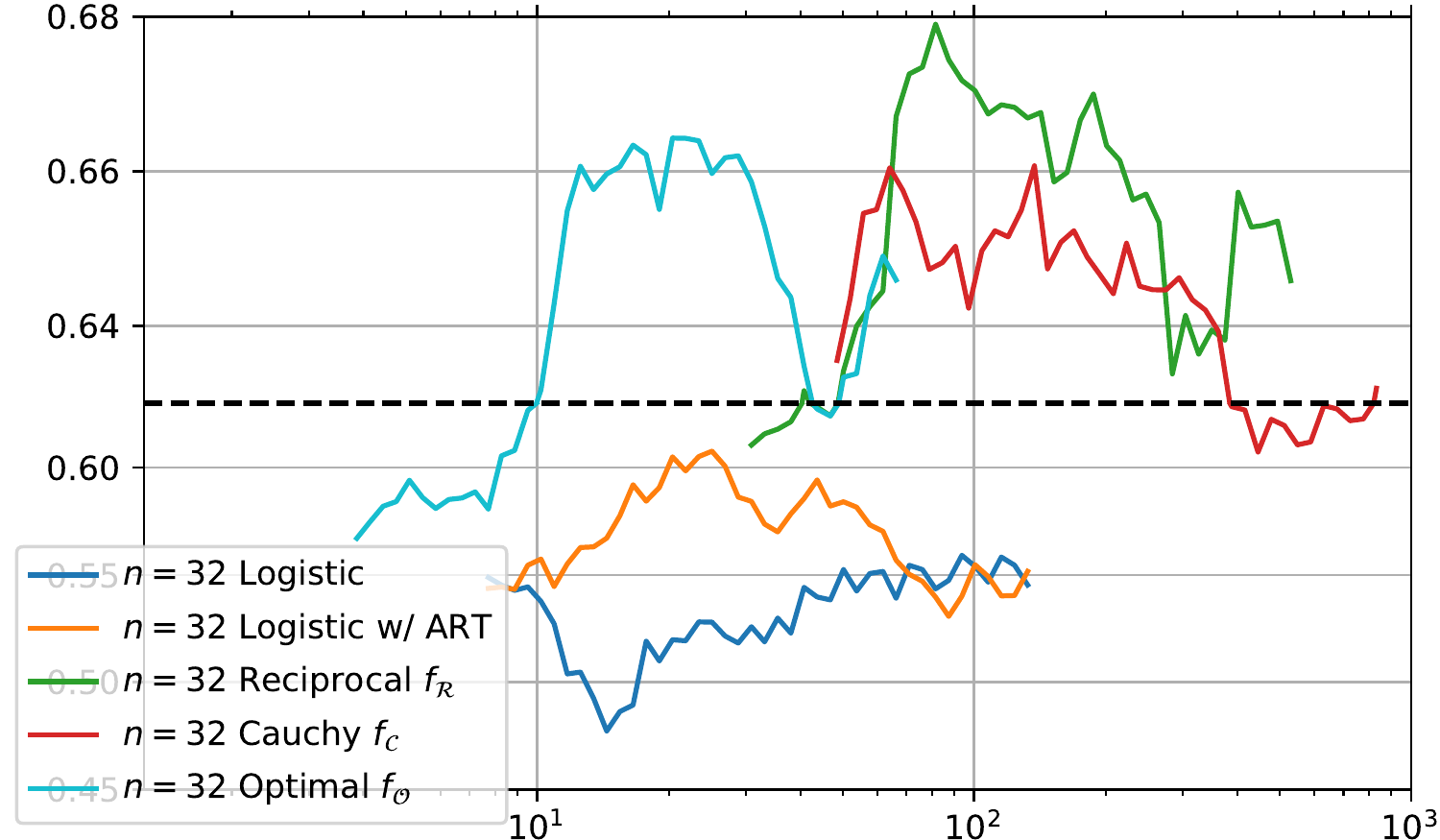}\\
    ~\hfill\resizebox{0.4em}{!}{$\beta$}\hfill\hfill\resizebox{0.4em}{!}{$\beta$}\hfill~\\
    \caption{
        Evaluating different sigmoid functions on the sorting MNIST task for ranges of different inverse temperatures $\beta$.
        The metric is the proportion of individual element ranks correctly identified (EW).
        In all settings, the monotonic sorting networks clearly outperform the non-monotonic ones.
        \textit{First 3 rows:} Odd-Even sorting networks with $n\in\{3,5,7,9,15,32\}$.
        \textit{Last row:} Bitonic network with $n\in\{16,32\}$.
        For small $n$ such as $3$, Cauchy performs best because it has a low error but is smooth at the same time.
        For larger $n$ such as $15$ and $32$, the optimal sigmoid function (wrt.~error) $f_\mathcal{O}$ performs better because it, while not being smooth, has the smallest possible approximation error which is more important for deeper networks.
        For the bitonic network with its more complex structure at $n=16$ and $n=32$ (last row), the reciprocal sigmoid $f_\mathcal{R}$ performs best.
    }
    \label{fig:sigmoid-behavior}
\end{figure*}

To understand the behavior of the proposed monotonic functions compared to the logistic sigmoid function, we evaluate all sigmoid functions for different inverse temperatures $\beta$ during training.
We investigate eight settings:
odd-even networks for $n\in\{3,5,7,9,15,32\}$ and a bitonic sorting network with $n\in\{15,32\}$ on the MNIST data set.
Notably, there are $15$ layers in the bitonic sorting networks with $n=32$, while the odd-even networks for $n=15$ also has $15$ layers.
We display the results of this evaluation in Figure \ref{fig:sigmoid-behavior}.
Note that, here, we train for $10^5$ training steps to reduce the computational cost.

We observe that the optimal inverse temperature depends on the number of layers, rather than the overall number of samples $n$.
This can be seen when comparing the peak accuracy of each function for the odd-even sorting network for different $n$ and thus for different numbers of layers.
The bitonic network for $n=32$ (last row, right) has the same number of layers as $n=15$ in the odd-even network (third row, left).
Here, the peak performances for each sigmoid function fall within the same range, whereas the peak performances for the odd-even network for $n=32$ (bottom left) are shifted almost an order of magnitude to the right.
For all configurations, the proposed sigmoid functions for monotonic sorting networks improve over the standard logistic sigmoid function, as well as the ART.

\begin{table*}[t]
    \caption{
    Results on the four-digit MNIST and SVHN tasks using the same architecture as previous works \cite{grover2019neuralsort, cuturi2019differentiable, blondel2020fast, petersen2021diffsort}.
    The metric is the proportion of rankings correctly identified (EM), and the value in parentheses is the proportion of individual element ranks correctly identified (EW).
    All results are averaged over $5$ runs. SVHN w/ $n=32$ is omitted to reduce the carbon impact of the evaluation. 
    We train each differentiable sorting network-based model for up to $2\cdot10^5$ steps for MNIST, and for up to $10^6$ steps for SVHN.
    }
    \label{table:results-mnist-svhn}
    \centering
    \newcommand{\emfivespacer}[0]{\phantom{\ 00.0}}
    \newcommand{\minispace}[0]{}
    \newcommand{\resFor}[2]{$#1\ (#2)$}
    \resizebox{\linewidth}{!}{
    \begin{tabular}{lcccccc|cc}
        \toprule
        \textbf{MNIST} & $n=3$ & $n=5$ & $n=7$ & $n=9$ & $n=15$ & $n=32$ & $n=16$ & $n=32$ \\
        &&&&&&&(bitonic)&(bitonic) \\
        \midrule
        NeuralSort         & \resFor{91.9}{94.5} & \resFor{77.7}{90.1} & \resFor{61.0}{86.2} & \resFor{43.4}{82.4} & \resFor{\phantom{0}9.7}{71.6}  & \resFor{\phantom{0}0.0}{38.8} & --- & --- \\
        Sinkhorn Sort       & \resFor{92.8}{95.0} & \resFor{81.1}{91.7} & \resFor{65.6}{88.2} & \resFor{49.7}{84.7} & \resFor{12.6}{74.2} & \resFor{\phantom{0}0.0}{41.2}  & --- & --- \\
        Fast Sort \& Rank       & \resFor{90.6}{93.5} & \resFor{71.5}{87.2} & \resFor{49.7}{81.3} & \resFor{29.0}{75.2} & \resFor{\phantom{0}2.8}{60.9} & --- & --- & --- \\
            Diffsort (Logistic) &  \resFor{92.0}{94.5}  &  \resFor{77.2}{89.8}  &  \resFor{54.8}{83.6}  &  \resFor{37.2}{79.4}  &  \resFor{4.7}{62.3}  &  \resFor{0.0}{56.3}  &  \resFor{10.8}{72.6}  &  \resFor{0.3}{63.2} \\
            Diffsort (Log.~w/~ART)  &  \resFor{94.3}{96.1}  &  \resFor{83.4}{92.6}  &  \resFor{71.6}{90.0}  &  \resFor{56.3}{86.7}  &  \resFor{23.5}{79.4}  &  \resFor{0.5}{64.9}  &  \resFor{19.0}{77.5}  &  \resFor{0.8}{63.0} \\
        \midrule
            $\sigmoidReciprocal$ : Reciprocal Sigmoid &  \resFor{94.4}{96.1}  &  \resFor{85.0}{93.3}  &  \resFor{73.4}{90.7}  &  \resFor{60.8}{88.1}  &  \resFor{30.2}{81.9}  &  \resFor{1.0}{66.8}  &  \resFor{28.7}{82.1}  &  \resFor{1.3}{68.0} \\
            $\sigmoidCauchy$ : Cauchy CDF &  \resFor{94.2}{96.0}  &  \resFor{84.9}{93.2}  &  \resFor{73.3}{90.5}  &  \resFor{63.8}{89.1}  &  \resFor{31.1}{82.2}  &  \resFor{0.8}{63.3}  &  \resFor{29.0}{82.1}  &  \resFor{1.6}{68.1} \\
            $\sigmoidOptimal$ : Optimal Sigmoid &  \resFor{94.6}{96.3}  &  \resFor{85.0}{93.3}  &  \resFor{73.6}{90.7}  &  \resFor{62.2}{88.5}  &  \resFor{31.8}{82.3}  &  \resFor{1.4}{67.9}  &  \resFor{28.4}{81.9}  &  \resFor{1.4}{67.7} \\[.25em]
        \toprule
        \textbf{SVHN} & $n=3$ & $n=5$ & $n=7$ & $n=9$ & $n=15$ & --- & $n=16$ & --- \\
        &&&&&&&(bitonic)&  \\
        \midrule
            Diffsort (Logistic) &  \resFor{76.3}{83.2}  &  \resFor{46.0}{72.7}  &  \resFor{21.8}{63.9}  &  \resFor{13.5}{61.7}  &  \resFor{0.3}{45.9}  &  ---  &  \resFor{1.2}{50.6}  &  --- \\
            Diffsort (Log.~w/~ART)  &  \resFor{83.2}{88.1}  &  \resFor{64.1}{82.1}  &  \resFor{43.8}{76.5}  &  \resFor{24.2}{69.6}  &  \resFor{2.4}{56.8}  &  ---  &  \resFor{3.4}{59.2}  &  --- \\
            \midrule
            $\sigmoidReciprocal$ : Reciprocal Sigmoid &  \resFor{85.7}{89.8}  &  \resFor{68.8}{84.2}  &  \resFor{53.3}{80.0}  &  \resFor{40.0}{76.3}  &  \resFor{13.2}{66.0}  &  ---  &  \resFor{11.5}{64.9}  &  --- \\
            $\sigmoidCauchy$ : Cauchy CDF &  \resFor{85.5}{89.6}  &  \resFor{68.5}{84.1}  &  \resFor{52.9}{79.8}  &  \resFor{39.9}{75.8}  &  \resFor{13.7}{66.0}  &  ---  &  \resFor{12.2}{65.6}  &  --- \\
            $\sigmoidOptimal$ : Optimal Sigmoid &  \resFor{86.0}{90.0}  &  \resFor{67.5}{83.5}  &  \resFor{53.1}{80.0}  &  \resFor{39.1}{76.0}  &  \resFor{13.2}{66.3}  &  ---  &  \resFor{10.6}{66.8}  &  --- \\
        \bottomrule
    \end{tabular}%
    }
\end{table*}
\begin{table}[t]
    \centering
    \caption{
    Inverse temperatures $\beta$ used for Table~\ref{table:results-mnist-svhn}, which correspond to the optima in Figure~\ref{fig:sigmoid-behavior}.
    }
    \label{tab:diffsort-beta}
{\footnotesize
\addtolength{\tabcolsep}{-3pt}
    \begin{tabular}{llcccccc|cc}
        \toprule
        Method & $n$ & $3$ & $5$ & $7$ & $9$ & $15$ & $32$ & $16$ & $32$ \\
        & odd-even / bitonic &oe&oe&oe&oe&oe&oe&bi&bi \\
        \midrule
            $\sigma$ & Logistic  &
  $79$ & $30$ & $33$ & $54$ & $32$ & $128$ & $43$ & $8$  \\
            \midrule
            $\sigma \circ \varphi$ & Log.~w/~ART   &
  $15$ & $20$ & $13$ & $34$ & $16$ & $29$ & $28$ & $26$  \\
            \midrule
            $\sigmoidReciprocal$ & Reciprocal Sigmoid  &
  $14$ & $60$ & $69$ & $44$ & $120$ & $1140$ & $124$ & $76$  \\
            \midrule
            $\sigmoidCauchy$ & Cauchy CDF  &
  $14.5\pi$ & $51\pi$ & $71\pi$ & $15\pi$ & $40\pi$ & $169\pi$ & $12\pi$ & $48.5\pi$  \\
&   &  $45.6$ & $160.2$ & $223.1$ & $47.1$ & $125.7$ & $531.0$ & $37.7$ & $152.4$  \\
            \midrule
            $\sigmoidOptimal$ & Optimal Sigmoid &
  $6$ & $20$ & $29$ & $32$ & $25$ & $124$ & $17$ & $25$  \\
        \bottomrule
    \end{tabular}%
}
\end{table}

\paragraph{Comparison to Other Differentiable Sorting Methods}
We compare the proposed functions to other state-of-the-art approaches using the same network architecture and training setup as used in previous works, as well as among themselves.
We report the results in Table~\ref{table:results-mnist-svhn}.
Here, for each setting, we use the inverse temperature, which performed best in Figure~\ref{fig:sigmoid-behavior}; the concrete choices of $\beta$ can be found in Table~\ref{tab:diffsort-beta}.
The proposed monotonic differentiable sorting networks outperform current state-of-the-art methods by a considerable margin.
Especially for those cases where more samples needed to be sorted, the gap between monotonic sorting nets and other techniques grows with larger $n$.

Comparing the three proposed functions among themselves, we observe that for odd-even networks on MNIST, the error-optimal function $\sigmoidOptimal$ performs best.
This is because here the approximation error is small.
However, for the more complex bitonic sorting networks, $\sigmoidCauchy$ (Cauchy) performs better than $\sigmoidOptimal$.
This is because $\sigmoidOptimal$ does not provide a higher-order smoothness and is only $C^1$ smooth, while the Cauchy function $\sigmoidCauchy$ is analytic and $C^\infty$ smooth.

\subsection{Runtime and Memory Analysis}
\label{sec:evaluation-runtime}
Finally, we report the runtime and memory consumption of differentiable sorting networks in Table~\ref{tab:runtime-memory}.
For GPU runtimes, we use a native CUDA implementation and measure the time and memory for sorting $n$ input elements including forward and backward pass.
For CPU runtimes, we use a PyTorch \cite{paszke2019pytorch} implementation.
For a small number of input elements, the odd-even and bitonic sorting networks have around the same time and memory requirements, while for larger numbers of input elements, bitonic is much faster than odd-even.

The asymptotic runtime of differentiable odd-even sort is in $\mathcal{O}(n^3)$ and for bitonic sort the runtime is in $\mathcal{O}(n^2 (\log n)^2)$.
Note that, for this, the matrix multiplication in Equation~\eqref{eq:p-mult} is a sparse matrix multiplication.
We also report runtimes for other differentiable sorting and ranking methods.
For large $n$, we empirically confirm that FastRank \cite{blondel2020fast} is the fastest method, i.a., because it produces only output ranks / sorted output values and not differentiable permutation matrices.
Note that differentiable sorting networks also produce sorted output values.
Computing only sorted output values is significantly faster than computing the full differentiable permutation matrices; however, for the effective cross-entropy training objective, differentiable permutation matrices are necessary.

\begin{table*}[t]
    \caption{
    Runtimes, memory requirements, and number of layers for sorting $n$ elements. %
    Runtimes reported for an Nvidia GTX 1070. We include NeuralSort \cite{grover2019neuralsort}, FastRank \cite{blondel2020fast}, and OT Sort \cite{cuturi2019differentiable}.
    }
    \label{tab:runtime-memory}
    \centering
    \newcommand{\mBi}{{\footnotesize$\mathrm{B}$}}
    \newcommand{\mKB}{{\footnotesize$\mathrm{KB}$}}
    \newcommand{\mMB}{{\footnotesize$\mathrm{MB}$}}
    \newcommand{\mGB}{{\footnotesize$\mathrm{GB}$}}
    \newcommand{\rtbold}[1]{{\bm{#1}}}
    \resizebox{1.\linewidth}{!}{
    \begin{tabular}{crrrrrrrrrrrrrrrrrrr}
        \toprule
                & \multicolumn{4}{c}{Differentiable Odd-Even Sort}                    & \multicolumn{4}{c}{Differentiable Bitonic Sort}                    && \multicolumn{2}{c}{NeuralSort}                    & {FastRank}                    & {OT Sort}                          \\
            \cmidrule(r){2-5}\cmidrule(r){6-9}\cmidrule(r){11-12}\cmidrule(r){13-13}\cmidrule(r){14-14}
        $n$     & GPU & CPU & Memory & \# Layers                                          & GPU & CPU & Memory & \# Layers                                                  && GPU & CPU                                         & CPU                           & CPU                                            \\
        \midrule
        $4$      & $\rtbold{69\,\mathrm{ns}}$     &  $1.9\,\mu\mathrm{s}$    & $1$\mKB     & $4$        & $\rtbold{52\,\mathrm{ns}}$       & $1.3\,\mu\mathrm{s}$     & $840$\mBi  & $3$            && $\rtbold{145\,\mathrm{ns}}$       & $7.1\,\mu\mathrm{s}$  & $\rtbold{189\,\mu\mathrm{s}}$          & $\rtbold{1.0\,\mathrm{ms}}$   \\
        $16$     & $\rtbold{1.2\,\mu\mathrm{s}}$   &   $54\,\mu\mathrm{s}$   & $42$\mKB    & $16$       & $\rtbold{759\,\mathrm{ns}}$      & $40\,\mu\mathrm{s}$      & $28$\mKB  & $10$            && $\rtbold{396\,\mathrm{ns}}$       & $11\,\mu\mathrm{s}$   & $\rtbold{215\,\mu\mathrm{s}}$          & $\rtbold{7.5\,\mathrm{ms}}$   \\
        $32$     & $\rtbold{7.4\,\mu\mathrm{s}}$  &  $309\,\mu\mathrm{s}$    & $315$\mKB   & $32$       & $\rtbold{3.5\,\mu\mathrm{s}}$    & $159\,\mu\mathrm{s}$     & $152$\mKB & $15$            && $\rtbold{969\,\mathrm{ns}}$       & $13\,\mu\mathrm{s}$   & $\rtbold{303\,\mu\mathrm{s}}$          & $\rtbold{17\,\mathrm{ms}}$      \\
        $128$    & $\rtbold{493\,\mu\mathrm{s}}$  & $19\,\mathrm{ms}$        & $20.2$\mMB  & $128$      & $\rtbold{97\,\mu\mathrm{s}}$     & $5\,\mathrm{ms}$         & $4.1$\mMB & $28$            && $\rtbold{12\,\mu\mathrm{s}}$      & $177\,\mu\mathrm{s}$  & $\rtbold{834\,\mu\mathrm{s}}$          & $\rtbold{55\,\mathrm{ms}}$        \\
        $1\,024$ & $\rtbold{660\,\mathrm{ms}}$     & $31\,\mathrm{s}$         & $4.9$\mGB  & $1\,024$   & $\rtbold{15\,\mathrm{ms}}$       & $1.7\,\mathrm{s}$        & $549$\mMB & $55$            && $\rtbold{1.2\,\mathrm{ms}}$       & $11\,\mathrm{ms}$     & $\rtbold{4.8\,\mathrm{ms}}$            & $\rtbold{754\,\mathrm{ms}}$      \\
        \bottomrule
    \end{tabular}%
    }
\end{table*}

\subsection*{Conclusion}

In this chapter, we presented differentiable sorting networks for training based on sorting and ranking supervision.
To this end, we relaxed the discrete $\min$ and $\max$ operators necessary for pairwise swapping in traditional sorting network architectures via perturbation.
We proposed an activation replacement trick to avoid the problems of vanishing gradients as well as blurred values for deep sorting networks and showed that it is possible to robustly sort and rank even long sequences on large input sets of up to at least $1024$ elements.
Further, we addressed and analyzed monotonicity and error-boundedness in differentiable sorting and ranking operators.
Specifically, we focused on differentiable sorting networks and presented a family of sigmoid functions that preserve monotonicity and bound approximation errors in differentiable sorting networks.
This makes the sorting functions quasiconvex, and we empirically observe that the resulting method outperforms the state-of-the-art in differentiable sorting supervision.
In the next chapter, we build on these ideas to present differentiable top-$k$ learning.

\setchapterpreamble[u]{\pagelogo{difftopk}\margintoc}
\chapter[Differentiable Top-\texorpdfstring{$\protect k$}{k}]{Differentiable Top-\texorpdfstring{$\protect k$}{k}}
\label{ch:difftopk}

After discussing differentiable sorting and ranking in the previous chapter, this chapter covers the related top-$k$ operator and proposes novel relaxations of it.
We use differentiable top-$k$ in a new learning setting, which we term differentiable top-$k$ classification learning.

Classification is one of the core disciplines in machine learning and computer vision.
The advent of classification problems with hundreds or even thousands of classes let the top-$k$ classification accuracy (i.e., one of the top-$k$ classes has to be the correct class) be established as an important metric.
Usually, models are trained to optimize the top-$1$ accuracy; and top-$5$ etc.~are used for evaluation only.
Some works \cite{lapin2016loss,berrada2018smooth} have challenged this idea and proposed top-$k$ losses, such as a smooth top-$5$ margin loss.
These methods have demonstrated superior robustness over the established top-$1$ softmax cross-entropy in the presence of additional label noise \cite{berrada2018smooth}.
In standard classification settings, however, these methods have so far not shown improvements over the established top-$1$ softmax cross-entropy.

In this chapter, we challenge the idea of selecting a single top-$k$ metric such as top-$1$ or top-$5$ for defining the loss.
Instead,
we propose to specify $k$ to be drawn from a distribution $P_K$, which may or may not depend on the confidence of specific data points or on the class label.
Examples for distributions $P_K$ are $[.5, 0, 0, 0, .5]$ ($50\%$ top-$1$ and $50\%$ top-$5$), $[.1, 0, 0, 0, .9]$ ($10\%$ top-$1$ and $90\%$ top-$5$), and $[.2, .2, .2, .2, .2]$ ($20\%$ top-$k$ for each $k$ from $1$ to $5$).
Note that, when $k$ is drawn from a distribution, this is done sampling-free as we can compute the expectation value in closed form.

Conventionally, given scores returned by a neural network, softmax produces a probability distribution over the top-$1$ rank.
Recent advances in differentiable sorting and ranking, as discussed in the previous chapter, provide methods for generalizing this to probability distributions over all ranks represented by a matrix $\mP$.
Based on differentiable ranking, multiple differentiable top-$k$ operators have recently been proposed.
They found applications in differentiable $k$-nearest neighbor algorithms~\cite{shvetsova2022differentiable, grover2019neuralsort}, differentiable beam search~\cite{xie2020differentiable}, attention mechanisms~\cite{xie2020differentiable}, and differentiable image patch selection \cite{cordonnier2021differentiable}.
In these areas, integrating differentiable top-$k$ improved results considerably by creating a more natural end-to-end learning setting.
However, to date, none of the differentiable top-$k$ operators have been employed as neural network losses for top-$k$ classification learning with $k>1$.

Building on differentiable sorting and ranking methods, we propose a new family of differentiable top-$k$ classification losses where $k$ is drawn from a probability distribution.
We find that our top-$k$ losses improve not only top-$k$ accuracies, but
also top-$1$ accuracy on multiple learning tasks.

We empirically evaluate our method using four differentiable sorting and ranking methods on the CIFAR-100~\cite{krizhevsky2009cifar10}, ImageNet-1K~\cite{deng2009imagenet}, and the ImageNet-21K-P~\cite{ridnik2021imagenet} data sets.
Using CIFAR-100, we demonstrate the capabilities of our losses to train models from scratch.
On ImageNet-1K, we demonstrate that our losses are capable of fine-tuning existing models and achieve a new state-of-the-art for publicly available models on both top-$1$ and top-$5$ accuracy.
We benchmark our method on multiple recent models and demonstrate that our proposed method consistently outperforms the baselines for the best two differentiable sorting and ranking methods.
With ImageNet-21K-P, where many classes overlap (but only one is the ground truth), we demonstrate that our losses are scalable to more than $10\,000$ classes and achieve improvements of over $1\%$ with only last layer fine-tuning.

Overall, while the performance improvements on fine-tuning are rather limited (because we retrain only the classification head), they are consistent and can be achieved without the large costs of training from scratch.
The absolute $0.2\%$ improvement that we achieve on the ResNeXt-101 32x48d WSL top-$5$ accuracy corresponds to an error reduction by approximately $10\%$, and can be achieved at much less than
the computational cost of \hbox{(re-)training} the full model in the first place.

\section{Related Work}

We structure the related work into three broad sections: works that derive and apply differentiable top-$k$ operators, works that use ranking and top-$k$ training objectives in general, and works that present classic selection networks.

\subsection{Differentiable Top-\texorpdfstring{$\protect k$}{k} Operators}

As top-$k$ can be reduced to sorting and ranking, the methods presented in the previous chapter can be used to obtain a differentiable top-$k$ operator.
However, in this chapter, we discuss how to improve over the vanilla top-$k$ operators derived from differentiable sorting and ranking.

Grover~\etal~\cite{grover2019neuralsort} include an experiment where they use the NeuralSort differentiable top-$k$ operator for $k$-nearest-neighbor ($k$NN) learning.
Cuturi~\etal~\cite{cuturi2019differentiable}, Blondel~\etal~\cite{blondel2020fast}, and Petersen~\etal~\cite{petersen2021diffsort} each apply their differentiable sorting and ranking methods to top-$k$ supervision with $k=1$.

Xie~\etal~\cite{xie2020differentiable} propose a differentiable top-$k$ operator based on optimal transport and the Sinkhorn algorithm \cite{cuturi2013sinkhorn}.
They apply their method to $k$NN learning, differential beam search with sorted soft top-$k$, and top-$k$ attention for machine translation.
Cordonnier~\etal~\cite{cordonnier2021differentiable} use perturbed optimizers \cite{berthet2020learning} to derive a differentiable top-$k$ operator, which they use for differentiable image patch selection.
Lee~\etal~\cite{lee2021differentiable} propose using NeuralSort for a differentiable top-$k$ operator to produce differentiable ranking metrics for recommender systems.
Goyal~\etal~\cite{goyal2018continuous} propose a continuous top-$k$ operator for differentiable beam search.
Pietruszka~\etal~\cite{pietruszka2020successive} propose the differentiable successive halving top-$k$ operator to approximate the normalized Chamfer Cosine Similarity ($nCCS@k$).

\subsection{Ranking and Top-\texorpdfstring{$\protect k$}{k} Training Objectives}

Fan~\etal~\cite{fan2017learning} propose the ``average top-$k$'' loss, an aggregate loss that averages over the $k$ largest individual losses of a training data set.
They apply this aggregate loss to SVMs for classification tasks.
Note that this is not a differentiable top-$k$ loss.
Here, the top-$k$ operator is not differentiable, and instead used for deciding which data points' losses are aggregated into the loss.

Lapin~\etal~\cite{lapin2015topk,lapin2016loss} propose relaxed top-$k$ surrogate error functions for multiclass SVMs.
Inspired by learning-to-rank losses, they propose top-$k$ calibration, a top-$k$ hinge loss, a top-$k$ entropy loss, as well as a truncated top-$k$ entropy loss.
They apply their method to multiclass SVMs and learn via stochastic dual coordinate ascent (SDCA).

Berrada~\etal~\cite{berrada2018smooth} build on these ideas and propose smooth loss functions for deep top-$k$ classification.
Their differentiable surrogate top-$k$ loss achieves good performance on the CIFAR-100 and ImageNet1K tasks.
While their method does not improve performance on the raw data sets in comparison to the strong Softmax Cross-Entropy baseline, in settings of label noise and data set subsets, they improve classification accuracy.
Specifically, with label noise of $20\%$ or more on CIFAR-100, they improve top-$1$ and top-$5$ accuracy and for subsets of ImageNet1K of up to $50\%$ they improve top-$5$ accuracy.
In contrast to \cite{berrada2018smooth}, our method improves classification accuracy in unmodified settings.
In our experiments, for the special case of $k$ being a concrete integer and not being drawn from a distribution, we provide comparisons to the smooth top-$k$ surrogate loss.

Yang~\etal~\cite{yang2020consistency} provide a theoretical analysis of top-$k$ surrogate losses
as well as produce a new surrogate top-$k$ loss, which they evaluate in synthetic data experiments.

A related idea is set-valued classification, where a set of labels is predicted. We refer to Chzhen~\etal~\cite{chzhen2021set} for an extensive overview. We note that our goal is not to predict a set of labels, but instead we return a score for each class corresponding to a ranking, where only one class can correspond to the ground truth.

\subsection{Selection Networks}

Previous selection networks have been proposed by Wah~\etal~\cite{wah1984partitioning}, Zazon-Ivry~\etal~\cite{zazonivry2012pairwise}, and Karpi\'nski~\etal~\cite{karpinski2015smaller}, among others.
All of these are based on classic divide-and-conquer sorting networks, which recursively sort subsequences and merge them.
In selection networks, during merging, only the top-$k$ elements are merged instead of the full (sorted) subsequences.
In comparison to those earlier works, we propose a new class of selection networks, which achieve tighter bounds (for $k\ll n$), and relax them.

\section{Differentiable Top-\texorpdfstring{$\protect k$}{k}}

The differentiable sorting algorithms discussed in Chapter~\ref{ch:diffsort} produce relaxed permutation matrices of size $n\times n$.
However, for top-$k$ classification learning, we require only the top $k$ rows for the number $k$ of top-ranked classes to consider.
Here, $k$ is the largest $k$ that is considered for the objective, i.e., where $P_K(k)>0$.
As $n\gg k$, producing a $k\times n$ matrix instead of a $n\times n$ matrix is much faster.

\paragraph{NeuralSort and SoftSort}
For NeuralSort and SoftSort, it is possible to simply compute only the top rows, as the algorithm is defined row-wise.

\paragraph{Optimal Transport / Sinkhorn Sort}
For optimal transport / Sinkhorn sort~\cite{cuturi2019differentiable}, it is not directly possible to improve the runtime, as the full matrix is required in each Sinkhorn iteration. 
However, Xie~\etal~\cite{xie2020differentiable} proposed an optimal transport and Sinkhorn-based differentiable top-$k$ operator, which computes a $2\times n$ matrix where the first row corresponds to the top-$k$ elements and the second row correspond to the remaining elements.
As this formulation does not directly support distinguishing between the placements of the top-$k$ elements among each other, which is necessary for the proposed top-$k$ learning objective, we use the SinkhornSort algorithm by Cuturi~\etal~\cite{cuturi2019differentiable}, which we also used in the previous chapter.

\paragraph{Differentiable Sorting Networks}
For differentiable sorting networks, it is possible to reduce the cost from $\mathcal{O}(n^2 \log^2 (n))$ to $\mathcal{O}(nk \log^2 (n))$ via a bi-directional evaluation.
Here, it is important to note the shape and order of multiplications for obtaining $P$.
As we only need those elements, which are (after the last layer of the sorting network) at the top $k$ ranks that we want to consider, we can omit all remaining rows of the permutation matrix of the last layer (layer $t$) and thus it is only of size $(k\times n)$.
\begin{equation}
    \underbrace{(k\times n)}_{\text{$P$}} \  = \ \underbrace{(k\times n)}_{\text{layer $t$}}\ \underbrace{(n\times n)}_{\text{layer $t-1$}}\ \ ...\ \ \underbrace{(n\times n)}_{\text{layer $1$}}
    \label{eq:shapes-diffsort-topk}
\end{equation}
Note that during the execution of the sorting network, $P$ is conventionally computed from layer $1$ to layer $t$, i.e., from right to left.
If we computed it in this order, we would only save a tiny fraction of the computational cost and only during the last layer.
Thus, we propose to execute the differentiable sorting network, save the values that populate the (sparse) $n\times n$ layer-wise permutation matrices, and compute $P$ in a second pass from the back to the front, i.e., from layer $t$ to layer $1$, or from left to right in Equation~\eqref{eq:shapes-diffsort-topk}.
This allows executing $t$ dense-sparse matrix multiplications with dense $k\times n$ matrices and sparse $n\times n$ matrices instead of dense $n\times n$ and sparse $n\times n$ matrices.
With this, we reduce the asymptotic complexity from $\mathcal{O}(n^2 \log^2 (n))$ to $\mathcal{O}(nk \log^2 (n))$.

\subsection{Differentiable Top-\texorpdfstring{$\protect k$}{k} Networks}

As only the top-$k$ rows of a relaxed permutation matrix are required for top-$k$ classification learning, it is possible to further improve the efficiency of computing the top-$k$ probability distribution via differentiable sorting networks by improving the underlying sorting network or (in general) comparator network architecture.
Thus, we propose differentiable top-$k$ networks, which relax selection networks in analogy to how differentiable sorting networks relax sorting networks.
Selection networks are networks that select only the top-$k$ out of $n$ elements \cite{knuth1998sorting}.

\subsubsection{Splitter Selection Networks}

In this context, we propose splitter selection networks (SSN), a novel class of selection networks that requires only $\mathcal{O}(\log n)$ layers (instead of the $\mathcal{O}(\log^2 n)$ layers for sorting networks), which makes top-$k$ supervision with differentiable top-$k$ networks more efficient and reduces the error (which is introduced in each layer).
SSNs follow the idea that the input is split into locally sorted sublists and then all wires that are not candidates to be among the global top-$k$ can be eliminated.
For example, for $n=1024, k=5$, SSNs require only $22$ layers, while the best previous selection network requires $34$ layers and full sorting (with a bitonic network) even requires $55$ layers.
For $n=10450, k=5$ (i.e., for ImageNet-21K-P), SNNs require $27$ layers, the best previous selection network requires $50$ layers, and full sorting requires $105$ layers.
In addition, the layers of SSNs are less computationally expensive than those of the bitonic sorting network.
Further details on SSNs, as well as their full construction, can be found in~\cite{petersen2022topk}.

Concluding, the contribution of differentiable top-$k$ networks is two-fold: first, we propose a novel kind of selection networks that needs fewer layers, and second, we relax those similarly to differentiable sorting networks.

\section{Top-\texorpdfstring{$\protect k$}{k} Learning}

\begin{figure*}
    \centering
    \includegraphics{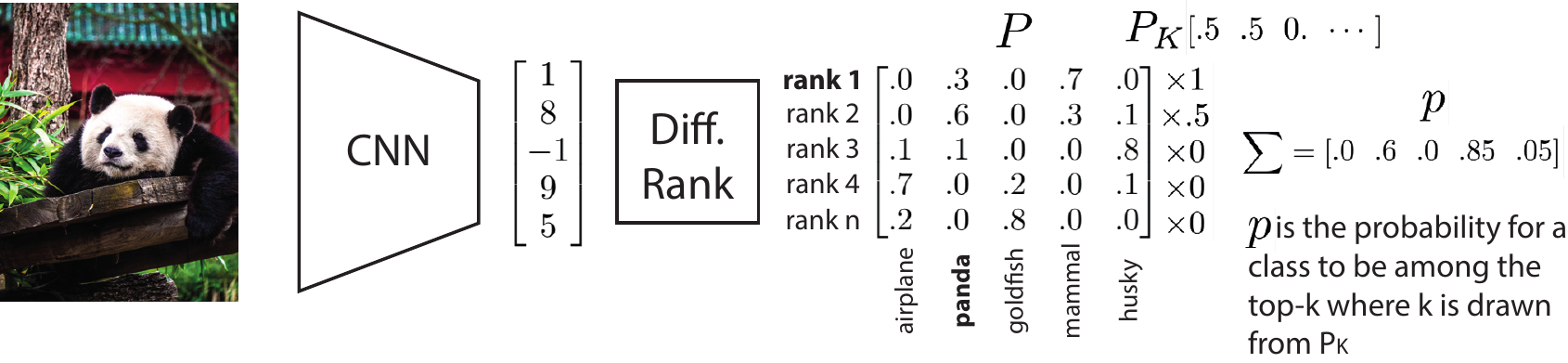}
    \caption{
        Overview of the proposed architecture:
        A CNN predicts scores for an image, which are then ranked by a differentiable ranking algorithm returning the probability distribution for each rank in matrix $\mP$.
        The rows of this distribution correspond to ranks, and the columns correspond to the respective classes.
        In the example, we use a $50\%$ top-$1$ and $50\%$ top-$2$ loss, i.e., $P_K=[.5, .5, 0, 0, 0]$. Here, the $k$th value refers to the top-$k$ component, which is satisfied if the prediction is at \textit{any} of \hbox{rank-$1$} to rank-$k$.
        Thus, the weights for the different ranks can be computed via a cumulative sum and are $[1, .5, 0, 0, 0]$.
        The correspondingly weighted sum of rows of $\mP$ yields the probability distribution~$p$, which can then be used in a cross-entropy loss.
    }
    \label{fig:panda-overview}
\end{figure*}

In this section, we start by introducing our objective, elaborate its exact formulation, and then build on differentiable sorting principles to efficiently approximate the objective.
A visual overview of the loss architecture is given in Figure~\ref{fig:panda-overview}.

The goal of top-$k$ learning is to extend the learning criterion from only accepting exact (top-$1$) predictions to accepting $k$ predictions among which the correct class has to be.
In its general form, for top-$k$ learning, $k$ may differ for each application, class, data point, or a combination thereof.
For example, in one case, one may want to rank $5$ predictions and assign a score that depends on the rank of the true class among these ranked predictions, while, in another case, one may want to obtain $5$ predictions but not care about their order.
In yet another case, such as image classification, one may want to enforce a top-$1$ accuracy on images from the ``person'' super-class, but resign to a top-$3$ accuracy for the ``animal'' super-class, as it may have more ambiguities in class-labels.
We model this by a random variable~$K$, following a distribution $P_K$ that describes the relative importance of different values~$k$.
The discrete distribution $P_K$ is either a marginalized distribution for a given setting (such as the discrete uniform distribution), or a conditional distribution for each class, data point, etc.
This allows specifying a marginalized or conditional distribution $k\sim P_K$.
This generalizes the ideas of conventional top-$1$ supervision (usually softmax cross-entropy) and top-$k$ supervision for a $k$ like $k=5$ (usually based on surrogate top-$k$ margin/hinge losses like \cite{lapin2016loss, berrada2018smooth}) and unifies them.

The objective of top-$k$ learning is maximizing the probability of accepted predictions of the model $f_\theta$ on data $X, y \sim \mathcal{D}$ given marginal distribution $P_K$ (or conditional $P_{K|X,y}$ if it depends on the class $y$ and/or data point $X$). In the following, $\mP_{k,y}$ is the predicted probability of $y$ being the $k$th-best prediction for data point $X$.
\begin{equation}
    \argmax_\theta\ \ \mathbb{E}_{X, y \sim \mathcal{D}} \left[\log\left(
    \mathbb{E}_{k \sim P_K} \left[
            {\textstyle\sum_{m=1}^k}
            \mP_{m, y}(f_\theta(X))
    \right]\right)
    \right]
\end{equation}
To evaluate the probability of $y$ to be the top-$1$ prediction, we can simply use $\operatorname{softmax}_y(f_\theta (X))$.
However, $k>1$ requires more consideration.
Here, we require probability scores $\mP_{k, c}$ for the $k$th prediction over classes $c\in\mathbb{C}$, where $\sum_{c=1}^n \mP_{k, c} = 1$ (i.e., $\mP$ is row-stochastic) and ideally additionally $\sum_{k=1}^n \mP_{k, c} = 1$ (i.e., $\mP$ is also column stochastic and thus doubly-stochastic.)
With this, we can optimize our model by minimizing the following loss
\begin{equation}
    \mathcal{L}(X, y)\,{=}\,{-} \log\!\left( \sum_{k=1}^n P_K(k) \!\left(\sum_{m=1}^k \mP_{m, y}(f_\theta (X)) \right)\!\!\right),
\end{equation}
which is the cross-entropy over the probabilities that the true class is among the top-$k$ class for each possible~$k$.
Note that $\sum_{k=1}^n P_K(k) = 1$.

To compute $\mP_{k, c}$, we require a function mapping from a vector of real-valued scores to an (ideally) doubly-stochastic matrix $\mP$.
The most suitable for this are the differentiable relaxations of the sorting and ranking functions, which produce differentiable permutation matrices $\mP$, which we introduced in Chapter~\ref{ch:diffsort}.
We build on these approximations to propose instances of top-$k$ learning losses and extend differentiable sorting networks to differentiable top-$k$ networks, as just finding the top-$k$ scores is computationally cheaper than sorting all elements and reduces the approximation error.

\subsection{Implementation Details}
\label{sec:implementation_details}

Despite those performance improvements, evaluating the differentiable ranking operators still requires a considerable amount of computational effort for large numbers of classes.
Especially if the number $n$ of elements to be ranked is $n=1\,000$ (ImageNet-1K) or even $n>10\,000$ (ImageNet-21K-P), the differentiable ranking operators can dominate the overall computational costs.
In addition, for large numbers $n$ of elements to be ranked, the performance of differentiable ranking operators decreases as differentially ranking more elements naturally introduces larger errors \cite{grover2019neuralsort,prillo2020softsort,cuturi2019differentiable,petersen2021diffsort}.
Thus, we reduce the number of outputs to be ranked differentially by only considering those classes (for each input) that have a score among the top-$m$ scores.
For this, we make sure that the ground truth class is among those top-$m$ scores, by replacing the lowest of the top-$m$ scores by the ground truth class, if necessary.
For $n=1000$, we choose $m=16$, and for $n>10\,000$, we choose $m=50$.
We find that this greatly improves training performance.

Because the differentiable ranking operators are (by their nature of being differentiable) only approximations to the hard ranking operator, they each have their characteristics and inconsistencies.
Thus, for training models from scratch, we replace the top-$1$ component of the loss by the regular softmax, which has a better and more consistent behavior.
This guides the other loss if the differentiable ranking operator behaves inconsistently.
To avoid the top-$k$ components affecting the guiding softmax component and avoid probabilities greater than $1$ in $p$, we can separate the cross-entropy into a mixture of the softmax cross-entropy (for the top-$1$ component) and the top-$k$ cross-entropy (for the top-$k\geq2$ components) as follows: %
\begin{align}
    &\underset{\mathrm{sm+top-}k}{\mathcal{L}(X, y)}
    = P_K(1) \cdot  \operatorname{SoftmaxCELoss}(f_\theta (X), y) \\[-0.7ex]
    &{-} (1-P_K(1)) \cdot \log\!\left( \sum_{k=2}^n P_K(k) \left(\sum_{m=1}^k \mP_{m, y}(f_\theta (X)) \right)\!\!\right) \notag %
\end{align}

\section{Experiments}

We evaluate the proposed top-$k$ classification loss for four differentiable ranking operators on CIFAR-100~\cite{krizhevsky2009cifar10}, ImageNet-1K~\cite{deng2009imagenet}, as well as the winter 2021 edition of ImageNet-21K-P~\cite{ridnik2021imagenet}.
We use CIFAR-100, which can be considered a small-scale data set with only $100$ classes, to train a ResNet18 model \cite{he2016deep} from scratch and show the impact of the proposed loss function on the top-$1$ and top-$5$ accuracy.
In comparison, ImageNet-1K and ImageNet-21K-P provide rather large-scale data sets with $1\,000$ and $10\,450$ classes, respectively.
To avoid the unreasonable carbon-footprint of training many models from scratch, we decided to exclusively use publicly available backbones for all ImageNet experiments.
This has the additional benefit of allowing more settings, making our work easily reproducible, and allowing to perform multiple runs on different seeds to improve the statistical significance of the results.
For ImageNet-1K, we use two publicly available state-of-the-art architectures as backbones:
First, the (four) ResNeXt-101 WSL architectures by Mahajan~\etal~\cite{mahajan2018exploring}, which were pretrained in a weakly-supervised fashion on a billion-scale data set from Instagram.
Second, the Noisy Student EfficientNet-L2~\cite{xie2020self}, which was pretrained on the unlabeled JFT-300M data set~\cite{sun2017revisiting}.
For ResNeXt-101 WSL, we extract $2\,048$-dimensional embeddings and for the Noisy Student EfficientNet-L2, we extract $5\,504$-dimensional embeddings of ImageNet-1K and fine-tune on them.

We apply the proposed loss in combination with various available differentiable sorting and ranking approaches, namely NeuralSort, SoftSort, SinkhornSort, and DiffSortNets.
To determine the optimal (inverse) temperature for each differentiable sorting method, we perform a grid search at a resolution of factor $2$.
For training, we use the Adam optimizer \cite{kingma2015adam}.
For training on CIFAR-100 from scratch, we train for up to $200$ epochs with a batch size of $100$ at a learning rate of $10^{-3}$.
For ImageNet-1K, we train for up to $100$ epochs at a batch size of $500$ and a learning rate of $10^{-4.5}$.
For ImageNet-21K-P, we train for up to $40$ epochs at a batch size of $500$ and a learning rate of $10^{-4}$.
We use early stopping and found that these settings lead to convergence in all settings (except those that also diverge with other settings).
As baselines, we use the respective original models, softmax cross-entropy, as well as learning with the smooth surrogate top-$k$ loss~\cite{berrada2018smooth}.

\begin{table}[t]
    \centering
    \addtolength{\tabcolsep}{-4pt}
    {\footnotesize
    \begin{tabular}{lcc}
\toprule
Method & $P_K$  &  {CIFAR-100} \\
\midrule
\underline{\textit{Baselines}}\\
Softmax                     & $([1,0,0,0,0])$     & $61.27\,|\,85.31$ \\[.2em]
Smooth\,top-$k$ loss \cite{berrada2018smooth}  & $([0,0,0,0,1])$     & $53.07\,|\,85.23$ \\
Top-$5$ NeuralSort          & $[0,0,0,0,1]$     & $22.58\,|\,84.41$ \\
Top-$5$ SoftSort            & $[0,0,0,0,1]$     & $1.01\,|\,5.09$   \\
Top-$5$ SinkhornSort        & $[0,0,0,0,1]$     & $55.62\,|\,\pmb{87.04}$ \\
Top-$5$ DiffSortNets        & $[0,0,0,0,1]$     & $52.81\,|\,84.21$ \\
\midrule
\underline{\textit{Ours}}\\
Top-$k$ NeuralSort          & $[.2,.2,.2,.2,.2]$   & $61.46\,|\,86.03$\\   %
Top-$k$ SoftSort            & $[.2,.2,.2,.2,.2]$   & $61.53\,|\,82.39$\\   %
Top-$k$ SinkhornSort        & $[.2,.2,.2,.2,.2]$   & $61.89\,|\,\pmb{86.94}$\\   %
Top-$k$ DiffSortNets        & $[.2,.2,.2,.2,.2]$   & $\pmb{62.00}\,|\,86.73$\\   %
\bottomrule
    \end{tabular}
    }
    \addtolength{\tabcolsep}{4pt}
    \caption{
        CIFAR-100 results for training a ResNet18 from scratch.
        The metrics are Top-$1\,|\,$Top-$5$ accuracy averaged over 2 seeds.
    }
    \label{tab:cifar100-main}
\end{table}

\subsection{Training from Scratch}

We start by demonstrating that the proposed loss can be used to train a network from scratch.
As a reference baseline, we train a ResNet18 from scratch on CIFAR-100.
In Table~\ref{tab:cifar100-main}, we compare the baselines (i.e., top-$1$ softmax, the smooth top-$5$ loss \cite{berrada2018smooth}, as well as ``pure'' top-$5$ losses using four differentiable sorting and ranking methods) with our top-$k$ loss with $k\sim[.2,.2,.2,.2,.2]$.

We find that training with top-$5$ alone---in some cases---slightly improves the top-$5$ accuracy but has a substantially worse top-$1$ accuracy.
Here, we note that the smooth top-$5$ loss \cite{berrada2018smooth}, top-$5$ Sinkhorn \cite{cuturi2019differentiable}, and top-$5$ DiffSort \cite{petersen2021diffsort} achieve moderate performance.
Notably, Sinkhorn \cite{cuturi2019differentiable} outperforms the softmax baseline on the top-$5$ metric, while NeuralSort and SoftSort are less stable and yield worse results especially on top-$1$ accuracy. %

By using our loss that corresponds to drawing $k$ from $[.2,.2,.2,.2,.2]$, we can achieve substantially improved results, especially also on the top-$1$ accuracy metric.
Using the DiffSortNets yields the best results on the top-$1$ accuracy and Sinkhorn yields the best results on the top-$5$ accuracy.
Note that, here, also NeuralSort and SoftSort achieve good results in this setting, which can be attributed to our loss with $k\sim[.2,.2,.2,.2,.2]$ being more robust to inconsistencies and outliers in the used differentiable sorting method.
Interestingly, top-$5$ SinkhornSort achieves the best performance on the top-$5$ metric, which suggests that SinkhornSort is a very robust differentiable sorting method as it does not require additional top-$k$ components.
Nevertheless, it is advisable to include other top-$k$ components as the model trained purely on top-$5$ exhibits poor top-$1$ performance.

\subsection{Fine-Tuning}

In this section, we discuss the results for fine-tuning on ImageNet-1K and ImageNet-21K-P.
In Table~\ref{tab:imagenet-main}, we find a very similar behavior to training from scratch on CIFAR-100.
Specifically, we find that the accuracy improves by drawing $k$ from a distribution for training.
An exception is (again) SinkhornSort, where focusing only on top-$5$ yields the best top-$5$ accuracy on ImageNet-1K, but the respective model exhibits poor top-$1$ accuracy.
Overall, we find that drawing $k$ from a distribution improves performance in all cases.

To demonstrate that the improvements also translate to different backbones, we show the improvements on all four model sizes of ResNeXt-101 WSL (32x8d, 32x16d, 32x32d, 32x48d) in Figure~\ref{fig:resnext-wsl-different-model-sizes-improvements-plot}.
Also, here, our method improves the model in all settings.

\begin{table}[t]
    \centering
    \addtolength{\tabcolsep}{-4pt}
    {\footnotesize
    \begin{tabular}{lccc}
\toprule
Method & $P_K$  &  {ImgNet-1K} & \kern-.25emImgNet-21K-P\\
\midrule
\underline{\textit{Baselines}}\\
Softmax                     & $([1,0,0,0,0])$     & $86.06\,|\,97.795$ & $39.29\,|\,69.63$ \\[.2em]
Smooth\,top-$k$ loss \cite{berrada2018smooth}  & $([0,0,0,0,1])$     & $85.15\,|\,97.540$ & $34.03\,|\,65.56$ \\
Top-$5$ NeuralSort          & $[0,0,0,0,1]$     & $33.37\,|\,94.748$ & $15.87\,|\,33.81$ \\
Top-$5$ SoftSort            & $[0,0,0,0,1]$     & $18.23\,|\,94.965$ & $33.61\,|\,69.82$ \\
Top-$5$ SinkhornSort        & $[0,0,0,0,1]$     & $85.65\,|\,\pmb{97.991}$ & $36.93\,|\,69.80$ \\
Top-$5$ DiffSortNets        & $[0,0,0,0,1]$     & $69.05\,|\,97.389$ & $35.96\,|\,69.76$ \\
\midrule
\underline{\textit{Ours}}\\
Top-$k$ NeuralSort          & $[.5,0,0,0,.5]$   & $86.30\,|\,97.896$        &  $37.85\,|\,68.08$ \\
Top-$k$ SoftSort            & $[.5,0,0,0,.5]$   & $86.26\,|\,97.963$        &  $39.93\,|\,70.63$ \\
Top-$k$ SinkhornSort        & $[.5,0,0,0,.5]$   & $\pmb{86.29}\,|\,\pmb{97.971}$  &  $39.85\,|\,70.56$ \\
Top-$k$ DiffSortNets        & $[.5,0,0,0,.5]$   & $86.24\,|\,97.937$        &  \pmb{$40.22\,|\,70.88$} \\
\bottomrule
    \end{tabular}
    }
    \addtolength{\tabcolsep}{4pt}
    \caption{
        ImageNet-1K and ImageNet-21K-P results for fine-tuning the head of ResNeXt-101 32x48d WSL~\cite{mahajan2018exploring}.
        The metrics are Top-$1\,|\,$Top-$5$ accuracy averaged over 10 seeds for ImageNet-1K and 2 seeds for ImageNet-21K-P.
    }
    \label{tab:imagenet-main}
\end{table}

\begin{figure}[t]
    \centering
    \resizebox{234.8775pt}{!}{\input{fig_difftopk/imagenet1k-top1-top5-2.pgf}}
    \caption{
        ImageNet-1K accuracy improvements for all ResNeXt-101 WSL model sizes (32x8d, 32x16d, 32x32d, 32x48d). %
        Green ($\bullet$) is the original model and red ($\blacktriangle$) is with top-$k$ fine-tuning.
        }
    \label{fig:resnext-wsl-different-model-sizes-improvements-plot}
\end{figure}

\begin{figure*}
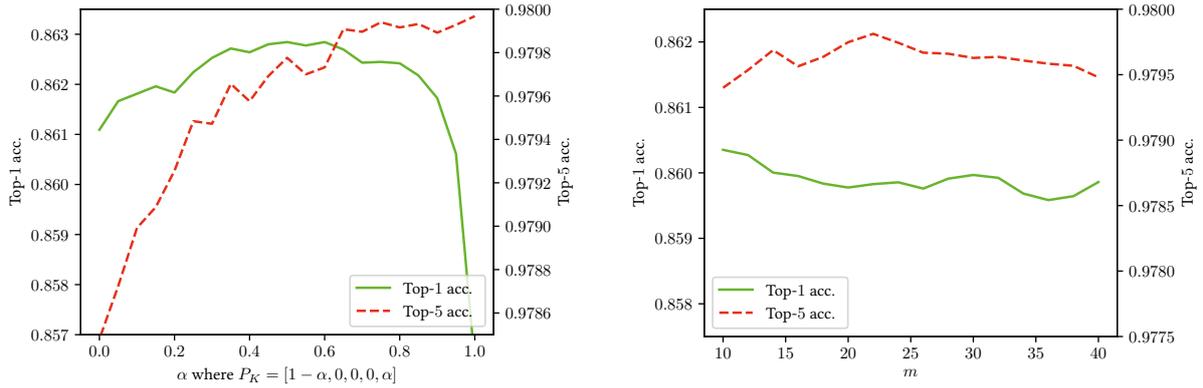

    \centering
    \resizebox{.49\linewidth}{!}{\input{fig_difftopk/imagenet1k-top1-vs-top5-loss-plot-sinkhorn.pgf}}\hfill
    \resizebox{.49\linewidth}{!}{\input{fig_difftopk/imagenet1k-varying-m-plot-sinkhorn_0_75.pgf}}
    \caption{
    Effects of varying the ratio between top-$1$ and top-$5$ \textit{(left)} and varying the size of differentially ranked subset $m$ \textit{(right)}.
    Both experiments build on the differentiable Sinkhorn ranking algorithm \cite{cuturi2019differentiable}.
    On the left, $m=16$, on the right, $\alpha=0.75$.
    Averaged over $5$ runs.
    }
    \label{fig:top1-vs-top5-loss-plot}
    \label{fig:different-ms-plot}
\end{figure*}

\subsection{Impact of the Distribution \texorpdfstring{$\protect P_K$}{PK} and Differentiable Sorting Methods}

We start by demonstrating the impact of $P_K$, which is the distribution from which we draw $k$.
Let us first consider the case where $k$ is $5$ with probability $\alpha$ and $1$ with probability $1-\alpha$, i.e., $P_K=[1-\alpha, 0, 0, 0, \alpha]$.
In Figure~\ref{fig:top1-vs-top5-loss-plot} (left), we demonstrate the impact that arises from changing $\alpha$, i.e., transitioning from a pure top-$1$ loss to a pure top-$5$ loss, in the case of fine-tuning ResNeXt-101 WSL with our loss using the SinkhornSort algorithm.
Increasing the weight of the top-$5$ component does not only increase the top-$5$ accuracy but also improves the top-$1$ accuracy up to around $60\%$ top-$5$; when using only $k=5$, the top-$1$ accuracy drastically decays as the incentive for the true class to be at the top-$1$ position vanishes (or is only indirectly given by being among the top-$5$.)
While the top-$5$ accuracy in this plot is best for a pure top-$5$ loss, this generally only applies to the Sinkhorn algorithm and overall training is more stable if a pure top-$5$ is avoided.
This can also be seen in Tables~\ref{tab:cifar100-main} and~\ref{tab:imagenet-main}.

In Tables~\ref{tab:cifar100-different-PK} and \ref{tab:imagenet-different-PK}, we consider additional settings with all differentiable ranking methods.
Specifically, we compare four notable settings:
$[.5,0,0,0,.5]$, i.e., equally weighted top-$1$ and top-$5$; $[.25,0,0,0,.75]$ and $[.1,0,0,0,.9]$, i.e., top-$5$ has larger weights; $[.2, .2, .2, .2, .2]$, i.e., the case of having an equal weight of $0.2$ for top-$1$ to top-$5$.
The $[.5,0,0,0,.5]$ setting is a rather canonical setting which usually performs well on both metrics, while the others tend to favor top-$5$.
In the $[.5,0,0,0,.5]$ setting, all sorting methods improve upon the softmax baseline on both top-$1$ and top-$5$ accuracy.
When increasing the weight of the top-$5$ component, the top-$5$ generally improves while top-$1$ decays.

Here we find a core insight of this chapter: the best performance cannot be achieved by optimizing top-$k$ for only a single $k$, but instead, drawing $k$ from a distribution improves performance on all metrics.

Comparing the differentiable ranking methods, we can find the overall trend that SoftSort outperforms NeuralSort, and that SinkhornSort as well as DiffSortNets perform best.
We can see that some sorting algorithms are more sensitive to the overall $P_K$ than others:
Whereas SinkhornSort \cite{cuturi2019differentiable} and DiffSortNets \cite{petersen2021diffsort} continuously outperform the softmax baseline, NeuralSort \cite{grover2019neuralsort} and SoftSort \cite{prillo2020softsort} tend to collapse when over-weighting the top-$5$ components.

Comparing the performance on the medium-scale ImageNet-1K to the larger ImageNet-21K-P in Table~\ref{tab:imagenet-main}, we observe a similar pattern.
Here, again, using the top-$k$ component alone is not enough to significantly increase accuracy, but combining top-$1$ and top-$k$ components helps to improve accuracy on both reported metrics.
While NeuralSort struggles in this large-scale ranking problem and stays below the softmax baseline, DiffSortNets~\cite{petersen2021diffsort} provide the best top-$1$ and top-$5$ accuracy with $40.22\%$ and $70.88\%$, respectively.

We note that we do not claim that all settings (especially all differentiable sorting methods) improve the classification performance on all metrics.
Instead, we include all methods and also additional settings to demonstrate the capabilities and limitations of each differentiable sorting method.

Overall, it is notable that SinkhornSort achieves the overall most robust training behavior, while also being by far the slowest sorting method and thus potentially slowing down training drastically, especially when the task is only fine-tuning.
SinkhornSort tends to require more Sinkhorn iterations towards the end of training.
DiffSortNets are considerably faster; especially, it is possible to only compute the top-$k$ probability matrices and because of our advances for more efficient selection networks.

\begin{table}[t]
    \centering
    \addtolength{\tabcolsep}{-4pt}
    {\footnotesize
    \begin{tabular}{lcccc}
\toprule
Method~/~$P_K$    & {\small$[.5,0,0,0,.5]$} & {\small$[.25,0,0,0,.75]$} & {\small$[.1,0,0,0,.9]$} & {\small$[.2,.2,.2,.2,.2]$}\\
\midrule
\textit{CIFAR-100}\kern-5em\\
NeuralSort          &  $61.12\,|\,86.47$  &  $61.07\,|\,87.23$  &  $52.57\,|\,85.76$  &  $61.46\,|\,86.03$\\   %
SoftSort            &  $61.17\,|\,83.95$  &  $61.05\,|\,83.10$  &  $58.16\,|\,79.26$  &  $61.53\,|\,82.39$\\   %
SinkhornSort        &  $61.34\,|\,86.38$  &  $61.50\,|\,86.68$  &  $57.35\,|\,86.34$  &  $61.89\,|\,86.94$\\   %
DiffSortNets        &  $60.07\,|\,86.44$  &  $61.57\,|\,86.51$  &  $61.74\,|\,\pmb{87.22}$  &  $\pmb{62.00}\,|\,86.73$\\   %
\bottomrule
    \end{tabular}
    }
    \addtolength{\tabcolsep}{4pt}
    \caption{
        CIFAR-100 results for different distributions $P_K$ for training a ResNet18 from scratch.
        The metrics are Top-$1\,|\,$Top-$5$ accuracy averaged over 2 seeds.
    }
    \label{tab:cifar100-different-PK}
\end{table}

\begin{table}[t]
    \centering
    \addtolength{\tabcolsep}{-4pt}
    {\footnotesize
    \begin{tabular}{lcccc}
\toprule
Method~/~$P_K$    & {\small$[.5,0,0,0,.5]$} & {\small$[.25,0,0,0,.75]$} & {\small$[.1,0,0,0,.9]$} & {\small$[.2,.2,.2,.2,.2]$}\\
\midrule
\textit{ImageNet-1K}\\
NeuralSort          & $86.30\,|\,97.896$  &  $34.26\,|\,95.410$  &  $34.32\,|\,94.889$  &  $85.75\,|\,97.865$\\
SoftSort            & $86.26\,|\,97.963$  &  $86.16\,|\,97.954$  &  $27.30\,|\,95.915$  &  $86.18\,|\,97.979$\\
SinkhornSort        & $\pmb{86.29}\,|\,97.971$  &  $86.24\,|\,97.989$  &  $86.18\,|\,97.987$  &  $86.22\,|\,97.989$\\  %
DiffSortNets        & $86.24\,|\,97.937$  &  $86.15\,|\,97.936$  &  $86.04\,|\,97.980$  &  $86.21\,|\,\pmb{98.003}$\\  %
\midrule
\textit{ImageNet-21K-P}\kern-5em\\
NeuralSort          & $37.85\,|\,68.08$  &  $36.16\,|\,67.60$  &  $33.02\,|\,67.29$  &  $37.09\,|\,67.90$\\
SoftSort            & $39.93\,|\,70.63$  &  $39.08\,|\,70.27$  &  $37.78\,|\,70.07$  &  $39.68\,|\,70.57$\\
SinkhornSort        & $39.85\,|\,70.56$  &  $39.21\,|\,70.41$  &  $38.42\,|\,70.12$  &  $39.22\,|\,70.49$\\
DiffSortNets        & \pmb{$40.22\,|\,70.88$}  &  $39.56\,|\,70.58$  &  $38.48\,|\,70.25$  &  $39.69\,|\,70.69$\\  %
\bottomrule
    \end{tabular}
    }
    \addtolength{\tabcolsep}{4pt}
    \caption{
        ImageNet-1K and ImageNet-21K-P results for different distributions $P_K$ for fine-tuning the head of ResNeXt-101 32x48d WSL \cite{mahajan2018exploring}.
        The metrics are Top-$1\,|\,$Top-$5$ accuracy averaged over 10 seeds for ImageNet-1K and 2 seeds for ImageNet-21K-P.
    }
    \label{tab:imagenet-different-PK}
\end{table}

\subsection{Differentiable Ranking Set Size \texorpdfstring{$\protect m$}{m}}
We consider how accuracy is affected by varying the number of scores $m$ to be differentially ranked.
Generally, the runtime of differentiable top-$k$ operators depends between linearly and cubic on $m$; thus it is important to choose an adequate value for $m$.
The choice of $m$ between $10$ and $40$ has only a moderate impact on the accuracy, as can be seen in Figure~\ref{fig:different-ms-plot} (right).
However, when setting $m$ to large values such as $1\,000$ or larger, we observe that the differentiable sorting methods tend to become unstable.
We note that we did not specifically tune $m$, and that better performance can be achieved by fine-tuning $m$, as displayed in the plot.

\subsection{Comparison to the State-of-the-Art}

We compare the proposed method to current state-of-the-art methods in Table~\ref{tab:sota}.
We focus on methods that are publicly available and build upon two of the best-performing models, namely Noisy Student EfficientNet-L2 \cite{xie2020self}, and ResNeXt-101 32x48d WSL \cite{mahajan2018exploring}.
Using both backbones, we achieve improvements on both metrics, and when fine-tuning on the Noisy Student EfficientNet-L2, we achieve a new state-of-the-art for publicly available models.

\paragraph{Significance Tests}
To evaluate the significance of the results, we perform a t-test (with a significance level of $0.01$).
We find that our model is significantly better than the original model on both top-$1$ and top-$5$ accuracy metrics.
Comparing to the observed accuracies of the baseline ($88.33\,|\,98.65$), DiffSortNets are significantly better (p=$0.00001\,|\,0.00005$).
Comparing to the reported accuracies of the baseline ($88.35\,|\,98.65$), DiffSortNets are also significantly better (p=$0.00087\,|\,0.00005$).

\begin{table}[t]
    \centering
    {\footnotesize
    \begin{tabular}{llrcc}
    \toprule
        Method & & \kern-1.5em Public & Top-$1$ & Top-$5$  \\
    \midrule
        ResNet50 &\cite{he2016deep}                                      & \cmark &  $79.26$  & $94.75 $ \\
        ResNet152 &\cite{he2016deep}                                     & \cmark &  $80.62$  & $95.51 $ \\
        ResNeXt-101 32x48d WSL &\cite{mahajan2018exploring}              & \cmark & $85.43$	 & $97.57$ \\
        ViT-L/16 &\cite{dosovitskiy2021image}                            & \cmark & $87.76 $  & ---   \\
        Noisy Student EfficientNet-L2 &\cite{xie2020self}                & \cmark & \pmb{$88.35$}   & \pmb{$98.65$} \\
        \midrule
        BiT-L   &\cite{kolesnikov2020big}                                & \xmark & $87.54 $  & $98.46$  \\
        CLIP (w/ Noisy Student EffNet-L2) &\cite{radford2021learning}    & \xmark & $\approx88.4$    & --- \\
        ViT-H/14 &\cite{dosovitskiy2021image}                            & \xmark & $88.55 $  & ---   \\
        ALIGN (EfficientNet-L2) &\cite{jia2021scaling}                   & \xmark & $88.64$   & \pmb{$98.67$} \\
        Meta Pseudo Labels (EfficientNet-L2) &\cite{pham2021meta}        & \xmark & $90.20 $  & $\approx98.8$  \\
        ViT-G/14 &\cite{zhai2021scaling}                                 & \xmark & $90.45 $  & ---   \\
        CoAtNet-7 &\cite{dai2021coatnet}                                 & \xmark & \pmb{$90.88$}   & --- \\
    \midrule
        ResNeXt-101 32x48d WSL &&& $86.06 $ & $97.80 $ \\
        Top-$k$ SinkhornSort                                                &&& \pmb{$86.22 $} & $97.99 $ \\ %
        Top-$k$ DiffSortNets                                                &&& $86.21 $ & \pmb{$98.00 $} \\
    \midrule
        Noisy Student EfficientNet-L2  &&& $88.33 $ & $98.65 $ \\
        Top-$k$ SinkhornSort                                                &&& $88.32 $ & $98.66 $ \\
        Top-$k$ DiffSortNets                                                &&& \pmb{$88.37$} & \pmb{$98.68$} \\
    \bottomrule
    \end{tabular}
    }
    \caption{
        ImageNet-1K result comparison to state-of-the-art.
        Among the overall best-performing differentiable sorting / ranking methods, almost all results in reasonable settings outperform their respective baseline on Top-$1$ and Top-$5$ accuracy.
        For publicly available models / backbones, we achieve a new state-of-the-art for top-$1$ and top-$5$ accuracy. Our results are averaged over $10$ runs.
    }
    \label{tab:sota}
\end{table}

\subsection*{Conclusion}

We discussed differentiable top-$k$ operators and proposed differentiable top-$k$ networks, which are more efficient than differentiable sorting networks through a computational trick and by using the proposed splitter selection network architecture.
Further, we presented a novel loss, which relaxes the assumption of using a fixed $k$ for top-$k$ classification learning.
We performed an array of experiments to explore different top-$k$ classification learning settings and achieved a state-of-the-art on ImageNet for publicly available models.

\setchapterpreamble[u]{\pagelogo{gendr}\margintoc}
\chapter{Differentiable Rendering}
\labch{gendr}

After covering differentiable sorting, ranking, and top-$k$, this chapter studies differentiable rendering and presents a generalized family of differentiable renderers.
We continue to build on the foundation presented in Chapter~\ref{ch:algovision}, especially Section~\ref{sec:exp-3d}.
We discuss from scratch which components are necessary for differentiable rendering and formalize the requirements for each component.
We instantiate our general differentiable renderer, which generalizes existing differentiable renderers like SoftRas and DIB-R, with an array of different smoothing distributions to cover a large spectrum of reasonable settings.
We evaluate an array of differentiable renderer instantiations on the popular ShapeNet 3D reconstruction benchmark and analyze the implications of our results.
Surprisingly, the simple uniform distribution yields the best overall results when averaged over 13 classes; however, in general, the optimal choice of distribution heavily depends on the task.

In the past years, many differentiable renderers have been published.
These include the seminal differentiable mesh renderer OpenDR~\cite{loper2014opendr}, the Neural 3D Mesh Renderer~\cite{kato2017neural}, and SoftRas~\cite{liu2019soft} among many others.
Using a differentiable renderer enables a multitude of computer vision applications, such as human pose estimation~\cite{bogo2016keep}, camera intrinsics estimation~\cite{palazzi2019end}, 3D shape optimization~\cite{kato2017neural}, 3D~reconstruction~\cite{kato2017neural,liu2019soft,chen2019learning}, and 3D style transfer~\cite{kato2017neural}.

A fundamental difference between different classes of differentiable renderers is the choice of the underlying 3D representation.
In this chapter, we focus on differentiable 3D mesh renderers~\cite{loper2014opendr,kato2017neural,liu2019soft,chen2019learning}; however, the aspects that we investigate could also be applied to other differentiable rendering concepts, such as rendering voxels~\cite{yan2016perspective}, point clouds~\cite{insafutdinov2018unsupervised}, surfels~\cite{yifan2019differentiable}, signed distance functions~\cite{jiang2020sdfdiff, vicini2022differentiable}, and other implicit representations~\cite{liu2019learning, sitzmann2019scene}.

Differentiable mesh renderers can be constructed in different ways:
either using an exact and hard renderer with approximate surrogate gradients or using an approximate renderer with natural gradients.
Loper~\etal~\cite{loper2014opendr} and Kato~\etal~\cite{kato2017neural} produce approximate surrogate gradients for their differentiable renderer, while their forward rendering is hard.
In contrast, other differentiable renderers approximate the forward rendering in such a way that they produce a natural gradient.
This can be achieved by modeling or approximating a renderer under a probabilistic perturbation, which is continuous and makes the renderer differentiable.
For that, in Chapter~\ref{ch:algovision}, we used the logistic distribution.
Rhodin~\etal~\cite{rhodin2015versatile} model it with a Gaussian distribution, while Liu~\etal~\cite{liu2019soft} model it with the square root of a logistic distribution, and Chen~\etal~\cite{chen2019learning} with the exponential distribution.
While this variational interpretation of perturbing by a respective distribution is not stressed in some of these papers~\cite{liu2019soft, chen2019learning}, we believe it is important because it explicitly allows comparing the characteristics of the differentiable renderers.
Moreover, the methods that only approximate gradients can also be seen as approximately modelling a perturbation: the gradient computed for the Neural 3D Mesh Renderer~\cite{kato2017neural} is approximately a perturbation by a uniform distribution.
Note that, here, the solutions for rendering under perturbations are obtained analytically in closed form and without sampling.\\

In this chapter, we introduce a generalized differentiable renderer (GenDR).
By choosing an appropriate probability distribution, we can (at least approximately) recover the above differentiable mesh renderers, which shows that a core distinguishing aspect of differentiable renderers is the type of distributions that they model.
The choice of probability distribution herein is directly linked to the sigmoid (i.e., S-shaped) function used for rasterization.
For example, a Heaviside sigmoid function corresponding to the Dirac delta distribution yields a conventional non-differentiable renderer, while a logistic sigmoid function of squared distances corresponds to the square root of a logistic distribution.
Herein, the sigmoid function is the CDF of the corresponding distribution.
In this chapter, we select and present an array of distributions and examine their theoretical properties.\\

Another aspect of approximate differentiable renderers is their aggregation function, i.e., the function that aggregates the occupancy probabilities of all faces for each pixel.
Existing differentiable renderers commonly aggregate the probabilities via the probabilistic sum ($\bot^P(a,b)=a+b-ab$ or $1-\prod_{t\in T} 1-p_t$),
which corresponds to the probability that at least one face covers the pixel
assuming that probabilities $p_t$ for each triangle $t$ are stochastically independent
(cf.~Eq.~4~in~\cite{liu2019soft} or Eq.~6~in~\cite{chen2019learning}).
In the field of real-valued logics and adjacent fields, this is well-known as a T-conorm, a relaxed form of the logical `or'.
Two examples of other T-conorms are the maximum T-conorm~$\bot^M(a,b)=\max(a,b)$~and
the Einstein sum~$\bot^E(a,b)=(a+b)/(1+ab)$,
which models the relativistic addition of velocities.
We generalize our differentiable renderer to use any continuous T-conorm and present a variety of suitable T-conorms.\\

In total, the set of resulting concrete instances arising from our generalized differentiable renderer and the proposed choices amounts to $\numrenderers$ concrete differentiable renderers.
We extensively benchmark all of them on a shape optimization task and a camera pose estimation task.
Further, we evaluate the best performing and most interesting instances on the popular ShapeNet~\cite{chang2015shapenet} 13~class single-view 3D~reconstruction experiment~\cite{kato2017neural}.
Here, we also include those instances that approximate other existing differentiable renderers.
We note that we do not introduce an alternative shading technique, and rely on existing blended shaders instead.\\

\marginnote{
In the domain of differentiable rendering, different from the work discussed in this chapter, we have also published two papers. 
Specifically, the differentiable renderer Pix2Vex~\cite{petersen2019pix2vex} and style-agnostic differentiable renderer-based 3D reconstruction via domain adaptation~\cite{petersen2022style}.
}
Summarizing the contributions of this chapter, 
(i) we propose a generalized differentiable mesh renderer;
(ii) we identify existing differentiable renderers (approximately) as instances of our generalized renderer;
(iii) we propose a variety of suitable sigmoid functions and T-conorms and group them by their characteristics;
(iv) we extensively benchmark $\numrenderers$ concrete differentiable renderers, analyze which characteristics and families of functions lead to a good performance, and find that the best choice heavily depends on the task, class, or characteristics of the data.

\section{Related Work}

The related work can be classified into those works that present differentiable renderers and those which apply them, although there is naturally also significant overlap.
For additional details on differentiable rendering approaches, cf.~the survey by Kato~\etal~\cite{kato2020differentiable}.

\subheading{Analytical Differentiable Renderers}
The first large category of differentiable renderers are those which produce approximate gradients in an analytical and sampling-free way.
This can either happen by surrogate gradients during backpropagation, as in \cite{kato2017neural}, or by making the forward computation naturally differentiable by perturbing the distances between pixels and triangles analytically in closed form \cite{liu2018adversarial, chen2019learning, petersen2019pix2vex}.
GenDR falls into this category and is of the second case.
Existing works each present their renderer for a specific distribution or sigmoid function.
We formally characterize the necessary functions for a differentiable renderer and present an array of options.

\subheading{Monte-Carlo Differentiable Renderers}
An alternative to analytical differentiable renderers are those which are based on Monte-Carlo sampling techniques.
The first example of this is the ``redner'' path tracer by Li~\etal~\cite{li2018differentiable}, who use edge sampling to approximate the gradients of their renderer.
Loubet~\etal~\cite{loubet2019reparameterizing} build on these ideas and reparameterize the involved discontinuous integrands yielding improved gradient estimates.
Zhang~\etal~\cite{zhang2020path} extend these ideas by differentiating the full path integrals which makes the method more efficient and effective.
Lidec~\etal~\cite{lidec2021differentiable} approach Monte-Carlo differentiable rendering by estimating the gradients of a differentiable renderer via the perturbed optimizers method \cite{berthet2020learning}.

\subheading{Applications}
Popular applications for differentiable renderers are pose~\cite{loper2014opendr, kato2017neural, liu2019soft, chen2019learning, palazzi2019end, lidec2021differentiable}, shape~\cite{kato2017neural, zhang2020path, petersen2019pix2vex}, material~\cite{liu2017material, shi2020match}, texture~\cite{liu2019soft, chen2019learning, loubet2019reparameterizing}, and lighting~\cite{zhang2020path} estimation.
Here, the parameters of an initial scene are optimized to match the scene in a reference image or a set of reference images.
Another interesting application is single-view 3D shape prediction without 3D supervision. Here, a neural network predicts a 3D representation from a single image, and the rendering of the image is compared to the original input image. This learning process is primarily guided by supervision of the object silhouette.
It is possible to omit this supervision via adversarial style transfer as demonstrated by Petersen~\etal~\cite{petersen2022style}.
Other applications are generating new 3D shapes that match a data set~\cite{henzler2018escaping, henderson2020leveraging} as well as adversarial examples in the real world~\cite{liu2019beyond}.

In our experiments, we use optimization for pose and shape to benchmark \textit{all} proposed differentiable renderers.
As the single-view 3D mesh reconstruction is a complex experiment requiring training a neural network, we benchmark our method on this task only for a selected subset of differentiable renderers.

\subheading{T-norms and T-conorms}
T-norms and T-conorms (triangular norms and conorms) are binary functions that generalize the logical conjunction (`and') and disjunction (`or'), respectively, to real-valued logics or probability spaces \cite{klement2013triangular,van2022analyzing}.
A generalization of `or' is necessary for a differentiable renderer to aggregate the occlusion caused by faces.
The existing analytical differentiable renderers all use the probabilistic T-conorm.

\section{The Generalized Differentiable Renderer}
\label{sec:gendr:method}

In this section, we present our generalized differentiable mesh renderer.
With a differentiable renderer, we refer to a renderer that is continuous everywhere and differentiable almost everywhere (a.e.).
Note that, in this context, continuity is a stricter criterion than differentiable a.e.~because formally (i)~conventional renderers are already differentiable a.e.~(which does not mean that they can provide any meaningful gradients), and (ii)~most existing ``differentiable'' renderers are not actually differentiable everywhere.

We start by introducing how a classic hard rendering algorithm operates.
The first step is to bring all objects into image space, which is typically a sequence of affine transformations followed by the camera projection.
This step is already differentiable.
The second step is rasterization:
For each pixel, we need to compute the set of faces (typically triangles) which cover it.
If the pixel is covered by at least one face, the face that is closest to the camera is displayed.

\subsection{Differentiable Occlusion Test}
To make the test whether a pixel~$p$ is occluded by a face~$t$ differentiable, we start by
computing the signed Euclidean distance~$d(p, t)$ between pixel and face boundary.
By convention, pixels inside the triangle have a positive distance, pixels outside the triangle have a negative distance.
For pixels exactly on the boundary, the distance to the face is~$0$.

For a hard occlusion test, we would just check whether~$d(p,t)$ is non-negative.
In a differentiable renderer, we instead introduce a perturbation in the form of a probability distribution with density~$\density$
together with a temperature or scale parameter~$\tau>0$.
We then evaluate the probability that the perturbed distance~$d(p,t)-\tau\epsilon$ is non-negative,
where~$\epsilon$ is distributed according to~$\density$.
Thus, we compute the probability that~$t$ occludes $p$ as
\begin{equation}
\begin{aligned}
    \mathbb{P}_{\epsilon \sim \density}( d(p, t) -\tau\epsilon\geq 0 )
    &= \mathbb{P}_{\epsilon \sim \density}{( \epsilon \leq d(p, t)/\tau ) }\\
    &= \int_{-\infty}^{d(p, t)/\tau} \kern-2.25em f(x)\, dx = \cdf\left(\frac{d(p, t)}{\tau}\right)
\end{aligned}
\label{eq:occlusion_test}
\end{equation}
where~$\cdf$ is the CDF of the distribution $\density$ and thus yields a closed-form solution for the desired probability (provided that $\cdf$ has a closed-form solution or can be appropriately approximated).
In a differentiable renderer, we require~$\cdf$ being continuous.
Typically, $\cdf$ has the S-shape of a sigmoid function, see Table~\ref{tab:vis-sigmoids}.
Therefore, in this chapter, we refer to CDFs as sigmoid functions and use both terms interchangeably.

Most existing differentiable renderers use sigmoid functions or transformations thereof, see Section~\ref{sec:instantiations},
to softly evaluate whether a pixel lies inside a triangle.
This accords with the probabilistic interpretation in Equation~(\ref{eq:occlusion_test}) where the probability distribution is defined via the sigmoid function used in each case.
Here, the logistic sigmoid function is a popular choice of such a sigmoid function.
Note that, recently, it has frequently been referred to as ``the'' sigmoid in the literature, which is not to be confused with the original and more
general terminology.
\begin{example}[Logistic Sigmoid]
    $\cdf_L(x)=1 / (1 + \exp(-x))$ is the logistic sigmoid function, which corresponds to the logistic distribution.
\end{example}

\vspace{-.75em}
\subsection{Aggregation}
\vspace{-.25em}
The second step to be made differentiable is the aggregation of multiple faces.
While this is conventionally done via a logical `or', the differentiable real-valued counterpart is a T-conorm.
T-conorms are formally defined as follows.
\begin{definition}[T-conorm]\label{def:t-conorm}
    A T-conorm is a binary operation~$\bot : [0,1] \times [0,1] \to [0,1]$, which satisfies
    \begin{itemize}
    \setlength\itemsep{0em}
        \item associativity: $\bot(a, \bot(b, c)) = \bot(\bot(a, b), c)$,
        \item commutativity: $\bot(a, b) = \bot(b, a)$,
        \item monotonicity: $(a\leq c) \land (b\leq d) \Rightarrow \bot(a, b) \leq \bot(c, d)$,
        \item $0$ is a neutral element $\bot(a, 0) = a$.
    \end{itemize}
\end{definition}
\begin{remark}[T-conorms and T-norms]
    While T-conorms~$\bot$ are the real-valued equivalents of the logical `or',
    so-called T-norms~$\top$ are the real-valued equivalents of the logical `and'. %
    Certain T-conorms and T-norms are dual in the sense that
    one can derive one from the other using a complement (typically $1-x$) and De~Morgan's laws ($\top(a, b) = 1-\bot(1-a, 1-b)$).
\end{remark}

We proceed by stating the T-conorm that is used in all applicable previous approximate differentiable renderers with natural gradients.
\begin{example}[Probabilistic Sum]
    The \emph{probabilistic sum} is a T-conorm that corresponds to the probability that at least one out of two independent events occurs. It is defined as \vspace{-.5em}\begin{equation}
        \bot^P(a, b) = a + b - ab.
    \end{equation}
\end{example}
An alternative to this is the Einstein sum, which is based on the relativistic addition of velocities.
\begin{example}[Einstein Sum]
    The \emph{Einstein sum} is a T-conorm that corresponds to the velocity addition under special relativity: \vspace{-.75em}\begin{equation}
        \bot^P(a, b) = \frac{a + b}{1 + ab}.
    \end{equation}
\end{example}

Combining the above concepts, we can compute the occupancy or coverage of a pixel $p$ given a set of faces $T$ as
\begin{equation}
    \mathcal{A}_O(p, T) = \underset{t\in T}{\mathlarger{\mathlarger{\mathlarger{\bot}}}}\; \cdf(\,d(p,t) / \tau\,)\,.
\end{equation}
In this chapter, we do not vary the shading aggregation and remain with the popular softmax choice.
This is equivalent to a Gumbel-Min perturbation of the distances between the camera and the faces.
We note that there are no closed-form solutions for adequate alternatives to the $n$-ary softmax in the literature that correspond to a probabilistic interpretation and could be expected to perform adequately.
In fact, this is an open research problem.
Therefore, we focus on varying the two components above and for shading we rely on Petersen~\etal~\cite{petersen2019pix2vex} and Liu~\etal~\cite{liu2019soft}.

\begin{figure}[]
    \centering
\forestset{
  squared just tree/.style={
    for tree={
      edge path={
        \noexpand\path [\forestoption{edge}] (!u.parent anchor) -- +(0,-5pt) -| (.child anchor)\forestoption{edge label};
      },
      align=center,
      if={(isodd(n_children))&&(n_children>2)}
      {
        for children={
          if={equal(n,((n_children("!u"))+1)/2)}
          {
            calign with current
          }{},
        }
      }{},
      if n children=0
      {
        before packing={tier=terminus}
      }{},
    },
  }
}
\resizebox{\linewidth}{!}{
    \begin{justtree}
      {%
        left justifications,
        squared just tree,
      }
      [Taxonomy of Distributions
        [Finite Support
          [Exact
            [Dirac Delta\\ (Heaviside)]
          ]
          [Continuous
            [Uniform\\ Cubic Hermite\\ Wigner Semicircle]
          ]
        ]
        [Infinite Support
          [Symmetrical
            [Exponential Conv.
              [Gaussian\\ Laplace\\ Logistic\\ \kern-4.5em Hyperbolic secant\hbox to0pt{ (Gudermannian)}
              ]
            ]
            [Linear Conv.
              [Cauchy\\ Reciprocal
              ]
            ]
          ]
          [Asymmetrical
            [Two-Sided
              [Gumbel-Max\\ Gumbel-Min
              ]
            ]
            [One-Sided
              [Exponential\\ Gamma\\ Levy
              ]
            ]
          ]
        ]
      ]
    \end{justtree}
}
\caption{
    Taxonomy of probability distributions corresponding to sigmoid functions.
    The subdivisions are chosen wrt.~properties that have a categorically different influence on the behavior of the corresponding renderer.
    The order of splits when going down in the tree (which could be chosen differently, e.g., symmetric/asymmetric could be the first split) reflects the importance of the properties.
}
\label{fig:sigmoid-taxonomy}
\end{figure}

\begin{table*}[]
    \centering
    \scriptsize
    \newcommand{\mktriangle}[3]{\kern-3em\includegraphics[width=4em]{fig_gendr/triangles/t_#1_0_#3_s#2.png}}
    \newcommand{\diagswidth}{0.82in}
    \begin{tabular}{cccccc}
        \toprule
        \includegraphics[width=\diagswidth]{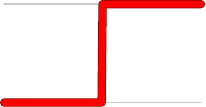}\mktriangle{0}{1.0}{0} &
        \includegraphics[width=\diagswidth]{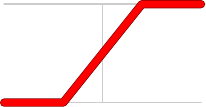}\mktriangle{1}{1.0}{0} &
        \includegraphics[width=\diagswidth]{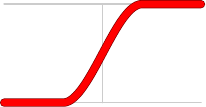}\mktriangle{3}{0.8}{0} &
        \includegraphics[width=\diagswidth]{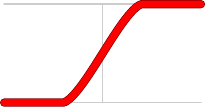}\mktriangle{5}{0.8}{0} &
        \includegraphics[width=\diagswidth]{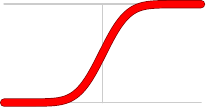}\mktriangle{7}{1.0}{0} &
        \includegraphics[width=\diagswidth]{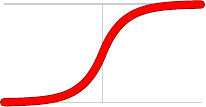}\mktriangle{9}{1.0}{0}
        \\
        Heaviside &
        Uniform &
        Cubic Hermite &
        Wigner Semicircle &
        Gaussian &
        Laplace
        \\
        \midrule
        \includegraphics[width=\diagswidth]{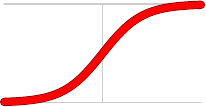}\mktriangle{11}{1.0}{0} &
        \includegraphics[width=\diagswidth]{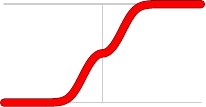}\mktriangle{12}{2.2}{1} &
        \includegraphics[width=\diagswidth]{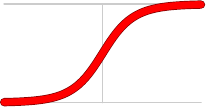}\mktriangle{13}{1.0}{0} &
        \includegraphics[width=\diagswidth]{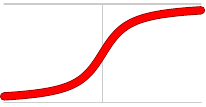}\mktriangle{15}{1.0}{0} &
        \includegraphics[width=\diagswidth]{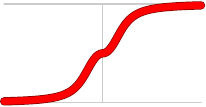}\mktriangle{16}{2.0}{1} &
        \includegraphics[width=\diagswidth]{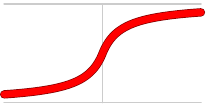}\mktriangle{17}{1.0}{0}
        \\
        Logistic &
        Logistic (squares) &
        Hyperbolic secant &
        Cauchy &
        Cauchy (squares) &
        Reciprocal
        \\
        \midrule
        \includegraphics[width=\diagswidth]{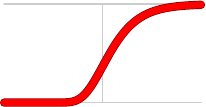}\mktriangle{19}{1.0}{0} &
        \includegraphics[width=\diagswidth]{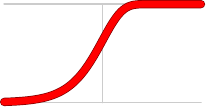}\mktriangle{21}{1.0}{0} &
        \includegraphics[width=\diagswidth]{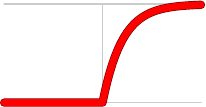}\mktriangle{23}{1.0}{0} &
        \includegraphics[width=\diagswidth]{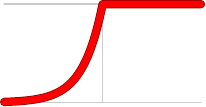}\mktriangle{27}{1.0}{0} &
        \includegraphics[width=\diagswidth]{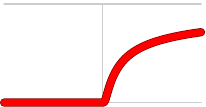}\mktriangle{47}{1.5}{0} &
        \includegraphics[width=\diagswidth]{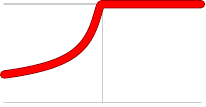}\mktriangle{51}{1.5}{0}
        \\
        Gumbel-Max &
        Gumbel-Min &
        Exponential &
        Exponential (Rev.) &
        Levy &
        Levy (Rev.)
        \\
        \midrule
        \includegraphics[width=\diagswidth]{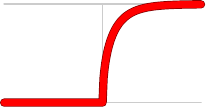}\mktriangle{37}{1.0}{0} &
        \includegraphics[width=\diagswidth]{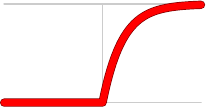}\mktriangle{35}{1.0}{0} &
        \includegraphics[width=\diagswidth]{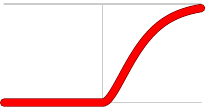}\mktriangle{33}{1.0}{0} &
        \includegraphics[width=\diagswidth]{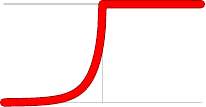}\mktriangle{45}{1.0}{0} &
        \includegraphics[width=\diagswidth]{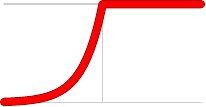}\mktriangle{43}{1.0}{0} &
        \includegraphics[width=\diagswidth]{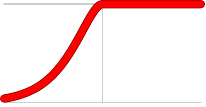}\mktriangle{41}{1.0}{0}
        \\
        Gamma ($p=0.5$) &
        Gamma ($p=1$) &
        Gamma ($p=2$) &
        Gamma ($p=.5$) (Rev.) &
        Gamma ($p=1$) (Rev.) &
        Gamma ($p=2$) (Rev.)
        \\
        \midrule
        \includegraphics[width=\diagswidth]{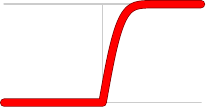}\mktriangle{38}{2.0}{1} &
        \includegraphics[width=\diagswidth]{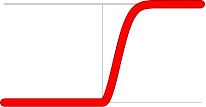}\mktriangle{36}{2.0}{1} &
        \includegraphics[width=\diagswidth]{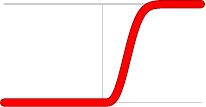}\mktriangle{34}{2.0}{1} &
        \includegraphics[width=\diagswidth]{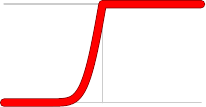}\mktriangle{46}{2.0}{1} &
        \includegraphics[width=\diagswidth]{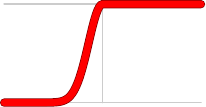}\mktriangle{44}{2.0}{1} &
        \includegraphics[width=\diagswidth]{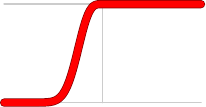}\mktriangle{42}{2.0}{1}
        \\
        Gamma ($p=0.5$) &
        Gamma ($p=1$) &
        Gamma ($p=2$) &
        Gamma ($p=.5$) (Rev.) &
        Gamma ($p=1$) (Rev.) &
        Gamma ($p=2$) (Rev.)
        \\
        (squares) &
        (squares) &
        (squares) &
        (squares) &
        (squares) &
        (squares)
        \\
        \bottomrule
    \end{tabular}
    \caption{Visualization of a selection of sigmoid functions, which are the CDFs of probability distributions. For each distribution, we display a single rendered triangle to demonstrate their different effects.}
    \label{tab:vis-sigmoids}
\end{table*}

\vspace{-1em}
\subsection{Shading}
\vspace{-.5em}
\label{sec:gendr:shading}
The coloring of faces is handled via the Phong model or any other shading model, which is already differentiable.
In the literature, Chen~\etal~\cite{chen2019learning} compare different choices.
Finally, to aggregate the coloring of each pixel depending on the distance of the face to the camera (depth), there are two popular choices in the literature:
no depth perturbations and taking the closest triangle (like \cite{loper2014opendr, kato2017neural, chen2019learning}) and Gumbel depth perturbations (like \cite{liu2019soft, petersen2019pix2vex}).
Only the latter choice is truly continuous, and the closed-form solution for Gumbel depth perturbations is the well-known $\operatorname{softmin}$.
As there are (i) no closed-form solutions for adequate alternatives to Gumbel perturbations in the literature, and (ii) these two options have been extensively studied in the literature~\cite{lidec2021differentiable, loper2014opendr, kato2017neural, chen2019learning, liu2019soft, petersen2019pix2vex}, in this chapter, we do not modify this component and focus on the differentiable silhouette computation and aggregation.
While we implement both options in GenDR, in our evaluation, we perform all experiments
agnostic to the choice of shading aggregation as the experiments rely solely on the silhouette.

\section{Instantiations of GenDR}
\label{sec:instantiations}

We proceed by discussing instantiations of the generalized differentiable renderer (GenDR).

\paragraph{Distributions}
Figure~\ref{fig:sigmoid-taxonomy} provides a taxonomy of the distributions and sigmoid functions that are visualized in Table~\ref{tab:vis-sigmoids}.
We classify the distributions into those with finite support as well as others with infinite support, where the support is the set of points for which the PDF is greater than zero.
Note that the CDFs are constant outside the support region.
Among the distributions with \textit{finite support}, there is the \textit{exact} Dirac delta distribution corresponding to the Heaviside function, which yields a discrete renderer, i.e., not a differentiable renderer.
There are also \textit{continuous} distributions allowing meaningful gradients, but (due to finite support) only in a limited proximity to each face.
Here, we have, among others, the uniform distribution, which corresponds to a piecewise linear step function.
The derivative of the uniform distribution is equivalent or very similar (due to minor implementation aspects) to the surrogate gradient of the Neural 3D Mesh Renderer~\cite{kato2017neural}.
The distributions with \textit{infinite support} can be categorized into symmetrical and asymmetrical.
Among the symmetrical distributions, the Gaussian, the Laplace, the logistic, and the hyperbolic secant have an \textit{exponential convergence} behavior or exponential decay of probability density.
On the other hand, there is also the Cauchy distribution which has a \textit{linear convergence}. This yields a significantly different behavior.
We include the algebraic function~$x\mapsto x / (2 + 2|x|) + 1/2$, called reciprocal sigmoid, which we introduced in Chapter~\ref{ch:diffsort}.
This also has a \textit{linear convergence}.
Finally, we consider \textit{asymmetrical} distributions with infinite support.
The Gumbel-Max and Gumbel-Min are extreme value distributions~\cite{coles2001introduction} and \textit{two-sided}, which means that their support covers both positive and negative arguments.
The exponential, Gamma, and Levy distributions are one-sided distributions. Here, it is important to not only consider the original distributions but also their mirrored or reversed variants, as well as shifted variations as can be seen in the last three rows of Table~\ref{tab:vis-sigmoids}.

SoftRas~\cite{liu2019soft}
squares the absolute part of the distance
before applying the logistic sigmoid function
and thus models the square roots of logistic perturbations. %
Instead of modifying the argument of~$\cdf$, we instead interpret it as applying a transformed
counterpart CDF~$\cdfsq$, which is more in line with the probabilistic interpretation
in Equation~(\ref{eq:occlusion_test}).
More precisely, we compute the occlusion probability as
\begin{equation}
\cdfsq( d(p,t) / \tau ) := \cdf ( |d(p, t)|\cdot d(p,t) / \tau)\,.
\label{eq:cdfsq}
\end{equation}
That means that for each choice of~$\cdf$, we obtain a counterpart~$\cdfsq$.
A selection of these for different CDFs~$\cdf$ is visualized in Table~\ref{tab:vis-sigmoids} denoted by ``(squares)''.
For a mathematical definition of each sigmoid function, see Supplementary Material~\ref{sm:dist}.

\begin{table*}[h!]
    \centering
    {\footnotesize
    \addtolength{\tabcolsep}{-2pt}
    \begin{tabular}{lllccccccc}
        \toprule
        T-conorm &             & equal to / where                       & continuous    & contin.~diff. & strict    & idempotent    & nilpotent & Archimedean   & $\uparrow / \downarrow$~wrt.~$p$\\
        \midrule
        (Logical `or') & $\lor$ &                                       & (\xmark)      & (\xmark)      & ---       & (\cmark)      & ---       & ---           & --- \\
        Maximum & $\bot^M$ &                                            & \cmark        & \xmark        & \xmark    & \cmark        & \xmark    & \xmark        & --- \\
        Probabilistic & $\bot^P$ & $=\bot^H_1 =\bot^A_1$                & \cmark        & \cmark        & \cmark    & \xmark        & \xmark    & \cmark        & --- \\
        Einstein & $\bot^E$      & $=\bot^H_0$                          & \cmark        & \cmark        & \cmark    & \xmark        & \xmark    & \cmark        & --- \\
        \midrule
        Hamacher & $\bot^H_p$ & $p\in(0, \infty)$                       & \cmark        & \cmark        & \cmark    & \xmark        & \xmark    & \cmark        & $\downarrow$ \\
        Frank & $\bot^F_p$ & $p\in(0, \infty)$                          & \cmark        & \cmark        & \cmark    & \xmark        & \xmark    & \cmark        & $\downarrow$ \\
        Yager & $\bot^Y_p$ & $p\in(0, \infty)$                          & \cmark        & \xmark        & \xmark    & \xmark        & \cmark    & \cmark        & $\uparrow$ \\
        Acz\'el-Alsina & $\bot^A_p$ & $p\in(0, \infty)$                 & \cmark        & \cmark        & \cmark    & \xmark        & \xmark    & \cmark        & $\uparrow$ \\
        Dombi & $\bot^D_p$ & $p\in(0, \infty)$                          & \cmark        & \cmark        & \cmark    & \xmark        & \xmark    & \cmark        & $\uparrow$ \\
        Schweizer-Sklar & $\bot^{SS}_p$\kern-.25em & $p\in(-\infty, 0)$ & \cmark        & \cmark        & \cmark    & \xmark        & \xmark    & \cmark        & --- \\
        \bottomrule
    \end{tabular}
    }
    \caption{Overview of a selection of suitable T-conorms, which we also benchmark.}
    \label{tab:t-conorms}
\end{table*}
\begin{figure*}[h!]
  \centering
  \includegraphics[width=.23\linewidth,trim={0 0 0 1.6cm},clip]{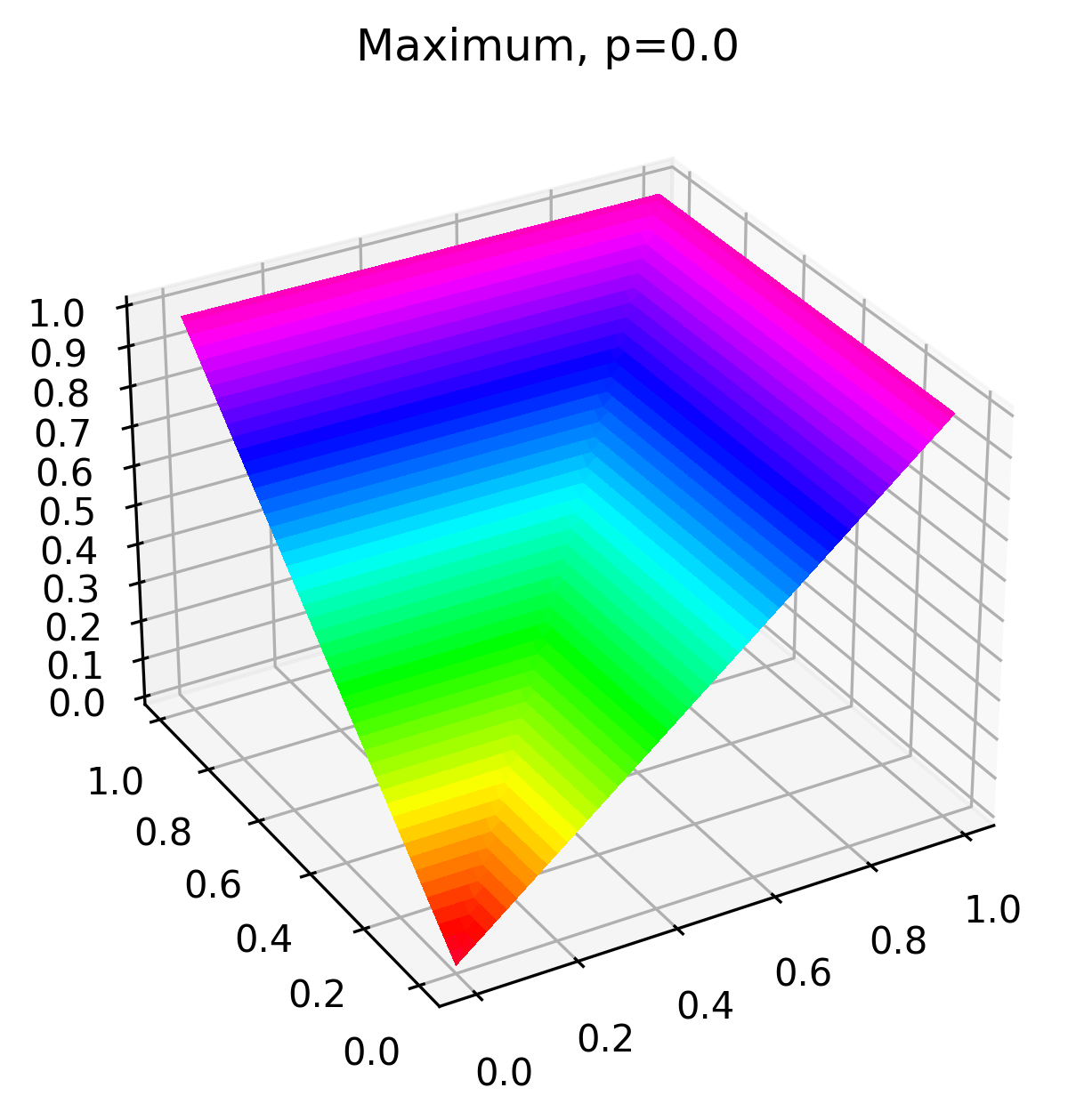}\hfill
  \includegraphics[width=.23\linewidth,trim={0 0 0 1.6cm},clip]{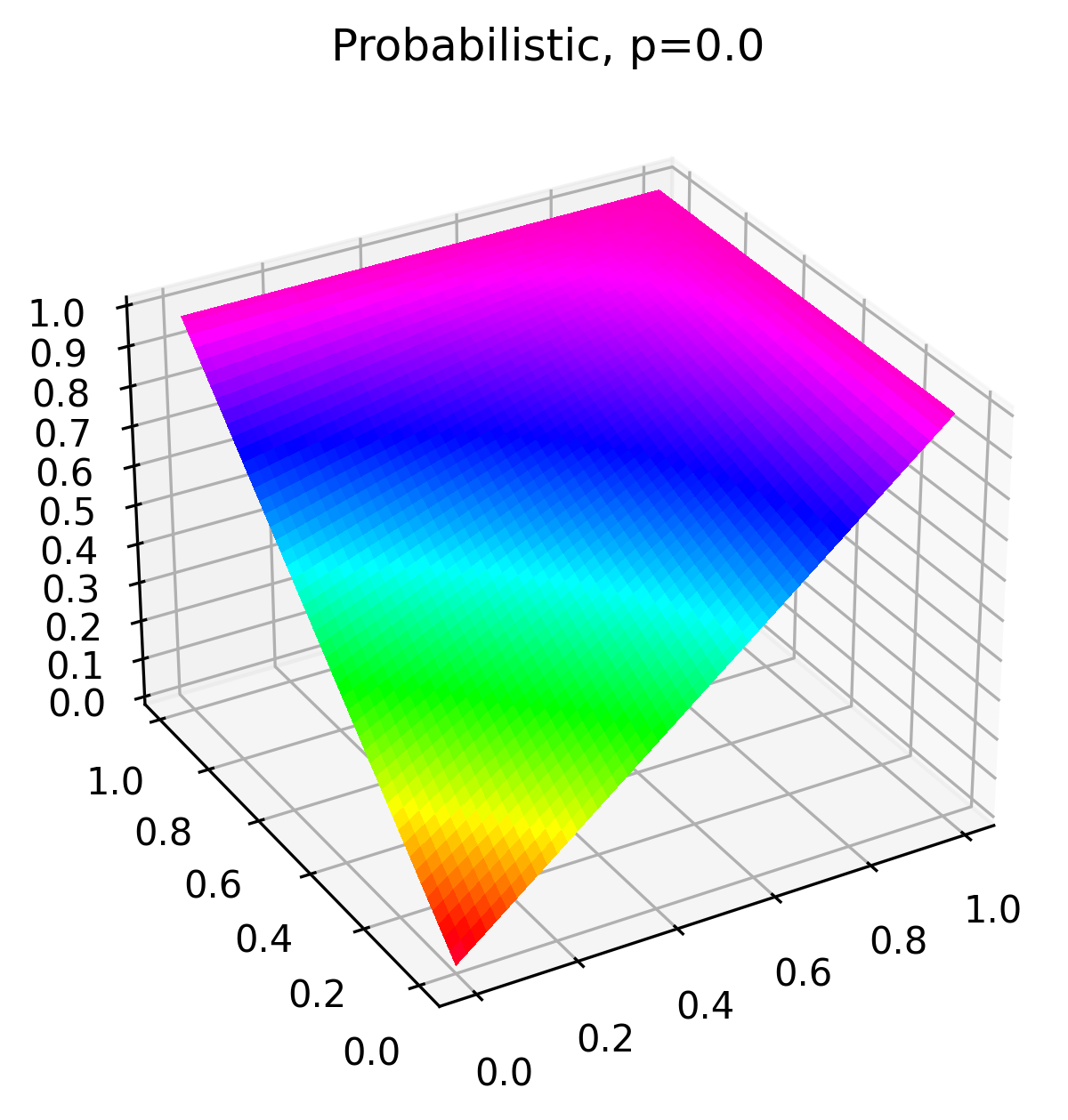}\hfill
  \includegraphics[width=.23\linewidth,trim={0 0 0 1.6cm},clip]{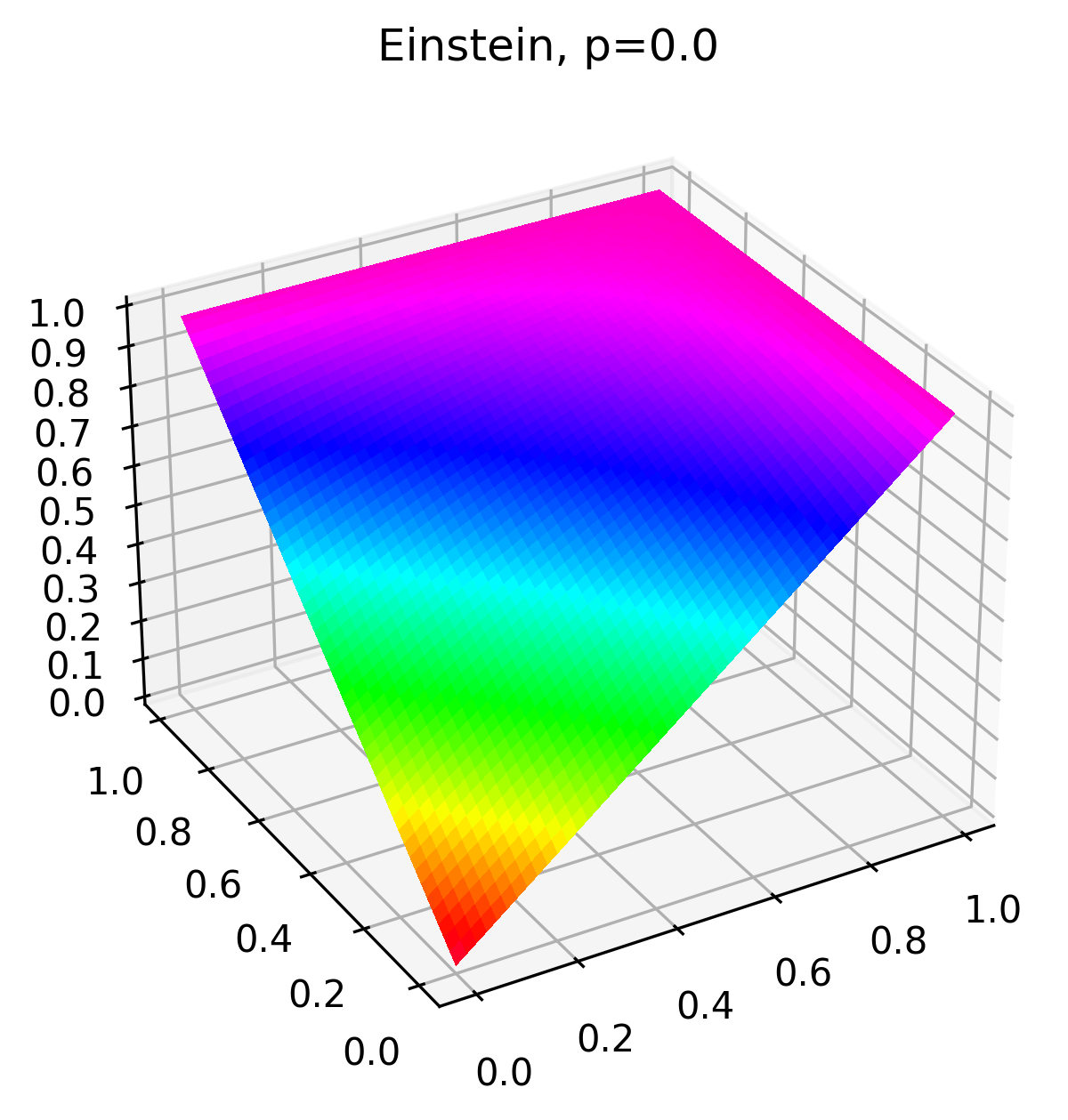}\hfill
  \includegraphics[width=.23\linewidth,trim={0 0 0 1.6cm},clip]{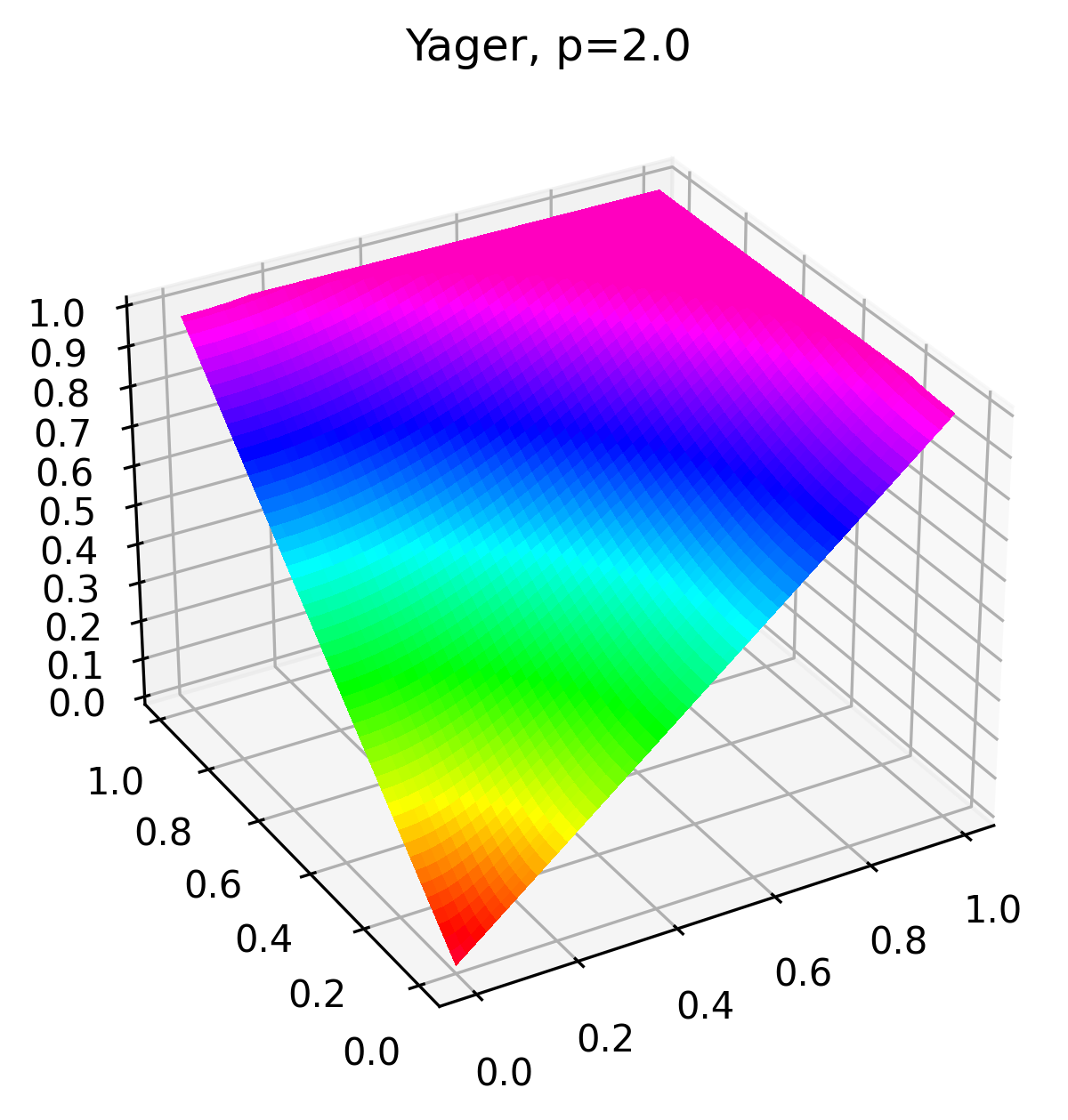}\\[-.25em]
  \noindent
  {\small \hfill (a) \hfill\hspace{.0267\linewidth}\hfill (b) \hfill\hspace{.0267\linewidth}\hfill (c) \hfill\hspace{.0267\linewidth}\hfill (d) \hfill}
  \caption{Plot of four selected T-conorms. \textit{From left to right:} Maximum, Probabilistic, Einstein, and Yager (w/ $p=2$).
    While (b) and (c) are smooth, the Yager T-conorm (d) is non-smooth, it plateaus and the value is constant outside the unit circle.
  }
  \label{fig:t-conorms}
\end{figure*}
\begin{figure}[b!]
    \centering
    \includegraphics[width=.49\linewidth]{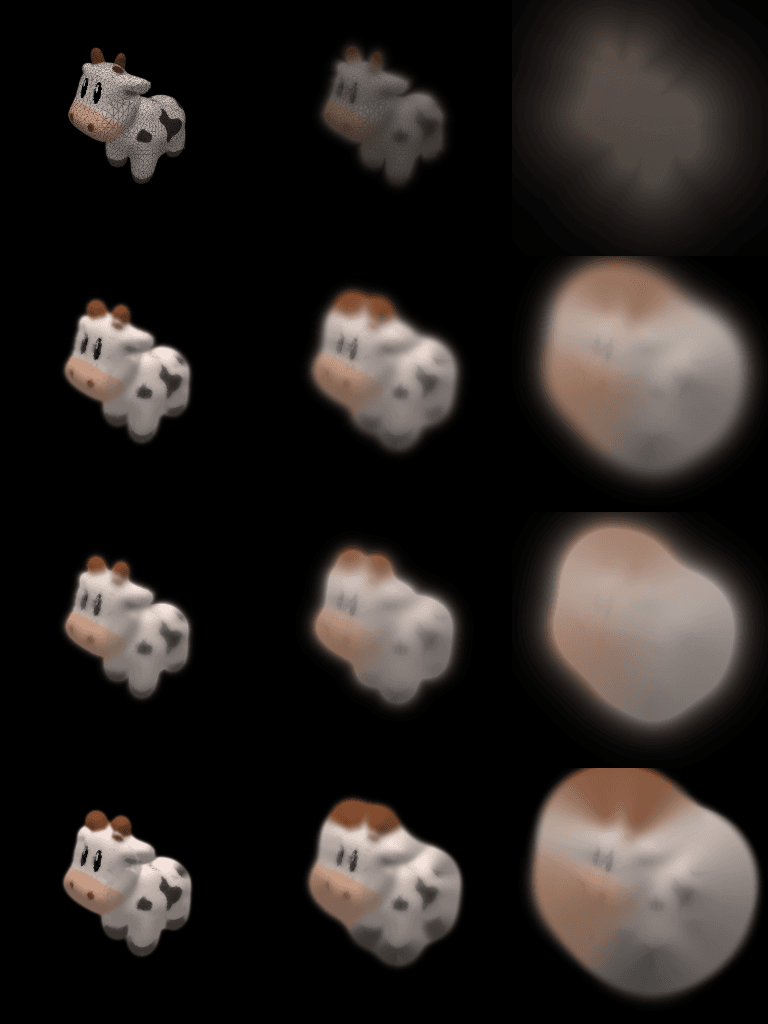}
    \hfill
    \includegraphics[width=.49\linewidth]{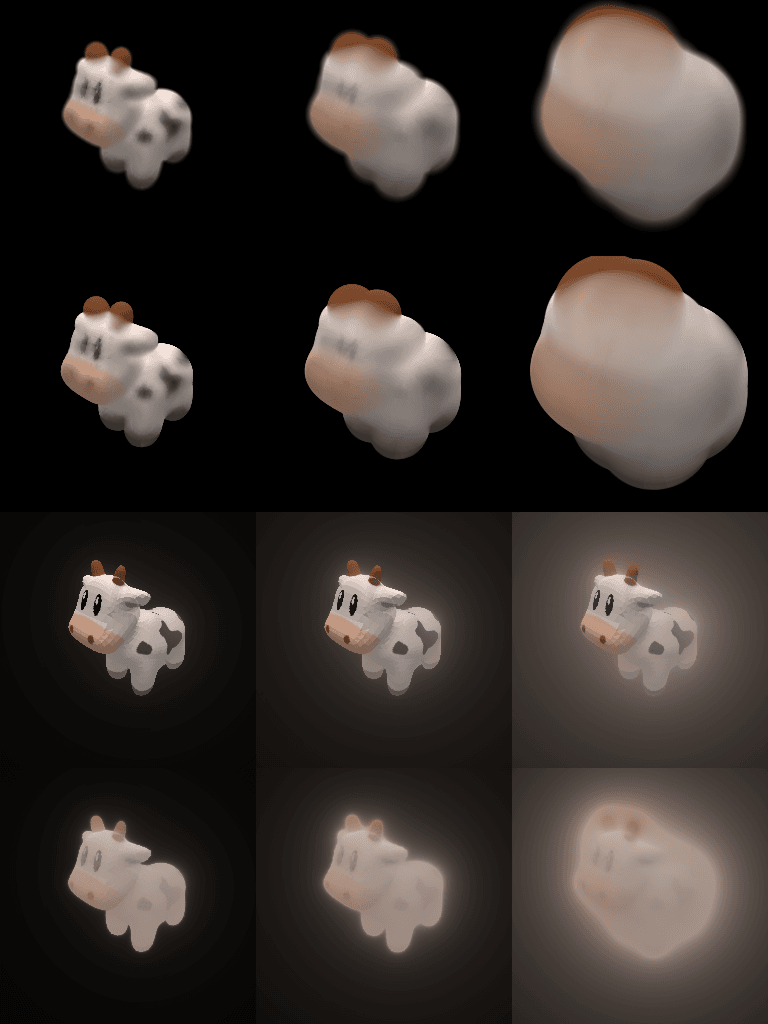}
    \caption{
        Visual comparison of different instances of GenDR.
        In each image, moving from left to right increases the temperature or scale~$\tau$ of the distribution.
        \textit{Left:} we use a logistic distribution to perturb the triangles and use different T-norms for aggregation (top to bottom: $\bot^M, \bot^P, \bot^Y_2, \bot^A_{0.5}$).
        \textit{Right:} for the two first rows, we use a uniform distribution and use $\bot^Y_2$ and $\bot^A_{0.5}$. For the last two rows, we use a Cauchy distribution and use $\bot^P$ and $\bot^Y_2$.
    }
    \label{fig:example-render}
\end{figure}

\paragraph{Aggregations}
Table~\ref{tab:t-conorms} provides an overview of selected T-conorms and displays their properties.
The logical `or' is not a T-conorm but the discrete and discontinuous equivalent, which is why we include it here.
While %
there are also discontinuous T-conorms such as the drastic T-cornom, these are naturally
not suitable for a differentiable renderer, which is why we exclude them.
All except for Max and Yager are continuously differentiable. %

The top four rows in Table~\ref{tab:t-conorms} contain individual T-conorms, and the remainder are families of T-conorms.
Here, we selected only suitable ranges for the parameter~$p$.
Note that there are some cases in which the T-conorms coincide, e.g, $\bot^P=\bot^H_1 =\bot^A_1$.
A discussion of the remaining properties and a mathematical definition of each T-conorm can be found in Supplementary Material~\ref{sm:tcn}.
Figure~\ref{fig:t-conorms} displays some of the T-conorms and illustrates different properties.
In Figure~\ref{fig:example-render}, we display example renderings with different settings and provide a visual comparison of how the aggregation function affects rendering.

\begin{table}
    \centering
    {\footnotesize
    \begin{tabular}{llc}
        \toprule
        Renderer & Distribution & T-conorm \\
        \midrule
        OpenDR~\cite{loper2014opendr} & Uniform (backward) & --- \\
        N3MR~\cite{kato2017neural} & Uniform (backward) & --- \\
        Rhodin~\etal~\cite{rhodin2015versatile} & Gaussian & $\bot^P$ \\
        SoftRas~\cite{kato2017neural} & Square-root of Logistic & $\bot^P$ \\
        Log.~Relax~(Ch.~\ref{ch:algovision}) & Logistic & $\bot^P$ \\
        DIB-R~\cite{chen2019learning} & Exponential & $\bot^P$ \\
        \bottomrule
    \end{tabular}
    }
    \caption{Differentiable renderers that are (approximately) special cases of GenDR. OpenDR and N3MR do not use a specific T-conorm as their forward computation is hard.
    }
    \label{tab:existing-renderers-characterization}
\end{table}

\subheading{Existing Special Cases of GenDR}
In Table~\ref{tab:existing-renderers-characterization}, we list which existing differentiable renderers are conceptually instances of GenDR.
These renderers do each have some other differences, but one key difference lies in the type of distribution employed.
Differences regarding shading were discussed in Section~\ref{sec:gendr:shading}.

\vspace{-.25em}
\section{Experiments}
\vspace{-.25em}

\begin{figure*}[t]
  \centering
  \includegraphics[height=.465\linewidth,trim={0.5cm 0.0cm 1.80cm 0.5cm},clip]{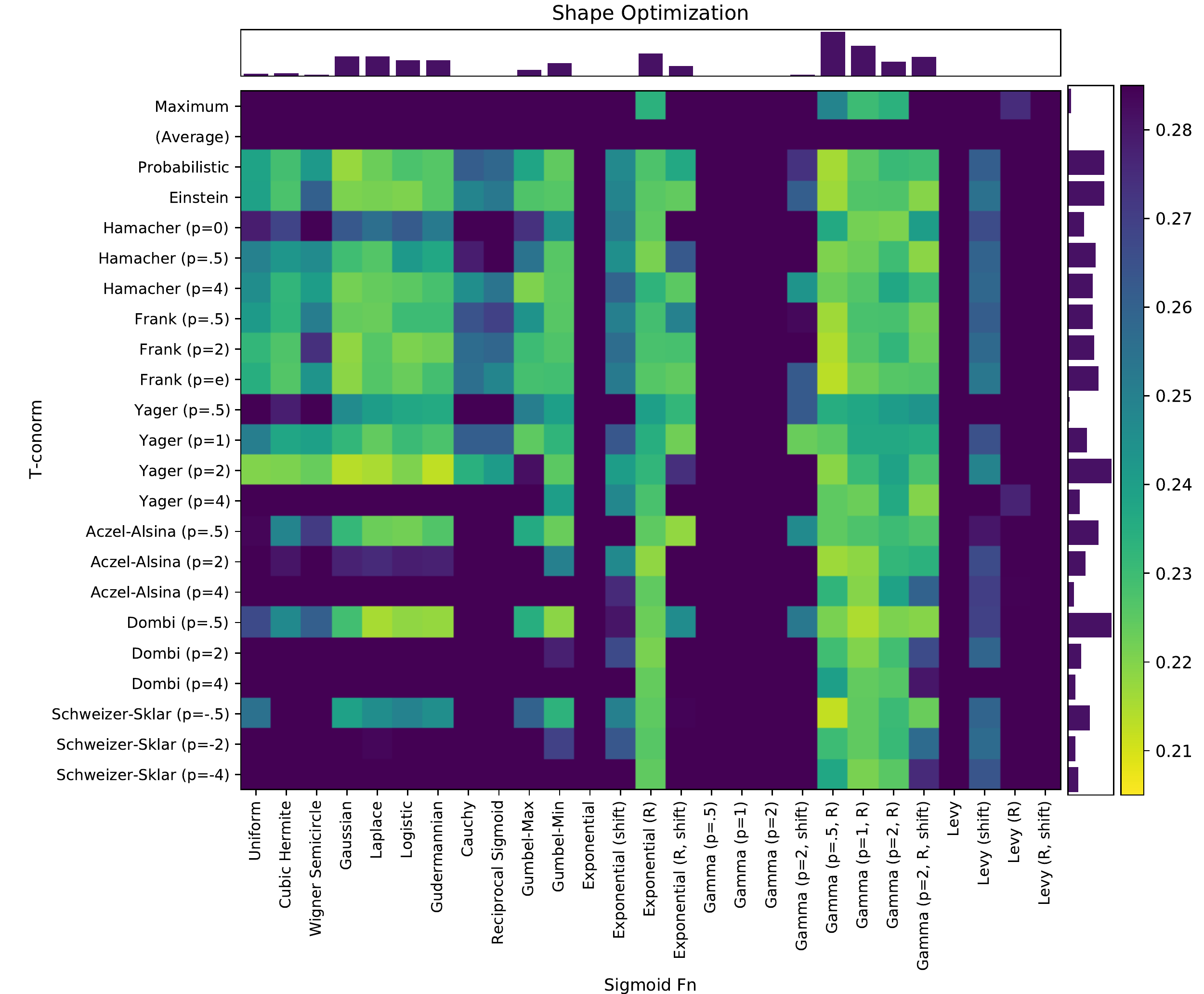}
  \includegraphics[height=.465\linewidth,trim={5.cm 0.0cm 0.25cm 0.5cm},clip]{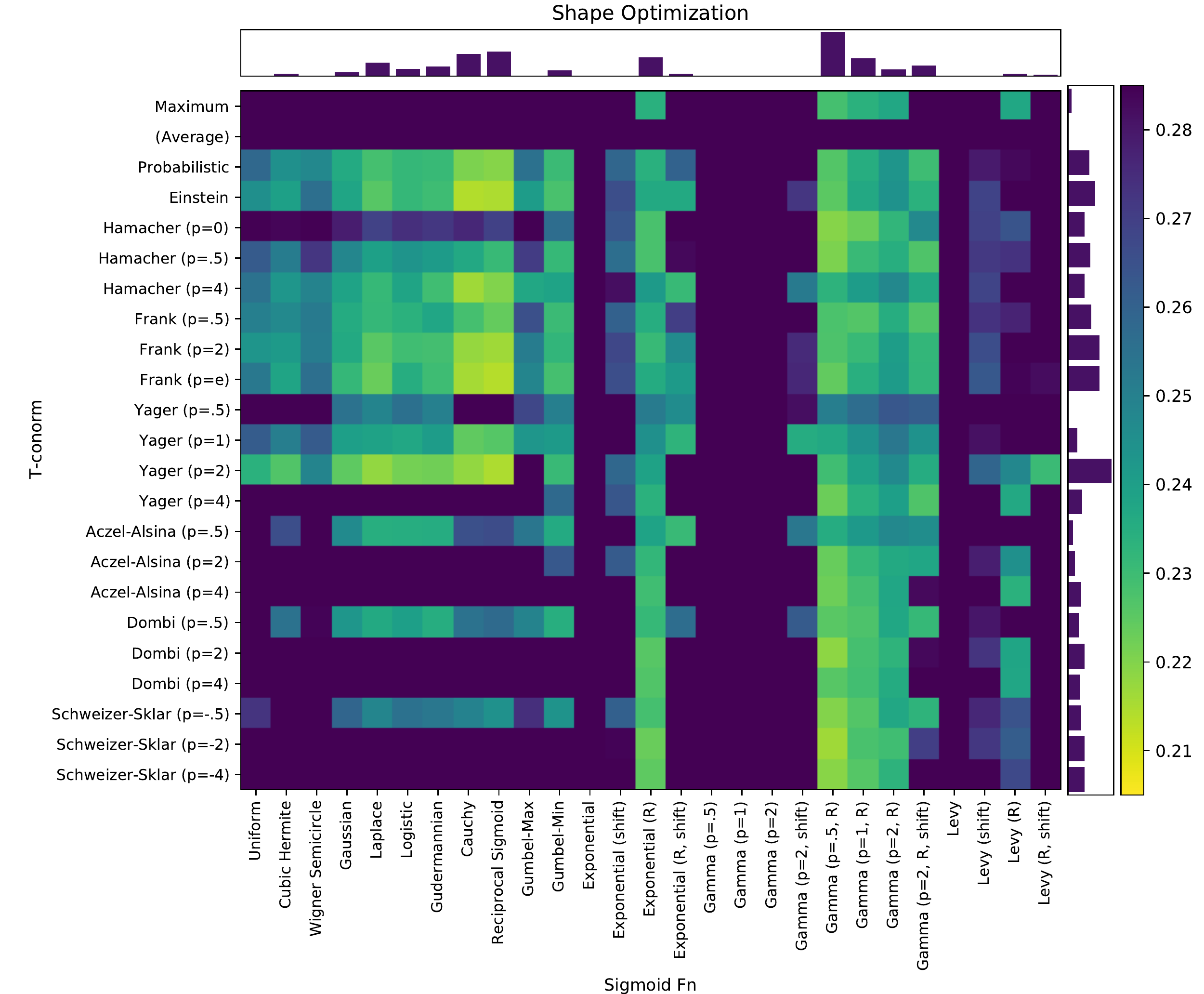}
  \caption{
  Results for the 24-view airplane shape optimization task. The optimization is done within a tight budget of 100 steps and the metric is the loss, i.e., lower (=yellow) is better.
  \textit{Left:} original set of distributions~$\cdf$. 
  \textit{Right:} the respective counter-parts~$\cdfsq$ in the same location.
  The marginal histograms display participation in the top $10\%$ combinations.
  }
  \label{fig:shape-opt-plot}
\end{figure*}

\subsection{Shape Optimization}
\label{sec:shape-opt}
Our first experiment is a shape optimization task.
Here, we use the mesh of an \textit{airplane} and render it from $24$ azimuths using a hard renderer.
The task is to optimize a mesh (initialized as a sphere) to fit the silhouette of the airplane within $100$ optimization steps.
Limiting the task to $100$ optimization steps is critical for two reasons:
(i) The task can be considered to be solved perfectly with any differentiable renderer that produces the correct gradient sign within a large number of steps, but we are interested in the quality of the gradients for the optimization task and how efficient each renderer is.
(ii) The total evaluation is computationally expensive because we evaluate a total of $\numrenderers$ renderers and perform a grid search over the distribution parameters for each one to provide a fair and reliable comparison.

\subheading{Setup}
For optimization, we use the Adam optimizer~\cite{kingma2015adam} with parameters $\beta_1=0.5, \beta_2=0.95$.
For each setting, we perform a grid search over three learning rates ($\lambda\in\{10^{-1.25}, 10^{-1.5}, 10^{-1.75}\}$) and temperatures $\tau \in \{10^{-0.1\cdot n}\,|\,n\in\mathbb{N}, 0\leq n\leq 80\}$.
Here, $\lambda = 10^{-1.5}\approx0.03$ performs best in almost all cases. %
As for the scale hyperparameter, it is important to use a fine-grained as well as large grid because this behaves differently for each distribution.
Here, we intentionally chose the grid larger than the range of reasonable values to ensure that the best choice is used for each setting; the extreme values for the scale were never optimal.
We perform this evaluation from five different elevation angles $\{-60^\circ,-30^\circ,0^\circ,30^\circ,60^\circ\}$ as independent runs and average the final results for each renderer instance.

\subheading{Results}
In Figure~\ref{fig:shape-opt-plot}, we display the results of our evaluation.
We can observe that the regular distributions~$\cdf$ typically perform better
than the counterpart~$\cdfsq$,
except for the case of Cauchy and reciprocal sigmoid, which are those with a linear convergence rate.
We explain this by the fact that by squaring the distance before applying the sigmoid function,
the function has a quadratic convergence rate instead.
As the linearly converging functions also perform poorly in comparison to the exponentially converging functions
(Gaussian, Laplace, Logistic, Gudermannian), we conclude that linear convergence is inferior to
quadratic and exponential convergence.
Columns~$1-3$ contain the distributions with finite support, and these do not perform very well on this task.
The block of exponentially decaying distributions (columns $4-7$) performs well.
The block of linearly decaying distributions (columns $8-9$) performs badly, as discussed above.
The block of Levy distributions (last $4$ columns) performs even worse because it has an even slower convergence.
Here, it also becomes slightly better in the squared setting, but it still exhibits worse performance than for linear convergence.\marginnote{
We note that this result also demonstrates that there is no one optimal distribution for all problems and algorithms, e.g., consider it in relation to differentiable sorting networks where linear convergence is highly desirable. 
}

\subheading{Comparison of Distributions}
Gumbel, exponential, and gamma distributions do not all perform equally well, but Gumbel-Min, the reversed exponential, and the reversed gamma are all competitive.
Confer Table~\ref{tab:vis-sigmoids} where it becomes clear that this is because Gumbel-Max, exponential and gamma have all of their mass
inside the triangle, i.e., they yield smaller faces.
This is problematic because in this case, it can cause gaps between neighboring triangles, which hinders optimization.
These gaps can also be seen in Figure~\ref{fig:example-render} (top left-most rendering).
As the reverse counterparts yield larger faces and do not suffer from this problem, they perform better.
Note that, in this respect, the asymmetrical distributions have an advantage over the symmetrical distributions
because symmetrical distributions always have an accumulated density of~$0.5$ at the edge,
and thus the size of the face stays the same.
We can see that, among the asymmetrical distributions, Gamma performs best.

\subheading{Comparison of T-conorms}
We find that $\bot^M$ and ``average'' (which is not a T-conorm but was used as a baseline in~\cite{liu2019soft}) perform poorly.
Also, $\bot^Y_4$, $\bot^A_2$, $\bot^A_4$, $\bot^D_2$, $\bot^D_4$, $\bot^{SS}_{-2}$, and $\bot^{SS}_{-4}$ perform poorly overall.
This can be explained as they are rather extreme members of their respective T-conorm families; in all of them, the $p$th power is involved, which can become a problematic component, e.g., $x^4$ is vanishingly small for~$x=0.5$.
Interestingly, the gamma and exponential distributions still perform well with these,
likely since they are not symmetric and have an accumulated probability of~$1$ on the edge.
Notably, the Yager T-conorm ($p=2$) performs very well, although having a plateau
and thus zero gradients outside the unit disc\,%
(Figure\,\ref{fig:t-conorms}\,(d)). %

Finally, we compute histograms of how many times each respective distribution and T-conorm is involved in the best~$10\%$ of overall results.
This is independent for the left and right plots.
We can observe that Gamma ($p=0.5$, Reversed) performs the best overall (because it is more robust to the choice of T-conorm).
Among the T-conorms, we find that $\bot^Y_2$ and $\bot^D_{0.5}$ perform best.
The probabilistic and Einstein sums perform equally, and share the next place.

\begin{figure*}[t]
  \centering
  \includegraphics[height=.45\linewidth,trim={0.5cm 0.15cm 2.00cm 0.6cm},clip]{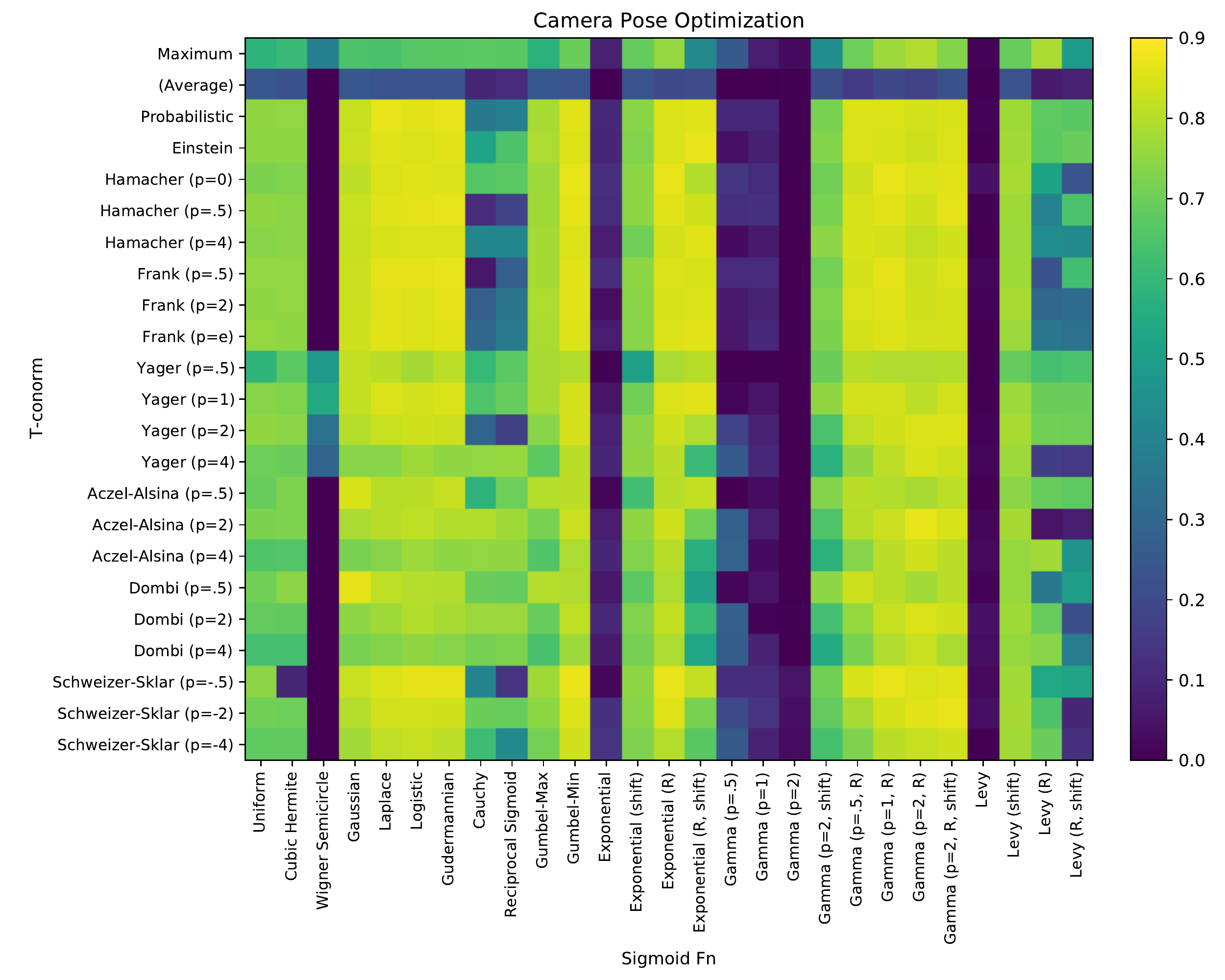}
  \includegraphics[height=.45\linewidth,trim={5.cm 0.15cm 0.25cm 0.6cm},clip]{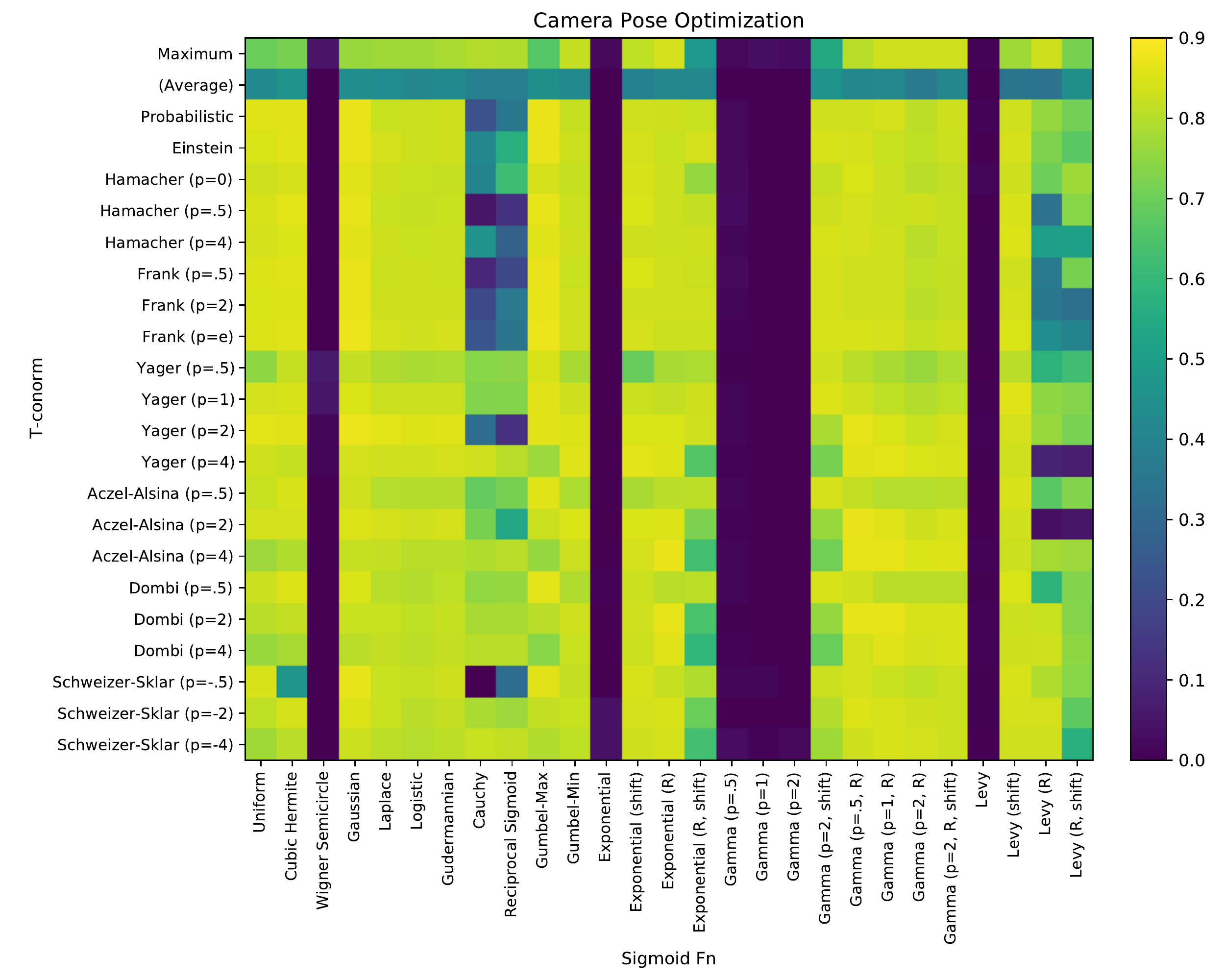}
  \caption{
  Results for the teapot camera pose optimization task. The optimization is done with a temperature $\tau$ that is scheduled to decay.
  The metric is the fraction of camera poses recovered, while the initialization angle errors are uniformly sampled from $[15^\circ, 75^\circ]$.
  The figure shows the original set of distributions~$\cdf$ \textit{(left)} and the respective square-root distribution~$\cdfsq$ \textit{(right)}.
  }
  \label{fig:camera-opt-plot}
  \label{fig:camera-opt-plot-squares}
\end{figure*}

\subsection{Camera Pose Optimization}
In our second experiment, the goal is to find the camera pose for a model of a \textit{teapot} from a reference image.
The angle is randomly modified by an angle uniformly drawn from $[15^\circ, 75^\circ]$, and the distance and camera view angle are also randomized.
We sample~$600$ pairs of a reference image and an initialization and use this set of settings for each method.
For optimization, we use Adam with a learning rate of either~$0.1$ or~$0.3$ (via grid search)
and optimize for~$1000$ steps.
During the optimization, we transition an initial scale of~$\tau=10^{-1}$ logarithmically to a final value of~$\tau=10^{-7}$.
This allows us to avoid a grid search for the optimal scale, and makes sense since an initially large~$\tau$ is beneficial for pose optimization,
because a smoother model has a higher probability of finding the correct orientation of the object.
This contrasts with the setting of shape estimation,
where this would be fatal because the vertices would collapse to the center.

\subheading{Results}
In Figure~\ref{fig:camera-opt-plot}, we display the results of this experiment.
The metric is the fraction of settings which achieved matching the ground truth pose up to $3^\circ$.
We find that in this experiment, the results are similar to those in the shape optimization experiment.
Note that there are larger yellow areas because the color map ranges from $0\%$ to $90\%$, while in the shape optimization plot the color map ranges in a rather narrow loss range.
Additional results for the model of a chair can be found in~\cite{petersen2022gendr}.

\begin{figure*}
    \centering
    \includegraphics[width=\linewidth,trim={0.5cm .0cm 0.5cm 0.cm}]{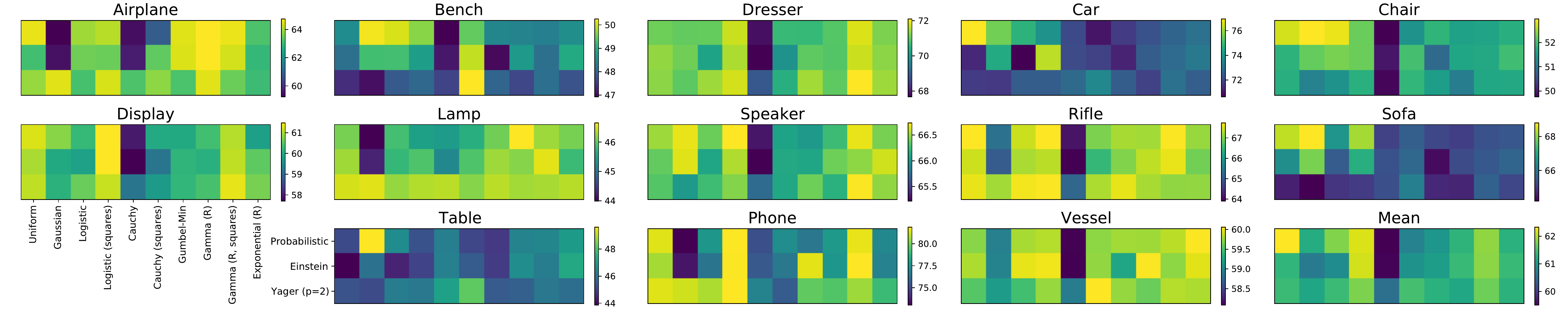}

    \caption{
    Single-view reconstruction results for each of the $30$ selected renderers as a 3D IoU (in \%) heatmap for each class.
    While the uniform distribution (w/ $\bot^P$) performs best on average and the square root of logistic (w/ $\bot^P, \bot^E$) performs second-best on average, the optimal setting depends on the characteristics of the respective classes.
    For the `Airplane' class, the Gamma distribution performed best, and this is also the distribution that performed best in the airplane shape optimization experiment in Section~\ref{sec:shape-opt}.
    For classes of furniture with legs, such as `Bench', `Chair', and `Table', we find that the Gaussian distribution consistently performs best.
    The pairs of similar classes `Display'+`Phone', `Dresser'+`Speaker', and `Vessel'+`Rifle' also show similar performance patterns.
    For example, dresser and speakers tend to be cuboid, while rifles and vessels tend to be rather long and slim.
    Considering the Guassian distribution, it is interesting to see that for some classes $\bot^P$ and $\bot^E$ perform better, while for other classes $\bot^Y_2$ performs much better.
    }
    \label{fig:recon-plots}
\end{figure*}
\begin{table*}[]
    \centering
    \resizebox{\linewidth}{!}{
\setlength{\tabcolsep}{4pt}
\newcommand{\fa}[1]{\textbf{#1}}
\begin{tabular}{lllllllllllllll}
\toprule
Method    & Airplane\kern-1em & Bench  & Dresser\kern-1em & Car    & Chair  & Display\kern-1em & Lamp & Speaker\kern-1em & Rifle & Sofa & Table & Phone & Vessel & \textit{Mean}       \\
\midrule
Kato \textit{et al.}
\cite{kato2017neural} N3MR (Uniform Backward)                          & 0.6172   & 0.4998 & 0.7143  & 0.7095 & 0.4990 & 0.5831  & 0.4126 & 0.6536   & 0.6322 & 0.6735  & 0.4829 & 0.7777 & 0.5645  & 0.6015       \\
Liu \textit{et al.}
\cite{liu2019soft} SoftRas (Square-root of Logistic)          & 0.6419   & 0.5080 & 0.7116  & 0.7697 & 0.5270 & 0.6156  & {0.4628} & 0.6654   & {0.6811} & 0.6878  & 0.4487 & 0.7895 & 0.5953  &  0.6234       \\
Chen~\etal~\cite{chen2019learning} DIB-R (Exponential)                     & 0.570    & 0.498  & 0.763   & 0.788  & 0.527  & 0.588   & 0.403  & 0.726    & 0.561  & 0.677   & 0.508  & 0.743  & 0.609   & 0.612        \\
\midrule
Probabilistic        + Uniform~~($\approx$\cite{kato2017neural, loper2014opendr})              & 0.6456 & 0.4855 & 0.7113 & \fa{0.7696} & 0.5276 & 0.6126 & 0.4611 & 0.6651 & \fa{0.6773} & 0.6835 & 0.4514 & 0.8148 & 0.5971 & \fa{0.6232} \\
Probabilistic        + Logistic~~($=$Ch.~\ref{ch:algovision})       & 0.6396 & 0.5005 & 0.7105 & 0.7471 & 0.5288 & 0.6022 & 0.4586 & 0.6639 & 0.6742 & 0.6660 & 0.4666 & 0.7771 & 0.5980 & 0.6179 \\
Probabilistic        + Logistic (squares)~~($=$\cite{liu2019soft})  & 0.6416 & 0.4966 & 0.7175 & 0.7386 & 0.5224 & \fa{0.6147} & 0.4550 & \fa{0.6673} & \fa{0.6771} & 0.6818 & 0.4529 & \fa{0.8186} & 0.5984 & 0.6217 \\
Probabilistic        + Exponential (R)~~($=$\cite{chen2019learning})         & 0.6321 & 0.4857 & 0.7123 & 0.7298 & 0.5178 & 0.5983 & 0.4611 & 0.6642 & 0.6713 & 0.6546 & 0.4700 & 0.7717 & \fa{0.6005} & 0.6130 \\
Probabilistic        + Gaussian~~($\approx$\cite{rhodin2015versatile})  & 0.5922 & \fa{0.5020} & 0.7104 & 0.7561 & \fa{0.5297} & 0.6080 & 0.4399 & \fa{0.6668} & 0.6533 & \fa{0.6879} & \fa{0.4961} & 0.7301 & 0.5894 & 0.6125 \\
Probabilistic        + Gamma (R)                                        & \fa{0.6473} & 0.4842 & 0.7093 & 0.7220 & 0.5159 & 0.6033 & \fa{0.4665} & 0.6626 & 0.6719 & 0.6505 & 0.4642 & 0.7778 & 0.5978 & 0.6133 \\
Einstein             + Gamma (R, squares)                               & 0.6438 & 0.4816 & \fa{0.7174} & 0.7284 & 0.5170 & 0.6111 & 0.4654 & 0.6647 & 0.6760 & 0.6546 & 0.4626 & \fa{0.8189} & 0.5973 & 0.6184 \\
Yager (p=2)          + Cauchy (squares)                                 & 0.6380 & \fa{0.5026} & 0.7047 & 0.7359 & 0.5188 & 0.5976 & 0.4617 & 0.6612 & 0.6726 & 0.6619 & 0.4819 & 0.7560 & \fa{0.6006} & 0.6149 \\
\bottomrule
\end{tabular}}
    \caption{
    Selected single-view reconstruction results measured in 3D IoU.
    }
\label{tab:recon}
\end{table*}

\subsection{Single-View 3D Reconstruction}
\subheading{Setup}
Finally, we reproduce the popular ShapeNet single-view 3D reconstruction experiment from Section~\ref{sec:exp-3d}.
We select three T-conorms ($\bot^P,\allowbreak \bot^E,\allowbreak \bot^Y_2$) and~$10$ distributions (Uniform, Gaussian, Logistic, Logistic (squares), Cauchy, Cauchy (squares), Gumbel-Min, Gamma (R, $p=0.5$), Gamma (R, $p=0.5$, squares), and Exponential (R)).
These have been selected because they have been used in previous works, are notable (Cauchy, Gumbel-Min, Einstein), or have performed especially well in the aircraft shape optimization experiment (Gamma, Yager).
For each setting, we perform a grid search of~$\tau$ at resolution~$10^{0.5}$.
We use the same model architecture as in Chapter~\ref{ch:algovision} and train with a batch size of $64$ for $250\,000$ steps using the Adam optimizer~\cite{kingma2015adam}.
This also corresponds to the setup by Liu~\etal~\cite{liu2019soft}.
We also schedule the learning rate to $10^{-4}$ for the first $150\,000$ steps and use a learning rate of $3\cdot10^{-5}$ for the remaining training.
At this point (after the first $150\,000$ steps), we also decrease the temperature $\tau$ by a factor of $0.3$.
Using different learning rates (which we did as an ablation) did not improve the results.

\subheading{Results}
In Figure~\ref{fig:recon-plots}, we display and discuss the class-wise results for all $30$ selected renderers.
In Table~\ref{tab:recon}, we show the (self-) reported results for existing differentiable renderers in the top block.
In the bottom block, we display our results for the methods that are equivalent ($=$) or very similar ($\approx$) to the six existing differentiable renderers.
The differences for equivalent methods can be explained with small variations in the setting and minor implementation and framework differences.
Additionally, we include three noteworthy alternative renderers, such as the one that also performed best on the prior airplane shape optimization task.
We conclude that the optimal choice of renderer heavily depends on the characteristics of the 3D models and the task.
Surprisingly, we find that the simple uniform method achieves consistently good results and the best average score.

\subsection*{Conclusion}
In this chapter, we generalized differentiable mesh renderers and explored a large space of instantiations of our generalized renderer GenDR, extending our work in Chapter~\ref{ch:algovision}.
We found that there are significant differences between different distributions for the occlusion test, but also between different T-conorms for the aggregation.
In our experiments, we observed that the choice of renderer has a large impact on the kind of models that can be rendered most effectively.
We find that the uniform distribution outperforms the other tested distributions on average, which is surprising considering its simplicity.
Remarkably, the uniform distribution had already been used implicitly for the early surrogate gradient renderers but was later discarded for the approximate differentiable renderers.

\setchapterpreamble[u]{\pagelogo{difflogic}\margintoc}
\chapter{Differentiable Logic}
\labch{difflogic}

In this chapter, we cover differentiable logic, which is an integral part of differentiable algorithms.
In contrast to the previous chapters, the primary goal of this chapter is not to relax an existing algorithm to apply algorithmic supervision but instead to find an algorithm or logical expression via a differentiable relaxation of the space of logical expressions.
Specifically, we consider the space of logic gate networks, and relax it to differentiable logic gate networks, such that the logic gate network or algorithm can be learned.
Differentiable logic gate networks are a novel kind of neural networks, and in their discretized (non-relaxed) form, they operate on logic gates only, which allows for very fast inference.

With the success of neural networks, there has also always been strong interest in research and industry in making the respective computations as fast and efficient as possible, especially at inference time.
Various techniques have been proposed to solve this problem, including reduced computational precision~\cite{choi2018pact, gupta2015deep}, binary~\cite{qin2020binary} and sparse~\cite{hoefler2021sparsity} neural nets.
In this chapter, we want to train a different kind of architecture, which is well known in the domain of computer architectures: logic (gate) networks.

The problem in training networks of discrete components like logic gates, is that they are non-differentiable and, therefore, conventionally, cannot be optimized via standard methods such as gradient descent \cite{rumelhart1986learning}.
One approach for this would be gradient-free optimization methods such as evolutionary training \cite{telikani2021evolutionary, rapin2018nevergrad}, which works for small models but becomes infeasible for larger ones.

In this chapter, we propose an approach for gradient-based training of logic gate networks.
Logic gate networks are based on binary logic gates, such as ``and'' and ``xor'' (see Table~\ref{tab:operators}).
For training logic gate networks, we continuously relax them to differentiable logic gate networks, which allows efficiently training them with gradient descent.
For this, we use real-valued logic and learn which logic gate to use at each neuron.
After training, the resulting network is binarized to a (hard) logic gate network by choosing the logic gate with the highest probability.
As the (hard) logic gate network comprises logic gates only, it can be executed very fast.
Additionally, as the logic gates are binary, every neuron / logic gate has only 2 inputs, and the network is extremely sparse. 

In contrast to binary neural networks, logic gate networks do not have weights, are intrinsically sparse as they have only 2 inputs to each neuron, and are not simply a form of low precision (wrt.~weights and/or activations) neural networks.

They also differ from current sparse neural network approaches, as our goal is to learn which logic gate operators are present at each neuron while the connections between neurons are (pseudo-)randomly initialized and remain fixed.
The network is, thus, parameterized by the choice of the binary function for each neuron. 
As there is a total of $16$ functions of signature $f: \{0,1\}\times\{0,1\}\to\{0,1\}$, only $4$ bits are required to store the information about which operation a neuron executes.
The objective is to learn which of those $16$ operations is optimal for each neuron.
Specifically, for each neuron, we learn a probability distribution over possible logic gates, which we parameterize via softmax.
We find that this approach allows learning logic gate networks very effectively via gradient descent.

Logic gate networks allow for very fast classification, with speeds beyond a million images per second on a single CPU core (for MNIST at $>97.5\%$ accuracy).
The computational cost of a layer with $n$ neurons is $\Theta(n)$ with very small constants (as only logic gates of Booleans are required), while, in comparison, a fully connected layer (with $m$ input neurons) requires $\Theta(n\cdot m)$ computations with significantly larger constants (as it requires floating-point arithmetic).
While the training can be more expensive than for regular neural networks (however, just by a constant and asymptotically less expensive), to our knowledge, the proposed method is the fastest available architecture at inference time.
Overall, our method accelerates inference speed (in comparison to fully connected ReLU neural networks) by around two orders of magnitude. 
In the experiments, we scale the training of logic gate networks up to 5 million parameters, which can be considered relatively small in comparison to other architectures.
In comparison to the fastest neural networks at $98.4\%$ on MNIST, our method is more than $12\times$ faster than the best binary neural networks and $2-3$ orders of magnitude faster than the theoretical speed of sparse neural networks.

\section{Related Work}
\label{sec:difflogic:rel-work}

In this section, we discuss related work on differentiable logics, learning logic gate networks, and methods with methodological or conceptual similarity.
In addition, we discuss binary and sparse neural networks, which are baselines because they are very fast network architectures, but which are not conceptually similar to differentiable logic gate networks.

\subsection{Differentiable Logics and Triangular Norms}

Differentiable logics (aka.~real-valued logics, or infinite-valued logic) are well-known in the fields of fuzzy logics~\cite{klir1997fuzzy} and probabilistic metric spaces~\cite{menger1942statistical, klement2013triangular}.
In Chapter~\ref{ch:gendr}, we provide an introduction to differentiable logics and triangular norms (T-norms), and in Supplementary Material~\ref{sm:tcn} we give examples for T-norms and T-conorms.
An additional reference for differentiable real-valued logics is Van~\etal~\cite{van2022analyzing}.

\subsection{Learning Logic Gate Networks}

Chatterjee~\cite{chatterjee2018learning} explored ``memorization'', a method for memorizing binary classification data sets with a network of binary lookup tables.
He does this to explore principles of learning and memorization, as well as their trade-off and generalization capabilities.
He constructs the networks of lookup tables by counting conditional frequencies of data points.
We mention this work here because binary logic gates may be seen as the special case of $2$-input lookup tables.
That is, his method has some similarities to our resulting networks.
However, as he memorizes the data set, while this leads to some generalization, this generalization is limited.
In his experiments, he considers the binary classification task of distinguishing the combined classes `0'--`4' from the combined classes `5'--`9' of MNIST and achieves a test accuracy of $90\%$.

Brudermueller~\etal~\cite{brudermueller2020making} propose a method where they train a neural network on a classification task and then translate it, first into random forests, and then into networks of AND-Inverter logic gates, i.e., networks based only on ``and'' and ``not'' logical gates.
They evaluate their approach on the ``gastrointestinal bleeding'' and ``veterans aging cohort study'' data sets and argue for the verifiability and interpretability of small logical networks in patient care and clinical decision-making.

\subsection{Relaxed Connectivity in Networks}

For differentiable logic gate network, we relax \textit{which logic operator} is applied at each node, while the connections are predefined.

Zimmer~\etal~\cite{zimmer2021differentiable} propose differentiable logic machines for inductive logic programming. 
For this, they propose logic modules, which contain one level of logic and for which they predefine that the first half of operators are fuzzy ``and''s and the second half are fuzzy ``or''s.
They relax \textit{which nodes} are the inputs to the ``and''s and ``or''s of their logic modules.

Similarly, Chen~\cite{chen2020learning} proposes Gumbel-Max Equation Learner Networks, where he predefines a set of arithmetic operations in each layer and learns via Gumbel-Softmax~\cite{maddison2017concrete, jang2017categorical}, \textit{which outputs} of the previous layer should be used as inputs of a respective arithmetic operation.
He uses this to learn symbolic expressions from data.

While these works relax which nodes are connected to which nodes, this is fixed in our method, and we relax which operator is at which node.

\subsection{Evolutionary Learning of Networks}

Mocanu~\etal~\cite{mocanu2018scalable} propose training neural networks with sparse evolutionary training inspired by network science.
Their method evolves an initial sparse topology of two consecutive layers of neurons into a scale-free topology.
On MNIST, they achieve (with $89\,797$ parameters) an accuracy of $98.74\%$.

Gaier~\etal~\cite{gaier2019weight} propose learning networks of operators such as ReLU, sin, inverse, absolute, step, and tanh using evolutionary strategies.
Specifically, they use the population-based neuroevolution algorithm NEAT.
They achieve learning those floating-point function--based networks and achieve an accuracy of $94.2\%$  on MNIST with a total of $1\,849$ connections.

\subsection{Learning of Decision Trees}

Zantedeschi~\etal~\cite{zantedeschi2021learning} propose to learn decision trees by quadratically relaxing the decision trees from mixed-integer programs that learn the discrete parameters of the tree (input traversal and node pruning).
This allows them to differentiate in order to simultaneously learn the continuous parameters of splitting decisions. 

\marginnote{
While decision trees are different from logic gate networks, the methods presented in Chapter~\ref{ch:algovision} could be used to continuously relax decision trees.
}
Logic gate networks can be seen as collections of binary (logic gate-based) trees, but binary (logic gate-based) trees are conceptually vastly different from \textit{decision} trees:
decision trees rely on splitting decisions instead of logical operations, and the tree structure of decision trees and logic gate-based trees are in the opposite directions~\cite{clark1989cn2}.
Logic gate-based trees begin with a number of inputs (leafs) and apply logic gates to aggregate them to a binary value (root).
Decision trees begin at the root and apply splitting decisions (for which they consider an external input) to decide between children, such that they end up at a leaf node corresponding to a value.

\subsection{Binary Neural Networks}
Binary neural networks (BNNs)~\cite{qin2020binary} are conceptually very different from logic gate networks.
For binary neural networks, ``binary'' refers to representing activations and weights of a neural network with binary states (e.g., $\{-1, +1\}$).
This allows approximating the expensive matrix multiplication by faster XNOR and bitcount (popcount) operations.
The logical operations involved in BNNs are not learned but instead predefined to approximate floating-point operations, and, as such, a regular weight-based neural network.
This is not the case for logic gate network, where we learn the logic operations, we do not approximate weight-based neural networks, and do not have weights.
While BNNs are defined via their weights and not via their logic operations, logic gate networks do not have weights and are purely defined via their logic operations.
We include BNNs as baselines in our experiments because they achieve the best inference speed.

\subsection{Sparse Neural Networks}

Sparse neural networks~\cite{hoefler2021sparsity} are neural networks where only a selected subset of connections is present, i.e., instead of fully-connected layers, the layers are \textit{sparse}.
In the literature of sparse neural networks, usually, the task is to distill a sparse neural network from a dense neural network and the choice of connections is important.
While logic gate networks are sparse by definition, their connections remain fixed and their initialization is random.
We include sparse nets in our experiments because they can provide efficient inference.

\section{Logic Gate Networks}

Logic gate networks are networks similar to neural networks where each neuron is represented by a binary logic gate like `and', `nand', and `nor' and accordingly has only two inputs (instead of all neurons in the previous layer as it is the case in fully-connected neural networks).
Given a binary vector as input, pairs of Boolean values are selected, binary logic gates are applied to them, and their output is then used as input for layers further downstream.
Logic gate networks do not use weights. Instead, they are parameterized via the choice of logic gate at each neuron.
In contrast to fully connected neural networks, binary logic gate networks are sparse because each neuron has only $2$ instead of $n$ inputs, where $n$ is the number of neurons per layer.

In logic gate networks, we do not need activation functions as they are intrinsically non-linear.

While it is possible to make a prediction simply with a single binary output or $k$ binary outputs for $k$ classes, this is not ideal.
This is because in the crisp case, we only get $0$s or $1$s and no graded prediction, which would be necessary for a ``greatest activation'' classification scheme.
By using multiple neurons per class and aggregating them by summation, even the crisp case allows for grading, with as many levels as there are neurons per class.
Each of these neurons could capture a different piece of evidence for a class, and this allows finer grade predictions.

\begin{figure*}[t]
    \centering
    \includegraphics[width=\linewidth]{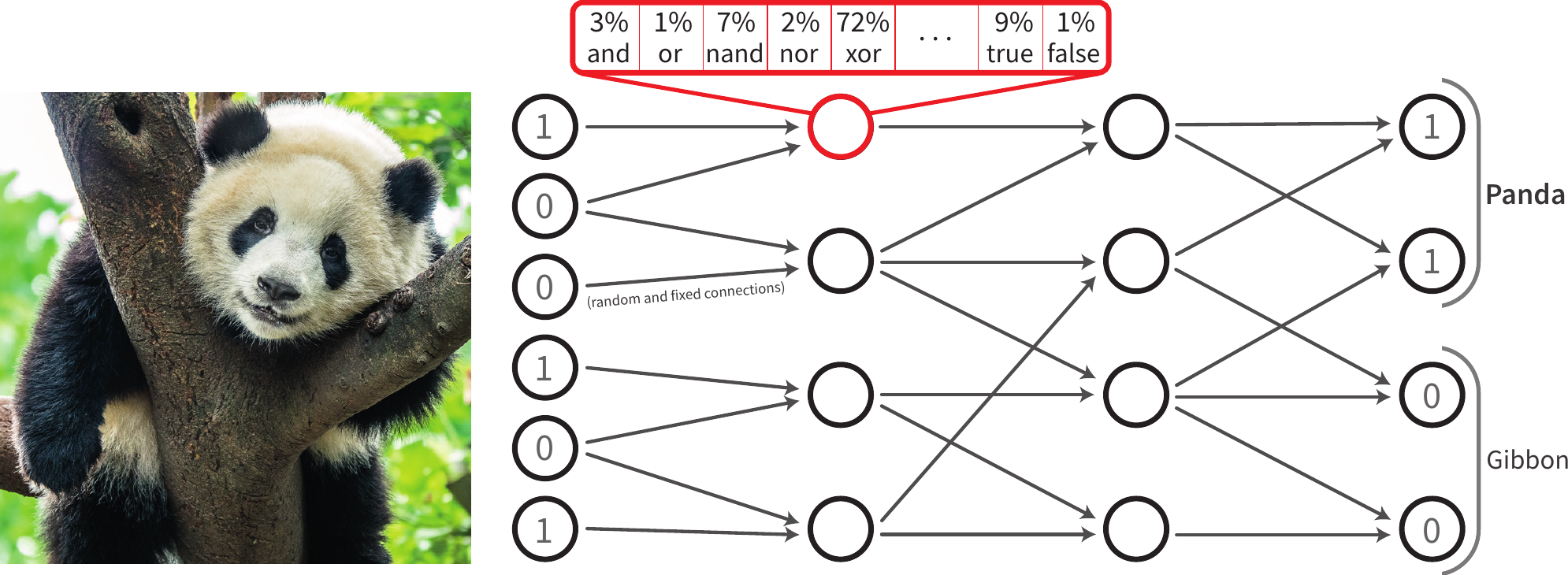}
    \caption{
    Overview of the proposed differentiable logic gate networks: Boolean valued inputs are processed by a layer of neurons such that each neuron receives two inputs.
    The connectivity of neurons remains fixed after an initial pseudo-random initialization.
    Each neuron is continuously parameterized by a distribution over logical operators.
    During training, this distribution is learned for each neuron, and, during inference, the most likely operator is used for each neuron.
    There are multiple outputs per class, which are aggregated by bit-counting, which yields the class scores. %
    }
    \label{fig:overview-small-net}
\end{figure*}

Figure~\ref{fig:overview-small-net} illustrates a small logic gate network.
In the illustration, each node corresponds to a single logic operator.
Note that the distribution over operators (red) is part of the differentiable relaxation discussed in the next section.

\section{Differentiable Logic Gate Networks}

Training binary logic gate networks is hard because they are not differentiable, and thus no gradient descent-based training is conventionally possible.
Thus, we propose relaxing logic gate networks to differentiable logic gate networks to allow for gradient-based training.

\paragraph{Differentiable Logics}
To make binary logic networks differentiable, we leverage the following relaxation.
First, instead of hard binary activations / values $a \in \{0,1\}$, we relax all values to probabilistic activations $a \in [0,1]$.
Second, we replace the logic gates by computing the expected value of the activation given probabilities of independent inputs $a_1$ and $a_2$.
For example, the probability that two independent events with probabilities $a_1$ and $a_2$ both occur is $a_1\cdot a_2$.
We report differentiable real-valued relaxations of all binary logic gate operators in Table~\ref{tab:operators}.
These operators correspond to the probabilistic T-norm and T-conorm.\\

Accordingly, we define the activation of a neuron with the $i$th operator as
\begin{equation}
    a' = f_i(a_1, a_2)\,,
\end{equation}
where $f_i$ is the $i$th real-valued operator corresponding to Table~\ref{tab:operators} and $a_1, a_2$ are the inputs to the neuron.
There are also alternative real-valued logics like the Hamacher T-(co)norm, the relativistic Einstein sum, and the Łukasiewicz T-(co)norm.
While, in this chapter, we use the probabilistic interpretation, we review an array of possible alternative T-norms and T-conorms in Supplementary Material~\ref{sm:tcn}.

\begin{table}[h]
    \centering
    \caption{
        Differentiable real-valued binary logic gate operators corresponding to the probabilistic T-norm and T-conorm.
        \label{tab:operators}
    }
    {
    \scriptsize
    \addtolength{\tabcolsep}{-1pt}
    \begin{tabular}{rclccccccccc}
        \toprule
        ID    & Operator                    & real-valued   & 00 & 01 & 10 & 11 \\
        \midrule
           0  & False                       & $0$                       & 0     & 0     & 0     & 0     \\
           1  & $A\land B$                  & $A\cdot B$                & 0     & 0     & 0     & 1     \\
           2  & $\neg(A \Rightarrow B)$     & $A-AB$                    & 0     & 0     & 1     & 0     \\
           3  & $A$                         & $A$                       & 0     & 0     & 1     & 1     \\
           4  & $\neg(A \Leftarrow B)$      & $B-AB$                    & 0     & 1     & 0     & 0     \\
           5  & $B$                         & $B$                       & 0     & 1     & 0     & 1     \\
           6  & $A \oplus B$                & $A + B - 2AB$             & 0     & 1     & 1     & 0     \\
           7  & $A \lor B$                  & $A + B - AB$              & 0     & 1     & 1     & 1     \\
           8  & $\neg(A \lor B)$            & $1 - (A + B - AB)$        & 1     & 0     & 0     & 0     \\
           9  & $\neg(A \oplus B)$          & $1 - (A + B - 2AB)$       & 1     & 0     & 0     & 1     \\
           10 & $\neg B$                    & $1 - B$                   & 1     & 0     & 1     & 0     \\
           11 & $A \Leftarrow B$            & $1-B+AB$                  & 1     & 0     & 1     & 1     \\
           12 & $\neg A$                    & $1-A$                  & 1     & 1     & 0     & 0     \\
           13 & $A \Rightarrow B$           & $1-A+AB$                  & 1     & 1     & 0     & 1     \\
           14 & $\neg(A \land B)$           & $1 - AB$                  & 1     & 1     & 1     & 0     \\
           15 & True                        & $1$                       & 1     & 1     & 1     & 1     \\
        \bottomrule
    \end{tabular}
    }
\end{table}

\paragraph{Differentiable Choice of Operator} 
While real-valued logics allow differentiation, they do not allow training as the operators are not continuously parameterized, and thus (under hard binary inputs) the activations in the network will always be $a \in \{0, 1\}$.
Thus, we propose to represent the choice of \textit{which} logic gate is present at each neuron by a categorical probability distribution.
For this, we parameterize each neuron with $16$ floats (i.e., $\vw\in\sR^{16}$), which, by softmax, map to the probability simplex (i.e., a categorical probability distribution such that all entries sum up to $1$ and it has only non-negative values).
That is, $\vp_i = e^{\vw_i} / (\sum_j e^{\vw_j})$, and thus $\vp$ lies in the probability simplex $\vp\in\Delta^{15}$.
During training, we evaluate for each neuron all $16$ relaxed binary logic gates and use the categorical probability distribution to compute their weighted average.
Thus, we define the activation $a'$ of a differentiable logic gate neuron as
\begin{equation}
    a' = \sum_{i=0}^{15} \vp_i \cdot {f}_i(a_1, a_2) = \sum_{i=0}^{15} \frac{e^{\vw_i}}{\sum_j e^{\vw_j}} \cdot {f}_i(a_1, a_2)\,.
\end{equation}

\paragraph{Aggregation of Output Neurons}
Now, we may have $n$ output neurons $a_1, a_2, ..., a_n\in[0,1]$, but we may want the logic gate network to only predict $k<n$ values of a larger range than $[0,1]$.
Further, we may want to be able to produce graded outputs.
Thus, we can aggregate the outputs as
\begin{equation}
    \hat y_i = {\sum_{j=i\cdot n/k + 1}^{(i+1)\cdot n/k}}  a_j\, /\, \tau + \beta\,
\end{equation}
where $\tau$ is a normalization temperature and $\beta$ is an optional offset.

\subsection{Training Considerations}

\paragraph{Training}
For learning, we randomly initialize the connections and the parameterization of each neuron.
For the initial parameterization of each neuron, we draw elements of $q$ independently from a standard normal distribution.
In all reported experiments, we use the same number of neurons in each layer (except for the input) and between $4$ and $8$ layers, which we call straight network.
We train all models with the Adam optimizer \cite{kingma2015adam} at a constant learning rate of $0.01$.

\paragraph{Discretization}
After training, during inference, we discretize the probability distributions by only taking their mode (i.e., their most likely value), and thus the network can be computed with Boolean values, which makes inference very fast.
In practice, we observe that most neurons converge to one logic gate operation; therefore, the discretization step introduces only a small error.
We note that all reported results are accuracies after discretization.

\paragraph{Classification}
In the application of a classification learning setting with $k$ classes (e.g., $10$) and $n$ output neurons (e.g., $1\,000$), we group the output into $k$ groups of size $n/k$ (e.g., $100$).
Then, we count the number of $1$s which corresponds to the classification score such that the predicted class can be retrieved via the $\argmax$ of the class scores.
During differentiable training, we sum up the probabilities of the outputs in each group instead of counting the $1$s, and we can train the model using a softmax cross-entropy classification loss.
For a reference on choosing the hyperparameter $\tau$, see Section~\ref{sec:difflogic:hyper-model-arch}; the offset $\beta$ is not relevant for the classification setting, as $\argmax$ is shift-invariant.
A heuristic for choosing $\tau$ is that when increasing $n$, we have to reduce $\tau$.
Empirically, when increasing $\tau$ by a factor of $10$, $\tau$ should be decreased by a factor of around $2$ to $\sqrt{10}$.

\paragraph{Regression}
For regression learning, let us assume that we need to predict a $k$-dimensional output vector.
Here, $\tau$ and $\beta$ play the role of an affine transformation to transform the range of possible predictions from $0$ to $n/k$ to an application specific and more suitable range.
Here, the optional bias $\beta$ is important, e.g., if we want to predict values outside the range of $[0, n/k/\tau]$.
In some cases, it is desirable to cover the entire range of real numbers, which may be achieved using a logit transform $\operatorname{logit}(x)=\sigma^{-1}(x) = \log \frac{x}{1-x}$ in combination with $\tau=n/k, \beta=0$.
During differentiable training, we sum up the probabilities of the outputs in each group instead of counting the $1$s, and we can train the model, e.g., using an MSE loss.

\subsection{Remarks}

\paragraph{Boolean Vectorization via Larger Data Types}
One important computational detail for inference time is that we do not use Boolean data types but instead use larger data types such as, e.g., int64 for a batch size of $64$, and thus perform bit-wise logics on larger batches which significantly improves speed on current hardware.
For int64, we batch $64$ data points such that the $i$th Boolean value of the $j$th data point is the $j$th bit in the $i$th int64 integer.
Thus, it is possible to compute on average around $250$ binary logic gates on each core in each CPU clock cycle (i.e., per Hz) on a typical desktop / notebook computer.
This is the case because modern CPUs execute many instructions per clock cycle even on a single core, and additionally (for Booleans) allow single-instruction multiple-data (SIMD) by batching bits from multiple data points into one integer (e.g., int64).
Using advanced vector extensions (AVX), even larger speedups would be possible.
On GPU, this computational speedup is also available in addition to typical GPU parallelization.

\paragraph{Aggregation of Output Neurons via Binary Adders}
In addition, during inference, we aggregate the output neurons directly using logic gate nets that make up respective adders, as writing all outputs to memory would constitute a bottleneck and aggregating them using logic gate networks is fast.
Specifically, using logic gates, we construct adders that can add exactly one bit to a binary number.
Thus, the logic gate networks return the aggregated results in the form of $\lceil\log_2 (n/k+1)\rceil$-bit binary numbers.

\paragraph{Memory Considerations}
Since we pseudo-randomly initialize the connections in binary logic gate networks, i.e., which are the two inputs for each neuron, we do not need to store the connections as they can be reproduced from a single seed.
Thus, it suffices to store the $4$-bit information which of the $16$ logic gate operators is used for each neuron.
Thus, the memory footprint of logic gate networks is drastically reduced in comparison to neural networks, binary neural networks, and sparse neural networks.

\paragraph{Pruning the Model}
An additional speedup for the inference of logic gate networks is available by pruning neurons that are not used, or by simplifying logical expressions.
However, this requires storing the connections, posing a (minor) trade-off between memory and speed.

\paragraph{Subset of Operators}
We investigated reducing the set of operators; however, we found that, in all settings, the more expressive full set of $16$ operators performed better.
Nevertheless, a smaller set of operators could be a good trade-off for reducing the model size.

\paragraph{Half Precision}
We also investigated training with half precision (float16).
In our experiments, half precision (in comparison to full precision) did not degrade training performance; nevertheless, all reported results were trained with full precision (float32).

\subsection{Current Limitations and Opportunities}

\paragraph{Expensive Training}
A limitation of differentiable logic gate networks is their relatively higher training cost compared to (performance-wise) comparable conventional neural networks.
The higher training cost is because multiple differentiable operators need to be evaluated for each neuron, and in their real-valued differentiable form, most of these operators require floating-point value multiplications.
However, the practical computational cost can be reduced through improved implementations.
Additionally, differentiable logic gate networks are asymptotically cheaper to train compared to conventional neural networks due to their sparsity.

\paragraph{Convolutions and Other Architectures}
Convolutional logic gate networks and other architectural components such as residual connections are interesting and important directions for future research.

\paragraph{Edge Computing and Embedded Machine Learning}
We would like to emphasize that the current limitations to rather small architectures (compared to large deep learning architectures) does not need to be a limitation:
For example, in edge computing and embedded machine learning~\cite{murshed2021machine, ajani2021overview, seng2021embedded, branco2019machine}, models are already limited to tiny architectures because they run, e.g., on mobile CPUs, microcontrollers, or IoT devices.
In these cases, training cost is not a concern because it is done before deployment.

We also note that there are many other applications in industry where the training cost is negligible in comparison to the inference cost.

\section{Experiments}

To empirically validate our method, we perform an array of experiments.
We start with the three MONK data sets and continue to the Adult Census and Breast Cancer data sets.
For each experiment, we compare our method to other methods in terms of model memory footprint, evaluation speed, and accuracy.
To demonstrate that our method also performs well on small-scale image recognition, we benchmark it on the MNIST as well as the CIFAR-10 data sets.
We benchmark the speeds and computational complexity of our method in comparison to baselines.
Finally, we investigate the distributions of logic gates for each layer.

\subsection{MONK's Problems}

The MONK's problems \cite{thrun1991monk} are 3 classic machine learning tasks that have been used to benchmark learning algorithms.
They consist of 3 binary classification tasks on a data set with 6 attributes with $2-4$ possible values each.
Correspondingly, the data points can be encoded as binary vectors of size 17.
In Table~\ref{tab:monk}, we show the performance of our method, a regular neural network, and a few of the original learning methods that have been benchmarked.
We give the prediction speed for a single CPU thread, the number of parameters, and storage requirements.

On all three data sets, our method performs better than logistic regression and on MONK-3 (which is the data set with label noise) our method even outperforms the much larger neural network.

\begin{table}[h]
    \centering
    \caption{
        Results on the MONK data sets. The inference times are per data point for 1 CPU thread. Averaged over $10$ runs.
        For Diff Logic Nets, $\#$ Parameters and Space vary between the MONK data sets as we use different architectures.
    }
    {\footnotesize
    \begin{tabular}{lccc}
    \toprule
        Method & MONK-1 & MONK-2 & MONK-3 \\
    \midrule
        Decision Tree Learner (ID3) \cite{quinlan1986induction}     & $98.6\%$ & $67.9\%$ & $94.4\%$ \\
        Decision Tree Learner (C4.5) \cite{salzberg1994c4}     & $100\%$ & $70.4\%$ & $100\%$ \\
        Rule Learner (CN2) \cite{clark1989cn2}             & $100\%$ & $69.0\%$ & $89.1\%$ \\
        Logistic Regression             & $71.1\%$ & $61.4\%$ & $97.0\%$ \\
        Neural Network                  & $100\%$ & $100\%$ & $93.5\%$ \\
        Diff Logic Net (\textit{ours})   & $100\%$ & $90.9\%$ & $97.7\%$ \\
    \toprule
         & \# Parameters & Inf.~Time & Space \\
    \midrule
        Decision Tree Learner   & $\approx30$ & $49\textrm{ns}$ & $\approx60\textrm{B}$ \\
        Logistic Regression     & $20$  &   $68\textrm{ns}$     & $80\textrm{B}$ \\
        Neural Network          & $162$ &   $152\textrm{ns}$     & $648\textrm{B}$ \\
        Diff Logic Net (\textit{ours}) & $144\,|\,72\,|\,72$ & $18\textrm{ns}$ & $72\textrm{B}\,|\,36\textrm{B}\,|\,36\textrm{B}$ \\
    \bottomrule
    \end{tabular}
    }
    \label{tab:monk}
\end{table}

\subsection{Adult and Breast Cancer Data Sets}

For our second set of experiments, we consider the Adult Census \cite{kohavi1996uci} and the Breast Cancer data set \cite{zwitter1988uci}.
We find that our method performs very similar to neural networks and logistic regression on the Adult data set while achieving a much faster inference speed.
On the Breast Cancer data set, our method achieves the best performance while still being the fastest model.
We present the results in Table~\ref{tab:adult-breast}.

\begin{table}[h]
    \centering
    \caption{Results for the Adult and Breast Cancer data sets averaged over 10 runs.}
    {\footnotesize
    \begin{tabular}{lcccc}
    \toprule
        \textbf{Adult} & Acc. & \# Param. & Infer.~Time & Space \\
    \midrule
        Decision Tree Learner           & $79.5\%$ & \kern-1.5ex$\approx50$ & $86$\textrm{ns} & \kern-2.3ex$\approx130$B \\
        Logistic Regression             & $84.8\%$ & $234$ & $63\textrm{ns}$ & $936$B      \\
        Neural Network                  & $84.9\%$ & $3810$ & $635\textrm{ns}$ & $15$KB           \\
        Diff Logic Net (\textit{ours})   & $84.8\%$ & $1280$ & $5.1\textrm{ns}$ & $640$B           \\
    \toprule
        \textbf{Breast Cancer} & Acc. & \# Param. & Infer.~Time & Space \\
    \midrule
        Decision Tree Learner           & $71.9\%$ & \kern-2.3ex$\approx 100$ & $82$\textrm{ns} & \kern-2.3ex$\approx 230$B  \\
        Logistic Regression             & $72.9\%$ & $104$ & $34$ns & $416$B \\
        Neural Network                  & $75.3\%$ & $434$ & $130$ns & $1.4$KB   \\
        Diff Logic Net (\textit{ours})   & $76.1\%$ & $640$ & $2.8$ns & $320$B   \\
    \bottomrule
    \end{tabular}
    }
    \label{tab:adult-breast}
\end{table}

\begin{table*}[t!]
    \centering
    \caption{
        Results for MNIST, all of our results are averaged over $10$ runs. Times (T.) are inference times per image, the GPU is an NVIDIA A6000, and the CPU is a single thread at $2.5$~GHz. For our experiments, i.e., the top block, we use binarized MNIST.
    }
    {\footnotesize
    \begin{tabular}{llrrccrccc}
    \toprule
        \textbf{MNIST} & Acc. & \# Param. & Space & T.~[CPU] & T.~[GPU] & OPs & FLOPs  \\
    \midrule
        Linear Regression               & $91.6\%$ & $4\,010$           & $16$KB    & $3\mu$s       & $2.4$ns       & ($4$M)    & $4$K               \\  %
        Neural Network (\textit{small}) & $97.92\%$ & $118\,282$        & $462$KB   & $14\mu$s      & $12.4$ns      & ($236$M)  & $236$K                    \\
        Neural Network                  & $98.40\%$ & $22\,609\,930$    & $86$MB    & $2.2$ms       & $819$ns       & ($45$G)   & $45$M             \\
        \midrule
        Diff Logic Net (\textit{small})  & $97.69\%$ & $48\,000$         & $23$KB    & $625$ns       & $6.3$ns       & $48$K     & |                   \\
        Diff Logic Net                   & $98.47\%$ & $384\,000$        & $188$KB   & $7\mu$s       & ($50$ns)      & $384$K    & |                 \\
        \midrule
        \midrule
        \textit{Binary Neural Networks} &&&&& T.~[FPGA] \\
        FINN \cite{umuroglu2017finn}         & $98.40\%$ &              &           & ($96\mu$s)    & $641$ns       & $5.28$M   &                \\
        BinaryEye~\cite{jokic2018binaryeye}  & $98.40\%$ &              &           &               & $50\mu$s     &           &                  \\
        ReBNet~\cite{ghasemzadeh2018rebnet}  & $98.29\%$ &              &           &               & $3\mu$s     &           &                  \\
        LowBitNN~\cite{zhan2020field}        & $99.2\%$  &              &           &               & $152\mu$s     &           &                  \\
        \midrule
        \textit{Sparse Neural Networks} &&&&& Sparsity \\
        Var.~Dropout \cite{molchanov2017variational}    & $98.08\%$ & $4\,000$           &           &               & $98.5\%$      &  ($8$M)  & $8$K  &       \\
        $L_0$ regularization \cite{louizos2018learning} & $98.6\%$  &                    &           &               & $2/3$         &  ($200$M)  & $200$K          &        \\
        SET-MLP \cite{mocanu2018scalable}               & $98.74\%$ &  $89\,797$         &           &               & $96.8\%$      &  ($180$M)  & $180$K          &        \\
        \midrule
        Sparse Function Net \cite{gaier2019weight} & $94.2\%$ & $3\times1\,849$ &&&&& $>2$K \\  %
    \bottomrule
    \end{tabular}
    }
    \label{tab:mnist}
\end{table*}
\begin{table*}[t]
    \centering
    \caption{
    Results on CIFAR-10.
    Times (T.) are inference times per image, the GPU is an NVIDIA A6000, and the CPU is a single thread at $2.5$~GHz. For our experiments, i.e., the top block, we use a color-channel resolution of $4$ for the first $3$ lines and a color-channel resolution of $32$ for the \textit{large} models.
    The other baselines were provided with the full resolution of $256$ color-channel values.
    The numbers in parentheses are extrapolated / estimated.
    }
    {\footnotesize
    \begin{tabular}{llrrccrccc}
    \toprule
        \textbf{CIFAR-10} & Acc. & \#\,Param & Space & T.~[CPU] & T.~[GPU] & OPs & FLOPs \\
    \midrule
        Neural Network (color-channel res.~= 4)        & $50.79\%$ & $12.6$M      & $48$MB    & $1.2$ms     & $370$ns  & ($25$G)    & $25$M     \\ %
        \midrule
        Diff Logic Net (\textit{small})              & $51.27\%$ & $48$K        & $24$KB    & $1.3\mu$s   & $19$ns   & $48$K      & |                          \\
        Diff Logic Net (\textit{medium})             & $57.39\%$ & $512$K       & $250$KB   & $7.3\mu$s   & $29$ns   & $512$K     & |                        \\
        Diff Logic Net (\textit{large})              & $60.78\%$ & $1.28$M      & $625$KB   & ($18\mu$s)  & ($73$ns) & $1.28$M    & |                         \\ %
        Diff Logic Net (\textit{large}$\times2$)     & $61.41\%$ & $2.56$M      & $1.22$MB  & ($37\mu$s)  & ($145$ns) & $2.56$M    & |                       \\
        Diff Logic Net (\textit{large}$\times4$)     & $62.14\%$ & $5.12$M      & $2.44$MB  & ($73\mu$s)  & ($290$ns) & $5.12$M    & |                       \\
        \midrule
        \midrule
        \textit{Best Fully-Connected Baselines~~(color-channel res.~= 256)}\kern-12em & \\
        Regularized SReLU NN \cite{mocanu2018scalable} & $68.70\%$  & $20.3$M        & $77$MB     & $1.9$ms     & $565$ns & ($40$G)    & $40$M           \\
        Student-Teacher NN \cite{urban2016deep}     & $65.8\%$  & $1$M                  & $4$MB      & $112\mu$s   & $243$ns & ($2$G)    & $2$M           \\
        Student-Teacher NN \cite{urban2016deep}     & $74.3\%$  & $31.6$M               & $121$MB    & $2.9$ms     & $960$ns & ($63$G)    & $63$M           \\
        \midrule
        \textit{Sparse Neural Networks} &&&&& Sparsity \\
        PBW (ResNet32) \cite{han2016deep}                           & $38.64\%$  &                &           &                     & $99.9\%$   & ($140$M)   & ($140$K)        \\ %
        MLPrune (ResNet32) \cite{zeng2019mlprune}                       & $36.09\%$  &                &           &                     & $99.9\%$   & ($140$M)   & ($140$K)       \\
        ProbMask (ResNet32) \cite{zhou2021effective}                      & $76.87\%$  &                &           &                     & $99.9\%$   & ($140$M)   & ($140$K)       \\
        SET-MLP \cite{mocanu2018scalable}                            & $74.84\%$  & $279$K         & $4.7$MB   &                     & $98.6\%$   & ($558$M)   & $558$K       \\
    \bottomrule
    \end{tabular}
    }
    \label{tab:cifar}
\end{table*}

\subsection{MNIST}

For our comparison to the fastest neural networks, we start by considering MNIST~\cite{lecun2010mnist}.

Qin~\etal~\cite{qin2020binary} give a current overview of binary neural networks in their survey.
They discuss the challenges of training binary neural networks or translating existing neural networks into their binarized counterparts.
They identify FINN by Umuroglu~\etal~\cite{umuroglu2017finn} as the fastest method for classifying MNIST at an accuracy of $98.4\%$ at a frame rate of $1\,561\,000$ images per second on specialized FPGA hardware.
FPGAs (field-programmable gate arrays) are configurable hardware accelerated processors that can achieve extreme speeds for fixed, predefined tasks that are expressed via logic gates.
Note that, in the context of FINN, these logic gates are not learned but rather used as a low-precision approximation of neural networks.
In comparison to FINN, our logic network achieves better performance while requiring less than $10\%$ of the number of binary operations.
That is, our model is objectively more than $10\times$ cheaper to evaluate.
When comparing real times, for an NVIDIA A6000 GPU, our model is $12\times$ faster than the model by Umuroglu~\etal~\cite{umuroglu2017finn} on their specialized FPGA hardware, even though our model only achieves a $7\%$ utilization of the GPU.
For the other BNNs, OPs have not been reported, but their inference speed is also substantially slower than FINN.

\marginnote{
A FLOP generally corresponds to many binary OPs.
Specifically, a float32 adder / multiplier requires usually at least $1\,000$ logical gates or look up tables and usually has a delay of tens of logical levels.
Practically, float32 adders / multipliers are implemented directly in hardware in CPUs and GPUs, as it is an essential operation on such platforms.
Nevertheless, also in practice, a float32 adder / multiplier is much more expensive than performing a bit-wise logical operation on int64 data types (even on float32 and int32 focused GPUs).
On CPUs, around $3-10$ int64 bit-wise operations can be performed per cycle, while floating-point operations usually require a full clock cycle.
To convert a non-sparse model, we assume a very conservative $100$ OPs per $1$ FLOP.
Note that speeds for sparse neural networks are also only theoretical because sparse execution usually brings an overhead of factor $10-100\times$.
So overall, in practice, $1\,000$ (binary) OPs per $1$ sparse (float32) FLOP is a very conservative estimate in favor of sparse float32 models.
Further, in theory, $1\,000$ OPs per $1$ FLOP is an accurate estimate (assuming sparsity to come without cost and assuming floating-point operations to not be hardware accelerated).
}

Specifically, the binary FINN MNIST model by Umuroglu~\etal~\cite{umuroglu2017finn} requires $5.82$ MOPs (Mega binary OPerations) per frame, which means that their FPGA achieves around $5.82\cdot10^6\cdot1.561\cdot10^6 = 9.09 \cdot10^{12}$ binary operations per second, i.e., $9.09$ TOPS (Tera binary OPerations per Second).
On different FPGAs, Ghasemzadeh~\etal~\cite{ghasemzadeh2018rebnet} achieve $330\,000$ images per second on MNIST at an accuracy of $98.29\%$, and Jokic~\etal~\cite{jokic2018binaryeye} propose an FPGA based embedded camera system achieving $20\,000$ images per second at $98.4\%$ accuracy.

When compared to the smallest sparse neural network, our model requires substantially fewer operations than each of the baselines.
Specifically, Hoefler~\etal~\cite{hoefler2021sparsity} give an overview of sparsity on deep learning in their recent literature review.
They identify Molchanov~\etal~\cite{molchanov2017variational} to achieve the sparsest (originally fully-connected) model on MNIST with a sparsity of $98.5\%$ achieving an accuracy of $98.08\%$.
For this, Molchanov~\etal~\cite{molchanov2017variational} propose variational dropout with unbounded dropout rates to sparsify neural networks.
Louizos~\etal~\cite{louizos2018learning} propose sparsification via $L_0$ regularization and report an MNIST accuracy of $98.6\%$ for a model with around $2\cdot 10^{5}$~FLOPs (FLoating point OPerations), which corresponds to a sparsity of around $2/3$.

Zhan~\etal~\cite{zhan2020field} concentrate on deploying Low-Bit Neural Networks (LBNNs) on FPGAs and achieve an accuracy of $99.2\%$ on MNIST at $6\,580$ images per second.
Shani~\etal~\cite{shani2019dynamics} explore analog logic gate networks (a physical approximation to boolean networks) and achieve accuracies up to $89\%$ on MNIST.

Sparse function networks~\cite{gaier2019weight}, which have been learned evolutionarily, achieve an accuracy of $94.2\%$.

\newpage
\subsection{CIFAR-10}

In addition to MNIST, we also benchmark our method on CIFAR-10 \cite{krizhevsky2009cifar10}.
For CIFAR-10, we reduce the color-channel resolution of the CIFAR-10 images and employ a binary embedding:
For a color-channel resolution of $4$ (the first three rows of Table~\ref{tab:cifar}), we use three binary values with the three thresholds $0.25, 0.5,$ and $0.75$.
For a color-channel resolution of $32$ (the large models, i.e., the last three rows of the top block of Table~\ref{tab:cifar}), we use 31 binary values with thresholds $(i/32)_{i\in\{1..31\}}$.
We do not apply data augmentation / dropout for our experiments, which could additionally improve performance.
For all baselines, we copied the reported accuracies, and thus those results are with data augmentation and the original color-channel resolution, and may include dropout~\cite{srivastava2014dropout} as well as other techniques such as student-teacher learning with a convolutional teacher~\cite{urban2016deep}.

\marginnote{
The results in parentheses are estimated because compilation to binaries did not finish / for GPU the largest models could also not be compiled due to compiler limitations.
This can be resolved if desired with moderate implementation effort, e.g., compiling the model directly to PTX (CUDA Assembly) without compiling via \texttt{gcc} and \texttt{nvcc}.
The actual problem is that the compilers used by the implementation have a quadratic compile time in the number of lines of code / statements.
}

The results are displayed in Table~\ref{tab:cifar}.
We find that our method outperforms neural networks in the first setting (color-channel resolution of $4$) by a small margin, while requiring less than $0.1\%$ of the memory footprint and (with a larger model) by a large margin while requiring less than $1\%$ of the memory footprint.
In comparison to the best fully-connected neural network baselines, which are trained with various tricks such as student-teacher learning and obtain the full color-channel resolution, our model does not achieve the same performance, while it is also much smaller and has access to fewer data.
With $1$ million parameters, the student-teacher model \cite{urban2016deep} has a footprint that is about $64\%$ larger than the footprint of our largest model (\textit{large}$\times4$), and achieves an accuracy which is only $3.7\%$ better than ours.
It is important to note that this models requires $2$ million floating-point operations, while our model requires $5$ million bit-wise logic operations (before pruning/optimization).
On float-arithmetic hardware-accelerated integrated circuits (as current GPUs and many CPUs), the $2$ million floating-point operations are around $100\times$ slower than $5$ million bit-wise logic operations.
On general purpose hardware (i.e., without float acceleration) the speed difference would be one order of magnitude larger, i.e., $1\,000\times$.

More competitive with respect to speed are sparse neural networks.
In the final block of Table~\ref{tab:cifar}, we report the sparsest models for CIFAR-10.
Note that two of the methods resulted in performances below $50\%$, and even those methods which achieve around $75\%$ accuracy require a significantly more expensive inference.

Zhou~\etal~\cite{zhou2021effective} proposed ProbMask and give an overview over the sparsest CIFAR-10 models.
Specifically, they report up to a sparsity of $99.9\%$, which corresponds to around $140$ kFLOPs for their smallest network architecture (ResNet32, which has a base cost of $~140$~MFLOPs \cite{lym2019prunetrain}.) %
This model is (depending on hardware) $1-2$ orders of magnitude more expensive than our largest model.
Other architectures reported in the literature are substantially less sparse and, therefore, more expensive.
While these models have the advantage of being based on the VGG and ResNet CNN architectures, our models are still very competitive, especially considering that our models are much smaller than their smallest reported results.

Blalock~\etal~\cite{blalock2020state} report in their survey that for CIFAR-10 with a VGG, to achieve a theoretical speedup of $32\times$, all evaluated methods drop significantly below $70\%$ test accuracy.
Note that this speedup corresponds to a sparsity of around $97\%$, which makes up a much larger model than the models considered in this chapter.

Mocanu~\etal~\cite{mocanu2018scalable} propose training neural networks with sparse evolutionary training inspired by network science.
Their method evolves an initial sparse topology of two consecutive layers of neurons into a scale-free topology.
They achieve an accuracy of $74.84\%$ on CIFAR-10 with $278\,630$ floating-point parameters.
We estimate this to correspond to a theoretical cost of around $550$ kFLOPs (multiplication + addition), corresponding to $550$ MOPs as per our conservative estimate.
On MNIST, they achieve (with $89\,797$ parameters) an accuracy of $98.74\%$.
The model is by two orders of magnitude more expensive to evaluate than the largest logic gate network considered in this chapter.

\begin{figure*}[t]
    \centering
    \includegraphics[width=\linewidth, trim={0 9.5cm 0 .25cm},clip]{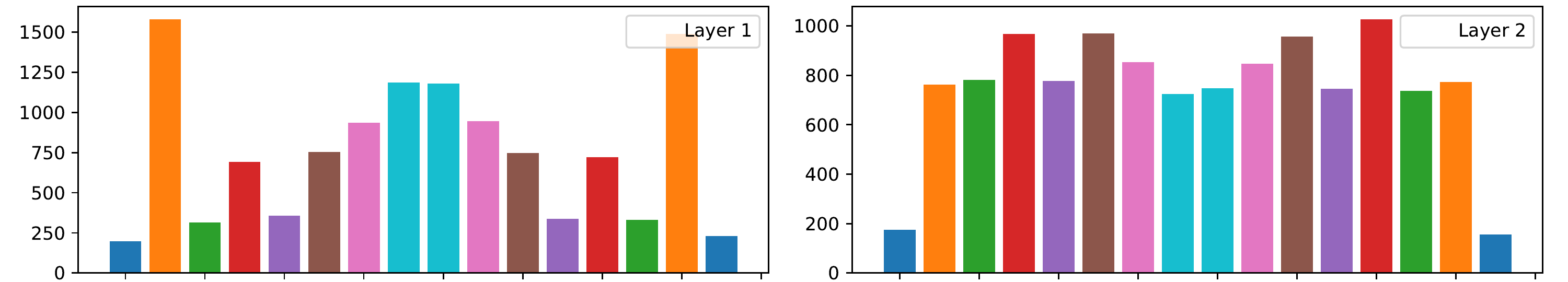}
    \includegraphics[width=\linewidth, trim={0 0 0 6.5cm},clip]{fig_logic/operator_distributions_00320550.pdf}
    \vspace{-1.5em}
    \caption{Distribution of logic gates in a trained four layer logic network.}
    \label{fig:operator-distributions}
\end{figure*}

\subsection{Distribution of Logic Gates}

To gain additional insight into learned logic gate networks, we consider histograms of operators present in each layer of a trained model.
Specifically, we consider a $4$ layer CIFAR-10 model with $12\,000$ neurons per layer in Figure~\ref{fig:operator-distributions}.

We observe that, generally, the constant $0/1$ ``operator'' is learned to be used only very infrequently as it does not actually provide value to the model.
Especially interesting is that it does not occur at all in the last layer.
In the first layer, we observe a stronger presence of `and', `nand', `or', and `nor'.
In the second and third layers, there are more `$A$', `$B$', `$\neg A$', and `$\neg B$'s, which can be seen as a residual / direct connection.
This enables the network to model lower-order dependencies more efficiently by expressing it with fewer layers than the predefined number of layers.
In the last layer, the most frequent operations are `xor' and `xnor', which can create conditional dependencies of activations of the previous layers.
Interestingly, however, implications (e.g., $A\Rightarrow B$) are only infrequently used.

\subsection{Hyperparameters and Model Architectures}
\label{sec:difflogic:hyper-model-arch}
In each experiment, we used so-called \textit{straight} architectures, i.e., architectures with the same number of neurons per layer.
In Tables~\ref{tab:network-architectures-logic} and~\ref{tab:network-architectures-mlp}, we display the numbers of layers numbers of neurons per layer for each network architecture.
The architecture search for all models was performed via grid search with the number of layers in $\{2, 3, 4, 5, 6, 7, 8, 9, 10\}$ and number of neurons per layer with a resolution of factor $2$.

\begin{table}[h!]
    \centering
    \caption{Logic gate network architectures.}
    {\footnotesize
    \addtolength{\tabcolsep}{-2pt}
    \begin{tabular}{lcrrrr}
    \toprule
        Data set     & Model     & Layers & Neurons / layer & Total num.~of p. & $\tau$  \\
    \midrule
        MONK-1      & ---       & $6$ & $24$ & $144$ & $1$ \\
        MONK-2      & ---       & $6$ & $12$ & $ 72$ & $1$ \\
        MONK-3      & ---       & $6$ & $12$ & $ 72$ & $1$ \\
    \midrule
        Adult       & ---       & $5$ & $256$ & $1\,280$ & $1/0.075$\\
        Breast Cancer & ---     & $5$ & $128$ & $640$ & $1/0.1$ \\
    \midrule
        MNIST       & small     & $6$ & $8\,000$ & $48\,000$ & $1/0.1$\\
                    & normal    & $6$ & $64\,000$ & $384\,000$ & $1/0.03$ \\
    \midrule
        CIFAR-10    & small     & $4$ & $12\,000$  & $48\,000$ & $1/0.03$ \\
                    & medium    & $4$ & $128\,000$ & $512\,000$ & $1/0.01$ \\
                    & large     & $5$ & $256\,000$ & $1\,280\,000$ & $1/0.01$ \\
                    & large$\times2$ & $5$ & $512\,000$ & $2\,560\,000$ & $1/0.01$ \\
                    & large$\times4$ & $5$ & $1\,024\,000$ & $5\,120\,000$ & $1/0.01$ \\
    \bottomrule
    \end{tabular}
    }
    \label{tab:network-architectures-logic}
\end{table}

For all models, we use the Adam optimizer~\cite{kingma2015adam}.
For all neural networks, we use a learning rate of $0.001$ and for all logic gate networks, we use a learning rate of $0.01$.
We train all models up to $200$ epochs at a batch size of $100$.
The softmax temperature $\tau$ was searched over a grid of $\{1, 1/0.3, 1/0.1, 1/0.03,\allowbreak 1/0.01\}$ (except for Adult).
The optimal temperature primarily depends on the number of outputs per class.
If there are more outputs per class, the range of predictions is larger, and thus, we use a larger temperature to counter this effect.

\clearpage
\begin{table}[t]
    \centering
    \caption{Multi-layer perceptron / neural network baseline architectures. All architectures are ReLU activated.}
    {\footnotesize
    \addtolength{\tabcolsep}{-2pt}
    \begin{tabular}{lcrrr}
    \toprule
        Data set     & Model     & Layers & Neurons / layer & Total num.~of parameters  \\
    \midrule
        MONK-1      & ---       & $2$ & $8$ & $162$ \\
        MONK-2      & ---       & $2$ & $8$ & $162$ \\
        MONK-3      & ---       & $2$ & $8$ & $162$ \\
    \midrule
        Adult       & ---       & $2$ & $32$ & $3\,810$ \\
        Breast Cancer & ---     & $2$ & $8$ & $434$ \\
    \midrule
        MNIST       & small     & $3$ & $128$ & $118\,282$ \\
                    & normal    & $7$ & $2\,048$ & $22\,609\,930$ \\
    \midrule
        CIFAR-10    & ---       & $5$ & $1\,024$ & $12\,597\,258$ \\
    \bottomrule
    \end{tabular}
    }
    \label{tab:network-architectures-mlp}
\end{table}

\vspace{3em}
\subsection*{Conclusion}

In this chapter, we leveraged real-valued logics and continuous relaxations to present a novel approach to training logic gate networks, which allows us to effectively train extremely efficient neural networks that---for their level of accuracy---are one or more orders of magnitude more efficient than the state-of-the-art.
At the moment, the capabilities of logic gate networks are mostly limited to tasks that can be sufficiently solved with fully-connected neural networks.
However, one direction of future research could be convolutional differentiable logic gate networks. %
A limitation of differentiable logic gate networks is their relatively higher training cost compared to (performance-wise) comparable conventional neural networks; however, since their original development, we have already reduced their training cost by up~to~$50$ times.

\setchapterpreamble[u]{\pagelogo{splitprop}\margintoc}
\chapter[Alternative Optimization Methods]{Alternative\\ Optimization Methods}
\labch{alternative}

\newcommand{\xb}{{\mathbf{x}}}%
\newcommand{\zb}{{\mathbf{z}}}%
\newcommand{\yb}{{\mathbf{y}}}%

After discussing the construction of differentiable algorithms and their applications, in this chapter, we present alternative optimization methods.
Specifically, we introduce splitting backpropagation, a general two-stage optimization method for combining different iterative optimization methods.
Further, we present two instances of splitting backpropagation:
First, Newton losses, a method for combining second-order optimization of the loss function and first-order optimization of the model.
Second, RESGRO, a simple stochastic optimization method which allows optimizing non-differentiable objectives and can be combined with first-order optimization of the model.
In our empirical evaluation, we find that both instances deliver strong empirical performance on algorithmic supervision tasks.

Neural network training has gained a tremendous amount of attention in machine learning in recent years.
This is primarily due to the success of backpropagation and stochastic gradient descent for first-order optimization.
However, there has also been a strong line of work on second-order optimization for neural network training; see the work by Agrawal~\etal~\cite{agrawal2017second} and references therein.
While these second-order optimization methods (such as Newton's method and natural gradient descent) exhibit improved convergence rates and therefore require fewer training steps, they have two major limitations~\cite{nocedal2006numerical}, namely
(i) computing the curvature (or its inverse) for a large and deep neural network is computationally substantially more expensive than simply computing the gradient with backpropagation, which makes second-order methods practically inapplicable in most cases; 
(ii) networks trained with second-order information exhibit reduced generalization capabilities~\cite{wadia2021whitening}.

When working with complicated and non-standard loss functions, the optimization problem for neural network training is often challenging. 
In these cases, a natural approach is to look for a decomposition, i.e., to break up the optimization problem into two (or more) sub-problems, which are then solved sequentially.
The idea of decomposing an optimization problem is not new, see \cite{bertsekas1999nonlinear} for an overview of decomposition methods.
A decomposition method, which has become particularly popular in machine learning, is the alternating direction method of multipliers \cite{boyd2011distributed}.

In this work, we propose a novel method for incorporating second-order information of the loss function into training while training the actual neural network with gradient descent.
As loss functions are usually substantially cheaper to evaluate than a neural network, the idea is to apply second-order optimization to the loss function while training the actual neural network with first-order optimization.
For this, we decompose the original iterative optimization problem into a two-stage iterative optimization problem, which leads to Newton losses. %
This is especially interesting for intrinsically hard-to-optimize loss functions, i.e., where second-order optimization of the inputs to the loss is superior to first-order optimization.
This includes the algorithmic supervision training objectives / algorithmic losses discussed in this thesis.

\newpage
We evaluate the proposed methods for various algorithmic losses on two benchmarks: the four-digit MNIST sorting benchmark and the Warcraft shortest-path benchmark, following up on Chapters~\ref{ch:algovision} and~\ref{ch:diffsort}. 
We find that they improve the performance in the case of hard-to-optimize losses and maintain the original performance in the case of easy-to-optimize losses.

The contributions of this chapter are 
(i) introducing a mathematical framework for splitting iterative optimization methods into two-stage schemes, which we show to be equal to the original optimization methods;
(ii) introducing Newton losses as combinations of first-order and second-order optimization methods;
(iii) introducing RESGRO, a stochastic method for optimizing non-differentiable objectives, which can be combined with gradient descent on differentiable sections of the architecture.

\section{Related Work}
\marginnote{
In the context of second-order methods, we would also like to mention ISAAC Newton by Petersen~\etal~\cite{petersen2022isaac}.
Newton Losses can be seen as a complement to ISAAC Newton.
While ISAAC Newton integrates second-order information from the input, Newton losses integrate second-order information from the loss into training.
}
The related work specific to this chapter primarily comprises second-order optimization methods.
Second-order methods have recently gained popularity in machine learning due to their fast convergence properties when compared to first-order methods~\cite{agrawal2017second}.
One alternative to vanilla Newton are quasi-Newton methods, which, instead of computing an inverse Hessian in the Newton step (which is expensive), approximate this curvature from the change in gradients~\cite{nocedal2006numerical}.
In addition, a number of new approximations to the pre-conditioning matrix have been proposed in the recent literature, i.a., \cite{martens2015optimizing,pilnaci2017newton,frantar2021mfac}. 
While the vanilla Newton methods rely on the Hessian, there are variants which use the empirical Fisher information matrix, which can coincide in specific cases with the Hessian, but generally exhibits somewhat different behavior. For an overview and discussion of Fisher-based methods (sometimes referred to as natural gradient descent), see~\cite{kunstner2019limitations, martens2020new}.
To the best of our knowledge, this is the first work combining second-order optimization of loss functions with first-order optimization of neural networks, especially for algorithmic losses.

\section[Splitting Backpropagation: A Two-Stage Optimization Method]{Splitting Backpropagation:\texorpdfstring{\\}{} A Two-Stage Optimization Method} \label{sec:Newton:loss}
We consider the training of a neural network $f(x; \theta)$, where $x\in\mathbb{R}^n$ is the vector of inputs, $\theta\in\mathbb{R}^d$ is the vector of weights and $y=f(x; \theta)\in\mathbb{R}^m$ is the vector of outputs. 
We assume that we have access to a data set of $N$ samples drawn from the input distribution, which describes the empirical input $\xb=[x_1, \dots,x_N]^\top\in\mathbb{R}^{N\times n}$. 
As per vectorization, we denote $\yb = f(\xb; \theta)\in\mathbb{R}^{N\times m}$ as the matrix describing the outputs of the neural network corresponding to the empirical inputs. 
Further, let $\ell:\mathbb{R}^{N\times m}\to\mathbb{R}$ denote the loss function, and let the ground truth output be implicitly encoded in~$\ell$ (because for many algorithmic losses, it is not simply a label, but could be ordinal information). 
In a general setting, the training of a neural network can be expressed as the optimization problem
\begin{equation}   \label{eq:training:problem}
    \arg\min_{\theta\in\Theta} \ell(f(\xb; \theta)) \,,
\end{equation}
where $\Theta\subseteq \mathbb{R}^d$ is the domain of the parameters $\theta$, and $f$ and $\ell$ are such that the minimum in \eqref{eq:training:problem} exists.
Note that the formulation~\eqref{eq:training:problem} is extremely general and includes, e.g., optimization of non-decomposable loss functions (i.e., not composed of individual losses per training sample), which is relevant for some algorithmic losses like ranking losses. \\

Typically, the optimization problem \eqref{eq:training:problem} is solved by using some iterative algorithm like gradient descent (or Newton's method) updating the weights $\theta$ by repeatedly applying the following step:
\begin{align}
    \theta &\leftarrow \text{One optimization step of }\ell(f(\xb; \theta))\text{ wrt.~}\theta \,. \label{eq:training:update}
\end{align}
However, in this work, we consider decomposing the optimization problem \eqref{eq:training:problem} into two problems, which may be solved by applying the following two updates in an alternating fashion:  
\begin{subequations} \label{eq:two:step:scheme}  
\begin{align}
    \zb^\star & \leftarrow \text{One optimization step of }\ell(\zb)\text{ wrt.~}\zb = f(\xb; \theta) \,, \label{eq:training:update-z} \\ 
    \theta &\leftarrow \text{One optimization step of }{\textstyle\frac12}\| \zb^\star - f(\xb; \theta)\|_2^2\text{ wrt.~}\theta \,. \label{eq:training:update-theta}
\end{align}%
\end{subequations}
\marginnote{
The split does not need to be between loss and neural network.
We choose this location to split because it is meaningful for algorithmic losses; however, this shall not restrict us to this case.
Further, we note that the method can be adapted to more than one split, and the theory for this case is analogous, as can be seen seen via induction.
}%
This split allows us later to use two different iterative optimization algorithms for \eqref{eq:training:update-z} and \eqref{eq:training:update-theta}, respectively.
This is especially interesting for optimization problems where the loss function $\ell$ is non-convex and its minimization is a difficult optimization problem itself, and as such those problems, where a stronger optimization method exhibits a superior rate of convergence.

We can express individual optimization steps (corr.~to~\eqref{eq:training:update} and~\eqref{eq:training:update-z}) via 
\begin{align}
\label{eq:omega-use-1}
   \theta &\leftarrow \arg\min_{\theta'\in\Theta} \ell(f(\xb;\theta')) + \Omega(\theta', \theta)\,, \\
\label{eq:omega-use-2}
   \zb^\star &\leftarrow \arg\min_{\zb\in\mathcal{Y}} \ell(\zb) + \Omega(\zb, f(\xb;\theta))
\end{align}
where $\Omega$ is a regularizer such that one step of a respective optimization method corresponds to the global optimum of the regularized optimization problems in~\eqref{eq:omega-use-1} and~\eqref{eq:omega-use-2}. 
The regularizer $\Omega$ has the standard property that $\Omega(a, b) = 0$ for any $a = b$.
Note that the explicit form of the regularizer $\Omega$ does not need to be known. 
Nevertheless, in Supplementary Material~\ref{sm:regularizer}, we discuss explicit choices of $\Omega$.

This allows us to express the set of points of convergence for the iterative optimization methods. 
Recall that an iterative optimization method has converged if it has arrived at a fixed point, i.e, the parameters do not change when applying an update. 
The set of points of convergence for \eqref{eq:training:update} is
\begin{equation}
    \mathcal{A} = \Big\{ \theta \ | \ \theta\in\arg\min_{\theta'} \ell(f(\xb;\theta')) + \Omega(\theta',\theta)\Big\}\,,
\end{equation}
i.e., those points at which the update does not change $\theta$. For the two-stage optimization method \eqref{eq:two:step:scheme}, the set of points of convergence is
\begin{equation}
    \mathcal{B} = \Big\{ \theta \ | \ f(\xb;\theta)=\zb^\star\in\arg\min_{\zb} \ell(\zb) + \Omega(\zb,f(\xb;\theta)\Big\}
\end{equation}
as the method has converged if the update \eqref{eq:training:update-z} yields $\zb^\star=\zb$ because the subsequent update \eqref{eq:training:update-theta} will not change $\theta$ as $\zb^\star=\zb=f(\xb;\theta)$ already holds, and thus ${\textstyle\frac12}\| \zb^\star - f(\xb; \theta)\|_2^2=0$.
Now, we show that the iterative method \eqref{eq:training:update} and the alternating method~\eqref{eq:two:step:scheme} lead to the same sets of convergence points.
\clearpage

\begin{widepar}
\begin{lemma}[Equality of the Sets of Convergence Points]\label{lem:equivalence}
    The set $\mathcal{A}$ of points of convergence obtained by the iterative optimization method \eqref{eq:training:update} is equal to the set $\mathcal{B}$ of points of convergence obtained by the two-step iterative optimization method \eqref{eq:two:step:scheme}.
\vspace{-.5em}
\begin{proof}
$(\mathcal{A}\subset \mathcal{B})\quad$ First, we show that any point in $\mathcal{A}$ also lies in $\mathcal{B}$.
By definition, for each point in $\mathcal{A}$, the optimization step \eqref{eq:training:update} does not change $\theta$, i.e., $\theta'=\theta$.
Thus, $f(\xb;\theta) = f(\xb;\theta') \in \arg\min_{\zb} \ell(\zb) + \Omega(\zb,f(\xb;\theta))$, and therefore $\theta\in\mathcal{B}$.

$(\mathcal{B}\subset \mathcal{A})\quad$
Second, we show that any point in $\mathcal{B}$ also lies in $\mathcal{A}$.
For each $\theta\in\mathcal{B}$, we know that, by definition, $f(\xb;\theta)=\zb^\star\in\arg\min_{\zb} \ell(\zb) + \Omega(\zb,f(\xb;\theta))$, therefore $\theta \in\arg\min_{\theta'} \ell(f(\xb;\theta')) + \Omega(f(\xb;\theta'),f(\xb;\theta))$ where $\Omega(f(\xb;\theta),f(\xb;\theta)) = 0 = \Omega(\theta, \theta)$, and, therefore $\theta\in\mathcal{A}$.
\end{proof}
\end{lemma}
\end{widepar}

While Lemma~\ref{lem:equivalence} states the equivalence of the original training~\eqref{eq:training:update} and its counterpart~\eqref{eq:two:step:scheme} wrt.~their possible points of convergence (i.e., solutions) for an arbitrary choice of the iterative method, the two approaches are also equal when applying standard first-order or second-order optimization schemes. 
In other words, running a gradient descent step according to \eqref{eq:training:update} coincides with two gradient steps of the alternating scheme \eqref{eq:training:update-z} and \eqref{eq:training:update-theta}, namely one step for \eqref{eq:training:update-z} and one step for \eqref{eq:training:update-theta}.

\vspace{-.5em}
\begin{robustwidepar}
\begin{theorem}[Gradient Descent Step Equality between \eqref{eq:training:update} and \eqref{eq:training:update-z} + \eqref{eq:training:update-theta}] \label{lem:gradient:equiv}
A gradient descent step according to \eqref{eq:training:update} with arbitrary step size $\eta$ coincides with two gradient descent steps according to \eqref{eq:training:update-z} and \eqref{eq:training:update-theta}, where the optimization over $\theta$ has a step size of\/ $\eta$ and the optimization over $z$ has a unit step size.
\vspace{-.5em}
\begin{proof}
Let $\theta\in\Theta$ be the current parameter vector and let $\zb = f(\xb;\theta)$. Then, the gradient descent steps according to \eqref{eq:training:update-z} and \eqref{eq:training:update-theta} with step sizes $1$ and $\eta>0$ are expressed as
\begin{align}
    \kern-.5em\zb &\leftarrow \zb - \nabla_{\zb}\ell(\zb) = f(\xb;\theta) - \nabla_f \,\ell(f(\xb;\theta)) \label{eq:gd:z} \\
    \kern-.5em\theta &\leftarrow \theta - \eta\, \nabla_\theta \, {\textstyle\frac{1}{2}}\|\zb - f(\xb; \theta)\|^2_2 
    = \theta - \eta\, \frac{\partial\,f(\xb; \theta)}{\partial\,\theta} \cdot (f(\xb; \theta) - \zb) \,. \label{eq:gd:theta}
\end{align}
Combining \eqref{eq:gd:z} and \eqref{eq:gd:theta} eventually leads to
\begin{align}
    \theta \leftarrow &\,\theta - \eta\, \frac{\partial\,f(\xb; \theta)}{\partial\,\theta} \cdot (f(\xb; \theta) - f(\xb;\theta) + \nabla_f \,\ell(f(\xb;\theta))) = \theta - \eta\, \nabla_{\theta}\ell(f(\xb;\theta)),
\end{align}
which is exactly a gradient descent step of problem \eqref{eq:training:problem} starting at $\theta\in\Theta$ with step size $\eta$.
\end{proof}
\end{theorem}

\begin{theorem}[Newton Step Equality between \eqref{eq:training:update} and \eqref{eq:training:update-z}+\eqref{eq:training:update-theta}] \label{lem:Newton:equiv}
In the case of $m=1$, a Newton step according to \eqref{eq:training:update} with arbitrary step size $\eta$ coincides with two Newton steps according to \eqref{eq:training:update-z} and \eqref{eq:training:update-theta}, where the optimization over $\theta$ has a step size of\/ $\eta$ and the optimization over $z$ has a unit step size.
\vspace{-.5em}
\begin{proof}
Let $\theta\in\Theta$ be the current parameter vector and let $\zb = f(\xb;\theta)$. 
Then, applying Newton steps according to \eqref{eq:training:update-z} and \eqref{eq:training:update-theta} leads to
\begin{align}
    \zb &\leftarrow \zb - (\nabla^2_{\zb}\ell(\zb))^{-1} \nabla_{\zb} \ell(\zb) =f(\xb;\theta) - (\nabla_f^2\ell(f(\xb;\theta)))^{-1} \nabla_f \ell(f(\xb;\theta)) \label{eq:newton:pf:1}\\
    \theta &\leftarrow \theta - \eta \left( \nabla^2_\theta \frac{1}{2}\|\zb - f(\xb; \theta)\|^2_2 \right)^{\!\!-1}\nabla_\theta \frac{1}{2}\|\zb - f(\xb; \theta)\|^2_2\\ 
    &= \theta - \eta\left( \frac{\partial}{\partial\theta}\left[   \frac{\partial\,f(\xb; \theta)}{\partial\,\theta} \cdot (f(\xb; \theta) - \zb)    \right] \right)^{\!\!-1}   \frac{\partial\,f(\xb; \theta)}{\partial\,\theta} \cdot (f(\xb; \theta) - \zb) \label{eq:newton:pf:2}
    \end{align}
    \begin{align}
    &= \theta - \eta \left( \frac{\partial}{\partial\theta}\left[\frac{\partial\,f(\xb; \theta)}{\partial\,\theta} \right] (f(\xb; \theta) - \zb) + \left( \frac{\partial\,f(\xb; \theta)}{\partial\,\theta} \right)^{\!2} \right)^{\!\!-1} \cdot \frac{\partial\,f(\xb; \theta)}{\partial\,\theta} \cdot (f(\xb; \theta) - \zb) 
\end{align}
Inserting \eqref{eq:newton:pf:1}, we can rephrase the update above as
\begin{equation}\label{eq:newton:pf:3}
\begin{aligned}
    \theta &\leftarrow  \theta - \eta \left( \frac{\partial}{\partial\theta}\left[\frac{\partial\,f(\xb; \theta)}{\partial\,\theta} \right] (\nabla_f^2\ell(f(\xb;\theta)))^{-1}  \nabla_f \ell(f(\xb;\theta)) + \left( \frac{\partial\,f(\xb; \theta)}{\partial\,\theta} \right)^{2} \right)^{-1} \\ 
    &\qquad \qquad \cdot \frac{\partial\,f(\xb; \theta)}{\partial\,\theta} \cdot (\nabla_f^2\ell(f(\xb;\theta)))^{-1} \nabla_f \ell(f(\xb;\theta)) 
\end{aligned}
\end{equation}
By applying the chain rule twice, we further obtain
\begin{align*}
    \nabla^2_{\theta} \ell(f(\xb;\theta))
    &= \frac{\partial}{\partial\theta} \left[\frac{\partial\,f(\xb; \theta)}{\partial\,\theta}   \right]\nabla_f\ell(f(\xb;\theta)) + \left(\frac{\partial\,f(\xb; \theta)}{\partial\,\theta}\right)^{\!2}   \nabla^2_f\ell(f(\xb;\theta)),
\end{align*}
which allows us to rewrite \eqref{eq:newton:pf:3} as
\begin{align*}
    \theta' &= \theta - \left( (\nabla_f^2 \ell(f(\xb;\theta)))^{-1} \nabla^2_{\theta} \ell(f(\xb;\theta)) \right)^{-1} (\nabla_f^2\ell(f(\xb;\theta)))^{-1} \nabla_{\theta}\ell(f(\xb;\theta)) \\
    &=\theta - (\nabla^2_{\theta} \ell(f(\xb;\theta)))^{-1} \nabla_{\theta}\ell(f(\xb;\theta)),
\end{align*}
which is precisely one Newton step of problem \eqref{eq:training:problem} starting at $\theta\in\Theta$.
\end{proof}
\end{theorem}
\end{robustwidepar}
\vspace{-2.5em}
\subsection*{Explicit Forms of the Regularization Term $\protect\Omega$}
\vspace{-.5em}
\label{sm:regularizer}
For the regularization term $\Omega$ in \eqref{eq:omega-use-1} and \eqref{eq:omega-use-2}, an explicit form does not need to be known nor computable.
For the cases of a gradient descent step as well as a Newton step, we discuss explicit forms of $\Omega$:

For an update on $\zb$ \eqref{eq:training:update-z} corresponding to a \textit{gradient descent step}, i.e., 
\vspace{-.25em}
\begin{align}
    \zb^\star & \leftarrow \zb - \eta \nabla \ell(\zb), \quad \zb = f(\xb; \theta) \,,
\end{align}\\[-1.75em]
we can explicitly choose the regularization term $\Omega$ as
\vspace{-.25em}
\begin{equation}
\begin{aligned}
    \Omega(\zb,f(\xb;\theta)) &= \ell (f(\xb;\theta)) + \eta \nabla_f \ell(f(\xb;\theta))^\top(\zb - f(\xb;\theta)) - \ell(\zb) \\
    &\qquad + {\textstyle\frac{1}{2}}(\zb - f(\xb;\theta))^\top (\zb - f(\xb;\theta)) . 
\end{aligned}\label{eq:Omega:grad}
\end{equation}\\[-1.25em]
Here, the first-order optimality conditions for $\min_{\zb} \ell (\zb) {+} \Omega(\zb, f(\xb;\theta))$ are%
\vspace{-1.5em}
\begin{equation}
    \nabla_z\left(\ell (\zb) + \Omega(\zb, f(\xb;\theta))\right) = 
    \eta \nabla_f \ell(f(\xb;\theta)) + \zb - f(\xb;\theta)=0,
\end{equation}\\[-1.75em]
leading to the gradient step $\zb = f(\xb;\theta) - \eta \nabla_f \ell(f(\xb;\theta))$.

\vspace{.5em}
In the case, where the update on $\zb$ \eqref{eq:training:update-z} corresponds to a \textit{Newton step}, i.e.,
\vspace{-.25em}
\begin{align}
    \zb^\star & \leftarrow \zb - \eta (\nabla^2 \ell(\zb))^{-1}\nabla \ell(\zb), \quad \zb = f(\xb; \theta)\,,
\end{align}\\[-1.75em]
we can explicitly choose the regularization term $\Omega$ as
\vspace{-.25em}
\begin{align}
    \kern-.5em \Omega(\zb,f(\xb;\theta)) &= \ell(f(\xb;\theta)) + \eta(\zb - f(\xb;\theta))^\top \nabla_f \ell(f(\xb;\theta)) 
    \label{eq:Omega:Newton} \\ 
    &\quad + {\textstyle\frac12} (\zb - f(\xb;\theta))^\top \nabla^2_f \ell(f(\xb;\theta)) (\zb - f(\xb;\theta)) - \ell(\zb)\,.  \notag
\end{align}\\[-1.75em]
The first-order optimality conditions for $\min_{\zb} \ell (\zb) + \Omega(\zb, f(\xb;\theta))$ are
\vspace{-.25em}
\begin{align}
   &\nabla_z\left(\ell (\zb) + \Omega(\zb, f(\xb;\theta))\right) \\
   &\qquad = \eta \nabla_f \ell(f(\xb;\theta)) + \nabla^2_f\ell(f(\xb;\theta))(\zb - f(\xb;\theta))  = 0, \notag
\end{align}\\[-1.75em]
leading to the Newton step $\zb = f(\xb;\theta) - \eta (\nabla^2 \ell(f(\xb;\theta)))^{-1}\allowbreak \nabla \ell(f(\xb;\theta))$.

\section{Newton Losses}
In this section, we build on the two-stage optimization method \eqref{eq:training:update-z}+\eqref{eq:training:update-theta}, and propose optimizing \eqref{eq:training:update-z} with Newton's method while optimizing \eqref{eq:training:update-theta} with stochastic gradient descent.
Let us begin by considering the quadratic approximation of the loss function at the location $\yb=f(\xb;\theta)$, i.e.,
\begin{align}
    \tilde\ell_\yb(\zb) &= \ell(\yb) + (\zb - \yb)^\top \nabla_\yb \ell(\yb) + {\textstyle\frac12} (\zb - \yb)^\top \nabla^2_\yb \ell(\yb) (\zb - \yb)\,.
\end{align}
To find the location~$\zb^\star$ of the minimum of $\tilde\ell_\yb(\zb)$, we set its derivative to $0$:
\begin{equation}
\begin{aligned}
    \nabla_\zb \tilde\ell_\yb(\zb^\star) = 0 ~~
    &\Leftrightarrow~~ \nabla_\yb \ell(\yb) + \nabla^2_\yb \ell(\yb) (\zb^\star - \yb) = 0 \\
    &\Leftrightarrow~~ \nabla_\yb \ell(\yb) = - \nabla^2_\yb \ell(\yb) (\zb^\star - \yb) \\
    &\Leftrightarrow~~ {-} \left(\nabla^2_\yb \ell(\yb)\right)^{-1} \nabla_\yb \ell(\yb) = \zb^\star - \yb 
\end{aligned}\label{eq:newton-step}
\end{equation}
Thus, the minimum of $\tilde\ell_\yb(\zb)$ is
\begin{equation} \label{eq:newton:step:z:star}
    \zb^\star = \arg\min_\zb \tilde\ell_\yb(\zb) 
    = \yb - (\nabla^2_\yb  \ell(\yb))^{-1} \nabla_\yb \ell (\yb)\,.
\end{equation}

When $\ell$ is quadratic, %
it can be readily seen that $\zb^\star$ is independent of the choice of $\yb$.
For non-quadratic functions, we heuristically assume independence as $\zb^\star$ is the projected optimum / goal of the function~$\ell$.
In implementations, this independence can be achieved using \texttt{.detach()} or \texttt{.stop\_gradient()}.
Using $\zb^\star$, we can derive the Newton loss $\ell^*$ as 
\begin{align} \label{eq:newton:loss:felix}
    \ell_{\zb^\star}^*(\yb) &= {\textstyle\frac12} (\zb^\star - \yb)^\top (\zb^\star - \yb) = {\textstyle\frac12} \left\| {\zb^\star - \yb} \right\|_2^2 %
\end{align}
where
\vspace*{-1em}
\begin{align} 
    \zb^\star = \arg\min_\zb \tilde\ell_\yb(\zb)
\end{align}
and its derivative as 
\vspace*{-1em}
\begin{align}
    \nabla_\yb \ell_{\zb^\star}^*(\yb) &= \yb - \zb^\star \,. \label{eq:newton-loss-grad}
\end{align}
With this construction, we obtain the Newton loss $\ell_{\zb^\star}^*$, a new convex loss, which has a gradient that corresponds to the Newton step of the original loss.

Note that \eqref{eq:newton:loss:felix} is an instance of \eqref{eq:omega-use-2} for a regularization term describing the quadratic approximation error, i.e., $\Omega(\zb,f(\xb;\theta)) = \tilde\ell_{f(\xb;\theta)}(\zb) - \ell(\zb)$. 
As $\Omega$ is already implicitly part of the Newton step, we do not need to evaluate it.

In general, $\ell_{\zb^\star}^*$ exhibits more desirable behavior than $\ell$, as a single gradient descent step can solve any quadratic problem, and it possesses the same convergence properties as the Newton method when optimizing $\yb$.
In the case of non-convex $\ell$, as it is common for many algorithmic losses, the incorporation of second-order information also substantially improves empirical performance. 
Note that, in the case of non-convex or ill-conditioned settings, using Tikhonov regularization~\cite{tikhonov1977solutions} stabilizes~$\ell_{z^\star}^*$.

In the following, we define Newton losses and use $x$ and $z^\star$ to denote samples / rows of $\xb$ and $\zb^\star$.
\begin{definition}[Element-wise Hessian-based Newton Losses]
\label{def:elem-wise-newton-loss}
    For a loss function $\ell$, and a given current parameter vector $\theta$, we define the element-wise Hessian-based Newton loss as
    $\ell^*_{z^\star}(y) = \frac{1}{2}\|z_E^\star - y\|^2_2\,$, where
    \begin{align*}
           z_E^\star = \bar y - (\nabla^2_{\bar y}  \ell(\bar y))^{-1} \nabla_{\bar y} \ell (\bar y) \quad \text{and} \quad  \bar y = f(x;\theta) \,.
    \end{align*}
\end{definition}
However, instead of using the element-wise Hessian-based Newton loss, it is typically more stable to use the empirical Hessian-based Newton loss.
\begin{definition}[Empirical Hessian-based Newton Losses]
\label{def:hessian-newton-loss}
    For a loss function $\ell$ and a given current parameter vector $\theta$, we define the empirical Hessian-based Newton loss as
    $\ell_{z^\star}^*(y) = \frac12 \left\| z_H^\star - y\right\|_2^2$, where
    \begin{align*}
        z_H^\star ={\bar y} - \Big(\mathbb{E}_{\bar y}\Big[\nabla^2_{\bar y} \ell({\bar y})\Big]\Big)^{\!-1} \nabla_{\bar y} \ell({\bar y}) \quad \text{and} \quad \bar y = f(x;\theta) \,.
    \end{align*}
\end{definition}
Instead of using the Hessian, it is also possible to use the Fisher information matrix for second-order information.
While this coincides with the Hessian in certain cases, it yields different results in most cases. %
The Fisher-based Newton loss can be seen as using natural gradient descent for optimizing the loss, while using regular gradient descent for optimizing the neural network.
\begin{definition}[Fisher-based Newton Losses]
\label{def:fisher-newton-loss}
    For a loss function $\ell$, and a given current parameter vector $\theta$, we define the Fisher-based Newton loss as
    $\ell^*_{z^\star}(y) = \frac{1}{2}\|z_F^\star - y\|^2_2\,$, where
        \begin{align*}
           z_F^\star = \bar y - \Big( \mathbb{E}_{\bar y}\Big[\nabla_{\bar y} \ell({\bar y})\nabla_{\bar y} \ell({\bar y})^\top\Big] \Big)^{\!-1} \nabla_{\bar y} \ell({\bar y}) ~~ \text{and} ~~  \bar y = f(x;\theta) \,.
    \end{align*}
\end{definition}
After presenting the formulations of different types of Newton losses, we would like to give some remarks on the computational cost and the derivative of Newton losses.

\begin{remark}[Computational Considerations]
    The Hessian of the loss function $\nabla^2_y \ell(y)$ may be approximated using the empirical Fisher matrix $F = \mathbb{E}\left[ \nabla_y \ell(y) \nabla_y \ell(y)^\top \right]$.
    However, as only the Hessian of the loss function (and not the Hessian of the neural network) needs to be computed, computing the exact Hessian $\nabla^2_y \ell(y)$ is usually also fast.
\end{remark}

\vspace{.75em}
\begin{remark}[Derivative of the Newton Loss]
    The derivative of the Newton loss is 
\begin{align}
    \frac{\partial}{\partial\,y} \frac12 \left\| z^\star - y\right\|_2^2 = y - z^\star\,. %
\end{align}
    Note that the derivative of $z^\star$ wrt.~$y$ is zero because it is the projected optimum of the original loss.
\end{remark}

\subsection{Examples} \label{ssec:examples:losses}

We have seen in \eqref{eq:newton:loss:felix} how a given loss function $\ell$ induces a corresponding Newton loss $\ell^*$. For specific loss functions, the Newton loss can be explicitly computed. We begin with the trivial example of the MSE loss. 
For notational simplicity we often drop the subscript $z^\star$ in the definition of the Newton loss \eqref{eq:newton:loss:felix}.

\vspace{1em}
\begin{example}[MSE Loss] \label{ex:MSE:loss}
Consider the classical MSE loss, i.e., $\ell(y) = \frac{1}{2}\|y - y^\star\|^2_2$, where $y^\star$ denotes the ground truth. 
Then,  $z^\star=y^\star$ and accordingly the Newton loss is given as
\begin{equation*}
    \ell^*_{z^\star}(y) = \frac{1}{2}\|z^\star - y\|_2^2 =  \frac{1}{2}\|y^\star - y\|_2^2 = \ell(y)\,.
\end{equation*}
Therefore, the MSE loss $\ell$ and its induced Newton loss $\ell^*$ are equivalent.
\end{example}

\vspace{2.5em}
A popular loss function for classification is the softmax cross-entropy (SMCE) loss, defined as
\begin{align}
    \ell_{\mathrm{SMCE}}(y) = \sum_{i=1}^k - p_i \log q_i\, , \qquad \mathrm{where} \qquad q_i = \frac{\exp(y_i)}{\sum_{j=1}^k \exp(y_j)}\,. \label{sq:softmax-ce}
\end{align}
\begin{example}[Softmax Cross-Entropy Loss]
    For the SMCE loss, the induced Newton loss is given as
    \begin{align}
        \ell_{\mathrm{SMCE}}^*(y) = \frac12 \left\| z^\star - y\right\|_2^2
    \end{align}
    where
    the element-wise Hessian variant is
    \begin{align}
        z_E^\star &= - \left(\operatorname{diag}(q) - qq^\top\right)^{-1} (q - p) + y \,,
    \end{align}
    the empirical Hessian variant is 
    \begin{align}
        z_{H}^\star &= - \mathbb{E}\left[\operatorname{diag}(q) - qq^\top\right]^{-1} (q - p) + y \,,
    \end{align}
    and the empirical Fisher variant is 
    \begin{align}
        z_{F}^\star &= - \mathbb{E}\left[(q - p) (q - p)^\top\right]^{-1} (q - p) + y\,.
    \end{align}
\end{example}
In the experiments, we use the SMCE loss for a classification experiment.

\vspace{2.5em}
A less trivial example is the binary cross-entropy (BCE) loss with
\begin{align}
    \ell_{\mathrm{BCE}}(y) = \mathrm{BCE}(y,p) = - \sum_{i=1}^m p_i \log y_i + (1-p_i) \log (1- y_i)\,,
\end{align}
where $p\in\Delta_m$ is a probability vector encoding of the ground truth.\\

\begin{example}[Binary Cross-Entropy Loss]
    For the BCE loss, the induced Newton loss is given as
    \begin{align}
        \ell_{\mathrm{BCE}}^*(y) = \frac12 \left\| z^\star - y\right\|_2^2\, ,
    \end{align}
    where the element-wise Hessian variant is
    \begin{equation}
    \begin{aligned}
        z_E^\star &= - \operatorname{diag}\left(- p \oslash y^2 + (1-p) \oslash (1-y)^2 \right)^{-1} \\
        & \qquad\qquad\quad \left(p \oslash y - (1-p) \oslash (1-y) \right) + y,
    \end{aligned}
    \end{equation}
    the empirical Hessian variant is 
    \begin{equation}
    \begin{aligned}
        z_{H}^\star &= - \operatorname{diag}\left(\mathbb{E}\left[- p \oslash y^2 + (1-p) \oslash (1-y)^2 \right]\right)^{-1} \\
        & \qquad\qquad\qquad \left(p \oslash y - (1-p) \oslash (1-y) \right) + y,
    \end{aligned}
    \end{equation}
    the empirical Fisher variant is 
    \begin{align}
        z_{F}^\star &= - \mathbb{E}\!\left[ \left(p \oslash y - (1-p) \oslash (1-y) \right) \left(p \oslash y - (1-p) \oslash (1-y) \right)^\top\! \right]^{\!-1} \notag \\
        & \qquad\quad \left(p \oslash y - (1-p) \oslash (1-y) \right) + y,
    \end{align}
    and $\odot$ and $\oslash$ are element-wise operations.
\end{example}

\vspace{1em}
The BCE loss is often extended using the logistic sigmoid function to what is called the sigmoid binary cross-entropy loss (SBCE), defined as
\begin{align}
    \ell_{\mathrm{SBCE}}(y) = \mathrm{BCE}(\sigma(y),p)  \qquad\mathrm{where}\qquad\sigma(x) = \frac{1}{1+\exp(-x)}\,.
\end{align}
\begin{example}[Sigmoid Binary Cross-Entropy Loss]
    For the SBCE loss, the induced Newton loss is given as
    \begin{align}
        \ell_{\mathrm{SBCE}}^*(y) = \frac12 \left\| z^\star - y\right\|_2^2
    \end{align}
    where the element-wise Hessian variant is
    \begin{align}
        z_E^\star &= - \operatorname{diag}\big(\sigma(y) - \sigma(y)^2\big)^{-1} (\sigma(y) - p) + y \,,
    \end{align}
    the empirical Hessian variant is 
    \begin{align}
        z_{H}^\star &= - \operatorname{diag}\big(\mathbb{E}\left[\sigma(y) - \sigma(y)^2\right]\big)^{-1} (\sigma(y) - p) + y  \,,
    \end{align}
    and the empirical Fisher variant is 
    \begin{align}
        z_{F}^\star &= - \mathbb{E}\left[ (\sigma(y) - p) (\sigma(y) - p)^\top \right]^{-1} (\sigma(y) - p) + y  \,,
    \end{align}
    and where $\odot$ and $\oslash$ are element-wise operations.
\end{example}

\clearpage
\section{Regularized Sampling Greedy Optimizer (RESGRO)}

In this section, we introduce the REgularized Sampling GReedy Optimizer (RESGRO), an optimization algorithm for our two-stage optimization setting, which can be used for step~\eqref{eq:training:update-z}.
Importantly, RESGRO does not require differentiability of $\ell$; therefore, it provides an alternative to the differentiable relaxations presented in this thesis.

The key idea behind RESGRO is replacing the difficult optimization of $\ell(z)$ simply by \textit{sampling} a number of samples around the current value and taking the value leading to the best accuracy / objective function value.
This procedure is done for each optimization step~\eqref{eq:training:update-z}, which makes the algorithm \textit{greedy}.
The core ingredient is limiting the number of samples and sampling only around the current value, which \textit{regularizes} RESGRO.
We note that RESGRO can be seen as a form of Hill Climbing.
We can formally introduce RESGRO as follows:
Let $\mathcal{A}\subset\mathcal{Y}$ be a random set, where each element $z\in\mathcal{A}$ is drawn from $z\sim y+\epsilon$ with $y\in\mathcal{Y}$. 
The random set $\mathcal{A}$ consists of $K$ elements, i.e., $|\mathcal{A}|=K$.
$\epsilon$ may be drawn from a Gaussian or a Cauchy distribution, among others.
Then, we can find a regularizer $\Omega$ such that 
\begin{align}
    z^\star &= \mathop{\arg\min}_{z\in\mathcal{Y}} \ell(z) + \Omega(z, f(x; \theta)) = \mathbb{E}_{\mathcal{A}} \left[ \mathop{\arg\min}_{z\in\mathcal{A}} \ell(z) \right] \label{eq:full-resgro}\\
    & = f(\xb;\theta) + \mathbb{E}_{\epsilon_1,...,\epsilon_K}\left[ \mathop{\arg\min}_{\epsilon_i\in\{\epsilon_1,...,\epsilon_K\}} \ell(f(\xb;\theta)+ \epsilon_i) \right].
\end{align}
Identifying an explicit form of the regularizer $\Omega$ is not necessary in this case, as we only require its implicit form, and $\Omega$ does not need to be computable because the minimum can be found through the application of RESGRO.
While the right-hand side of \eqref{eq:full-resgro} is the expectation value of the optimal input among $K$ samples, in many cases, we can simply approximate it via a single set of samples using
\begin{equation} 
    z^\star \triangleq \mathop{\arg\min}_{z\in\mathcal{A}} \ell(z)\,.
\end{equation}
In other cases, however, we want to average over multiple sets of samples to obtain a better and less stochastic training response, for which we could simply compute an empirical estimate of \eqref{eq:full-resgro}.
As, e.g., for shortest-path problems, the algorithm is quite costly, we want to avoid sampling large numbers of samples.
Thus, we can use bootstrapping to more efficiently use the samples, i.e., \textit{bootstrapped RESGRO}.
Also, large numbers of samples reduce the implicit regularization $\Omega$ induced by limited local sampling, an essential component to RESGRO.
Thus, we propose using bootstrapping to effectively reuse samples to approximate $z^\star$ as
\begin{equation} \label{eq:bootstrap:RESGRO}
    z^\star \triangleq \mathbb{E}_{\mathcal{A}\subset\mathcal{B}}\left[ \mathop{\arg\min}_{z\in\mathcal{A}} \ell(z) \right],
\end{equation}
where $\mathcal{B}$ is a set of $M>K$ uniform samples from $y+\epsilon$ and $\mathcal{A}\subset\mathcal{B}$ such that $|\mathcal{A}|=K$.

\clearpage

\subsection{Discussion}
The reason why this simple algorithm performs well when used in our proposed two-stage optimization scheme is that the space of outputs $\mathcal{Y}$ is moderately low dimensional.
In contrast, the space of neural network parameters $\Theta$ is typically extremely high-dimensional, which would make the algorithm non-applicable.
However, it is plausible to apply split backpropagation with \textit{RESGRO} and without classical gradient-based backpropagation, by splitting at each layer, and dividing the neurons of each layer into smaller groups, which are handled independently.

We postulate that RESGRO may also be used for reinforcement learning as a replacement for stochastic smoothing / REINFORCE~\cite{williams1992simple}. 
Further, RESGRO may be combined with REINFORCE, as the samples for which the score / loss function $\ell$ has to be evaluated may be from the same distribution and, therefore, could be shared between RESGRO and REINFORCE, i.e., we can have two training objectives at the computational cost of one (assuming that cost is measured in function evaluations of $\ell$).

\section{Experiments}

For the experiments, we start with a simple classification problem and then extend the evaluation to two applications of algorithmic supervision.
The first algorithmic supervision task is sorting and ranking supervision, where only the relative order of a set of samples is known, while their absolute values remain unsupervised.
The second algorithmic supervision task is shortest-path supervision, where only the shortest path is supervised, while the underlying cost matrix remains unsupervised.
These algorithmic supervision tasks were introduced in Chapter~\ref{ch:algovision}.

\subsection{Classification}

In this section, we explore the utility of Newton losses for the simple case of MNIST classification with a softmax cross-entropy loss.
Note that, as softmax is already a well-behaved objective, we cannot expect large improvements, and the purpose of the experiment is rather to demonstrate that no loss in performance is induced through the Newton losses.

To facilitate a fair and extensive comparison, we benchmark training on $5$~models and with $2$~optimizers:
We use $5$-layer fully connected ReLU networks with $100$ (M1), $400$ (M2) and $1\,600$ (M3) neurons per layer, as well as the convolutional LeNet-5 with sigmoid activations (M4) and LeNet-5 with ReLU activations (M5).
Further, we use SGD and Adam as optimizers.
To evaluate both early performance and full training performance, we test after $1$ and $200$ epochs.
As computing the Hessian inverse is trivial for these settings, we also include the element-wise Hessian (e.w.~H); 
however, as it performs (expectedly) poorly and as it would also be expensive in the following experiments, we do not include it for the algorithmic losses experiments in the following sections.

We run each experiment with $20$ seeds, which allows us to perform significance tests (significance level $0.05$).
As displayed in Table~\ref{tab:mnist-classification}, we find that the element-wise Hessian (e.w.~H) performs similar to regular training.
Using the empirical Hessian (H) is indistinguishable from regular training; specifically, in $12$ out of $20$ cases it is better and significantly better in one $1$ out of $20$ cases, which is to be expected from equal methods (on average $1/20$ tests will be significant at a significance level of $0.05$).
Finally, we find that the Fisher-based Newton losses (F), perform better than regular training.
Specifically, with the SGD optimizer, in $9$ out of $10$ settings, it is significantly better and on the remaining setting, it has a higher mean.
Using the Adam optimizer~\cite{kingma2015adam}, both methods perform similarly.

\begin{table*}[]
    \centering
    \caption{
        MNIST classification learning results.
        The models are a $5$-layer fully connected ReLU networks with $100$ (M1), $400$ (M2) and $1\,600$ (M3) neurons per layer, as well as the convolutional LeNet-5 with sigmoid activations (M4) and LeNet-5 with ReLU activations (M5).
        The results are averaged over $20$ seeds, and significance tests between regular training and the Newton methods are conducted. A better mean is indicated by a gray bold-face number, and a significantly better result is indicated by a black bold-face number. 
    }
    {\footnotesize
    \addtolength{\tabcolsep}{-1.1pt}
    \begin{tabular}{llcccccccccc}
    \toprule
        &                      \hspace{4.8em}Optim. & \multicolumn{5}{c}{SGD Optimizer} & \multicolumn{5}{c}{Adam Optimizer} \\ 
        \cmidrule(r){3-7}\cmidrule(r){8-12}
        Ep. & Method~~~/~~~Model                & M1 & M2 & M3 & M4 & M5    & M1 & M2 & M3 & M4 & M5 \\
    \midrule
        1     & Regular                     & $93.06\%$ & $94.26\%$ & $94.77\%$ & $10.57\%$ & $96.25\%$ & $94.75\%$ & ${\color{gray}\pmb{96.36\%}}$ & $95.90\%$ & $90.60\%$ & $97.56\%$ \\
        1     & Newton L.~(e.w.~H)          & $92.95\%$ & $94.23\%$ & $94.74\%$ & $10.57\%$ & $96.14\%$ & $94.63\%$ & $96.30\%$ & ${\color{gray}\pmb{96.06\%}}$ & $90.32\%$ & ${\color{gray}\pmb{97.63\%}}$ \\
        1     & Newton L.~(H)               & $93.06\%$ & ${\color{gray}\pmb{94.28\%}}$ & ${\color{gray}\pmb{94.77\%}}$ & $10.57\%$ & $96.23\%$ & ${\color{gray}\pmb{94.75\%}}$ & $96.36\%$ & ${\color{gray}\pmb{95.91\%}}$ & ${\color{gray}\pmb{90.63\%}}$ & ${\color{gray}\pmb{97.63\%}}$ \\
        1     & Newton L.~(F)               & $\pmb{94.56\%}$ & $\pmb{95.47\%}$ & $\pmb{95.36\%}$ & ${\color{gray}\pmb{10.64\%}}$ & $\pmb{97.77\%}$ & ${\color{gray}\pmb{94.86\%}}$ & $96.28\%$ & ${\color{gray}\pmb{95.95\%}}$ & $90.57\%$ & $97.49\%$ \\
    \midrule
        200   & Regular                     & $98.13\%$ & $98.40\%$ & $98.46\%$ & $99.06\%$ & $99.07\%$ & $98.12\%$ & $98.46\%$ & $98.62\%$ & $98.95\%$ & ${\color{gray}\pmb{99.23\%}}$ \\
        200   & Newton L.~(e.w.~H)          & $98.11\%$ & $98.39\%$ & $98.44\%$ & $99.02\%$ & ${\color{gray}\pmb{99.09\%}}$ & ${\color{gray}\pmb{98.16\%}}$ & $98.44\%$ & $98.54\%$ & $98.95\%$ & $99.22\%$ \\
        200   & Newton L.~(H)               & $98.11\%$ & ${\color{gray}\pmb{98.42\%}}$ & ${\color{gray}\pmb{98.46\%}}$ & $99.04\%$ & $\pmb{99.11\%}$ & $98.12\%$ & ${\color{gray}\pmb{98.50\%}}$ & ${\color{gray}\pmb{98.63\%}}$ & ${\color{gray}\pmb{98.97\%}}$ & $99.20\%$ \\
        200   & Newton L.~(F)               & $\pmb{98.22\%}$ & $\pmb{98.56\%}$ & $\pmb{98.68\%}$ & $\pmb{99.11\%}$ & $\pmb{99.23\%}$ & $98.09\%$ & $\pmb{98.53\%}$ & $\pmb{98.66\%}$ & ${\color{gray}\pmb{98.98\%}}$ & $99.21\%$ \\
    \bottomrule
    \end{tabular}
    }
    \label{tab:mnist-classification}
\end{table*}

\subsection{Sorting and Ranking Supervision}

After exploring the simple softmax cross-entropy classification loss in the previous section, in this section, we explore the more complex sorting and ranking supervision setting with an array of differentiable sorting-based losses.
Here, we perform the four-digit MNIST sorting benchmark as in Chapter~\ref{ch:diffsort} with differentiable sorting networks, NeuralSort~\cite{grover2019neuralsort}, and SoftSort~\cite{prillo2020softsort}.
Since, in this experiment, the objectives are harder to optimize, we can achieve substantial improvements over the baselines.
We train the CNN using the Adam optimizer~\cite{kingma2015adam} at a learning rate of $10^{-3}$ for $100\,000$ steps and with a batch size of $100$.

We explore NeuralSort, SoftSort, and differentiable sorting networks (DSNs) with logistic and Cauchy distributions in Table~\ref{tab:mnist-softsort-neuralsort-diffsort}.
For NeuralSort and SoftSort, we find that using Newton losses, either based on the Hessian or based on the Fisher matrix, improves performance substantially.
In this case, using the Hessian performs better than using the Fisher matrix.
For DSNs, we find that for logistic DSNs, the improvements are substantial.
In Chapter~\ref{ch:diffsort}, we have shown that monotonic differentiable sorting networks, i.e., the Cauchy DSNs, provide an improved variant of differentiable sorting networks.
Thus, in this case, for $n=5$ and Cauchy DSNs, there is no improvement over the default loss. 
However, for the somewhat harder setting of $n=10$, we can observe an improvement of more than $1\%$ using the Hessian-based Newton loss, even in the Cauchy DSN case.

In summary, we obtain strong improvements on losses that are difficult to optimize, while on well-behaving losses only small or no improvements can be achieved.
This aligns with our goal of improving performance on losses that are hard to optimize.

\subsubsection{RESGRO}

\begin{margintable}
    \centering
    \caption{
    Sorting results for RESGRO.
    The metric is the percentage of rankings correctly identified (EM) (and individual element ranks correctly identified (EW)) averaged over $10$ seeds.
    }
    \setlength{\tabcolsep}{2.pt}
    {\footnotesize
    \begin{tabular}{lcc}
    \toprule
      & $n=5$ & $n=10$ \\
    \midrule
    RESGRO & $74.34\, (88.37)$ & $42.23\, (82.56)$ \\
    \bottomrule
    \end{tabular}
    }
    \label{tab:mnist-sorting-resgro}
\end{margintable}

In addition to Newton losses, we also explore RESGRO on the ranking supervision task.
Here, we apply RESGRO to maximizing the Kendall's $\tau$ coefficient between the ground truth and the predictions, which is a discontinuous metric.
RESGRO can be evaluated extremely fast even for large numbers of samples in this setting.
As shown in Table~\ref{tab:mnist-sorting-resgro}, RESGRO achieves competitive performance compared to differentiable sorting methods.
In fact, the only method that performs better than RESGRO is monotonic differentiable sorting networks (i.e., Cauchy).

\begin{table*}[t]
    \centering
    \caption{
    Differentiable sorting results. 
    The metric is the percentage of rankings correctly identified (EM) (and individual element ranks correctly identified (EW)) averaged over $10$ seeds.
    }
    {\footnotesize
    \setlength{\tabcolsep}{2.1pt}
    \begin{tabular}{lcccccccc}
    \toprule
    & \multicolumn{4}{c}{$n=5$} & \multicolumn{4}{c}{$n=10$}  \\
    \cmidrule(r){2-5}\cmidrule(r){6-9}
    & Cauchy\;DSN & Logistic\,DSN & NeuralSort & SoftSort & Cauchy\;DSN & Logistic\,DSN & NeuralSort & SoftSort   \\
\midrule
Regular         & $85.09\, (93.31)$ & $53.56\, (77.04)$ & $71.33\, (87.10)$ & $70.70\, (86.75)$ & $55.29\, (87.06)$ & $12.31\, (58.81)$ & $24.26\, (74.47)$ & $27.46\, (76.02)$        \\
NL (Hessian)    & $\pmb{85.11\, (93.31)}$ & $\pmb{75.02\, (88.53)}$ & $83.31\, (92.54)$ & $83.87\, (92.72)$ & $\pmb{56.49\, (87.44)}$ & $\pmb{42.14\, (75.35)}$ & $\pmb{48.76\, (84.83)}$ & $\pmb{55.07\, (86.89)}$     \\
NL (Fisher)     & $84.95\, (93.25)$ & $63.11\, (79.28)$ & $\pmb{83.93\, (92.80)}$ & $\pmb{84.03\, (92.82)}$ & $56.12\, (87.35)$ & $25.72\, (52.18)$ & $39.23\, (81.14)$ & $54.00\, (86.56)$   \\
\bottomrule
    \end{tabular}
    }
    \label{tab:mnist-softsort-neuralsort-diffsort}
\end{table*}

\subsection{Shortest-Path Supervision}

In this section, we apply Newton losses to the shortest-path supervision task of the $12\times12$ Warcraft shortest-path problem~\cite{vlastelica2019differentiation, berthet2020learning, petersen2021learning} as in Chapter~\ref{ch:algovision}.
Here, $12\times12$ Warcraft terrain maps are given as $96\times96$ RGB images, and the supervision is the shortest path from the top left to the bottom right according to a hidden cost embedding.
The goal is to predict $12\times12$ cost embeddings of the terrain maps such that the shortest path according to the predicted embedding corresponds to the ground truth shortest path.
For this task, we explore three approaches: the relaxed AlgoVision Bellman-Ford algorithm, stochastic smoothing, and perturbed optimizers.

\begin{table}[b]
    \centering
    \caption{
        Shortest-path benchmark results for different variants of the AlgoVision-relaxed Bellman-Ford algorithm.
        The displayed metric is the percentage of perfect matches averaged over $10$ seeds.
    }
    {\footnotesize
    \begin{tabular}{l|cc|cc}
    \toprule
Algorithm Loop  &     \multicolumn{2}{c}{\texttt{For}}  & \multicolumn{2}{c}{\texttt{While}} \\
Loss            &      $L_1$ & $L_2^2$                  &  $L_1$ & $L_2^2$   \\
    \midrule
Regular         &       94.19 & \textbf{95.90} & 94.30 & 95.77 \\
Fisher Newton   &       \textbf{94.52} & 95.37 & \textbf{94.47} & \textbf{95.93} \\
\bottomrule
    \end{tabular}
    }
    \label{tab:shortest-path-algovision}
\end{table}

For the relaxed AlgoVision Bellman-Ford algorithm, we explore two variants of the algorithm (an outer \texttt{For} loop and an outer \texttt{While} loop) and two losses ($L_1$ and $L_2^2$), i.e., a total of four settings.
As computing the Hessian of the AlgoVision Bellman-Ford algorithm is too expensive with the PyTorch implementation, we restrict this case to the Fisher-based Newton loss.
As displayed in Table~\ref{tab:shortest-path-algovision}, the Newton loss improves performance in three out of four settings, and the overall best performance is also provided by the Newton loss.

After discussing analytical relaxations, we continue with stochastic methods, the results of which are displayed in Table~\ref{tab:shortest-path-stochastic}.
For stochastic smoothing of the loss function, (i.e., stochastic smoothing applied to the algorithm and loss as one unit), we find that Newton losses improve the performance for $10$ and $30$ samples, while the regular training performs best if only $3$ samples can be drawn.
This makes sense as the estimation of Hessian or Fisher with stochastic smoothing is not good enough with too few samples, but as soon as we have at least $10$ samples, it is good enough to improve performance.
For stochastic smoothing of the algorithm, (i.e., stochastic smoothing applied only to the algorithm, and the gradient of the loss afterward computed using backpropagation), we can observe a very similar behavior. 
While estimating the Hessian is intractable in this case, we can see improvements using the Fisher for $\geq 10$ samples.
For perturbed optimizers with a Fenchel-Young loss~\cite{berthet2020learning}, we can confirm that the number of samples drawn barely affects performance. 
By extending the formulation to also computing the Hessian of the Fenchel-Young loss, we can compute the Newton loss, and find that we achieve improvements of more than $2\%$ in this case.

Additionally, we make two interesting observations:

We find that perturbed optimizers are more sample efficient but do not improve with more samples.
Thus, a rule of thumb is that if only a few samples can be afforded, perturbed optimizers are better, and if many samples can be computed, plain stochastic smoothing performs better.

An interesting comparison is also stochastic smoothing of the loss vs.\ stochastic smoothing of the algorithm:
Here, we find that stochastic smoothing of the loss, which computes a gradient, is more sample efficient for few samples.
On the other hand, stochastic smoothing of the algorithm, which computes a large Jacobian, requires more samples but performs better for $\geq 10$ samples.
This makes sense, as estimating a vector (in this case of length $144$) requires fewer samples than estimating a matrix (in this case of size $144\times144$).

\begin{table*}[t]
    \centering
    \caption{
        Shortest-path benchmark results for the stochastic smoothing of the loss (including the algorithm), stochastic smoothing of the algorithm (excluding the loss), and perturbed optimizers with the Fenchel-Young loss.
        The metric is the percentage of perfect matches averaged over $10$ seeds.
    }
    {\footnotesize
    \begin{tabular}{l|ccc|ccc|ccc}
    \toprule
Loss            & \multicolumn{3}{c}{SS of loss} & \multicolumn{3}{c}{SS of algorithm} & \multicolumn{3}{c}{PO~w/ FY loss} \\
\# Samples      & 3 & 10 & 30 & 3 & 10 & 30 & 3 & 10 & 30 \\
\midrule
Regular         & \textbf{62.83} & 77.01 & 85.48                & \textbf{57.55} & 78.70 & 87.26            & 80.64 & 80.39 & 80.71 \\ 
Hessian Newton  & 62.40 & \textbf{78.82} & \textbf{85.94}       & --- & --- & ---                           & \textbf{83.09} & \textbf{81.13} & \textbf{83.45} \\ 
Fisher Newton   & 58.80 & \textbf{78.74} & \textbf{86.10}       & 53.82 & \textbf{79.24} & \textbf{87.41}   & 80.70 & 80.37 & 80.45 \\ 
\bottomrule
    \end{tabular}
    }
    \label{tab:shortest-path-stochastic}
\end{table*}

\begin{table}[h]
    \centering
    \caption{
    Bootstrapped RESGRO applied to shortest-path supervision. 
    Displayed is the percentage of perfect matches on the test set averaged over $10$ seeds.
    }
    {\footnotesize
    \begin{tabular}{lcccccccccccccc}
    \toprule
    & 3 sets & 10 sets & 30 sets & 100 sets \\
    \midrule
6 samples   & 52.85 & 54.98 & 62.67 & 59.24 \\
10 samples  & 68.47 & 72.21 & 74.70 & 75.83 \\
20 samples  & 76.95 & 81.53 & 82.82 & 83.84 \\
30 samples  & 80.16 & 83.83 & 84.60 & 86.01 \\
60 samples  & 82.98 & 85.11 & 86.58 & 87.17 \\
100 samples & 83.63 & 86.21 & 87.58 & 88.11 \\
\bottomrule
    \end{tabular}
    }
    \label{tab:shortest-path-resgro-bootstrap}
\end{table}

\subsubsection{RESGRO}

Using RESGRO for the shortest-path supervision task, we achieve competitive performance compared to stochastic smoothing and perturbed optimizers.
In Table~\ref{tab:shortest-path-resgro-one-set}, we apply the vanilla one-set RESGRO variant with between $3$ and $100$ samples.
In Table~\ref{tab:shortest-path-resgro-bootstrap}, we extend this to the multi-set bootstrapped RESGRO variant, and achieve performance of up to $88.11\%$ with $100$ samples and $86.01\%$ with $30$ samples. 
In comparison, two out of the three methods in Table~\ref{tab:shortest-path-stochastic} underperform the bootstrapped RESGRO for $30$ samples.
Overall, we find that bootstrapping substantially improves performance for RESGRO in the shortest-path setting.

\begin{margintable}[-12em]
    \centering
    \caption{
    Vanilla one-set RESGRO applied to shortest-path supervision. 
    Displayed is the percentage of perfect matches on the test set averaged over $10$ seeds.
    }
    {\footnotesize
    \begin{tabular}{lc}
    \toprule
3 samples   & 36.34 \\
10 samples  & 72.02 \\
30 samples  & 77.22 \\
100 samples & 79.31 \\
\bottomrule
    \end{tabular}
    }
    \label{tab:shortest-path-resgro-one-set}
\end{margintable}

\vspace{1em}
As RESGRO is very competitive compared to stochastic smoothing, which is popular in reinforcement learning, we postulate that RESGRO may also be used for reinforcement learning.
Further, we postulate that Newton losses may also be applied for reinforcement learning.

\subsection*{Conclusion}
In this chapter, we proposed splitting backpropagation, a novel class of alternative optimization methods, which allows combining different optimizers. 
We proposed two instances of splitting backpropagation:
First, Newton losses, a method for combining second-order optimization of the loss function and first-order optimization of the model.
Second, RESGRO, a simple stochastic method for backpropagating through non-differentiable functions.
Both methods deliver strong empirical performance on algorithmic supervision tasks.

\chapter[Conclusion]{Conclusion}
\label{ch:conclusion}
\label{ch:perspectives}

To conclude this thesis, in this chapter, we discuss the implications of learning with differentiable algorithms on the field of machine learning, and discuss interesting avenues of future research.

In this thesis, we combined machine learning and classical algorithms.
First, we developed a general framework for differentiable algorithms.
Then, we focused on specific classes of algorithms, allowing us to cover them in greater detail.
Finally, we proposed alternative optimization methods for learning with algorithms.
These concepts are novel and have a broad range of applications and implications for different fields, as discussed in Section~\ref{sec:future:applications}.
Further, the methods presented in this work provide many direct paths of future research, as discussed in Section~\ref{sec:future:extensions}.

The author's future research will continue in the direction of learning with differentiable algorithms and will build upon the methods presented in this work.
In addition, the author will continue research in the areas of individual fairness, efficient and biologically plausible neural architectures, as well as optimization.

\section{Implications and Future Perspectives}
\label{sec:future:applications}

Learning with differentiable algorithms is a rapidly growing subject in machine learning.
Differentiable algorithms are the foundation for integrating algorithmic concepts into neural architectures, whether to enable algorithmic supervision or an algorithm being part of a neural network model.
Algorithmic supervision enables learning with limited information, reducing manual annotation requirements and reducing human bias in labeling.
In the following, we discuss the direct and indirect implications of differentiable algorithms for a selection of machine learning communities.

\paragraph{Weakly-Supervised Learning}
As algorithmically-supervised learning is a setting of weakly-supervised learning, this work directly contributes to weakly-supervised learning.
Apart from our direct contributions to weakly-supervised learning, in recent time, the number of weakly-supervised architectures that integrate differentiable algorithms is rapidly rising.

\paragraph{Self-Supervised Learning}
Very recently, the utility of differentiable sorting for self-supervised learning has been demonstrated for learning on audio~\cite{carr2021self}.
We have applied differentiable sorting networks for self-supervised representation learning on images~\cite{shvetsova2022differentiable}.

\paragraph{Algorithm-Enhanced Models}
In this work, we considered algorithm-enhanced models, e.g., in the setting of shortest-path supervision.
Another interesting example of algorithm-enhanced models is the image segmentation model by Cho~\etal~\cite{cho2021differentiable}, which they enhance by using differentiable splines.

\paragraph{Neuro-Symbolic Learning}
In the area of neuro-symbolic learning~\cite{besold2017neural, yi2020clevrer, asai2018classical}, the idea is to integrate symbolic knowledge and reasoning into neural architectures. 
Thus, differentiable algorithms and logic are natural candidates for advancing neuro-symbolic systems.
We believe that our alternative optimization methods could also be interesting for the neuro-symbolic community.

\paragraph{Edge Computing and Embedded Machine Learning}
In the growing field of machine learning for edge computing and embedded systems~\cite{murshed2021machine, ajani2021overview, seng2021embedded, branco2019machine}, the focus lies on inference at low computational cost, e.g., on a mobile CPU, a microcontroller, or an IoT device.
As differentiable logic gate networks provide great inference at a very low computational cost, their application in edge computing and embedded systems comes naturally.
While they are currently limited to moderately small architectures, in many edge computing applications, large neural architectures are not feasible in the first place, which substantially reduces the impact of this current limitation.
We believe that deep differentiable logic gate networks are a great opportunity for practitioners in edge computing and embedded systems.

\paragraph{Reinforcement Learning}
The domain of reinforcement learning has some intersections with learning with differentiable algorithms.
Specifically, the idea of stochastic smoothing is well known in reinforcement learning under the name REINFORCE~\cite{williams1992simple} among many other names.
As we propose alternative optimization strategies for learning with blackbox functions such as Newton losses and RESGRO, we postulate that these methods could also be used for reinforcement learning tasks.
Specifically, Newton losses could extend existing reinforcement learning objectives, and for stochastic smoothing, we have demonstrated the utility of Newton losses.
RESGRO could also be used for reinforcement learning and could be promising through its simplicity as well as competitiveness to stochastic smoothing.

\paragraph{Computer Vision and Computer Graphics}
Recently, in computer vision, Neural Radiance Fields (NeRFs) have been popularized.
They do not only deliver neural representations of 3D shapes, but also achieve stunning results on novel view generation tasks~\cite{mildenhall2020nerf}.
NeRFs and similar methods (like~\cite{liu2022neuray}) build on differentiable volume rendering and advances presented in this work could help improve these methods, e.g., interesting directions are using distributions besides the logistic distribution, or using aggregations via T-norms besides the probabilistic T-conorm.
In computer graphics, in recent years, not only have differentiable renderers become very relevant~\cite{tewari2021advances}, but also many other building blocks have been made differentiable, e.g., differentiable material graphs for procedural materials~\cite{shi2020match}, or differentiable simulations~\cite{coros2021differentiable}.

\clearpage
\section{Extensions and Future Work}
\label{sec:future:extensions}

Finally, after discussing implications of learning with differentiable algorithms, we would like to point out immediate extensions for and interesting directions of future work for the methodologies presented in this thesis.

\paragraph{Differentiable Algorithms}
For the proposed general method of differentiable algorithms, natural extensions are using alternative distributions and / or T-norms and T-conorms.
Due to the generality of the method, it can be readily used to relax other algorithms or non-differentiable expressions.
The framework already supports evaluating a neural network as a function as part of a differentiable algorithm; we have not empirically evaluated this case in this thesis, but it is a great direction for future exploration.

\paragraph{Differentiable Sorting and Ranking}
A natural application of differentiable sorting networks are classic learning-to-rank tasks in the domain of recommender systems~\cite{chapelle2011yahoo, qin2013introducing}.
Apart from this, there are also many other applications of differentiable sorting and ranking, as can be seen from the literature published in recent years~\cite{carr2021self, huang2022relational, goyal2018continuous, pobrotyn2021neuralndcg, lee2021differentiable, swezey2021pirank}.

\paragraph{Differentiable Top-$k$}
Differentiable top-$k$ classification learning may be applied in any classification task.
Herein, interesting settings are those with large label-noise and label-ambiguities or settings in which there is uncertainty about the ground truth label; 
however, the method also achieves improvements on vanilla classification settings.

\paragraph{Differentiable Rendering}
Our work on differentiable rendering could be extended by generalizing differentiable rendering also wrt.~other aspects, e.g., wrt.~shading.
Further, the ideas presented in this work could be applied to other forms of differentiable rendering, such as implicit differentiable rendering.

\paragraph{Differentiable Logic}
For differentiable logic gate networks, there are many opportunities for future work:
From an engineering perspective, there are opportunities for improving the computational efficiency of frameworks for training logic gate networks, as well as for improving inference speed.
Another direction is combinatorial optimization and pruning of trained logic gate networks to reduce the number of computations required for inference.
For practitioners, there is the opportunity of using differentiable logic gate networks in edge computing and embedded systems---both for replacing existing models by better and more efficient models, or enabling the use of machine learning models in extremely low-end embedded systems and microcontrollers in the first place.
From a research perspective, methods that have shown success for training conventional neural networks may be adapted for training differentiable logic gate networks.
Furthermore, developing more scalable architectures is an interesting research direction.

\paragraph{Alternative Optimization Methods}
The proposed two-stage optimization method can readily be extended to an $n$-stage optimization method by applying the split at multiple locations.
Moreover, splitting backpropagation can also be directly applied to other optimization methods, such as convex optimization.
We believe that splitting backpropagation can also improve performance in many other applications of differentiable algorithms, as we saw improvements over state-of-the-art methods in both of the empirically evaluated settings.
But we also believe that the method has many applications beyond learning with differentiable algorithms.

\appendix 

\pagelayout{wide} %
\addpart{Supplementary Materials}
\pagelayout{margin} %

\setchapterpreamble[u]{\margintoc}

\chapter{Distributions}
\label{sm:dist}

\begin{figure*}[]
    \centering
    \newcommand{\diagwidth}{.14\linewidth}
    \addtolength{\tabcolsep}{-3pt}
    \footnotesize
    \begin{tabular}{cccccc}
        \toprule
        \includegraphics[width=\diagwidth]{fig_gendr/diags/diags-0.pdf} &
        \includegraphics[width=\diagwidth]{fig_gendr/diags/diags-1.pdf} &
        \includegraphics[width=\diagwidth]{fig_gendr/diags/diags-3.pdf} &
        \includegraphics[width=\diagwidth]{fig_gendr/diags/diags-5.pdf} &
        \includegraphics[width=\diagwidth]{fig_gendr/diags/diags-7.pdf} &
        \includegraphics[width=\diagwidth]{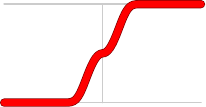}
        \\
        \includegraphics[width=\diagwidth]{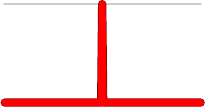} &
        \includegraphics[width=\diagwidth]{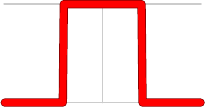} &
        \includegraphics[width=\diagwidth]{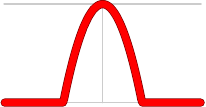} &
        \includegraphics[width=\diagwidth]{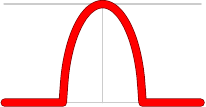} &
        \includegraphics[width=\diagwidth]{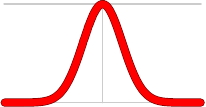} &
        \includegraphics[width=\diagwidth]{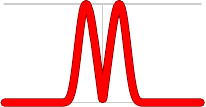}
        \\
        Heaviside &
        Uniform &
        Cubic Hermite &
        Wigner Semicircle &
        Gaussian &
        Gaussian (sq.)
        \\
        \midrule
        \includegraphics[width=\diagwidth]{fig_gendr/diags/diags-9.pdf} &
        \includegraphics[width=\diagwidth]{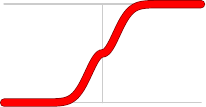} &
        \includegraphics[width=\diagwidth]{fig_gendr/diags/diags-11.pdf} &
        \includegraphics[width=\diagwidth]{fig_gendr/diags/diags-12.pdf} &
        \includegraphics[width=\diagwidth]{fig_gendr/diags/diags-13.pdf} &
        \includegraphics[width=\diagwidth]{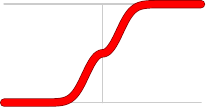}
        \\
        \includegraphics[width=\diagwidth]{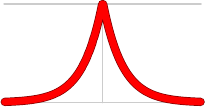} &
        \includegraphics[width=\diagwidth]{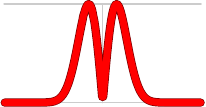} &
        \includegraphics[width=\diagwidth]{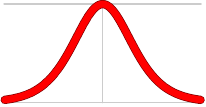} &
        \includegraphics[width=\diagwidth]{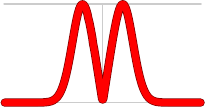} &
        \includegraphics[width=\diagwidth]{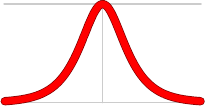} &
        \includegraphics[width=\diagwidth]{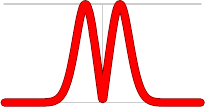}
        \\
        Laplace &
        Laplace (sq.) &
        Logistic &
        Logistic (sq.) &
        Hyperbolic secant &
        Hyperbolic secant (sq.)
        \\
        \midrule
        \includegraphics[width=\diagwidth]{fig_gendr/diags/diags-15.pdf} &
        \includegraphics[width=\diagwidth]{fig_gendr/diags/diags-16.pdf} &
        \includegraphics[width=\diagwidth]{fig_gendr/diags/diags-17.pdf} &
        \includegraphics[width=\diagwidth]{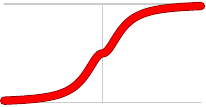} &
        \includegraphics[width=\diagwidth]{fig_gendr/diags/diags-19.pdf} &
        \includegraphics[width=\diagwidth]{fig_gendr/diags/diags-21.pdf}
        \\
        \includegraphics[width=\diagwidth]{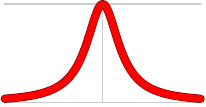} &
        \includegraphics[width=\diagwidth]{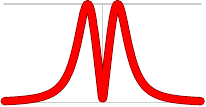} &
        \includegraphics[width=\diagwidth]{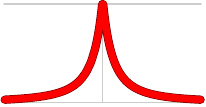} &
        \includegraphics[width=\diagwidth]{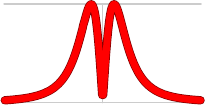} &
        \includegraphics[width=\diagwidth]{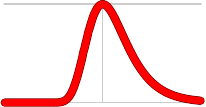} &
        \includegraphics[width=\diagwidth]{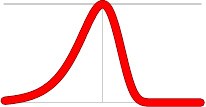}
        \\
        Cauchy &
        Cauchy (sq.) &
        Reciprocal &
        Reciprocal (sq.) &
        Gumbel-Max &
        Gumbel-Min
        \\
        \midrule
        \includegraphics[width=\diagwidth]{fig_gendr/diags/diags-23.pdf} &
        \includegraphics[width=\diagwidth]{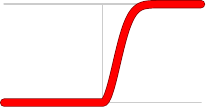} &
        \includegraphics[width=\diagwidth]{fig_gendr/diags/diags-27.pdf} &
        \includegraphics[width=\diagwidth]{fig_gendr/diags/diags-47.pdf} &
        \includegraphics[width=\diagwidth]{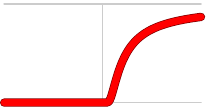} &
        \includegraphics[width=\diagwidth]{fig_gendr/diags/diags-51.pdf}
        \\
        \includegraphics[width=\diagwidth]{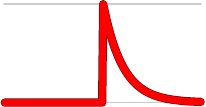} &
        \includegraphics[width=\diagwidth]{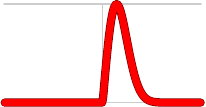} &
        \includegraphics[width=\diagwidth]{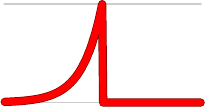} &
        \includegraphics[width=\diagwidth]{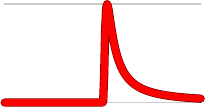} &
        \includegraphics[width=\diagwidth]{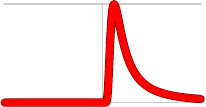} &
        \includegraphics[width=\diagwidth]{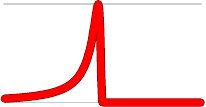}
        \\
        Exponential &
        Exponential (sq.) &
        Exponential (Rev.) &
        Levy &
        Levy (sq.) &
        Levy (Rev.)
        \\
        \midrule
        \includegraphics[width=\diagwidth]{fig_gendr/diags/diags-37.pdf} &
        \includegraphics[width=\diagwidth]{fig_gendr/diags/diags-35.pdf} &
        \includegraphics[width=\diagwidth]{fig_gendr/diags/diags-33.pdf} &
        \includegraphics[width=\diagwidth]{fig_gendr/diags/diags-45.pdf} &
        \includegraphics[width=\diagwidth]{fig_gendr/diags/diags-43.pdf} &
        \includegraphics[width=\diagwidth]{fig_gendr/diags/diags-41.pdf}
        \\
        \includegraphics[width=\diagwidth]{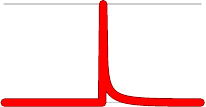} &
        \includegraphics[width=\diagwidth]{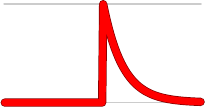} &
        \includegraphics[width=\diagwidth]{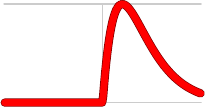} &
        \includegraphics[width=\diagwidth]{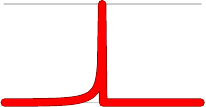} &
        \includegraphics[width=\diagwidth]{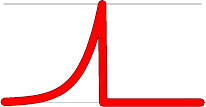} &
        \includegraphics[width=\diagwidth]{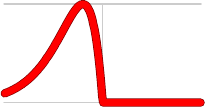}
        \\
        Gamma ($p{=}0.5$) &
        Gamma ($p{=}1$) &
        Gamma ($p{=}2$) &
        Gamma ($p{=}.5$, R.) &
        Gamma ($p{=}1$, R.) &
        Gamma ($p{=}2$, R.)
        \\
        \midrule
        \includegraphics[width=\diagwidth]{fig_gendr/diags/diags-38.pdf} &
        \includegraphics[width=\diagwidth]{fig_gendr/diags/diags-36.pdf} &
        \includegraphics[width=\diagwidth]{fig_gendr/diags/diags-34.pdf} &
        \includegraphics[width=\diagwidth]{fig_gendr/diags/diags-46.pdf} &
        \includegraphics[width=\diagwidth]{fig_gendr/diags/diags-44.pdf} &
        \includegraphics[width=\diagwidth]{fig_gendr/diags/diags-42.pdf}
        \\
        \includegraphics[width=\diagwidth]{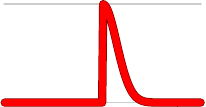} &
        \includegraphics[width=\diagwidth]{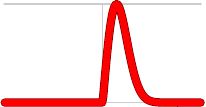} &
        \includegraphics[width=\diagwidth]{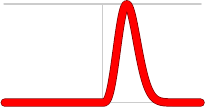} &
        \includegraphics[width=\diagwidth]{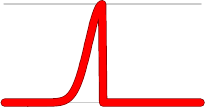} &
        \includegraphics[width=\diagwidth]{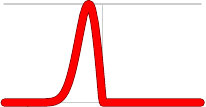} &
        \includegraphics[width=\diagwidth]{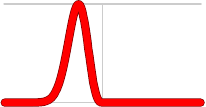}
        \\
        Gamma ($p{=}0.5$, sq.) &
        Gamma ($p{=}1$, sq.) &
        Gamma ($p{=}2$, sq.) &
        Gamma ($p{=}.5$, R., sq.) &
        Gamma ($p{=}1$, R., sq.) &
        Gamma ($p{=}2$, R., sq.)
        \\
        \bottomrule
    \end{tabular}
    \caption{Visualization of CDFs \textit{(top)} and PDFs \textit{(bottom)} for different distributions.}
    \label{tab:vis-pdf-cdf}
\end{figure*}

In this supplementary material, we define each of the presented distributions / sigmoid functions.
These are relevant to Chapters~\ref{ch:algovision}, \ref{ch:diffsort}, \ref{ch:difftopk}, and~\ref{ch:gendr}, as they are either used in or comprise extension to the methods in the respective chapters.
Figure~\ref{tab:vis-pdf-cdf} displays the respective CDFs and PDFs.

An extensive discussion, including a taxonomy of the presented distributions, can be found in Chapter~\ref{ch:gendr}, Section~\ref{sec:instantiations}.

Note that, for each distribution, the PDF $\density$ is defined as the derivative of the CDF $\cdf$.
Also, note that a reversed (Rev.) CDF is defined as $\cdfrev(x) = 1-\cdf(-x)$, which means that $\cdfrev = \cdf$ for symmetric distributions.
The square-root distribution~$\cdfsq$ is defined in terms of $\cdf$ as in Equation (\ref{eq:cdfsq}).
Therefore, in the following, we define the distributions via their CDFs~$\cdf$.\\

\newcommand{\disttitle}[1]{\bigskip\smallskip\noindent\textbf{#1}}

\disttitle{Heaviside}
\begin{equation}
    x\mapsto\begin{cases}
    0 & \mathrm{if}~x < 0 \\
    1 & \mathrm{otherwise}
    \end{cases}
\end{equation}

\disttitle{Uniform}
\begin{equation}
    x\mapsto\begin{cases}
    0 & \mathrm{if}~x < -1 \\
    0.5 \cdot (1 + x) & \mathrm{if}~-1 \leq x \leq 1 \\
    1 & \mathrm{otherwise}
    \end{cases}
\end{equation}

\disttitle{Cubic Hermite}
\begin{equation}
    x\mapsto\begin{cases}
    0 & \mathrm{if}~x < -1 \\
    3 y^2 - 2 y^3 & \mathrm{if}~-1 \leq x \leq 1 \\
    1 & \mathrm{otherwise}
    \end{cases}
\end{equation}
where $y:=(x+1)/2$.

\disttitle{Wigner Semicircle}
\begin{equation}
    x\mapsto\begin{cases}
    0 & \mathrm{if}~x < -1 \\
    \frac12+\frac{x\sqrt{1-x^2}}{\pi} + \frac{\arcsin (x)}{\pi} & \mathrm{if}~-1 \leq x \leq 1 \\
    1 & \mathrm{otherwise}
    \end{cases}
\end{equation}

\disttitle{Gaussian}
\begin{equation}
    x\mapsto
    \frac12\left(
        1 + \operatorname{erf} \left( \frac{x}{\sqrt{2}} \right)
    \right)
\end{equation}

\disttitle{Laplace}
\begin{equation}
    x\mapsto \begin{cases} \frac{1}{2} \exp \left(x\right) & \text{if }x\leq 0 \\ 1-{\frac {1}{2}}\exp \left(-x\right) & \text{if }x\geq 0 \end{cases}
\end{equation}

\disttitle{Logistic}
\begin{equation}
    x\mapsto \frac{1}{1 + \exp (-x)}
\end{equation}

\disttitle{Hyperbolic secant / Gudermannian}
\begin{equation}
    x\mapsto \frac{2}{\pi} \arctan \left( \exp \left( \frac{\pi}{2} x \right)\right)
\end{equation}

\disttitle{Cauchy}
\begin{equation}
    x\mapsto \frac{1}{\pi} \arctan\left( x \right) + \frac{1}{2}
\end{equation}

\disttitle{Reciprocal}
\begin{equation}
    x\mapsto \frac{x}{2 + 2|x|} + \frac12
\end{equation}

\disttitle{Gumbel-Max}
\begin{equation}
    x\mapsto e^{-e^{-x}}
\end{equation}

\disttitle{Gumbel-Min}
\begin{equation}
    x\mapsto e^{-e^x}
\end{equation}

\disttitle{Exponential}
\begin{equation}
    x\mapsto 1 -e^{-x}
\end{equation}

\disttitle{Levy}
\begin{equation}
    x\mapsto 2 -2\Phi \left(\sqrt{\frac{1}{x}}\right)
\end{equation}
where $\Phi$ is the CDF of the standard normal distribution.

\disttitle{Gamma}
\begin{equation}
    x\mapsto \frac{1}{\Gamma(p)} \gamma(p,x)
\end{equation}
where $\gamma(p,x)$ is the lower incomplete gamma function and $p>0$ is the shape parameter.

\chapter{T-Norms and T-Conorms}
\label{sm:tcn}

In this supplementary material, we cover various T-norms and T-conorms, which are integral to real-valued logic as used in the aggregation within differentiable rendering, in differentiable logic networks, as well as in general differentiable algorithms.

An extensive discussion, including a characterization of the presented families of T-norms and T-conorms, can be found in Chapter~\ref{ch:gendr}, Section~\ref{sec:instantiations}.

The axiomatic approach to multi-valued logics is based on defining reasonable properties for truth functions. 
We stated the axioms for multi-valued generalizations of the disjunction (logical ``or''), called T-conorms, in Definition~\ref{def:t-conorm}. 
Here, we complement this with the axioms for multi-valued generalizations of the conjunction (logical ``and''), called T-norms.
\begin{definition}[T-norm]
A T-norm (triangular norm) is a binary operation~$\top: [0,1] \times [0,1] \to [0,1]$, which satisfies
\begin{itemize}\itemsep0pt
\item associativity:
      $\top(a, \top(b,c)) = \top(\top(a,b), c)$,
\item commutativity:
      $\top(a,b) = \top(b,a)$,
\item monotonicity:
      $(a \leq c) \land (b \leq d)
       \Rightarrow \top(a,b) \leq \top(c,d)$,
\item $1$ is a neutral element:
      $\top(a,1) = a$.
\end{itemize}
\end{definition}
These axioms ensure that the corners of the unit square, that is, the value pairs considered in classical logic, are processed as with a standard conjunction: neutral element and commutativity imply that $(1,1) \mapsto 1$, $(0,1) \mapsto 0$, $(1,0) \mapsto 0$. 
From one of the latter two and monotonicity, it follows $(0,0) \mapsto 0$.
Analogously, the axioms of T-conorms ensure that the corners of the unit square are processed as with a standard disjunction. 
Actually, the axioms already fix the values not only at the corners but on the boundaries of the unit square. 
Only inside the unit square (that is, for $(0,1)^2$) T-norms (as well as T-conorms) can differ.

\begin{table}[b]
\centering
\begin{tabular}{|l|c@{~$=$~}l|}\hline\rule{0pt}{2.6ex}%
Minimum         & $\top^M(a,b)$
  & $\min(a,b)$                  \\[1.2ex]
Probabilistic   & $\top^P(a,b)$
  & $ab$                         \\[1.2ex]
Einstein        & $\top^E(a,b)$
  & $\frac{ab}{2-a-b+ab}$        \\[1.2ex]
Hamacher        & $\top^H_p(a,b)$
  & $\frac{ab}{p+(1-p)(a+b-ab)}$ \\[1.2ex]
Frank           & $\top^F_p(a,b)$
  & $\log_p\left(1+\frac{(p^a-1)(p^b-1)}{p-1}\right)$ \\[1.2ex]
Yager           & $\top^Y_p(a,b)$
  & $\max\left(0, 1-\left(\left(1-a\right)^p+\left(1-b\right)^p\right)^{\frac{1}{p}}\right)$ \\[1.2ex]
Acz\'el-Alsina  & $\top^A_p(a,b)$
  & $\exp\big(-\left(|\log(a)|^p+|\log(b)|^p
               \right)^{\frac{1}{p}}\big)$ \\[0.3ex]
Dombi           & $\top^D_p(a,b)$
  & $\Big(1+\left( \left(\frac{1-a}{a}\right)^p
                  +\left(\frac{1-b}{b}\right)^p
           \right)^{\frac{1}{p}}\Big)^{\!\!-1}$ \\[1.2ex]
Schweizer-Sklar & $\top^S_p(a,b)$
  & $(a^p+b^p-1)^{\frac{1}{p}}$ \\[1ex] \hline
\end{tabular}
\caption{\label{tab.t-norms}(Families of) T-norms.\kern-5.4em}
\end{table}

\begin{table}[t]
\centering
\resizebox{\linewidth}{!}{
\begin{tabular}{|l|c@{~$=$~}l|}\hline\rule{0pt}{2.6ex}%
Maximum         & $\bot^M(a,b)$
  & $\max(a,b)$                  \\[1.2ex]
Probabilistic   & $\bot^P(a,b)$
  & $a+b-ab$                     \\[1.2ex]
Einstein        & $\bot^E(a,b)$
  & $\bot^H_2(a,b)=\frac{a+b}{1+ab}$           \\[1.2ex]
Hamacher        & $\bot^H_p(a,b)$
  & $\frac{a+b+(p-2)ab}{1+(p-1)ab}$ \\[1.2ex]
Frank           & $\bot^F_p(a,b)$
  & $1-\log_p\left(1+\frac{(p^{1-a}-1)(p^{1-b}-1)}{p-1}\right)$ \\[1.2ex]
Yager           & $\bot^Y_p(a,b)$
  & $\min\left(1, (a^p+b\kern0.1pt^p)^{\frac{1}{p}}\right)$ \\[1.2ex]
Acz\'el-Alsina  & $\bot^A_p(a,b)$
  & $1 -\exp\big(-\left(|\log(1-a)|^p+|\log(1-b)|^p
                 \right)^{\frac{1}{p}}\big)$ \\[0.3ex]
Dombi           & $\bot^D_p(a,b)$
  & $\Big(1+\left( \left(\frac{1-a}{a}\right)^p
                  +\left(\frac{1-b}{b}\right)^p
           \right)^{-\frac{1}{p}}\Big)^{\!\!-1}$ \\[1.2ex]
Schweizer-Sklar & $\bot^S_p(a,b)$
  & $1-((1-a)^p+(1-b)^p-1)^{\frac{1}{p}}$ \\[1ex] \hline
\end{tabular}
}
\caption{\label{tab.t-conorms}(Families of) T-conorms.}
\end{table}

In the theory of multi-valued logics, and especially in fuzzy logic \cite{klir1997fuzzy}, it was established that the largest possible T-norm is the minimum and the smallest possible T-conorm is the maximum: for any
T-norm~$\top$ it is $\top(a,b) \le \min(a,b)$ and for any T-conorm~$\bot$ it is $\bot(a,b) \ge \max(a,b)$.
The other extremes, i.e., the smallest possible T-norm and the largest possible T-conorm are the so-called drastic T-norm, defined as $\top^\circ(a,b) = 0$ for $(a,b) \in (0,1)^2$, and the drastic T-conorm, defined as $\bot^\circ(a,b) = 1$ for $(a,b) \in (0,1)^2$. 
Hence, it is $\top(a,b) \ge \top^\circ(a,b)$ for any T-norm~$\top$ and $\bot(a,b) \le \bot^\circ(a,b)$ for any T-conorm~$\bot$.
We do not consider the drastic T-conorm because it clearly does not yield useful gradients.

As mentioned, it is common to combine a T-norm~$\top$, a T-conorm~$\bot$ and a negation~$N$ (or complement, most commonly $N(a) = 1-a$) so that DeMorgan's laws hold. 
Such a triplet is often called a {\em dual triplet}.
In Tables~\ref{tab.t-norms} and~\ref{tab.t-conorms} we show the formulas for the families of T-norms and T-conorms, respectively, where matching lines together with the standard negation $N(a) = 1-a$ form dual triplets.
Note that, for some families, we limited the range of values for the parameter~$p$ (see Table~\ref{tab:t-conorms}) compared to more general definitions~\cite{klir1997fuzzy}.

\begin{figure*}[h!]
  \centering
  \includegraphics[width=\linewidth,trim={0 5156px 0 60px},clip]{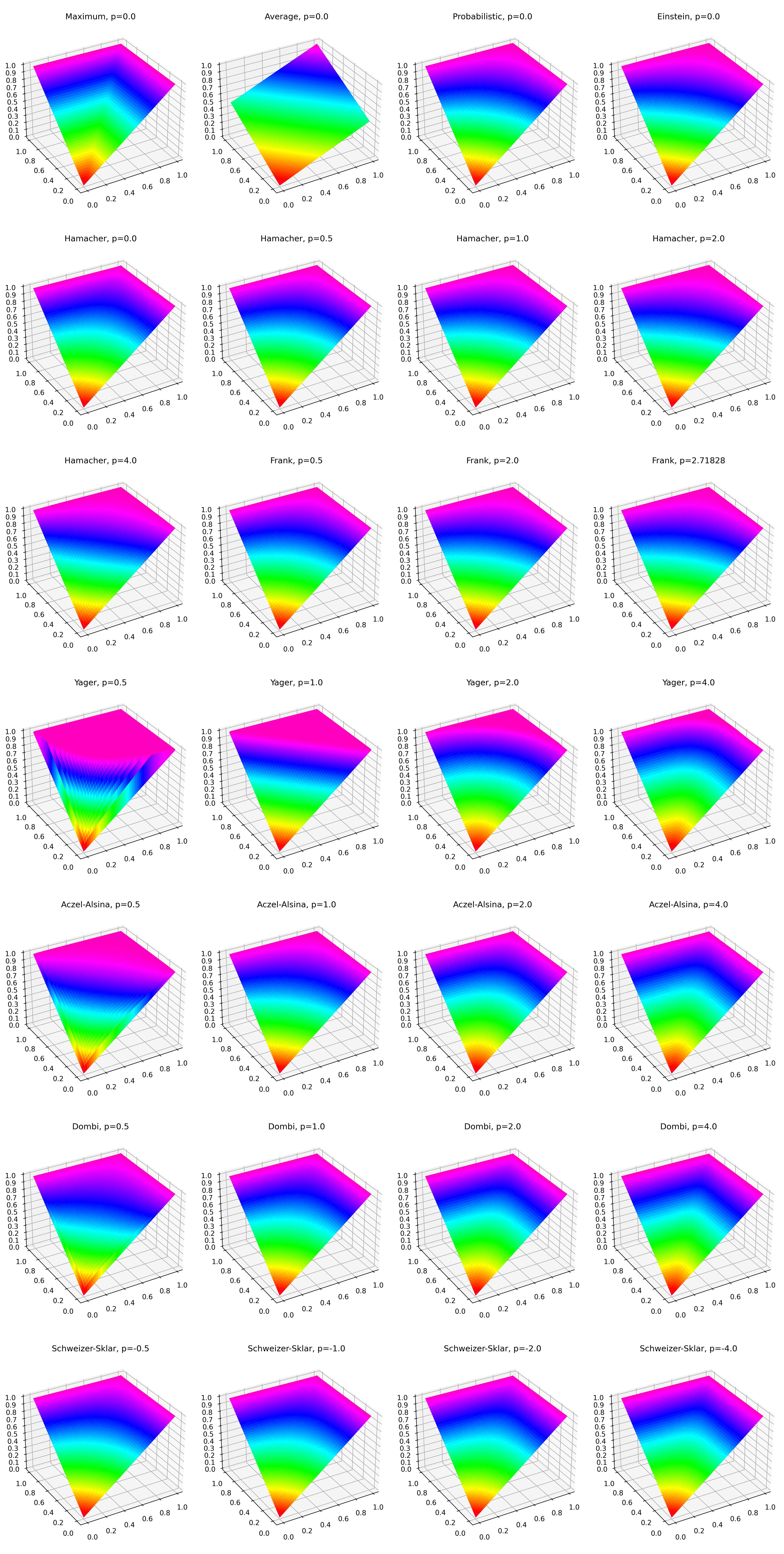}
  \caption{T-conorm plots (1/2). Note that `Average' is not a T-cornom and just included for reference. Also, Note how `Probabilistic' is equal to `Hamacher $p=1$' and `Einstein' is equal to `Hamacher $p=2$'.}
  \label{fig:t-plot-1}
\end{figure*}

Figures~\ref{fig:t-plot-1} and~\ref{fig:t-plot-2} illustrate the considered set of T-conorms.
\vspace{1.5em}

\begin{figure*}
  \centering
  \includegraphics[width=\linewidth,trim={0 0 0 2090px},clip]{fig_gendr/t_conorms.jpg}
  \caption{T-conorm plots (2/2).}
  \label{fig:t-plot-2}
\end{figure*}

\chapter[Relaxed Minimum and Maximum]{Relaxed Minimum\\ and Maximum}

\newcommand{\propertyFormatMinMax}[1]{\\[1.5em]\textbf{#1}\hspace{1em}}

\label{sec:properties-of-min-and-max}

A core element of differentiable algorithms is the relaxation of the $\min$ and $\max$ operators.
For example, for differentiable sorting networks, the relaxations of $\min$ and $\max$ (i.e., $\softmin$ and $\softmax$) are used to relax the conditional swap operation, allowing for a soft transition between passing through and swapping, such that the sorting operator becomes differentiable.
It is natural to try to achieve this by using soft versions of the minimum (denoted by $\softmin$) and maximum operators (denoted by $\softmax$).
However, before we consider concrete examples, let us collect some desirable properties that such relaxations should have.
Naturally, $\softmin$ and $\softmax$ should satisfy many properties that their crisp / hard counterparts $\min$ and $\max$ satisfy, as well as a few others (for $a,b,c \in \IR$):
\propertyFormatMinMax{Symmetry / Commutativity}
  Since $\min$ and $\max$ are symmetric/commutative, so should be their soft counterparts: $\softmin(a,b) = \softmin(b,a)$ and $\softmax(a,b) = \softmax(b,a)$.
\propertyFormatMinMax{Ordering}
  A (soft) maximum of two numbers should be at least as large as a (soft) minimum of the same two numbers:
  $\softmin(a,b) \le \softmax(b,a)$.
\propertyFormatMinMax{Continuity in Both Arguments}
  Both $\softmin$ and $\softmax$ should be continuous in both arguments.
\propertyFormatMinMax{Idempotency}
  If the two arguments are equal in value, this value should be the result of $\softmin$ and $\softmax$, that is, $\softmin(a,a) = \softmax(a,a) = a$.
\propertyFormatMinMax{Inversion}
  As for $\min$ and $\max$, the two operators $\softmin$ and $\softmax$ should be connected in such a way that the result of one operator equals the negated result of the other operator applied to negated arguments: $\softmin(a,b) = -\softmax(-a,-b)$ and $\softmax(a,b) = -\softmin(-a,-b)$.
\propertyFormatMinMax{Stability / Shift Invariance}
  Shifting both arguments by some value~$c \in \IR$ should shift each operator's result by the same value:
  $\softmin(a+c,b+c) = \softmin(a,b) +c$ and $\softmax(a+c,b+c) = \softmax(a,b) +c$.
  Stability implies that the values of $\softmin$ and $\softmax$ depend effectively only on the difference of their arguments.
  Specifically, choosing $c = -a$ yields
  $\softmin(a,b) = \softmin(0,b-a) +a$ and
  $\softmax(a,b) = \softmax(0,b-a) +a$, and $c = -b$ yields
  $\softmin(a,b) = \softmin(a-b,0) +b$ and
  $\softmax(a,b) = \softmax(a-b,0) +b$.
\propertyFormatMinMax{Sum preservation}
  The sum of $\softmin$ and $\softmax$ should equal the sum of $\min$ and $\max$:
  $\softmin(a,b) +\softmax(a,b) = \min(a,b) +\max(a,b) = a+b$.
  Note that sum preservation follows from stability, inversion and symmetry:
  $\softmin(a,b) = \softmin(a-b,0) +b
                 = b -\softmax(0,b-a)
                 = b -(\softmax(a,b) -a)
                 = a+b -\softmax(a,b)$
\propertyFormatMinMax{Bounded by Hard Versions}
  Soft operators should not yield values more extreme than their crisp / hard counterparts: $\min(a,b) \le \softmin(a,b)$ and $\softmax(a,b) \le \max(a,b)$.
  Note that together with ordering this property implies idempotency, vi\tironianEt.: $a = \min(a,a) \le \softmin(a,a) \le \softmax(a,a) \le \max(a,a) = a$.
  Otherwise, they cannot be defined via a convex combination of their inputs, making it impossible to define proper $\softargmin$ and $\softargmax$, and hence we could not compute, e.g., differentiable permutation matrices.
\propertyFormatMinMax{Monotonicity in Both Arguments}
  For any $c > 0$, it should be
  $\softmin(a+c,b) \ge \softmin(a,b)$,
  $\softmin(a,b+c) \ge \softmin(a,b)$,
  $\softmax(a+c,b) \ge \softmax(a,b)$, and
  $\softmax(a,b+c) \ge \softmax(a,b)$.
  Note that the second expression for each operator follows from the first with the help of symmetry / commutativity.
\propertyFormatMinMax{Bounded Error / Minimum Deviation from Hard Versions}
  Soft versions of minimum and maximum should differ as little as possible from their crisp / hard counterparts. 
  However, this condition needs to be made more precise to yield concrete properties (see below for details).

Note that $\softmin$ and $\softmax$ {\em cannot\/} satisfy associativity, as this would force them to be identical to their hard counterparts.
Associativity means that
$\softmax(a, \softmax(b,c)) = \softmax(\softmax(a,b),c)$ and that
$\softmin(a, \softmin(b,c)) =\allowbreak \softmin(\softmin(a,b),c)$.
Now, consider $a,b \in \mathbb{R}$ with $a < b$.
With associativity and idempotency $\softmax(a, \softmax(a,b)) = \softmax(\softmax(a,a),b) = \softmax(a,b)$ and hence $\softmax(a,b) = b = \max(a,b)$ (by comparison of the second arguments). 
Analogously, one can show that if associativity held, we would have $\softmin(a,b) = a = \min(a,b)$.
That is, one cannot have both associativity and idempotency.
Note that, without idempotency, the soft operators would not be bounded by their hard versions.
As idempotency is necessary, associativity has to be given up.

If $\softmin$ and $\softmax$ are to be bounded by the crisp / hard version and symmetry, ordering, inversion and stability (which imply sum preservation) hold, they must be convex combinations of the arguments~$a$ and $b$ with weights that depend only on the difference of $a$ and $b$.
That is,
\begin{eqnarray*}
\softmin(a,b) & = & \phantom{(1-{}} f(b-a)\phantom{)} \cdot a
                   +(1-f(b-a)) \cdot b \\
\softmax(a,b) & = & (1-f(b-a)) \cdot a
                   +\phantom{(1-{}} f(b-a) \phantom{)} \cdot b,
\end{eqnarray*}
where $f(x)$ yields a value in $[0,1]$ (due to the boundedness of $\softmin$ and $\softmax$ by their crisp / hard counterparts).
Due to inversion, $f$ must satisfy $f(x) = 1-f(-x)$ and hence $f(0) = \frac{1}{2}$. 
Monotonicity of $\softmin$ and $\softmax$ requires that $f$ is a monotonically increasing function. Continuity requires that $f$ is a continuous function. In summary, $f$ must be a continuous sigmoid function (in the older meaning of this term, i.e., an s-shaped function, of which the logistic function is only a special case) satisfying $f(x) = 1-f(-x)$.
Such a sigmoid function is the CDF of a corresponding probability distribution, and, therefore, CDFs of probability distributions are candidates for $f$. 
We note that we discuss a selection of distributions in Supplementary Material~\ref{sm:dist}.

As mentioned, the condition that the soft versions of minimum and maximum should deviate as little as possible from the crisp / hard versions causes a slight problem: this deviation can always be made smaller by making the sigmoid function steeper (reaching the crisp / hard versions in the limit for infinite inverse temperature, when the sigmoid function turns into the Heaviside step function). 
Hence, in order to find the best shape of the sigmoid function, we have to limit its inverse temperature.
Therefore, for the respective analysis, we require w.l.o.g. the sigmoid function to be Lipschitz-continuous with Lipschitz constant~$\alpha = 1$.

\label{apx:proofs:sec:monotonic}

\newpage
\begin{widepar}
\section{Monotonicity of the Reciprocal Relaxed Min and Max}
\begin{theorem}
    $\min_{f_\mathcal{R}}$ and $\max_{f_\mathcal{R}}$ are monotonic functions with the sigmoid function $f_\mathcal{R}$.
\begin{proof}
W.l.o.g., we assume $a_i = x$ and $a_j = 0$.
\begin{align}
    \softmin_{f_\mathcal{R}}(x, 0)
    = x \cdot f_\mathcal{R}(-x)
    = x \frac{1}{2} \left(\frac{x}{1+|x|} + 1 \right)
\end{align}
To show monotonicity, we consider its derivative / slope.
\begin{align}
    \frac{d}{dx} \softmin_{f_\mathcal{R}}(x, 0) 
    &= \frac{d}{dx}  \left( x \frac{1}{2} \left(\frac{x}{1+|x|} + 1 \right) \right) \\
    &= \frac{1}{2} \left(\frac{x}{1+|x|} + 1 \right) + x \frac{1}{2} \frac{d}{dx}  \left( \frac{x}{1+|x|} + 1  \right) \\
    &= \frac{1}{2} \left(\frac{x}{1+|x|} + 1 \right) + x \frac{1}{2} \frac{d}{dx}  \left( \frac{x}{1+|x|} \right) \\
    &= \frac{1}{2} \left(\frac{x}{1+|x|} + 1 \right) + x \frac{1}{2} \frac{ \frac{dx}{dx} \cdot (1+|x|) - x \cdot \frac{d |x|+1}{dx} }{ (1+|x|)^2 } \\
    &= \frac{1}{2} \left(\frac{x}{1+|x|} + 1 \right) + x \frac{1}{2} \frac{ (1+|x|) - x \operatorname{sgn}(x) }{ (1+|x|)^2 } \\
    &= \frac{1}{2} \left(\frac{x}{1+|x|} + 1 \right) + x \frac{1}{2} \frac{ 1+|x| - |x| }{ (1+|x|)^2 } \\
    &= \frac{1}{2} \left(\frac{x}{1+|x|} + 1 \right) + x \frac{1}{2} \frac{ 1 }{ 1 + 2|x| + |x|^2 } \\
    &= \frac{1}{2} \left(\frac{x}{1+|x|} + 1  +  \frac{ x }{ 1 + 2|x| + |x|^2 } \right)\\
    &= \frac{1}{2} \left(\frac{x (1+|x|)}{1 + 2|x| + |x|^2} + \frac{ 1 + 2|x| + |x|^2 }{ 1 + 2|x| + |x|^2 }  +  \frac{ x }{ 1 + 2|x| + |x|^2 } \right)\\
    &= \frac{1}{2} \left(\frac{2x + 2|x| + x|x| + |x|^2 + 1}{1 + 2|x| + |x|^2} \right)\\
    &= \frac{1}{2} \left(\frac{2(x + |x|) + |x|(x + |x|) + 1}{1 + 2|x| + |x|^2} \right)\\
    &\geq \frac{1}{2} \left(\frac{1}{1 + 2|x| + |x|^2} \right)
      \qquad \mbox{(because $x+|x| \ge 0$)}\\
    &> 0
\end{align}
$\max_{f_\mathcal{R}}$ is analogous.
\end{proof}
\end{theorem}
\end{widepar}

\newpage

\begin{widepar}
\section{Monotonicity of the Cauchy Relaxed Min and Max}
\begin{theorem}
    $\min_{f_\mathcal{C}}$ and $\max_{f_\mathcal{C}}$ are monotonic functions with the sigmoid function $f_\mathcal{C}$.
\begin{proof}
W.l.o.g., we assume $a_i = x$ and $a_j = 0$.
\begin{align}
    \softmin_{f_\mathcal{C}}(x, 0)
    = x \cdot f_\mathcal{C}(-x) = x \cdot \left( \frac{1}{\pi} \arctan (-\beta x) +\frac{1}{2} \right)
\end{align}
To show monotonicity, we consider its derivative.
\begin{align}
\frac{\partial}{\partial x} \softmin_{f_{\cal C}}(0,x)
& = \frac{\partial}{\partial x} (f_{\cal C}(-x) \cdot x)
~~=~~ x \cdot \frac{\partial}{\partial x} f_{\cal C}(-x)
      + f_{\cal C}(-x) \cdot \frac{\partial}{\partial x} x \notag \\
    & = x \cdot \frac{\partial}{\partial x}
      \left( \frac{1}{\pi} \arctan (-\beta x) +\frac{1}{2} \right)
      +\frac{1}{\pi} \arctan(-\beta x) +\frac{1}{2} \notag \\
    & = x \cdot \frac{1}{\pi} \frac{-\beta}{1+(\beta x)^2}
      -\frac{1}{\pi} \arctan(\beta x) +\frac{1}{2} \notag \\
    & = \frac{1}{2} -\frac{1}{\pi}\arctan(\beta x)
      -\frac{1}{\pi} \frac{\beta x}{1+(\beta x)^2} \notag \\
    &= \frac{1}{2} - \frac{1}{\pi} \arctan (z) - \frac{1}{\pi} \frac{z}{1 + z^2} \quad (\text{with }z=\beta x)
\end{align}
To reason about the derivative, we also consider the second derivative:
\begin{align}
    \lim_{z\to \infty} \frac{1}{2} - \frac{1}{\pi} \arctan (z) - \frac{1}{\pi} \frac{z}{1 + z^2}
    &= \frac{1}{2} - \lim_{z\to \infty} \frac{1}{\pi} \arctan (z) - \lim_{z\to \infty} \frac{1}{\pi} \frac{z}{1 + z^2}\notag \\
    &= \frac{1}{2} - \frac{1}{\pi}\frac{\pi}{2} - 0 = 0 \label{sm:proof:cauchy:derivative-cauchy-min-greater-0-converges}
\end{align}
For $z\in[0,\infty)$: The derivative of $\softmin_{f_\mathcal{C}} (0, x) $ converges to $0$ for $z\to \infty$.
\begin{align}
    \lim_{z\to -\infty} \frac{1}{2} - \frac{1}{\pi} \arctan (z) - \frac{1}{\pi} \frac{z}{1 + z^2}
    &= \frac{1}{2} - \lim_{z\to -\infty} \frac{1}{\pi} \arctan (z) - \lim_{z\to -\infty} \frac{1}{\pi} \frac{z}{1 + z^2}\notag \\
    &= \frac{1}{2} - \frac{1}{\pi}\frac{-\pi}{2} - 0 = \frac{1}{2} + \frac{1}{\pi}\frac{\pi}{2} - 0 = 1 \label{sm:proof:cauchy:derivative-cauchy-min-smaller-0-converges}
\end{align}
For $z\in(-\infty,0]$: The derivative of $\softmin_{f_\mathcal{C}} (0, x) $ converges to $1$ for $z\to -\infty$.

\begin{align}
    \frac{\partial}{\partial z} \frac{1}{2} - \frac{1}{\pi} \arctan (z) - \frac{1}{\pi} \frac{z}{1 + z^2}
    &= -\frac{2}{\pi ( 1+z^2 )^2} < 0 \label{sm:proof:cauchy:second-derivative-cauchy-min}
\end{align}
The second derivative of $\softmin_{f_\mathcal{C}} (0, x) $ is always negative.

Therefore, the derivative is always in $(0,1)$, and therefore always positive. 
Thus, $\softmin_{f_\mathcal{C}} (0, x)$ is strictly monotonic.
$\max_{f_\mathcal{C}}$ is analogous.
\end{proof}
\end{theorem}
\end{widepar}

\hyphenation{now-ro-uze-zah-rai}

\emergencystretch=1em

\printbibliography[heading=bibintoc, title=Bibliography]

@string{aaai = {AAAI Conference on Artificial Intelligence}}

@STRING{aistats = {International Conference on Artificial Intelligence and Statistics (AISTATS)}}

@string{cvpr = {Proc.~International Conference on Computer Vision and Pattern Recognition (CVPR)}}

@STRING{corr = {Computing Research Repository (CoRR) in arXiv}}

@string{eccv = {Proc.~European Conference on Computer Vision (ECCV)}}

@string{eccvw = {Proc.~European Conference on Computer Vision Workshops (ECCVW)}}

@string{iccv = {Proc.~International Conference on Computer Vision (ICCV)}}

@STRING{iclr = {International Conference on Learning Representations (ICLR)}}

@STRING{iclrw = {Workshops at the International Conference on Learning Representations}}

@STRING{icml = {Proc.~Machine Learning Research (PMLR), International Conference on Machine Learning (ICML)}}

@STRING{jmlr = {Journal of Machine Learning Research (JMLR)}}

@string{nat = {Nature}}

@string{natc = {Nature Communications}}

@string{nips = {Advances in Neural Information Processing Systems (NeurIPS)}}

@string{open = {OpenReview preprint}}

@string{siam = {SIAM Journal on Applied Mathematics}}

@string{tog = {ACM Transactions on Graphics}}

@string{togsg = {ACM Transactions on Graphics (Proc.~SIGGRAPH)}}

@string{togsgasia = {ACM Transactions on Graphics (Proc.~SIGGRAPH Asia)}}

@STRING{wacv = {IEEE Winter Conference on Applications of Computer Vision (WACV)}}

@inproceedings{petersen2018towards,
  title={Towards Formula Translation using Recursive Neural Networks},
  author={Petersen, Felix and Schubotz, Moritz and Gipp, Bela},
  booktitle={Proceedings of the 11th Conference on Intelligent Computer Mathematics (CICM), Work-in-Progress Paper Track},
  year={2018}
}

@article{petersen2019pix2vex,
  title={{Pix2Vex: Image-to-Geometry Reconstruction using a Smooth Differentiable Renderer}},
  author={Petersen, Felix and Bermano, Amit H and Deussen, Oliver and Cohen-Or, Daniel},
  journal=corr,
  year={2019}
}

@inproceedings{petersen2019algonet,
  title={{AlgoNet: $C^\infty$ Smooth Algorithmic Neural Networks}},
  author={Petersen, Felix and Borgelt, Christian and Deussen, Oliver},
  booktitle=corr,
  year={2019}
}

@inproceedings{petersen2019smooth,
  title={{$C^\infty$ Smooth Algorithmic Neural Networks for Solving Inverse Problems}},
  author={Petersen, Felix and Borgelt, Christian and Deussen, Oliver},
  booktitle={NeurIPS Deep Inverse Workshop},
  year={2019}
}

@inproceedings{petersen2021learning,
  title={{Learning with Algorithmic Supervision via Continuous Relaxations}},
  author={Petersen, Felix and Borgelt, Christian and Kuehne, Hilde and Deussen, Oliver},
  booktitle=nips,
  year={2021}
}

@inproceedings{petersen2021diffsort,
  title={{Differentiable Sorting Networks for Scalable Sorting and Ranking Supervision}},
  author={Petersen, Felix and Borgelt, Christian and Kuehne, Hilde and Deussen, Oliver},
  booktitle=icml,
  year={2021}
}

@inproceedings{petersen2021post,
  title={{Post-processing for Individual Fairness}},
  author={Petersen, Felix and Mukherjee, Debarghya and Sun, Yuekai and Yurochkin, Mikhail},
  booktitle=nips,
  year={2021}
}

@inproceedings{petersen2022style,
  title={Style Agnostic 3D Reconstruction via Adversarial Style Transfer},
  author={Petersen, Felix and Goldluecke, Bastian and Deussen, Oliver and Kuehne, Hilde},
  booktitle=wacv,
  year={2022}
}

@inproceedings{petersen2022monotonic,
  title={{Monotonic Differentiable Sorting Networks}},
  author={Petersen, Felix and Borgelt, Christian and Kuehne, Hilde and Deussen, Oliver},
  booktitle=iclr,
  year={2022}
}

@inproceedings{petersen2022gendr,
  title={{GenDR: A Generalized Differentiable Renderer}},
  author={Petersen, Felix and Goldluecke, Bastian and Borgelt, Christian and Deussen, Oliver},
  booktitle=cvpr,
  year={2022}
}

@inproceedings{petersen2022topk,
  title={{Differentiable Top-k Classification Learning}},
  author={Petersen, Felix and Borgelt, Christian and Kuehne, Hilde and Deussen, Oliver},
  booktitle=icml,
  year={2022}
}

@inproceedings{petersen2022difflogic,
  title={{Deep Differentiable Logic Gate Networks}},
  author={Petersen, Felix and Borgelt, Christian and Kuehne, Hilde and Deussen, Oliver},
  booktitle={under review.},
  year={2022}
}

@inproceedings{petersen2022newton,
  title={{Newton Losses: Efficiently Including Second-Order Information into Gradient Descent}},
  author={Petersen, Felix and Sutter, Tobias and Borgelt, Christian and Kuehne, Hilde and Deussen, Oliver},
  booktitle={under review.},
  year={2022}
}

@inproceedings{petersen2022propagating,
  title={{Propagating Distributions through Neural Networks}},
  author={Petersen, Felix and Borgelt, Christian and Yurochkin, Mikhail and Kuehne, Hilde and Deussen, Oliver},
  booktitle={under review.},
  year={2022}
}

@inproceedings{petersen2022ape,
  title={{APE-VAE: Training Variational Auto-Encoders with Approximate Evidence}},
  author={Petersen, Felix and Mukherjee, Debarghya and Borgelt, Christian and Kuehne, Hilde and Deussen, Oliver},
  booktitle={under review.},
  year={2022}
}

@inproceedings{petersen2022isaac,
  title={{ISAAC Newton: Input-based Approximate Curvature for Newton's Method}},
  author={Petersen, Felix and Sutter, Tobias and Borgelt, Christian and Dongsung, Huh and Kuehne, Hilde and Sun, Yuekai and Deussen, Oliver},
  booktitle={under review.},
  year={2022}
}

@inproceedings{mukherjee2022domain,
  title={{Domain Adaptation meets Individual Fairness. And they get along.}},
  author={Mukherjee, Debarghya and Petersen, Felix and Yurochkin, Mikhail and Sun, Yuekai},
  booktitle=corr,
  year={2022}
}

@inproceedings{petersen2022distributional,
  title={{Distributional Quantization}},
  author={Petersen, Felix and Sutter, Tobias},
  booktitle={under review.},
  year={2022}
}

@inproceedings{shvetsova2022differentiable,
  title={{Differentiable K-Nearest Neighbor Sorting for Self-supervised Learning}},
  author={Shvetsova, Nina and Petersen, Felix and Feris, Rogerio and Kuehne, Hilde},
  booktitle={under review.},
  year={2022}
}

@inproceedings{petersen2022neural,
  title={Neural Machine Translation for Mathematical Formulae},
  author={Petersen, Felix and Schubotz, Moritz and Greiner-Petter, Andr\'e and Gipp, Bela},
  booktitle={under review.},
  year={2022}
}

@inproceedings{abadi2016tensorflow,
  title={TensorFlow: A system for Large-scale Machine Learning},
  author={Martin Abadi and Paul Barham and Jianmin Chen and Zhifeng Chen and Andy Davis and Jeffrey Dean
          and Matthieu Devin and Sanjay Ghemawat and Geoffrey Irving and Michael Isard and Manjunath Kudlur
          and Josh Levenberg and Rajat Monga and Sherry Moore and Derek G. Murray and Benoit Steiner and Paul Tucker
          and Vijay Vasudevan and Pete Warden and Martin Wicke and Yuan Yu and Xiaoqiang Zheng},
  year={2016},
  booktitle={12th USENIX Symposium on Operating Systems Design and Implementation (OSDI 16)}
}

@article{abernethy2016perturbation,
  title={Perturbation techniques in online learning and optimization},
  author={Abernethy, Jacob and Lee, Chansoo and Tewari, Ambuj},
  journal={Perturbations, Optimization, and Statistics},
  year={2016}
}

@article{adams2011ranking,
  title={Ranking via Sinkhorn Propagation},
  author={Adams, Ryan Prescott and Zemel, Richard S},
  journal=corr,
  year={2011}
}

@article{agrawal2017second,
  author={Agarwal, Naman and Bullins, Brian and Hazan, Elad},
  title={{Second-Order Stochastic Optimization for Machine Learning in Linear Time}},
  journal=jmlr,
  year={2017}
}

@article{agrawal2019differentiable,
  title={{Differentiable Convex Optimization Layers}},
  author={Agrawal, Akshay and Amos, Brandon and Barratt, Shane and Boyd, Stephen and Diamond, Steven and Kolter, J Zico},
  journal=nips,
  year={2019}
}

@article{ajani2021overview,
  title={An overview of machine learning within embedded and mobile devices--optimizations and applications},
  author={Ajani, Taiwo Samuel and Imoize, Agbotiname Lucky and Atayero, Aderemi A},
  journal={Sensors},
  volume={21},
  number={13},
  year={2021},
  publisher={Multidisciplinary Digital Publishing Institute}
}

@inproceedings{ajtai1983sorting,
  title={An 0(n Log n) Sorting Network},
  author={Ajtai, M. and Koml\'{o}s, J. and Szemer\'{e}di, E.},
  booktitle={Proceedings of the Fifteenth Annual ACM Symposium on Theory of Computing},
  year={1983}
}

@inproceedings{amos2017optnet,
  title={Optnet: Differentiable optimization as a layer in neural networks},
  author={Amos, Brandon and Kolter, J Zico},
  booktitle=icml,
  year={2017}
}

@inproceedings{asai2018classical,
  title={Classical planning in deep latent space: Bridging the subsymbolic-symbolic boundary},
  author={Asai, Masataro and Fukunaga, Alex},
  booktitle=aaai,
  year={2018}
}

@article{astrachan2003bubble,
  title={Bubble Sort: an Archaeological Algorithmic Analysis},
  author={Astrachan, Owen},
  journal={ACM Sigcse Bulletin},
  volume={35},
  number={1},
  pages={1--5},
  year={2003},
  publisher={ACM New York, NY, USA}
}

@book{baddar2012designing,
  title={Designing sorting networks: A new paradigm},
  author={Baddar, Sherenaz W Al-Haj and Batcher, Kenneth E},
  publisher={Springer Science \& Business Media},
  year={2012}
}

@inproceedings{bahdanau2015neural,
  title={Neural Machine Translation by jointly Learning to Align and Translate},
  author={Bahdanau, Dzmitry and Cho, Kyunghyun and Bengio, Yoshua},
  booktitle=iclr,
  year={2015}
}

@inproceedings{bahdanau2016task,
  title={Task loss estimation for sequence prediction},
  author={Bahdanau, Dzmitry and Serdyuk, Dmitriy and Brakel, Phil{\'e}mon and Ke, Nan Rosemary and Chorowski, Jan and Courville, Aaron and Bengio, Yoshua},
  booktitle=iclrw,
  year={2016}
}

@article{bangaru2021systematically,
  title={Systematically differentiating parametric discontinuities},
  author={Bangaru, Sai Praveen and Michel, Jesse and Mu, Kevin and Bernstein, Gilbert and Li, Tzu-Mao and Ragan-Kelley, Jonathan},
  journal=tog,
  year={2021}
}

@inproceedings{batcher1968sorting,
  title={Sorting networks and their applications},
  author={Batcher, Kenneth E},
  booktitle={Proc.\ AFIPS Spring Joint Computing Conference (Atlantic City, NJ)},
  pages={307--314},
  year={1968}
}

@article{bellman1958routing,
  title={On a routing problem},
  author={Bellman, Richard},
  journal={Quarterly of Applied Mathematics},
  volume={16},
  number={1},
  pages={87--90},
  year={1958}
}

@inproceedings{berrada2018smooth,
  title={Smooth loss functions for deep top-k classification},
  author={Berrada, Leonard and Zisserman, Andrew and Kumar, M Pawan},
  booktitle=iclr,
  year={2018}
}

@article{besold2017neural,
  title={Neural-symbolic learning and reasoning: A survey and interpretation},
  author={Besold, Tarek R and Garcez, Artur d'Avila and Bader, Sebastian and Bowman, Howard and Domingos, Pedro and Hitzler, Pascal and K{\"u}hnberger, Kai-Uwe and Lamb, Luis C and Lowd, Daniel and Lima, Priscila Machado Vieira and others},
  journal=corr,
  year={2017}
}

@book{bertsekas1999nonlinear,
  author = {Bertsekas, D.P.},
  publisher = {Athena Scientific},
  title = {Nonlinear Programming},
  year = {1999}
}

@inproceedings{berthet2020learning,
  title={{Learning with Differentiable Perturbed Optimizers}},
  author={Berthet, Quentin and Blondel, Mathieu and Teboul, Olivier and Cuturi, Marco and Vert, Jean-Philippe and Bach, Francis},
  booktitle=nips,
  year={2020}
}

@article{bidlo2019evolutionary,
  title={Evolutionary Development of Growing Generic Sorting Networks by Means of Rewriting Systems},
  author={Bidlo, Michal and Dobe{\v{s}}, Michal},
  journal={IEEE Transactions on Evolutionary Computation},
  year={2019}
}

@inproceedings{blalock2020state,
  title={What is the State of Neural Network Pruning?},
  author={Blalock, Davis and Ortiz, Jose Javier Gonzalez and Frankle, Jonathan and Guttag, John},
  booktitle={Proceedings of the 3rd MLSys Conference, Austin, TX, USA},
  year={2020}
}

@article{blondel2020learning,
  title={Learning with {F}enchel-{Y}oung Losses},
  author={Blondel, Mathieu and Martins, Andr{\'e} FT and Niculae, Vlad},
  journal=jmlr,
  year={2020}
}

@inproceedings{blondel2020fast,
  title={{Fast Differentiable Sorting and Ranking}},
  author={Blondel, Mathieu and Teboul, Olivier and Berthet, Quentin and Djolonga, Josip},
  booktitle=icml,
  year={2020}
}

@inproceedings{blondel2021differentiable,
  title={Differentiable Divergences Between Time Series},
  author={Blondel, Mathieu and Mensch, Arthur and Vert, Jean-Philippe},
  booktitle=aistats,
  year={2021}
}

@article{blondel2021efficient,
  title={Efficient and Modular Implicit Differentiation},
  author={Blondel, Mathieu and Berthet, Quentin and Cuturi, Marco and Frostig, Roy and Hoyer, Stephan and Llinares-L{\'o}pez, Felipe and Pedregosa, Fabian and Vert, Jean-Philippe},
  journal=corr,
  year={2021}
}

@inproceedings{bogo2016keep,
  title={Keep it SMPL: Automatic Estimation of 3D Human Pose and Shape from a Single Image},
  author={Bogo, Federica and Kanazawa, Angjoo and Lassner, Christoph and Gehler, Peter and Romero, Javier and Black, Michael J},
  booktitle=eccv,
  year={2016}
}

@inproceedings{bosnjak2017programming,
  title={Programming with a Differentiable Forth Interpreter},
  author={Matko Bo{\v{s}}njak and Tim Rockt{\"a}schel and Jason Naradowsky and Sebastian Riedel},
  booktitle=icml,
  year={2017}
}

@article{boyd2011distributed,
  author = {Boyd, Stephen and Parikh, Neal and Chu, Eric and Peleato, Borja and Eckstein, Jonathan},
  title = {Distributed Optimization and Statistical Learning via the Alternating Direction Method of Multipliers},
  year = {2011},
  publisher = {Now Publishers Inc.},
  volume = {3},
  number = {1},
  journal = {Foundations and Trends in Machine Learning},
}

@software{bradbury2018jax,
  author = {James Bradbury and Roy Frostig and Peter Hawkins and Matthew James Johnson and Chris Leary and Dougal Maclaurin and George Necula and Adam Paszke and Jake Vander{P}las and Skye Wanderman-{M}ilne and Qiao Zhang},
  title = {{JAX}: composable transformations of {P}ython+{N}um{P}y programs},
  url = {http://github.com/google/jax},
  year = {2018},
}

@article{branco2019machine,
  title={Machine learning in resource-scarce embedded systems, FPGAs, and end-devices: A survey},
  author={Branco, S{\'e}rgio and Ferreira, Andr{\'e} G and Cabral, Jorge},
  journal={Electronics},
  volume={8},
  number={11},
  pages={1289},
  year={2019},
  publisher={Multidisciplinary Digital Publishing Institute}
}

@article{brudermueller2020making,
  title={Making Logic Learnable With Neural Networks},
  author={Brudermueller, Tobias and Shung, Dennis L and Stanley, Adrian J and Stegmaier, Johannes and Krishnaswamy, Smita},
  journal=corr,
  year={2020}
}

@inproceedings{burges2005ranknet,
  title={Learning to Rrank using Gradient Descent},
  author={Burges, Chris and Shaked, Tal and Renshaw, Erin and Lazier, Ari and Deeds, Matt and Hamilton, Nicole and Hullender, Greg},
  booktitle=icml,
  year={2005}
}

@inproceedings{burges2007lambdarank,
  title={Learning to rank with nonsmooth cost functions},
  author={Burges, Christopher J and Ragno, Robert and Le, Quoc V},
  booktitle=nips,
  year={2007}
}

@article{carr2021self,
  title={Self-supervised learning of audio representations from permutations with differentiable ranking},
  author={Carr, Andrew N and Berthet, Quentin and Blondel, Mathieu and Teboul, Olivier and Zeghidour, Neil},
  journal={IEEE Signal Processing Letters},
  year={2021}
}

@inproceedings{ceterchi2008spiking,
  title={Spiking Neural P Systems – A Natural Model for Sorting Networks},
  author={Rodica Ceterchi and Alexandru I. Tomescu},
  booktitle={Proc. of the Sixth Brainstorming Week on Membrane Computing},
  year={2008}
}

@article{chang2015shapenet,
  title={{ShapeNet}: An Information-Rich {3D} Model Repository},
  author={Chang, Angel X. and Funkhouser, Thomas and Guibas, Leonidas and Hanrahan, Pat and Huang, Qixing
          and Li, Zimo and Savarese, Silvio and Savva, Manolis and Song, Shuran and Su, Hao and Xiao, Jianxiong
          and Yi, Li and Yu, Fisher},
  journal=corr,
  year={2015},
  arxivId={1512.03012}
}

@inproceedings{charpentier2022differentiable,
  title={Differentiable DAG Sampling},
  author={Charpentier, Bertrand and Kibler, Simon and G{\"u}nnemann, Stephan},
  booktitle=iclr,
  year={2022}
}

@inproceedings{chapelle2011yahoo,
  title={Yahoo! Learning to Rank Challenge Overview},
  author={Chapelle, Olivier and Chang, Yi},
  booktitle={Proceedings of the Learning to Rank Challenge},
  year={2011},
  organization={PMLR}
}

@inproceedings{chatterjee2018learning,
  title={Learning and Memorization},
  author={Chatterjee, Satrajit},
  booktitle=icml,
  year={2018}
}

@inproceedings{chaudhuri2010smooth,
  title={Smooth Interpretation},
  author={Chaudhuri, Swarat and Solar-Lezama, Armando},
  booktitle={Proceedings of the 31st ACM SIGPLAN Conference on Programming Language Design and Implementation},
  year={2010}
}

@inproceedings{chaudhuri2011smoothing,
  title={Smoothing a Program Soundly and Robustly},
  author={Chaudhuri, Swarat and Solar-Lezama, Armando},
  booktitle={Proceedings of the 23rd International Conference on Computer Aided Verification},
  year={2011}
}

@inproceedings{chen2019learning,
  title={Learning to Predict {3D} Objects with an Interpolation-based Differentiable Renderer},
  author={Chen, Wenzheng and Gao, Jun and Ling, Huan and Smith, Edward J. and Lehtinen, Jaakko and Jacobson, Alec and Fidler, Sanja},
  booktitle=nips,
  year={2019}
}

@article{chen2020learning,
  title={Learning Symbolic Expressions via Gumbel-Max Equation Learner Network},
  author={Chen, Gang},
  journal=corr,
  year={2020}
}

@inproceedings{chen2022semi,
  title={{Semi-Discrete Normalizing Flows through Differentiable Voronoi Tessellation}},
  author={Chen, Ricky TQ and Amos, Brandon and Nickel, Maximilian},
  booktitle={ICLR Workshop on Deep Generative Models for Highly Structured Data},
  year={2022}
}

@article{cho2021differentiable,
  title={Differentiable Spline Approximations},
  author={Cho, Minsu and Balu, Aditya and Joshi, Ameya and Deva Prasad, Anjana and Khara, Biswajit and Sarkar, Soumik and Ganapathysubramanian, Baskar and Krishnamurthy, Adarsh and Hegde, Chinmay},
  journal=nips,
  year={2021}
}

@article{choi2018pact,
  title={PACT: Parameterized Clipping Activation for Quantized Neural Networks},
  author={Choi, Jungwook and Wang, Zhuo and Venkataramani, Swagath and Chuang, Pierce I-Jen and Srinivasan, Vijayalakshmi and Gopalakrishnan, Kailash},
  journal=corr,
  year={2018}
}

@article{chzhen2021set,
  title={Set-valued Classification--Overview via a Unified Framework},
  author={Chzhen, Evgenii and Denis, Christophe and Hebiri, Mohamed and Lorieul, Titouan},
  journal=corr,
  year={2021}
}

@article{clark1989cn2,
  title={The CN2 Induction Algorithm},
  author={Clark, Peter and Niblett, Tim},
  journal={Machine Learning},
  year={1989}
}

@article{cody1969rational,
  title={Rational {C}hebyshev approximations for the error function},
  author={Cody, William J},
  journal={Mathematics of Computation},
  volume={23},
  number={107},
  pages={631--637},
  year={1969}
}

@article{cohen2017emnist,
  author={Gregory Cohen and Saeed Afshar and Jonathan Tapson and Andr{\'{e}} van Schaik},
  title={{EMNIST:} an extension of {MNIST} to handwritten letters},
  journal=corr,
  year={2017}
}

@book{coles2001introduction,
  title={An Introduction to Statistical Modeling of Extreme Values},
  author={Coles, Stuart},
  series={Springer Series in Statistics},
  year={2001}
}

@inproceedings{collobert2019fully,
    title={A Fully Differentiable Beam Search Decoder},
    author={Ronan Collobert and Awni Hannun and Gabriel Synnaeve},
    booktitle=icml,
    year={2019}
}

@inproceedings{cordonnier2021differentiable,
  title={Differentiable Patch Selection for Image Recognition},
  author={Cordonnier, Jean-Baptiste and Mahendran, Aravindh and Dosovitskiy, Alexey and Weissenborn, Dirk and Uszkoreit, Jakob and Unterthiner, Thomas},
  booktitle=cvpr,
  year={2021}
}

@inproceedings{corenflos2021differentiable,
  title={Differentiable Particle Filtering via Entropy-Regularized Optimal Transport},
  author={Corenflos, Adrien and Thornton, James and Deligiannidis, George and Doucet, Arnaud},
  booktitle=icml,
  year={2021}
}

@incollection{coros2021differentiable,
  title={Differentiable Simulation},
  author={Coros, Stelian and Macklin, Miles and Thomaszewski, Bernhard and Th{\"u}rey, Nils},
  booktitle={ACM SIGGRAPH Asia 2021 Courses},
  year={2021}
}

@inproceedings{corro2019differentiable,
  title={Differentiable Perturb-and-Parse: Semi-Supervised Parsing with a Structured Variational Autoencoder},
  author={Corro, Caio and Titov, Ivan},
  booktitle=iclr,
  year={2019}
}

@inproceedings{cuturi2013sinkhorn,
  title={Sinkhorn Distances: Lightspeed Computation of Optimal Transport},
  author={Cuturi, Marco},
  booktitle=nips,
  year={2013}
}

@inproceedings{cuturi2017soft,
  title={{Soft-DTW: A Differentiable Loss Function for Time-Series}},
  author={Cuturi, Marco and Blondel, Mathieu},
  booktitle=icml,
  year={2017},
}

@inproceedings{cuturi2019differentiable,
  title={Differentiable Ranking and Sorting using Optimal Transport},
  author={Cuturi, Marco and Teboul, Olivier and Vert, Jean-Philippe},
  booktitle=nips,
  year={2019}
}

@article{dai2021coatnet,
  title={CoAtNet: Marrying Convolution and Attention for All Data Sizes},
  author={Dai, Zihang and Liu, Hanxiao and Le, Quoc V and Tan, Mingxing},
  journal=corr,
  year={2021}
}

@inproceedings{dasgupta2020improving,
  title={Improving Local Identifiability in Probabilistic Box Embeddings},
  author={Dasgupta, Shib Sankar and Boratko, Michael and Zhang, Dongxu and Vilnis, Luke and Li, Xiang Lorraine and McCallum, Andrew},
  booktitle=nips,
  year={2020}
}

@article{de2018end,
  title={End-to-End Differentiable Physics for Learning and Control},
  author={de Avila Belbute-Peres, Filipe and Smith, Kevin and Allen, Kelsey and Tenenbaum, Josh and Kolter, J Zico},
  journal=nips,
  year={2018}
}

@article{degrave2019differentiable,
  title={{A Differentiable Physics Engine for Deep Learning in Robotics}},
  author={Degrave, Jonas and Hermans, Michiel and Dambre, Joni and others},
  journal={Frontiers in Neurorobotics},
  year={2019},
  publisher={Frontiers}
}

@inproceedings{deng2009imagenet,
  title={ImageNet: A Large-Scale Hierarchical Image Database},
  author={Deng, Jia and Dong, Wei and Socher, Richard and Li, Li-Jia and Li, Kai and Fei-Fei, Li},
  booktitle=cvpr,
  year={2009}
}

@inproceedings{deng2020cvxnet,
  title={Cvxnet: Learnable convex decomposition},
  author={Deng, Boyang and Genova, Kyle and Yazdani, Soroosh and Bouaziz, Sofien and Hinton, Geoffrey and Tagliasacchi, Andrea},
  booktitle=cvpr,
  year={2020}
}

@article{dijkstra1959note,
  title={A note on two problems in connexion with graphs},
  author={Dijkstra, Edsger W and others},
  journal={Numerische Mathematik},
  volume={1},
  number={1},
  pages={269--271},
  year={1959}
}

@article{djolonga2017differentiable,
  title={{Differentiable Learning of Submodular Models}},
  author={Djolonga, Josip and Krause, Andreas},
  journal=nips,
  year={2017}
}

@article{domke2010implicit,
  title={Implicit differentiation by perturbation},
  author={Domke, Justin},
  journal=nips,
  year={2010}
}

@inproceedings{dosovitskiy2021image,
  title={An image is worth 16x16 words: Transformers for image recognition at scale},
  author={Dosovitskiy, Alexey and Beyer, Lucas and Kolesnikov, Alexander and Weissenborn, Dirk and Zhai, Xiaohua
          and Unterthiner, Thomas and Dehghani, Mostafa and Minderer, Matthias and Heigold, Georg and Gelly, Sylvain
          and others},
  booktitle=iclr,
  year={2021}
}

@inproceedings{engel2020ddsp,
  title={DDSP: Differentiable digital signal processing},
  author={Engel, Jesse and Hantrakul, Lamtharn and Gu, Chenjie and Roberts, Adam},
  booktitle=iclr,
  year={2020}
}

@inproceedings{fan2017learning,
  title={Learning with Average Top-k Loss},
  author={Fan, Yanbo and Lyu, Siwei and Ying, Yiming and Hu, Bao-Gang},
  booktitle=nips,
  year={2017}
}

@inproceedings{feser2017neural,
  title={Neural Functional Programming},
  author={Feser, John K and Brockschmidt, Marc and Gaunt, Alexander L and Tarlow, Daniel},
  booktitle=iclrw,
  year={2017}
}

@inproceedings{frantar2021mfac,
  title={{M-FAC}: Efficient Matrix-Free Approximations of Second-Order Information}, 
  author={Elias Frantar and Eldar Kurtic and Dan Alistarh},
  booktitle=nips,
  year={2021}
}

@article{frey1999variational,
  title={Variational Learning in Nonlinear Gaussian Belief Networks},
  author={Brendan J. Frey and Geoffrey E. Hinton},
  journal={Neural Computation},
  volume={11},
  issue={1},
  pages={193--213},
  year={1999}
}

@inproceedings{fu2022differentiable,
  title={Differentiable Scaffolding Tree for Molecular Optimization},
  author={Fu, Tianfan and Gao, Wenhao and Xiao, Cao and Yasonik, Jacob and Coley, Connor W and Sun, Jimeng},
  booktitle=iclr,
  year={2022}
}

@article{fuegi2003lovelace,
  title={Lovelace \& Babbage and the creation of the 1843 'notes'},
  author={Fuegi, John and Francis, Jo},
  journal={IEEE Annals of the History of Computing},
  volume={25},
  number={4},
  pages={16--26},
  year={2003}
}

@article{fukushima1980self,
  title={Neocognitron: A self-organizing neural network model for a mechanism of pattern recognition unaffected by shift in position},
  author={Fukushima, Kunihiko},
  journal={Biological Cybernetics},
  volume={36},
  pages={193--202},
  year={1980}
}

@inproceedings{gaier2019weight,
  title={Weight Agnostic Neural Networks},
  author={Gaier, Adam and Ha, David},
  booktitle=nips,
  year={2019}
}

@inproceedings{gast2018lightweight,
  title={Lightweight Probabilistic Deep Networks},
  author={Jochen Gast and Stefan Roth},
  booktitle=cvpr,
  year={2018}
}

@inproceedings{gaunt2017differentiable,
  title={Differentiable Programs with Neural Libraries},
  author={Alexander L. Gaunt and Marc Brockschmidt and Nate Kushman and Daniel Tarlow},
  booktitle=icml,
  year={2017}
}

@article{goodfellow2013svhn,
  title={Multi-digit number recognition from street view imagery using deep convolutional neural networks},
  author={Goodfellow, Ian J and Bulatov, Yaroslav and Ibarz, Julian and Arnoud, Sacha and Shet, Vinay},
  journal=corr,
  year={2013}
}

@book{goodfellow2016deep,
  title={Deep Learning},
  author={Ian J. Goodfellow and Yoshua Bengio and Aaron Courville},
  publisher={MIT Press},
  year={2016}
}

@inproceedings{ghasemzadeh2018rebnet,
  title={ReBNet: Residual Binarized Neural Network},
  author={Mohammad Ghasemzadeh, Mohammad Samragh, Farinaz Koushanfar},
  booktitle={Proceedings of the 26th IEEE International Symposium on Field-Programmable Custom Computing Machines (FCCM)},
  year={2018}
}

@article{glynn1990likelihood,
  title={Likelihood Ratio Gradient Estimation for Stochastic Systems},
  author={Glynn, Peter W},
  journal={Communications of the ACM},
  volume={33},
  number={10},
  pages={75--84},
  year={1990}
}

@inproceedings{govindaraju2006gputerasort,
  title={GPUTeraSort: High Performance Graphics Co-Processor Sorting for Large Database Management},
  author={Govindaraju, Naga K. and Gray, Jim and Kumar, Ritesh and Manocha, Dinesh},
  booktitle={SIGMOD Conference},
  year={2006}
}

@inproceedings{goyal2018continuous,
  title={A continuous relaxation of beam search for end-to-end training of neural sequence models},
  author={Goyal, Kartik and Neubig, Graham and Dyer, Chris and Berg-Kirkpatrick, Taylor},
  booktitle=aaai,
  year={2018}
}

@article{gowanlock2019hybrid,
  title={A hybrid CPU GPU approach for optimizing sorting throughput},
  author={Gowanlock, Michael and Karsin, Ben},
  journal={Parallel Computing},
  volume={85},
  year={2019}
}

@inproceedings{grathwohl2018backpropagation,
  title={Backpropagation through the Void: Optimizing control variates for black-box gradient estimation},
  author={Grathwohl, Will and Choi, Dami and Wu, Yuhuai and Roeder, Geoffrey and Duvenaud, David},
  booktitle=iclr,
  year={2018}
}

@article{graves2014neural,
  title={Neural Turing Machines},
  author={Graves, Alex and Wayne, Greg and Danihelka, Ivo},
  journal=corr,
  year={2014}
}

@inproceedings{grover2019neuralsort,
  title={{Stochastic Optimization of Sorting Networks via Continuous Relaxations}},
  author={Grover, Aditya and Wang, Eric and Zweig, Aaron and Ermon, Stefano},
  booktitle=iclr,
  year={2019}
}

@inproceedings{gupta2015deep,
  title={Deep Learning with Limited Numerical Precision},
  author={Gupta, Suyog and Agrawal, Ankur and Gopalakrishnan, Kailash and Narayanan, Pritish},
  booktitle=icml,
  year={2015}
}

@article{habermann1972parallel,
  title={Parallel Neighbor-Sort (or the Glory of the Induction Principle)},
  author={Habermann, A Nico},
  publisher={Carnegie Mellon University},
  year={1972}
}

@inproceedings{han2016deep,
  title={Deep Compression: Compressing Deep Neural Networks with Pruning, Trained Quantization and Huffman Coding},
  author={Han, Song and Mao, Huizi and Dally, William J},
  booktitle=iclr,
  year={2016}
}

@inproceedings{he2016deep,
  title={Deep Residual Learning for Image Recognition},
  author={He, Kaiming and Zhang, Xiangyu and Ren, Shaoqing and Sun, Jian},
  booktitle=cvpr,
  year={2016}
}

@inproceedings{henderson2020leveraging,
  title={Leveraging 2D Data to Learn Textured 3D Mesh Generation},
  author={Paul Henderson, Vagia Tsiminaki, Christoph H. Lampert},
  booktitle=cvpr,
  year={2020}
}

@inproceedings{henzler2018escaping,
  title={Escaping Plato's Cave using Adversarial Training: {3D} Shape From Unstructured {2D} Image Collections},
  author={Henzler, Philipp and Mitra, Niloy and Ritschel, Tobias},
  booktitle=iccv,
  year={2019}
}

@article{hoefler2021sparsity,
  title={Sparsity in Deep Learning: Pruning and Growth for Efficient Inference and Training in Neural Networks},
  author={Hoefler, Torsten and Alistarh, Dan and Ben-Nun, Tal and Dryden, Nikoli and Peste, Alexandra},
  journal=corr,
  year={2021}
}

@inproceedings{holl2020phiflow,
  title={phiflow: A Differentiable PDE Solving Framework for Deep Learning via Physical Simulations},
  author={Holl, Philipp and Koltun, Vladlen and Um, Kiwon and Thuerey, Nils},
  booktitle={Workshop on Differentiable Computer Vision, Graphics, and Physics in Machine Learning at NeurIPS 2020},
  year={2020}
}

@inproceedings{holl2020learning,
  title={Learning to Control PDEs with Differentiable Physics},
  author={Holl, Philipp and Koltun, Vladlen and Thuerey, Nils},
  booktitle=iclr,
  year={2020}
}

@inproceedings{hu2019difftaichi,
  title={{DiffTaichi: Differentiable Programming for Physical Simulation}},
  author={Hu, Yuanming and Anderson, Luke and Li, Tzu-Mao and Sun, Qi and Carr, Nathan and Ragan-Kelley, Jonathan and Durand, Fr{\'e}do},
  booktitle=iclr,
  year={2020}
}

@inproceedings{huang2022relational,
  title={{Relational Surrogate Loss Learning}},
  author={Huang, Tao and Li, Zekang and Lu, Hua and Shan, Yong and Yang, Shusheng and Feng, Yang and Wang, Fei and You, Shan and Xu, Chang},
  booktitle=iclr,
  year={2022}
}

@article{hullermeier2008label,
  title={Label ranking by learning pairwise preferences},
  author={H{\"u}llermeier, Eyke and F{\"u}rnkranz, Johannes and Cheng, Weiwei and Brinker, Klaus},
  journal={Artificial Intelligence},
  volume={172},
  number={16-17},
  pages={1897--1916},
  year={2008},
  publisher={Elsevier}
}

@inproceedings{ingraham2018learning,
  title={Learning protein structure with a differentiable simulator},
  author={Ingraham, John and Riesselman, Adam and Sander, Chris and Marks, Debora},
  booktitle=iclr,
  year={2018}
}

@inproceedings{insafutdinov2018unsupervised,
  title={Unsupervised Learning of Shape and Pose with Differentiable Point Clouds},
  author={Insafutdinov, Eldar and Dosovitskiy, Alexey},
  booktitle=nips,
  year={2018}
}

@inproceedings{jang2017categorical,
  title={Categorical Reparameterization with Gumbel-Softmax},
  author={Jang, Eric and Gu, Shixiang and Poole, Ben},
  booktitle=iclr,
  year={2017}
}

@inproceedings{jia2014caffe,
  title={Caffe: Convolutional Architecture for Fast Feature Embedding},
  author={Jia, Yangqing and Shelhamer, Evan and Donahue, Jeff and Karayev, Sergey and Long, Jonathan and Girshick, Ross and Guadarrama, Sergio and Darrell, Trevor},
  booktitle={Proceedings of the 22nd ACM International Conference on Multimedia},
  year={2014}
}

@inproceedings{jia2021scaling,
  title={Scaling up visual and vision-language representation learning with noisy text supervision},
  author={Jia, Chao and Yang, Yinfei and Xia, Ye and Chen, Yi-Ting and Parekh, Zarana and Pham, Hieu and Le, Quoc V and Sung, Yunhsuan and Li, Zhen and Duerig, Tom},
  booktitle=icml,
  year={2021}
}

@inproceedings{jiang2020sdfdiff,
  title={{SDFDiff}: Differentiable Rendering of Signed Distance Fields for {3D} Shape Optimization},
  author={Jiang, Yue and Ji, Dantong and Han, Zhizhong and Zwicker, Matthias},
  booktitle=cvpr,
  year={2020}
}

@inproceedings{jokic2018binaryeye,
  title={Binaryeye: A 20 kfps Streaming Camera System on FPGA with Real-Time On-Device Image Recognition Using Binary Neural Networks},
  author={Jokic, Petar and Emery, Stephane and Benini, Luca},
  booktitle={IEEE 13th International Symposium on Industrial Embedded Systems (SIES)},
  year={2018}
}

@inproceedings{karpinski2015smaller,
  title={Smaller Selection Networks for Cardinality Constraints Encoding},
  author={Micha{\l} Karpi\'nski and Marek Piotr\'ow},
  booktitle={Proc.\ Principles and Practice of Constraint Programming (CP 2015, Cork, Ireland)},
  year={2015}
}

@inproceedings{kato2017neural,
  title={Neural {3D} Mesh Renderer},
  author={Kato, Hiroharu and Ushiku, Yoshitaka and Harada, Tatsuya},
  booktitle=cvpr,
  year={2018}
}

@article{kato2020differentiable,
  title={Differentiable Rendering: A Survey},
  author={Kato, Hiroharu and Beker, Deniz and Morariu, Mihai and Ando, Takahiro and Matsuoka, Toru and Kehl, Wadim and Gaidon, Adrien},
  journal=corr,
  year={2020}
}

@inproceedings{kingma2015adam,
  title={{Adam}: A Method for Stochastic Optimization},
  author={Kingma, Diederik and Ba, Jimmy},
  booktitle=iclr,
  year={2015}
}

@book{kirk2016programming,
  title={{Programming Massively Parallel Processors: A Hands-on Approach}},
  author={Kirk, David B and Wen-Mei, W Hwu},
  publisher={Morgan Kaufmann},
  year={2016}
}

@article{kiwiel2001convergence,
  title={Convergence and Efficiency of Subgradient Methods for Quasiconvex Minimization},
  author={Kiwiel, Krzysztof C},
  journal={Mathematical Programming},
  volume={90},
  year={2001}
}

@article{kleijnen1996optimization,
  title={Optimization and Sensitivity Analysis of Computer Simulation Models by the Score Function Method},
  author={Kleijnen, Jack PC and Rubinstein, Reuven Y},
  journal={European Journal of Operational Research},
  volume={88},
  number={3},
  pages={413--427},
  year={1996},
  publisher={Elsevier}
}

@book{klement2013triangular,
  title={Triangular Norms},
  author={Klement, Erich Peter and Mesiar, Radko and Pap, Endre},
  publisher={Springer Science \& Business Media},
  year={2013}
}

@book{klir1997fuzzy,
  author={George J. Klir and Bo Yuan},
  title={Fuzzy Sets and Fuzzy Logic: Theory and Applications},
  publisher={Prentice Hall},
  year={1997}
}

@article{knight2008sinkhorn,
  title={The Sinkhorn--Knopp algorithm: convergence and applications},
  author={Knight, Philip A},
  journal={SIAM Journal on Matrix Analysis and Applications},
  volume={30},
  number={1},
  pages={261--275},
  year={2008},
  publisher={SIAM}
}

@book{knuth1998sorting,
  title={The Art of Computer Programming, Volume 3: Sorting and Searching (2nd Ed.)},
  author={Knuth, Donald E.},
  publisher={Addison Wesley},
  year={1998}
}

@article{kohavi1996uci,
  title={UCI Machine Learning Repository: Adult Data Set},
  author={Kohavi, Ronny and Becker, Barry},
  url={https://archive.ics.uci.edu/ml/machine-learning-databases/adult},
  year={1996}
}

@inproceedings{kolesnikov2020big,
  title={Big Transfer (BiT): General Visual Representation Learning},
  author={Kolesnikov, Alexander and Beyer, Lucas and Zhai, Xiaohua and Puigcerver, Joan and Yung, Jessica and Gelly, Sylvain and Houlsby, Neil},
  booktitle=eccv,
  year={2020}
}

@article{kong2020rankmax,
  title={Rankmax: An Adaptive Projection Alternative to the Softmax Function},
  author={Kong, Weiwei and Krichene, Walid and Mayoraz, Nicolas and Rendle, Steffen and Zhang, Li},
  journal=nips,
  year={2020}
}

@inproceedings{krizhevsky2012imagenet,
  title={{ImageNet Classification with Deep Convolutional Neural Networks}},
  author={Krizhevsky, Alex and Sutskever, Ilya and Hinton, Geoffrey E},
  booktitle=nips,
  year={2012}
}

@inproceedings{kunstner2019limitations,
  title={Limitations of the empirical {F}isher approximation for natural gradient descent},
  author={Kunstner, Frederik and Balles, Lukas and Hennig, Philipp},
  booktitle=nips,
  year={2019}
}

@article{krizhevsky2009cifar10,
  title={CIFAR-10 (Canadian Institute for Advanced Research)},
  author={Alex Krizhevsky and Vinod Nair and Geoffrey Hinton},
  url={http://www.cs.toronto.edu/~kriz/cifar.html},
  year={2009}
}

@inproceedings{lapin2015topk,
  title={Top-k Multiclass SVM},
  author={Lapin, Maksim and Hein, Matthias and Schiele, Bernt},
  booktitle=nips,
  year={2015}
}

@inproceedings{lapin2016loss,
  title={Loss Functions for Top-k Error: Analysis and Insights},
  author={Lapin, Maksim and Hein, Matthias and Schiele, Bernt},
  booktitle=cvpr,
  year={2016}
}

@incollection{lecun1999object,
  title={{Object Recognition with Gradient-Based Learning}},
  author={LeCun, Yann and Haffner, Patrick and Bottou, L{\'e}on and Bengio, Yoshua},
  booktitle={Shape, contour and grouping in computer vision},
  pages={319--345},
  year={1999},
  publisher={Springer}
}

@article{lecun2010mnist,
  title={MNIST Handwritten Digit Database},
  author={LeCun, Yann and Cortes, Corinna and Burges, CJ},
  url={http://yann.lecun.com/exdb/mnist},
  year={2010}
}

@article{lee2021differentiable,
  title={Differentiable Ranking Metric Using Relaxed Sorting for Top-K Recommendation},
  author={Lee, Hyunsung and Cho, Sangwoo and Jang, Yeongjae and Kim, Jaekwang and Woo, Honguk},
  journal={IEEE Access},
  year={2021}
}

@inproceedings{levenshtein1966binary,
  title={Binary Codes Capable of Correcting Deletions, Insertions, and Reversals},
  author={Levenshtein, Vladimir I},
  booktitle={Soviet physics doklady},
  volume={10},
  number={8},
  pages={707--710},
  year={1966}
}

@article{li2018differentiable,
  title={Differentiable {Monte Carlo} Ray Tracing Through Edge Sampling},
  author={Li, Tzu-Mao and Aittala, Miika and Durand, Fredo and Lehtinen, Jaakko},
  journal=togsgasia,
  year={2018}
}

@article{li2020differentiable,
  title={Differentiable Vector Graphics Rasterization for Editing and Learning},
  author={Li, Tzu-Mao and Luk{\'a}{\v{c}}, Michal and Gharbi, Micha{\"e}l and Ragan-Kelley, Jonathan},
  journal=tog,
  year={2020}
}

@inproceedings{lidec2021differentiable,
  title={Differentiable Rendering with Perturbed Optimizers},
  author={Lidec, Quentin Le and Laptev, Ivan and Schmid, Cordelia and Carpentier, Justin},
  booktitle=nips,
  year={2021}
}

@article{liu2011learning,
  title={Learning to Rank for Information Retrieval},
  author={Liu, Tie-Yan},
  publisher={Springer Science \& Business Media},
  year={2011}
}

@inproceedings{liu2017material,
  title={Material Editing using a Physically based Rendering Network},
  author={Liu, Guilin and Ceylan, Duygu and Yumer, Ersin and Yang, Jimei and Lien, Jyh-Ming},
  booktitle=iccv,
  year={2017}
}

@article{liu2018adversarial,
  title={Adversarial Geometry and Lighting using a Differentiable Renderer},
  author={Liu, Hsueh-Ti Derek and Tao, Michael and Li, Chun-Liang and Nowrouzezahrai, Derek and Jacobson, Alec},
  journal=corr,
  year={2018},
}

@inproceedings{liu2019beyond,
  title={Beyond Pixel Norm-Balls: Parametric Adversaries using an Analytically Differentiable Renderer},
  author={Liu, Hsueh-Ti Derek and Tao, Michael and Li, Chun-Liang and Nowrouzezahrai, Derek and Jacobson, Alec},
  booktitle=iclr,
  year={2019}
}

@inproceedings{liu2019soft,
  title={{Soft Rasterizer: A Differentiable Renderer for Image-based 3D Reasoning}},
  author={Liu, Shichen and Li, Tianye and Chen, Weikai and Li, Hao},
  booktitle=iccv,
  year={2019}
}

@inproceedings{liu2019learning,
  title={Learning to Infer Implicit Surfaces without {3D} Supervision},
  author={Liu, Shichen and Saito, Shunsuke and Chen, Weikai and Li, Hao},
  booktitle=nips,
  year={2019},
}

@inproceedings{liu2022neuray,
  title={Neural Rays for Occlusion-aware Image-based Rendering},
  author={Liu, Yuan and Peng, Sida and Liu, Lingjie and Wang, Qianqian and Wang, Peng and Theobalt, Christian and Zhou, Xiaowei and Wang, Wenping},
  booktitle=cvpr,
  year={2022}
}

@inproceedings{loper2014opendr,
  title={{OpenDR}: An approximate differentiable renderer},
  author={Loper, Matthew M. and Black, Michael J.},
  booktitle=eccv,
  year={2014}
}

@article{lorch2021dibs,
  title={DiBS: Differentiable Bayesian Structure Learning},
  author={Lorch, Lars and Rothfuss, Jonas and Sch{\"o}lkopf, Bernhard and Krause, Andreas},
  journal=nips,
  year={2021}
}

@inproceedings{louizos2018learning,
  title={Learning Sparse Neural Networks through $L_0$ Regularization},
  author={Louizos, Christos and Welling, Max and Kingma, Diederik P},
  booktitle=iclr,
  year={2018}
}

@article{loubet2019reparameterizing,
  title={Reparameterizing Discontinuous Integrands for Differentiable Rendering},
  author={Loubet, Guillaume and Holzschuch, Nicolas and Wenzel, Jakob},
  journal=togsgasia,
  number={6},
  volume={38},
  year={2019}
}

@inproceedings{lym2019prunetrain,
  title={PruneTrain: Fast Neural Network Training by Dynamic Sparse Model Reconfiguration},
  author={Lym, Sangkug and Choukse, Esha and Zangeneh, Siavash and Wen, Wei and Sanghavi, Sujay and Erez, Mattan},
  booktitle={Proceedings of the International Conference for High Performance Computing, Networking, Storage and Analysis},
  year={2019}
}

@inproceedings{maddison2017concrete,
  title={The Concrete Distribution: A Continuous Relaxation of Discrete Random Variables},
  author={Maddison, Chris J and Mnih, Andriy and Teh, Yee Whye},
  booktitle=iclr,
  year={2017}
}

@inproceedings{mahajan2018exploring,
  title={Exploring the Limits of Weakly Supervised Pretraining},
  author={Mahajan, Dhruv and Girshick, Ross and Ramanathan, Vignesh and He, Kaiming and Paluri, Manohar and Li, Yixuan and Bharambe, Ashwin and Van Der Maaten, Laurens},
  booktitle=eccv,
  year={2018}
}

@inproceedings{martens2015optimizing,
  title={Optimizing Neural Networks with {K}ronecker-Factored Approximate Curvature},
  author={Martens, James and Grosse, Roger},
  booktitle=icml,
  year={2015}
}

@article{martens2020new,
  title={New Insights and Perspectives on the Natural Gradient Method},
  author={James Martens},
  journal=jmlr,
  year={2020}
}

@article{mcculloch1943logical,
  title={A logical calculus of the ideas immanent in nervous activity},
  author={McCulloch, Warren S and Pitts, Walter},
  journal={{The Bulletin of Mathematical Biophysics}},
  volume={5},
  number={4},
  pages={115--133},
  year={1943},
  publisher={Springer}
}

@inproceedings{mena2018learning,
  title={Learning Latent Permutations with Gumbel-Sinkhorn Networks},
  author={Gonzalo Mena and David Belanger and Scott Linderman and Jasper Snoek},
  booktitle=iclr,
  year={2018}
}

@article{menabrea1843sketch,
  title={{Sketch of the Analytical Engine, invented by Charles Babbage, Esq., by LF Menabrea, of Turin, officer of the Military Engineers}},
  author={Menabrea, Luigi F and Lovelace, Ada A},
  journal={Translated and with notes by AA, L.~Taylor's Scientific Memoirs},
  volume={3},
  pages={666--731},
  year={1843}
}

@article{menger1942statistical,
  title={Statistical metrics},
  author={Menger, Karl},
  journal={Proceedings of the National Academy of Sciences of the United States of America},
  volume={28},
  number={12},
  pages={535},
  year={1942},
  publisher={National Academy of Sciences}
}

@inproceedings{mensch2018differentiable,
  title={Differentiable Dynamic Programming for Structured Prediction and Attention},
  author={Mensch, Arthur and Blondel, Mathieu},
  booktitle=icml,
  year={2018}
}

@inproceedings{metta2015sorting,
  title={Sorting Using Spiking Neural P Systems with Anti-spikes and Rules on Synapses},
  author={Metta, Venkata Padmavati and Kelemenova, Alica},
  booktitle={International Conference on Membrane Computing},
  year={2015}
}

@article{mihai2021differentiable,
  title={Differentiable Drawing and Sketching},
  author={Mihai, Daniela and Hare, Jonathon},
  journal=corr,
  year={2021}
}

@article{mihai2021learning,
  title={Learning to Draw: Emergent Communication through Sketching},
  author={Mihai, Daniela and Hare, Jonathon},
  journal=nips,
  year={2021}
}

@inproceedings{mildenhall2020nerf,
  title={NeRF: Representing Scenes as Neural Radiance Fields for View Synthesis},
  author={Mildenhall, Ben and Srinivasan, Pratul P and Tancik, Matthew and Barron, Jonathan T and Ramamoorthi, Ravi and Ng, Ren},
  booktitle=eccv,
  year={2020}
}

@article{mocanu2018scalable,
  title={Scalable training of artificial neural networks with adaptive sparse connectivity inspired by network science},
  author={Mocanu, Decebal Constantin and Mocanu, Elena and Stone, Peter and Nguyen, Phuong H and Gibescu, Madeleine and Liotta, Antonio},
  journal=natc,
  year={2018}
}

@inproceedings{molchanov2017variational,
  title={Variational Dropout Sparsifies Deep Neural Networks},
  author={Molchanov, Dmitry and Ashukha, Arsenii and Vetrov, Dmitry},
  booktitle=icml,
  year={2017}
}

@article{murshed2021machine,
  title={Machine Learning at the Network Edge: A Survey},
  author={Murshed, MG Sarwar and Murphy, Christopher and Hou, Daqing and Khan, Nazar and Ananthanarayanan, Ganesh and Hussain, Faraz},
  journal={ACM Computing Surveys (CSUR)},
  year={2021}
}

@article{netzer2011svhn,
  title={Reading digits in natural images with unsupervised feature learning},
  author={Netzer, Yuval and Wang, Tao and Coates, Adam and Bissacco, Alessandro and Wu, Bo and Ng, Andrew Y},
  year={2011}
}

@book{neugebauer1969exact,
  title={{The Exact Sciences in Antiquity}},
  author={Neugebauer, O.},
  series={Acta historica scientiarum naturalium et medicinalium},
  publisher={Dover Publications},
  year={1969}
}

@inproceedings{niemeyer2020differentiable,
  title={Differentiable Volumetric Rendering: Learning Implicit 3D Representations without 3D Supervision},
  author={Niemeyer, Michael and Mescheder, Lars and Oechsle, Michael and Geiger, Andreas},
  booktitle=cvpr,
  year={2020}
}

@article{niepert2021implicit,
  title={Implicit MLE: Backpropagating through discrete exponential family distributions},
  author={Niepert, Mathias and Minervini, Pasquale and Franceschi, Luca},
  journal=nips,
  year={2021}
}

@article{nimier2019mitsuba,
  title={Mitsuba 2: A retargetable forward and inverse renderer},
  author={Nimier-David, Merlin and Vicini, Delio and Zeltner, Tizian and Jakob, Wenzel},
  journal={ACM Transactions on Graphics (TOG)},
  volume={38},
  number={6},
  pages={1--17},
  year={2019},
  publisher={ACM New York, NY, USA}
}

@book{nocedal2006numerical,
  author={Nocedal, J. and Wright, S.J.},
  title={{Numerical Optimization}},
  publisher={Springer New York},
  year={2006}
}

@inproceedings{palazzi2019end,
  title={End-to-End {6-DoF} Object Pose Estimation Through Differentiable Rasterization},
  author={Palazzi, Andrea and Bergamini, Luca and Calderara, Simone and Cucchiara, Rita},
  booktitle=eccvw,
  year={2019}
}

@incollection{paszke2019pytorch,
 title={PyTorch: An Imperative Style, High-Performance Deep Learning Library},
 author={Paszke, Adam and Gross, Sam and Massa, Francisco and Lerer, Adam and Bradbury, James and Chanan, Gregory
         and Killeen, Trevor and Lin, Zeming and Gimelshein, Natalia and Antiga, Luca and Desmaison, Alban
         and Kopf, Andreas and Yang, Edward and DeVito, Zachary and Raison, Martin and Tejani, Alykhan
         and Chilamkurthy, Sasank and Steiner, Benoit and Fang, Lu and Bai, Junjie and Chintala, Soumith},
 booktitle=nips,
 year={2019}
}

@inproceedings{patel2022recall,
  title={Recall@k Surrogate Loss with Large Batches and Similarity Mixup},
  author={Patel, Yash and Tolias, Giorgos and Matas, Jiri},
  booktitle=cvpr,
  year={2022}
}

@inproceedings{paulus2020gradient,
  title={Gradient estimation with stochastic softmax tricks},
  author={Paulus, Max B and Choi, Dami and Tarlow, Daniel and Krause, Andreas and Maddison, Chris J},
  booktitle=nips,
  year={2020}
}

@inproceedings{paulus2021comboptnet,
  title={CombOptNet: Fit the Right NP-Hard Problem by Learning Integer Programming Constraints},
  author={Paulus, Anselm and Rol{\'i}nek, Michal and Musil, V{\'i}t and Amos, Brandon and Martius, Georg},
  booktitle=icml,
  year={2021}
}

@article{peng2021shape,
  title={Shape As Points: A Differentiable Poisson Solver},
  author={Peng, Songyou and Jiang, Chiyu and Liao, Yiyi and Niemeyer, Michael and Pollefeys, Marc and Geiger, Andreas},
  journal=nips,
  year={2021}
}

@inproceedings{pham2021meta,
  title={Meta Pseudo Labels},
  author={Pham, Hieu and Dai, Zihang and Xie, Qizhe and Le, Quoc V},
  booktitle=cvpr,
  year={2021}
}

@article{pietruszka2020successive,
  title={Successive Halving Top-k Operator},
  author={Pietruszka, Micha{\l} and Borchmann, {\L}ukasz and Grali{\'n}ski, Filip},
  journal=corr,
  year={2020}
}

@article{pilnaci2017newton,
  author={Pilanci, Mert and Wainwright, Martin J.},
  title={{Newton Sketch: A Near Linear-Time Optimization Algorithm with Linear-Quadratic Convergence}},
  journal={SIAM Journal on Optimization},
  year={2017}
}

@article{plotz2018neural,
  title={Neural Nearest Neighbors Networks},
  author={Pl{\"o}tz, Tobias and Roth, Stefan},
  journal=nips,
  year={2018}
}

@article{pobrotyn2021neuralndcg,
  title={{NeuralNDCG: Direct Optimisation of a Ranking Metric via Differentiable Relaxation of Sorting}},
  author={Pobrotyn, Przemys{\l}aw and Bia{\l}obrzeski, Rados{\l}aw},
  journal=corr,
  year={2021}
}

@inproceedings{prillo2020softsort,
  title={SoftSort: A continuous relaxation for the argsort operator},
  author={Prillo, Sebastian and Eisenschlos, Julian},
  booktitle=icml,
  year={2020}
}

@article{qin2013introducing,
  title={Introducing LETOR 4.0 datasets},
  author={Qin, Tao and Liu, Tie-Yan},
  journal=corr,
  year={2013}
}

@article{qin2020binary,
  title={Binary Neural Networks: A Survey},
  author={Qin, Haotong and Gong, Ruihao and Liu, Xianglong and Bai, Xiao and Song, Jingkuan and Sebe, Nicu},
  journal={Pattern Recognition},
  year={2020}
}

@article{quinlan1986induction,
  title={Induction of Decision Trees},
  author={Quinlan, J. Ross},
  journal={Machine Learning},
  year={1986}
}

@article{radford2021learning,
  title={Learning transferable visual models from natural language supervision},
  author={Radford, Alec and Kim, Jong Wook and Hallacy, Chris and Ramesh, Aditya and Goh, Gabriel and Agarwal, Sandhini and Sastry, Girish and Askell, Amanda and Mishkin, Pamela and Clark, Jack and others},
  journal=corr,
  year={2021}
}

@article{rakotosaona2021differentiable,
  title={Differentiable Surface Triangulation},
  author={Rakotosaona, Marie-Julie and Aigerman, Noam and Mitra, Niloy J and Ovsjanikov, Maks and Guerrero, Paul},
  journal=tog,
  year={2021}
}

@misc{rapin2018nevergrad,
  author={J. Rapin and O. Teytaud},
  title={{Nevergrad - A gradient-free optimization platform}},
  journal={GitHub repository},
  url={https://GitHub.com/FacebookResearch/Nevergrad},
  year={2018}
}

@inproceedings{redmon2016yolo,
  author={Redmon, Joseph and Divvala, Santosh and Girshick, Ross and Farhadi, Ali},
  title={You Only Look Once: Unified, Real-Time Object Detection},
  booktitle=cvpr,
  year={2016}
}

@inproceedings{rhodin2015versatile,
  title={A Versatile Scene Model with Differentiable Visibility applied to Generative Pose Estimation},
  author={Rhodin, Helge and Robertini, Nadia and Richardt, Christian and Seidel, Hans-Peter and Theobalt, Christian},
  booktitle=cvpr,
  year={2015}
}

@inproceedings{riad2022learning,
  title={Learning Strides in Convolutional Neural Networks},
  author={Riad, Rachid and Teboul, Olivier and Grangier, David and Zeghidour, Neil},
  booktitle=iclr,
  year={2022}
}

@inproceedings{ridnik2021imagenet,
  title={Imagenet-21k Pretraining for the Masses},
  author={Ridnik, Tal and Ben-Baruch, Emanuel and Noy, Asaf and Zelnik-Manor, Lihi},
  booktitle=nips,
  year={2021}
}

@article{rippel2015spectral,
  title={Spectral representations for convolutional neural networks},
  author={Rippel, Oren and Snoek, Jasper and Adams, Ryan P},
  journal=nips,
  year={2015}
}

@inproceedings{rolinek2020optimizing,
  title={Optimizing rank-based metrics with blackbox differentiation},
  author={Rolinek, Michal and Musil, Vit and Paulus, Anselm and Vlastelica, Marin and Michaelis, Claudio and Martius, Georg},
  booktitle=cvpr,
  year={2020}
}

@inproceedings{rolinek2020deep,
  title={Deep Graph Matching via Blackbox Differentiation of Combinatorial Solvers},
  author={Rol{\'i}nek, Michal and Swoboda, Paul and Zietlow, Dominik and Paulus, Anselm and Musil, V{\'i}t and Martius, Georg},
  booktitle=eccv,
  year={2020}
}

@inproceedings{romero2021flexconv,
  title={FlexConv: Continuous Kernel Convolutions with Differentiable Kernel Sizes},
  author={Romero, David W and Bruintjes, Robert-Jan and Tomczak, Jakub M and Bekkers, Erik J and Hoogendoorn, Mark and van Gemert, Jan C},
  booktitle=iclr,
  year={2022}
}

@article{rumelhart1986learning,
  title={Learning representations by back-propagating errors},
  author={Rumelhart, David E and Hinton, Geoffrey E and Williams, Ronald J},
  journal={Nature},
  volume={323},
  number={6088},
  pages={533--536},
  year={1986},
  publisher={Nature Publishing Group}
}

@article{salzberg1994c4,
  title={C4. 5: Programs for machine learning by j. ross quinlan. morgan kaufmann publishers, inc., 1993},
  author={Salzberg, Steven L},
  publisher={Kluwer Academic Publishers},
  year={1994}
}

@article{senior2020improved,
  title={Improved Protein Structure Prediction using Potentials from Deep Learning},
  author={Senior, Andrew W. and Evans, Richard and Jumper, John and Kirkpatrick, James and Sifre, Laurent
          and Green, Tim and Qin, Chongli and {\v{Z}}{\'i}dek, Augustin and Nelson, Alexander W. R. and Bridgland, Alex
          and Penedones, Hugo and Petersen, Stig and Simonyan, Kareand Crossan, Steve and Kohli, Pushmeet
          and Jones, David T. and Silver, David and Kavukcuoglu, Koray and Hassabis, Demis},
  journal=nat,
  year={2020}
}

@article{scott2022differentiable,
  title={Differentiable IFS Fractals},
  author={Scott, Cory Braker},
  journal=corr,
  year={2022}
}

@article{seng2021embedded,
  title={Embedded intelligence on fpga: Survey, applications and challenges},
  author={Seng, Kah Phooi and Lee, Paik Jen and Ang, Li Minn},
  journal={Electronics},
  volume={10},
  number={8},
  pages={895},
  year={2021},
  publisher={Multidisciplinary Digital Publishing Institute}
}

@inproceedings{shah2020learning,
  title={Learning Differentiable Programs with Admissible Neural Heuristics},
  author={Shah, Ameesh and Zhan, Eric and Sun, Jennifer J and Verma, Abhinav and Yue, Yisong and Chaudhuri, Swarat},
  booktitle=nips,
  year={2020}
}

@article{shani2019dynamics,
  title={Dynamics of Analog Logic-Gate Networks for Machine Learning},
  author={Shani, Itamar and Shaughnessy, Liam and Rzasa, John and Restelli, Alessandro and Hunt, Brian R and Komkov, Heidi and Lathrop, Daniel P},
  journal={Chaos: An Interdisciplinary Journal of Nonlinear Science, AIP Publishing LLC},
  year={2019}
}

@inproceedings{shekhovtsov2019feedforward,
  title={Feed-Forward Propagation in Probabilistic Neural Networks with Categorical and Max Layers},
  author={Alexander Shekhovtsov and Boris Flach},
  booktitle=iclr,
  year={2019}
}

@inproceedings{shi2020match,
  title={{MATch}: Differentiable Material Graphs for Procedural Material Capture},
  author={Shi, Liang and Li, Beichen and Ha{\v{s}}an, Milo{\v{s}} and Sunkavalli, Kalyan and Boubekeur, Tamy and Mech, Radomir and Matusik, Wojciech},
  booktitle=togsgasia,
  year={2020}
}

@inproceedings{shirobokov2020black,
  title={Black-box Optimization with Local Generative Surrogates},
  author={Shirobokov, Sergey and Belavin, Vladislav and Kagan, Michael and Ustyuzhanin, Andrey and Baydin, At{\i}l{\i}m G{\"u}nes},
  booktitle={Workshop on Real World Experiment Design and Active Learning at International Conference on Machine Learning (ICML)},
  year={2020}
}

@article{sinkhorn1967concerning,
  title={Concerning nonnegative matrices and doubly stochastic matrices},
  author={Sinkhorn, Richard and Knopp, Paul},
  journal={Pacific Journal of Mathematics},
  volume={21},
  number={2},
  pages={343--348},
  year={1967},
  publisher={Mathematical Sciences Publishers}
}

@article{sitzmann2018end,
  title={End-to-end optimization of optics and image processing for achromatic extended depth of field and super-resolution imaging},
  author={Sitzmann, Vincent and Diamond, Steven and Peng, Yifan and Dun, Xiong and Boyd, Stephen and Heidrich, Wolfgang and Heide, Felix and Wetzstein, Gordon},
  journal=tog,
  year={2018}
}

@inproceedings{sitzmann2019scene,
  title={Scene Representation Networks: Continuous {3D}-Structure-Aware Neural Scene Representations},
  author={Sitzmann, Vincent and Zollh{\"{o}}fer, Michael and Wetzstein, Gordon},
  booktitle=nips,
  year={2019}
}

@article{srivastava2014dropout,
  title={Dropout: a Simple Way to Prevent Neural Networks from Overfitting},
  author={Srivastava, Nitish and Hinton, Geoffrey and Krizhevsky, Alex and Sutskever, Ilya and Salakhutdinov, Ruslan},
  journal=jmlr,
  year={2014}
}

@inproceedings{sun2017revisiting,
  title={Revisiting Unreasonable Effectiveness of Data in Deep Learning Era},
  author={Sun, Chen and Shrivastava, Abhinav and Singh, Saurabh and Gupta, Abhinav},
  booktitle=cvpr,
  year={2017}
}

@inproceedings{swezey2021pirank,
  title={{PiRank: Learning To Rank via Differentiable Sorting}},
  author={Swezey, Robin and Grover, Aditya and Charron, Bruno and Ermon, Stefano},
  booktitle=nips,
  year={2021}
}

@inproceedings{szegedy2015going,
  title={{Going Deeper with Convolutions}},
  author={Szegedy, Christian and Liu, Wei and Jia, Yangqing and Sermanet, Pierre and Reed, Scott and Anguelov, Dragomir and Erhan, Dumitru and Vanhoucke, Vincent and Rabinovich, Andrew},
  booktitle=cvpr,
  year={2015}
}

@inproceedings{taylor2008softrank,
  title={Softrank: Optimizing non-smooth Rank Metrics},
  author={Taylor, Michael and Guiver, John and Robertson, Stephen and Minka, Tom},
  booktitle={Proceedings of the 2008 International Conference on Web Search and Data Mining},
  year={2008}
}

@article{telikani2021evolutionary,
  title={Evolutionary Machine Learning: A Survey},
  author={Telikani, Akbar and Tahmassebi, Amirhessam and Banzhaf, Wolfgang and Gandomi, Amir H},
  journal={ACM Computing Surveys (CSUR)},
  year={2021}
}

@inproceedings{tewari2021advances,
    title = {Advances in Neural Rendering},
    author = {Tewari, A. and Fried, O. and Thies, J. and Sitzmann, V. and Lombardi, S. and Xu, Z. and Simon, T. and Nie\ss{}ner, M. and Tretschk, E. and Liu, L. and Mildenhall, B. and Srinivasan, P. and Pandey, R. and Orts-Escolano, S. and Fanello, S. and Guo, M. and Wetzstein, G. and Zhu, J.-Y. and Theobalt, C. and Agrawala, M. and Goldman, D. B and Zollh\"{o}fer, M.},
    booktitle = {ACM SIGGRAPH 2021 Courses},
    year = {2021}
}

@article{thandiackal2022differentiable,
  title={{Differentiable Zooming for Multiple Instance Learning on Whole-Slide Images}},
  author={Thandiackal, Kevin and Chen, Boqi and Pati, Pushpak and Jaume, Guillaume and Williamson, Drew FK and Gabrani, Maria and Goksel, Orcun},
  journal=corr,
  year={2022}
}

@article{thrun1991monk,
  title={The monk's problems: A performance comparison of different learning algorithms},
  author={Thrun, Sebastian B and Bala, Jerzy W and Bloedorn, Eric and Bratko, Ivan and Cestnik, Bojan and Cheng, John and De Jong, Kenneth A and Dzeroski, Saso and Fisher, Douglas H and Fahlman, Scott E and others},
  year={1991}
}

@book{tikhonov1977solutions,
  author = {Tikhonov, A. N. and Arsenin, V. Y.},
  publisher = {W.H.~Winston},
  title = {Solutions of Ill-posed problems},
  year = 1977
}

@article{tucker2017rebar,
  title={REBAR: Low-variance, unbiased gradient estimates for discrete latent variable models},
  author={Tucker, George and Mnih, Andriy and Maddison, Chris J and Lawson, John and Sohl-Dickstein, Jascha},
  journal=nips,
  year={2017}
}

@inproceedings{umuroglu2017finn,
  title={Finn: A Framework for Fast, Scalable Binarized Neural Network Inference},
  author={Umuroglu, Yaman and Fraser, Nicholas J and Gambardella, Giulio and Blott, Michaela and Leong, Philip and Jahre, Magnus and Vissers, Kees},
  booktitle={Proceedings of the 2017 ACM/SIGDA International Symposium on Field-Programmable Gate Arrays},
  year={2017}
}

@inproceedings{urban2016deep,
  title={Do Deep Convolutional Nets Really Need to be Deep and Convolutional?},
  author={Urban, Gregor and Geras, Krzysztof J and Kahou, Samira Ebrahimi and Aslan, Ozlem and Wang, Shengjie and Caruana, Rich and Mohamed, Abdelrahman and Philipose, Matthai and Richardson, Matt},
  booktitle=iclr,
  year={2018}
}

@article{van2022analyzing,
  title={Analyzing Differentiable Fuzzy Logic Operators},
  author={van Krieken, Emile and Acar, Erman and van Harmelen, Frank},
  journal={Artificial Intelligence},
  year={2022}
}

@inproceedings{vaswani2017attention,
  author={Vaswani, Ashish and Shazeer, Noam and Parmar, Niki and Uszkoreit, Jakob and Jones, Llion and Gomez, Aidan N
          and Kaiser, \L ukasz and Polosukhin, Illia},
  title={Attention is All you Need},
  booktitle=nips,
  year={2017}
}

@inproceedings{vicini2022differentiable,
  title={Differentiable Signed Distance Function Rendering},
  author={Vicini, Delio and Speierer, Sebastien and Jakob, Wenzel},
  booktitle=tog,
  year={2022}
}

@inproceedings{vinyals2016order,
  title={Order Matters: Sequence to sequence for sets},
  author={Oriol Vinyals and Samy Bengio and Manjunath Kudlur},
  booktitle=iclr,
  year={2016}
}

@inproceedings{vlastelica2019differentiation,
  title={Differentiation of blackbox combinatorial solvers},
  author={Vlastelica, Marin and Paulus, Anselm and Musil, Vit and Martius, Georg and Rolinek, Michal},
  booktitle=iclr,
  year={2020}
}

@inproceedings{wadia2021whitening,
  title={Whitening and second order optimization both make information in the dataset unusable during training, and can reduce or prevent generalization},
  author={Wadia, Neha and Duckworth, Daniel and Schoenholz, Samuel S and Dyer, Ethan and Sohl-Dickstein, Jascha},
  booktitle=icml,
  year={2021}
}

@article{wah1984partitioning,
  title={A Partitioning Approach to the Design of Selection Networks},
  author={Benjamin W.~Wah and Kuo-Liang Chen},
  journal={IEEE Transactions on Computers},
  volume={33},
  issue={3},
  pages={261--268},
  year={1984}
}

@inproceedings{wang2016natural,
  title={Natural-Parameter Networks: A Class of Probabilistic Neural Networks},
  author={Hao Wang and Xingjian Shi and Dit-Yan Yeung},
  booktitle=nips,
  year={2016}
}

@article{williams1992simple,
  title={Simple Statistical Gradient-Following Algorithms for Connectionist Reinforcement Learning},
  author={Williams, Ronald J},
  journal={Machine learning},
  volume={8},
  number={3},
  pages={229--256},
  year={1992},
  publisher={Springer}
}

@inproceedings{wu2019deterministic,
  title={Deterministic variational inference for robust bayesian neural networks},
  author={Wu, Anqi and Nowozin, Sebastian and Meeds, Edward and Turner, Richard E and Hernandez-Lobato, Jose Miguel and Gaunt, Alexander L},
  booktitle=iclr,
  year={2019}
}

@inproceedings{xie2020self,
  title={Self-training with noisy student improves imagenet classification},
  author={Xie, Qizhe and Luong, Minh-Thang and Hovy, Eduard and Le, Quoc V},
  booktitle=cvpr,
  year={2020}
}

@inproceedings{xie2020differentiable,
  title={Differentiable Top-k with Optimal Transport},
  author={Xie, Yujia and Dai, Hanjun and Chen, Minshuo and Dai, Bo and Zhao, Tuo and Zha, Hongyuan and Wei, Wei and Pfister, Tomas},
  booktitle=nips,
  year={2020}
}

@inproceedings{yan2016perspective,
  title={Perspective Transformer Nets: Learning Single-View {3D} Object Reconstruction without {3D} Supervision},
  author={Yan, Xinchen and Yang, Jimei and Yumer, Ersin and Guo, Yijie and Lee, Honglak},
  booktitle=nips,
  year={2016}
}

@inproceedings{yang2020consistency,
  title={On the Consistency of Top-k Surrogate Losses},
  author={Yang, Forest and Koyejo, Sanmi},
  booktitle=icml,
  year={2020}
}

@inproceedings{yang2022safe,
  title={Safe Neurosymbolic Learning with Differentiable Symbolic Execution},
  author={Yang, Chenxi and Chaudhuri, Swarat},
  booktitle=iclr,
  year={2022}
}

@article{yariv2020multiview,
  title={Multiview neural surface reconstruction by disentangling geometry and appearance},
  author={Yariv, Lior and Kasten, Yoni and Moran, Dror and Galun, Meirav and Atzmon, Matan and Ronen, Basri and Lipman, Yaron},
  journal=nips,
  year={2020}
}

@article{yi2018neural,
  title={Neural-symbolic vqa: Disentangling reasoning from vision and language understanding},
  author={Yi, Kexin and Wu, Jiajun and Gan, Chuang and Torralba, Antonio and Kohli, Pushmeet and Tenenbaum, Josh},
  journal=nips,
  year={2018}
}

@article{yi2020clevrer,
  title={Clevrer: Collision events for video representation and reasoning},
  author={Yi, Kexin and Gan, Chuang and Li, Yunzhu and Kohli, Pushmeet and Wu, Jiajun and Torralba, Antonio and Tenenbaum, Joshua B},
  journal=iclr,
  year={2020}
}

@article{yifan2019differentiable,
  title={Differentiable Surface Splatting for Point-based Geometry Processing},
  author={Yifan, Wang and Serena, Felice and Wu, Shihao and {\"{O}}ztireli, Cengiz and Sorkine-Hornung, Olga},
  volume=38,
  number=6,
  journal=tog,
  year={2019}
}

@inproceedings{zantedeschi2021learning,
  title={Learning Binary Decision Trees by Argmin Differentiation},
  author={Zantedeschi, Valentina and Kusner, Matt and Niculae, Vlad},
  booktitle=icml,
  year={2021}
}

@unpublished{zazonivry2012pairwise,
  title={Pairwise Networks are Superior for Selection},
  author={Moshe Zazon-Ivry and Michael Codish},
  note={Unpublished manuscript},
  year={2012}
}

@article{zhai2021scaling,
  title={Scaling vision transformers},
  author={Zhai, Xiaohua and Kolesnikov, Alexander and Houlsby, Neil and Beyer, Lucas},
  journal=corr,
  year={2021}
}

@article{zhan2020field,
  title={Field Programmable Gate Array-based All-Layer Accelerator with Quantization Neural Networks for Sustainable Cyber-Physical Systems},
  author={Zhan, Jinyu and Zhou, Xingzhi and Jiang, Wei},
  journal={Software: Practice and Experience, Wiley Online Library},
  year={2020}
}

@article{zeng2019mlprune,
    title={{MLP}rune: Multi-Layer Pruning for Automated Neural Network Compression},
    author={Wenyuan Zeng and Raquel Urtasun},
    journal=open,
    year={2019}
}

@inproceedings{zhang2020path,
  title={Path-Space Differentiable Rendering},
  author={Zhang, Cheng and Miller, Bailey and Yan, Kan and Gkioulekas, Ioannis and Zhao, Shuang},
  booktitle=togsg,
  year={2020}
}

@article{zhang2021antithetic,
  title={Antithetic sampling for Monte Carlo differentiable rendering},
  author={Zhang, Cheng and Dong, Zhao and Doggett, Michael and Zhao, Shuang},
  journal={ACM Transactions on Graphics (TOG)},
  volume={40},
  number={4},
  pages={1--12},
  year={2021},
  publisher={ACM New York, NY, USA}
}

@inproceedings{zhang2021image,
  title={{Image GANs meet Differentiable Rendering for Inverse Graphics and Interpretable 3D Neural Rendering}},
  author={Zhang, Yuxuan and Chen, Wenzheng and Ling, Huan and Gao, Jun and Zhang, Yinan and Torralba, Antonio and Fidler, Sanja},
  booktitle=iclr,
  year={2021}
}

@article{zheng2018dags,
  title={DAGs with NO TEARS: Continuous Optimization for Structure Learning},
  author={Zheng, Xun and Aragam, Bryon and Ravikumar, Pradeep K and Xing, Eric P},
  journal=nips,
  year={2018}
}

@inproceedings{zhou2021effective,
  title={Effective Sparsification of Neural Networks with Global Sparsity Constraint},
  author={Zhou, Xiao and Zhang, Weizhong and Xu, Hang and Zhang, Tong},
  booktitle=cvpr,
  year={2021}
}

@article{zimmer2021differentiable,
  title={Differentiable Logic Machines},
  author={Zimmer, Matthieu and Feng, Xuening and Glanois, Claire and Jiang, Zhaohui and Zhang, Jianyi and Weng, Paul and Dong, Li and Jianye, Hao and Wulong, Liu},
  journal=corr,
  year={2021}
}

@article{zwitter1988uci,
  title={UCI Machine Learning Repository: Breast Cancer Dataset},
  author={Zwitter, M and Soklic, M},
  year={1988}
}

\end{document}